\documentclass[12pt,oneside]{report}

\usepackage[margin=1in]{geometry}
\usepackage{graphicx}   %
\usepackage{booktabs}   %
\usepackage{hyperref}   %
\usepackage{lipsum}     %
\usepackage{titling}    %
\usepackage[numbers]{natbib}
\usepackage[table]{xcolor}
\usepackage{colortbl}
\usepackage{xspace}
\usepackage{amsmath,amssymb}
\usepackage{bm}
\usepackage{multirow}
\usepackage{wrapfig}
\usepackage{enumitem}
\usepackage{pifont}
\usepackage{cleveref}
\usepackage{tikz}
\usepackage{capt-of}
\usepackage{rotating}
\usepackage{silence}
\usepackage{setspace}

\renewcommand{\arraystretch}{0.95}

\newcommand{\cmark}{\ding{51}}
\newcommand{\xmark}{\ding{55}}
\newcommand{\etal}{\textit{et al.}}

\hypersetup{
  colorlinks=true,
  linkcolor=black,
  citecolor=black,
  urlcolor=blue,
  pdfauthor={Your Name},
  pdftitle={Your Thesis Title}
}
\title{Efficient 3D Content Reconstruction and Generation}
\author{Jiahao Li}
\date{\today} %

\begin{document}
\hypersetup{pageanchor=false}

\begin{titlepage}
  \centering

  {\huge \textbf{\thetitle}\par}\vspace{1.5em}

  {\large BY}\par\vspace{0.5em}
  {\large JIAHAO LI}\par\vspace{4em}

  {\large A thesis submitted\par
  in partial fulfillment of the requirements for\par
  the degree of\par
  Doctor of Philosophy in Computer Science\par
  at the\par
  TOYOTA TECHNOLOGICAL INSTITUTE AT CHICAGO}\par\vspace{10em}
  
  {\large Thesis Committee:}\par\vspace{0.25em}
  {\large Gregory Shakhnarovich (Advisor)}\par
  {\large Matthew R. Walter}\par
  {\large Michael Maire}\par

  \vfill

  {\large Chicago, United States}\\
  {\large \today} %
\end{titlepage}

\pagenumbering{roman} %

\begin{abstract}
\noindent

Automatic 3D content creation seeks to replace labor-intensive modeling and scanning pipelines with systems that can synthesize or recover 3D assets directly from text or images. Its applications span video games, virtual reality, robotics, and simulation, enabling rapid asset prototyping, diverse interactive world generation, and efficient 3D data collection for training foundation models. Contemporary solutions largely follow two complementary paradigms: (i) text- or image-to-3D generation, which learns priors over 3D geometry and appearance to create novel assets from natural language or a single view image; and (ii) 3D reconstruction, which estimates camera poses and geometry from RGB images. This thesis advances both directions. On the generation side, I introduce Instant3D, which combines multi-view diffusion with feed-forward sparse-view 3D reconstruction to produce high-quality assets in 5–20 seconds. On the reconstruction side, I develop FastMap, a structure-from-motion pipeline that achieves up to 10$\times$ speedup over prior state-of-the-art by using first-order optimization with fused GPU kernels extensively, while maintaining comparable pose accuracy and downstream novel view synthesis quality.

\end{abstract}

\chapter*{Acknowledgments}
\addcontentsline{toc}{chapter}{Acknowledgments}

\noindent
I am grateful to my advisor, Prof.\ Greg Shakhnarovich, for his guidance,
insight, and steady support throughout my PhD career. My parents, my brother,
and my girlfriend Feiran, kept me grounded with encouragement and
emotional support. I thank my thesis committee members
Matthew R. Walter and Michael Maire for their time, feedback, and guidance.
My labmates Haochen Wang, Xiaodan Du, Raymond Yeh, Marcelo
Sandoval-Castañeda, Shester Gueuwou and Anand Bhattad made the lab a supportive and
engaged place through discussions and day-to-day help. Collaborations
with Adobe (Sai Bi, Yicong Hong, Fujun Luan, Kalyan Sunkavalli, Hao Tan,
Desai Xie, Yinghao Xu, Zexiang Xu, Kai Zhang) and Toyota Research
Institute (Vitor Campagnolo Guizilini, Muhammad Zubair Irshad, Igor
Vasiljevic) were productive and full of good ideas.
Friends at TTIC and Uchicago (Joshua Ahn, Jiading Fang, Shengjie Lin,
Kevin Suk, David Yunis, Xiao Zhang) brought insightful discussions,
and friends during my Master's (Hongming Tan, Tong Wu, Minjie
Yang, Tianwei Zhou, Xiaochen Zhou) helped ease the stress of PhD
applications. Lastly, I thank my roommate Shinan Liu for tolerating my
unhealthy daily schedule for 4 years.

\tableofcontents
\listoffigures
\listoftables
\cleardoublepage
\hypersetup{pageanchor=true}

\clearpage
\pagenumbering{arabic} %

\chapter{Introduction}
\label{chap:intro}
Automatic 3D content creation seeks to replace labor-intensive modeling and scanning pipelines with systems that can synthesize or recover 3D assets directly from text or images. Instead of sculpting meshes by hand or running carefully calibrated multi-view capture rigs, creators increasingly want to describe a scene in natural language or upload a set of RGB photos and obtain a usable 3D object or environment. The same approach is equally valuable on the data side: large-scale 3D datasets underpin progress in 3D perception and embodied AI, but acquiring real-world assets with accurate geometry, texture, and camera poses remains expensive and slow.  These methods are already transforming downstream applications. In the video game industry, rapid asset prototyping significantly reduces the cost of the original labor-intensive development cycle. In robotics and autonomous driving, fast 3D reconstruction and generation enable controllable simulators that can be used to generate data of rare but safety-critical edge cases.

Contemporary solutions to automatic 3D content creation largely follow two complementary paradigms. The first is text- or image-to-3D generation, where generative models create novel assets from natural language prompts or a single or multiple input images. Early work in this direction trained 3D diffusion models such as Diffusion-SDF~\citep{li2023diffusionsdf} and LION~\citep{zeng2022lion} directly on collections of hand-made 3D assets. However, compared to 2D images, available 3D training data is scarce and biased: while datasets like LAION-5B~\citep{schuhmann2022laion5b} contain on the order of billions of text--image pairs, the largest public 3D repositories such as Objaverse-XL~\citep{deitke2023objaversexl} provide ``only'' tens of millions of assets with limited diversity. As a result, purely 3D-trained generative models tend to struggle with instruction following, visual fidelity, diversity, and compositional generalization.

A second, now dominant, line of work sidesteps the 3D data bottleneck by leveraging powerful 2D text-to-image diffusion models as priors. Score-distillation-based methods~\citep{poole2022dreamfusion,wang2023score} use a pretrained 2D diffusion model to calculate gradients on rendered images, and back-propagate the gradients through a differentiable 3D representation (typically a NeRF~\citep{mildenhall2020nerf}) so that its views are assigned high likelihood under the 2D model. Compared to feed-forward models trained on pure 3D data, this strategy inherits the semantic coverage and visual quality of large-scale 2D models, and can produce 3D shapes of complex geometry and detailed texture. However, it requires test-time optimization, where each prompt requires many steps of iterative rendering and backpropagation, leading to inference times from one to ten hours per asset even on modern GPUs. These methods also tend to exhibit characteristic artifacts such as over-saturated colors, Janus (``multi-face'') problems caused by biased 2D priors on different viewing angles, and limited diversity even when sampling with different seeds.

The second major paradigm for 3D content creation is 3D reconstruction, which jointly estimates camera poses and geometry from RGB images. Previous structure-from-motion (SfM) systems such as COLMAP~\citep{schoenberger2016sfm} remain the de-facto standard for annotating large image collections with camera poses and sparse 3D structure. These geometry-based methods are widely used to produce pseudo-ground-truth for monocular depth estimation, learned feed-forward reconstruction, and NeRF-style novel view synthesis. However, they are computationally expensive: processing a scene with thousands of images can take many hours or even days, even when using GPU-accelerated solvers. The core bottleneck lies in large nonlinear least-squares problems, especially bundle adjustment, that are typically solved with iterative second-order Gauss-Newton variants such as Levenberg-Marquardt. While techniques like the Schur complement and sparse Cholesky factorization exploit problem structure and lead to considerable speed-up, each iteration still carries a high wall-clock cost that scales poorly with the size and connectivity of the reconstruction graph.

This scalability problem directly limits our ability to train and evaluate modern 3D-aware models on diverse real-world data. Synthetic 3D datasets have scaled rapidly and now support tasks from visual SLAM to 3D asset generation, but scaling real-world 3D data remains difficult because accurate camera poses are expensive to obtain. Efficiently turning large, unstructured image collections into reliable camera trajectories and sparse 3D structure is therefore a crucial piece of infrastructure for both reconstruction and generation research. It is not enough for an SfM pipeline to be accurate; it must also be simple to deploy and fast enough to keep up with the data needs of learning systems.

My thesis advances both the generative and reconstruction fronts of automatic 3D content creation. On the generative side, I introduce Instant3D~\citep{li2024instant3d}, a two-stage feed-forward text-to-3D system that couples diffusion-based multi-view generation with a large, feed-forward sparse-view 3D reconstruction model. It fine-tunes a vanilla 2D text-to-image diffusion model to generate a structured $2\times 2$ grid of four consistent views of the same object. These sparse multi-view images are then lifted directly into a triplane-based NeRF representation by a transformer-based reconstruction model that encodes the input views into patch tokens and decodes them into triplane parameters. Trained on multi-view renderings of roughly 750k objects from Objaverse~\citep{deitke2023objaverse}, this two-stage architecture produces high-quality and diverse 3D assets from text in 5--20 seconds---over two orders of magnitude faster than optimization-based score-distillation methods, while achieving comparable or even better quality. Building on this design space, we further explore how to improve and simply the model: Carve3D~\citep{xie2024carve3d} improves multi-view consistency of diffusion models with reinforcement learning fine-tuning, and DMV3D~\citep{xu2023dmv3d} aims at single-stage 3D diffusion generation that directly generate NeRF representations from text or images.

On the reconstruction side, I introduce FastMap~\citep{li2026fastmap}, a new global SfM framework focused on speed and simplicity. FastMap re-examines the standard assumptions in SfM optimization, and shows that a purely first-order approach can be both scalable and accurate. The method identifies two key bottlenecks that arise when naively replacing second-order solvers with gradient descent: the dependence of per-step complexity on the number of 3D points, and poor GPU utilization due to kernel launch overheads and memory traffic in off-the-shelf autograd implementations. FastMap addresses the first issue by designing all major optimization sub-problems---global rotation, translation alignment and epipolar pose refinement---so that each gradient step has computational complexity independent of the number of 3D points, relying instead on point-free formulations and pre-aggregated compact information over image pairs. To tackle the second issue, it uses custom fused CUDA kernels that perform forward and backward computations in a single kernel, drastically reducing launch overhead and avoiding unnecessary movement between global and shared memory on GPUs. Through extensive experiments on eight diverse datasets, FastMap~\citep{li2026fastmap} demonstrates that this combination of first-order optimization and careful systems design can yield up to $10\times$ speedups over GPU-accelerated COLMAP~\citep{schoenberger2016sfm} and GLOMAP~\citep{pan2024glomap}, while maintaining comparable pose accuracy and novel view synthesis quality.

Taken together, these contributions aim to move automatic 3D content creation toward a regime where high-quality 3D assets can be generated or reconstructed efficiently at scale. On the generative side, Instant3D~\citep{li2024instant3d} explores how to best harness 2D diffusion priors and large reconstruction models to deliver fast and faithful text- and image-to-3D pipelines. On the reconstruction side, FastMap~\citep{li2026fastmap} revisits SfM through the lens of first-order optimization and GPU-efficient implementation, enabling scalable recovery of camera poses from RGB images.

\chapter{Related Work}
\label{chap:related}
\section{3D Generation}
\label{sec:related:3d-generation}

\subsection{Pre-diffusion 3D generation}
\label{sec:related:prediffusion}

Prior to diffusion-based models, early 3D generation research focused on directly training on 3D data, and explored a spectrum of shape representations, each trading off resolution, geometry flexibility, and rendering efficiency. The papers summarized below are organized primarily by the 3D representation they use.

\noindent\textbf{Voxels and volumetric grids.} Voxel grids offered a straightforward way to apply 3D convolutions at the cost of coarse resolution. 3D ShapeNets\cite{Wu:2015:3SA} modeled shapes as occupancy grids and used deep networks for recognition and shape completion from partial observations. 3D-GAN\cite{wu2016learning} extended volumetric modeling with adversarial training, learning a latent space that can sample diverse 3D shapes and provide discriminative features for recognition.

\noindent\textbf{Point clouds.} Point-based generation avoids voxel discretization and uses an unordered point set structure. PSGN\cite{Fan_2017_CVPR} predicted point clouds from a single image and explicitly handled ambiguity in the mapping from image to 3D shape. Achlioptas \emph{et al.}\cite{pmlr-v80-achlioptas18a} studied autoencoders and latent-space generative models for point clouds. FoldingNet\cite{Yang_2018_CVPR} introduced a folding-based decoder that deforms a 2D grid to reconstruct point clouds, enabling compact auto-encoding. PointFlow\cite{Yang_2019_ICCV} modeled point clouds as a distribution of distributions using continuous normalizing flows, providing exact likelihoods and faithful reconstruction. TreeGAN\cite{Shu_2019_ICCV} used a tree-structured graph convolution generator and introduced Fr\'echet point cloud distance for evaluation. PointGrow\cite{Sun_2020_WACV} framed generation as an autoregressive process with self-attention to capture long-range dependencies between points.

\noindent\textbf{Meshes.} Explicit meshes are directly usable in downstream graphics pipelines but are harder to optimize. AtlasNet\cite{Groueix_2018_CVPR} generated surfaces as a collection of parametric patches learned by deforming 2D primitives. Pixel2Mesh\cite{Wang_2018_ECCV} used a graph convolutional network to progressively deform an ellipsoid into a target mesh from a single RGB image, with multi-stage refinement losses. GET3D\cite{gao2022get3d} advanced mesh generation by producing textured meshes from 2D image collections using differentiable rendering and adversarial supervision, yielding high-fidelity geometry and texture.

\noindent\textbf{Implicit fields (occupancy and SDF).} Implicit functions represent geometry continuously without fixed resolution. Occupancy Networks\cite{Mescheder_2019_CVPR} represented shapes via a learned decision boundary in function space, enabling high-resolution reconstructions with low memory overhead. DeepSDF\cite{Park_2019_CVPR} modeled signed distance functions for continuous shape representation, interpolation, and completion. IM-NET\cite{Chen_2019_CVPR} introduced an implicit field decoder that improved generative modeling quality when paired with autoencoders or GANs. LDIF\cite{Genova_2020_CVPR} combined global implicit fields with local shape elements to capture finer details and structured decompositions.

\noindent\textbf{Tri-plane representations.} Tri-plane factorization offers an efficient middle ground between 3D volumes and implicit fields. EG3D\cite{chan2022eg3d} popularized tri-planes in 3D-aware GANs by decoupling feature generation and rendering for efficient, high-quality synthesis. Building on the same representation, Triplane Diffusion\cite{Shue2023triplanediffusion} showed that tri-planes can be used as a compact 3D neural field parameterization for diffusion-based generation.

\subsection{Optimization-based 3D generation via score distillation}
\label{sec:related:score-distillation}

Optimization-based text-to-3D methods distill guidance from pretrained 2D diffusion models into a differentiable 3D representation. DreamFusion\cite{poole2022dreamfusion} introduced score distillation sampling (SDS) to optimize a NeRF so that its rendered views match a text-conditioned diffusion prior, enabling per-prompt 3D synthesis without 3D training data. Score Jacobian Chaining (SJC)\cite{Wang_2023_CVPR} follows a similar vein and provides an alternative theoretical explanation. DreamTime\cite{huang2023dreamtime} studies timestep sampling in SDS and proposes time-prioritized sampling to better align diffusion sampling with 3D optimization, improving convergence, quality, and diversity.

Several works address diversity and variance in the distilled gradients. ProlificDreamer\cite{wang2023prolificdreamer} models the 3D parameters as random variables and derives variational score distillation (VSD) to improve diversity and reduce oversmoothing in SDS-based optimization. SteinDreamer\cite{pmlr-v258-wang25j} interprets SDS and VSD as high-variance Monte Carlo estimators and introduces Stein score distillation with control variates (instantiated with monocular depth) to reduce variance and improve visual quality for both objects and scenes. Collaborative Score Distillation (CSD)\cite{kim2023csd} extends distillation to sets of images by treating multiple samples as particles in an SVGD update, improving consistency for editing panoramas, videos, and multi-view 3D scenes.

Other works target higher fidelity, personalization, or efficiency. Magic3D\cite{Lin_2023_CVPR} adopts a two-stage coarse-to-fine pipeline that first optimizes a sparse neural representation under low-resolution diffusion guidance and then refines a textured mesh under a high-resolution latent diffusion model. Fantasia3D\cite{Chen_2023_ICCV} disentangles geometry and appearance by guiding geometry with normal maps and modeling appearance with BRDF-based materials for photorealistic rendering. DreamBooth3D\cite{Raj_2023_ICCV} personalizes text-to-3D generation from a small set of subject images via a multi-stage optimization that mitigates viewpoint overfitting. DreamGaussian\cite{DBLP:conf/iclr/TangRZ0Z24} and GaussianDreamer\cite{Yi_2024_CVPR} replace NeRFs with 3D Gaussian splatting, combining diffusion guidance with faster optimization and mesh refinement or point-cloud priors to speed up generation while keeping real-time rendering. For editing, Instruct-NeRF2NeRF\cite{Haque_2023_ICCV} iteratively edits input images with instruction-conditioned diffusion and jointly optimizes the NeRF to maintain multi-view consistency.

\subsection{3D scene generation}
\label{sec:related:scene-generation}
Scene generation methods target full rooms or environments rather than isolated objects. \cite{pmlr-v235-epstein24a} introduces layout-driven factorization to separate structural layout from scene appearance in generation. Ctrl-Room\cite{Fang_2025_3DV} adds explicit layout constraints for text-to-3D room mesh generation, while ControlRoom3D\cite{Schult_2024_CVPR} uses semantic proxy rooms to provide controllable room synthesis with semantic structure. \cite{Koh_2023_AAAI} focuses on indoor scene synthesis using reconstructed RGB-D cues, and iControl3D\cite{li2024icontrol3d} provides an interactive system for controllable 3D scene generation.

Diffusion-based pipelines often reconstruct or inpaint scenes through multi-view supervision. Text2Room\cite{Hollein_2023_ICCV} extracts textured meshes from 2D text-to-image models, and Text2NeRF\cite{zhang2023text2nerf} uses text guidance to build NeRF scenes. NeRFiller\cite{Weber_2024_CVPR} addresses missing regions via generative 3D inpainting, 3D-SceneDreamer\cite{Zhang_2024_CVPR} targets text-driven 3D-consistent scene generation, and RealmDreamer\cite{shriram2024realmdreamer} combines inpainting with depth diffusion for text-driven scenes. Recent Gaussian-splatting-based methods scale scene generation and interactivity: LucidDreamer\cite{chung2023luciddreamer} and Text2Immersion\cite{ouyang2023text} generate 3D Gaussian scenes from text, BloomScene\cite{hou2025bloomscene} proposes a lightweight structured Gaussian splatting formulation for cross-modal scene generation, and WonderWorld\cite{Yu_2025_CVPR} generates interactive scenes from a single image. Efficiency-focused systems such as Bolt3D\cite{szymanowicz2025bolt3d} and WonderTurbo\cite{Ni_2025_ICCV} emphasize fast scene generation, while SynCity\cite{Engstler_2025_ICCV} explores training-free 3D world generation.

\subsection{Feed-forward reconstruction}
\label{sec:related:feedforward}
Feed-forward reconstruction methods aim to predict 3D structure directly from one or a few images, often with learned priors that replace per-instance optimization. Diffusion-guided single-image pipelines such as Make-It-3D\cite{Tang_2023_ICCV} and Magic123\cite{Magic123} still optimize per instance, but they show how 2D and 3D diffusion priors can stabilize geometry and texture reconstruction from a single view. Make-It-3D\cite{Tang_2023_ICCV} uses a two-stage optimization pipeline (NeRF followed by textured point clouds) guided by a 2D diffusion prior, while Magic123\cite{Magic123} adopts a coarse-to-fine strategy with joint 2D and 3D diffusion guidance and a trade-off parameter between the priors. ATT3D\cite{Lorraine_2023_ICCV} shifts toward amortized inference by training a text-conditioned model to produce 3D objects without per-prompt optimization, trading per-instance fitting for a learned feed-forward generator.  One-2-3-45\cite{liu2023one2345} generates multi-view images with Zero123 and lifts them to a 360-degree mesh using an SDF-based generalizable surface reconstructor in a single feed-forward pass. LRM\cite{hong2023lrm} predicts a triplane representation from a single image using a large transformer-based reconstruction model trained on large-scale multi-view data. GS-LRM\cite{gslrm2024} extends this direction to 3D Gaussian splatting by predicting per-pixel Gaussians from 2--4 posed images with a simple transformer architecture. Splatter Image\cite{Szymanowicz_2024_CVPR} directly regresses one Gaussian per pixel from a single view for real-time reconstruction and can be extended to multi-view inputs via cross-view attention. ReconFusion\cite{Wu_2024_CVPR} leverages a diffusion prior for novel-view guidance to regularize few-view NeRF reconstruction, improving geometry and texture in underconstrained regions. Relatedly, diffusion-based monocular depth estimation repurposes pretrained image diffusion models (e.g., Marigold\cite{Ke_2024_CVPR} derived from Stable Diffusion) to infer affine-invariant depth with improved generalization, providing an additional feed-forward cue that can support reconstruction pipelines.

\subsection{Video diffusion for novel view synthesis and 3D generation}
\label{sec:related:video-diffusion-nvs}
Video diffusion models provide strong spatiotemporal priors that can be repurposed for multi-view generation. Recent work adapts video diffusion to synthesize camera trajectories for downstream 3D generation.

For single-image or sparse-view novel view synthesis, ViVid-1-to-3\cite{kwak2023vivid1to3novelviewsynthesis} reformulates the target view as part of a scanning video and combines view-conditioned and video diffusion denoising along a smooth camera path. MultiDiff\cite{Muller_2024_CVPR} incorporates monocular depth and video diffusion priors, jointly generating a sequence of views (rather than autoregressive frame-by-frame synthesis) and introducing structured noise to improve geometric stability and long-range consistency. ViewCrafter\cite{yu2024viewcraftertamingvideodiffusion} builds a coarse point cloud with dense stereo, uses a point-conditioned video diffusion model as a renderer, and iteratively expands the point cloud to enable long-range view synthesis and subsequent 3D Gaussian splatting optimization. ViewExtrapolator\cite{liu2024novelviewextrapolationvideo} leverages Stable Video Diffusion as a prior to refine artifact-prone renderings from radiance fields or point clouds without fine-tuning, enabling larger-baseline view extrapolation. SplatDiff\cite{zhang2025highfidelitynovelviewsynthesis} integrates pixel-splatting guidance with video diffusion, using an aligned synthesis strategy and a texture-bridge module to reduce texture hallucination while preserving geometry. Two generalist strategies include SEVA\cite{zhou2025stablevirtualcameragenerative}, which synthesizes consistent novel views from arbitrary numbers of input views and target cameras, and NVS-Solver\cite{you2025nvs}, which performs zero-shot synthesis by adaptively modulating the diffusion score with warped input views across single-view, multi-view, and monocular-video inputs.

Several methods use video diffusion outputs to drive explicit 3D reconstruction or 3D generation. SV3D\cite{voleti2024sv3d} adapts image-to-video diffusion with explicit camera control to generate multi-view orbit videos and then uses those views for downstream 3D optimization. V3D\cite{chen2024v3dvideodiffusionmodels} fine-tunes a video diffusion model with a geometric consistency prior to produce 360-degree orbit frames and reconstruct meshes or 3D Gaussian scenes, and extends to scene-level novel view synthesis with camera-path control. Scene123\cite{yang2024scene123prompt3dscene} combines video generation with a consistency-enhanced MAE that warps and inpaints adjacent views, then optimizes a NeRF for geometry consistency and uses a GAN-based loss to improve detail fidelity. ReconX\cite{liu2025reconxreconstructscenesparse} uses a global point cloud as 3D structural guidance injected into the diffusion process and reconstructs scenes with a confidence-aware 3D Gaussian splatting optimization from the synthesized video. Look Outside the Room\cite{Ren_2022_CVPR} addresses long-term camera motions by autoregressively predicting frames with a transformer and a camera-aware locality constraint for consistent long-range videos from a single image.

Video diffusion models also provide the advantage of generating dynamic contents beyond static 3D assets. SV4D\cite{xie2025sv4ddynamic3dcontent} generates temporally consistent novel-view videos for each frame of a monocular reference video and optimizes a dynamic NeRF for 4D content generation. Generative Camera Dolly (GCD)\cite{vanhoorick2024gcd} fine-tunes Stable Video Diffusion with relative camera pose controls to synthesize extreme monocular dynamic novel views along specified trajectories. StreetCrafter\cite{yan2024streetcrafter} focuses on street scenes by conditioning video diffusion on LiDAR-rendered pixel-level cues, enabling controllable view synthesis and distillation into a dynamic 3D Gaussian representation for real-time rendering.

\section{3D Reconstruction}
\label{sec:related:3d-reconstruction}

\subsection{Traditional Structure-from-Motion}
\label{sec:related:traditional-sfm}

Traditional structure-from-motion (SfM) estimates camera poses and a sparse 3D point cloud from multi-view feature correspondences, with bundle adjustment\cite{triggs2000bundle} as the standard nonlinear refinement step. Early large-scale systems demonstrated that Internet photo collections can be used for large scale 3D reconstruction. Photo Tourism\cite{Snavely_2006} reconstructs tourist sites from Internet photos, while Modeling the World\cite{Snavely_2007} extends the same paradigm to larger and more diverse collections. Skeletal Graphs\cite{Snavely_2008} introduce a compact connectivity structure to improve efficiency in large collections. Building Rome in a Day\cite{Agarwal_2009} demonstrates city-scale reconstruction from online photos using a highly parallel pipeline, and the subsequent CACM article\cite{Agarwal_2011} summarizes the system and its practical lessons. Discrete-continuous optimization\cite{Crandall_2011} addresses robustness and scalability in SfM formulations, and Reconstructing the World in Six Days\cite{Heinly_2015_CVPR} shows streaming reconstruction at Yahoo 100M scale.

Incremental pipelines remain common for unordered image sets. VisualSFM\cite{Wu_2013} targets linear-time incremental SfM, while COLMAP\cite{schoenberger2016sfm} revisits the incremental pipeline with a robust and complete system that is widely used as a reference implementation nowadays.

Global SfM instead optimizes all camera poses together for high parallelism and efficiency. \cite{Moulon_2013} aggregates relative pose estimates into a consistent global solution, and similarity averaging\cite{Cui_2015_ICCV} resolves unknown scale by averaging similarity transforms rather than pure translations. Stable SfM\cite{Olsson_2011} focuses on robustness for unordered collections. Open-source libraries such as Theia\cite{theia_website} and openMVG\cite{Moulon_2017} provide complete SfM toolchains that expose both incremental and global components for research and deployment. Recent global pipelines revisit scalability and parallelization, including GLOMAP\cite{pan2024glomap} and InstantSfM\cite{zhong2025instantsfmfullysparseparallel}.

Rotation averaging is a core subproblem in global SfM. Previous research includes robust large-scale methods\cite{Chatterjee_2013_ICCV}, theoretical and optimization treatments for rotation averaging\cite{Hartley_2013}, L1 formulations solved with the Weiszfeld algorithm\cite{Hartley_2011}, and robust relative rotation averaging under outlier contamination\cite{Chatterjee_2018}. The difficulty of rotation averaging under different regimes is analyzed in detail in later work\cite{Wilson_2016}.

Translation averaging comes after rotation averaging, and is usually the most difficult part of global SfM. Early two-view and Lie-algebraic formulations\cite{Govindu_2001,Govindu_2004} establish consistent motion averaging principles. 1DSfM\cite{wilson_eccv2014_1dsfm} reduces the problem to a sequence of 1D subproblems for outlier removal and global position recovery. ShapeFit and ShapeKick\cite{Goldstein_2016} provide robust and scalable formulations for global translation estimation.

Bundle adjustment is the gold standard of camera pose refinement and is a major computational bottleneck. Scaling strategies include large-scale problem formulations\cite{Agarwal_2010}, out-of-core optimization\cite{Ni_2007}, multicore parallelization\cite{Wu_2011}, and software packages that implement generic sparse bundle adjustment\cite{Lourakis_2009}.

\subsection{Learnable Components in Traditional SfM Pipelines}
Classical SfM pipelines decompose reconstruction into multiple stages, including feature extraction, feature matching, geometric verification, and multi-view optimization, etc. Many of these steps now have learned counterparts that can be used as drop-in replacements.

For learned local features, detectors and descriptors such as LIFT\cite{yi2016liftlearnedinvariantfeature}, SuperPoint\cite{DeTone_2018}, D2-Net\cite{dusmanu2019d2nettrainablecnnjoint}, R2D2\cite{revaud2019r2d2repeatablereliabledetector}, DISK\cite{tyszkiewicz2020disklearninglocalfeatures}, ASLFeat\cite{luo2020aslfeatlearninglocalfeatures}, Key.Net\cite{barrosolaguna2019keynetkeypointdetectionhandcrafted}, and HardNet\cite{HardNet} have been adopted in SfM pipelines as alternatives to handcrafted features. Learned two-view matching has similarly advanced with SuperGlue\cite{sarlin2020supergluelearningfeaturematching}, LightGlue\cite{Lindenberger_2023_ICCV}, LoFTR\cite{sun2021loftrdetectorfreelocalfeature}, ASpanFormer\cite{chen2022aspanformerdetectorfreeimagematching}, NCNet\cite{rocco2018neighbourhoodconsensusnetworks} and Sparse-NCNet\cite{rocco2020efficientneighbourhoodconsensusnetworks}, S2DNet\cite{germain2020s2dnetlearningaccuratecorrespondences}, Patch2Pix\cite{zhou2021patch2pixepipolarguidedpixellevelcorrespondences}, and COTR\cite{jiang2021cotrcorrespondencetransformermatching}, which either refine sparse matches or operate in a detector-free manner.

For pair selection, learned global or hybrid descriptors are commonly used for image retrieval and place recognition, including NetVLAD\cite{Arandjelovic_2016_CVPR}, DELF\cite{Noh_2017_ICCV}, DELG\cite{Cao_2020}, GeM-based retrieval\cite{radenović2018finetuningcnnimageretrieval}, Patch-NetVLAD\cite{hausler2021patchnetvladmultiscalefusionlocallyglobal}, CosPlace\cite{Berton_2022_CVPR}, MixVPR\cite{Ali-bey_2023_WACV}, and HF-Net\cite{sarlin2019coarsefinerobusthierarchical}.

After initial correspondences are formed, learned modules can improve geometric verification and two-view estimation. OANet\cite{zhang2019learningtwoviewcorrespondencesgeometry} predicts inliers while jointly estimating geometry, and correspondence pruning can be strengthened by consensus learning\cite{Zhao_2021_ICCV} and neighbor consistency mining\cite{Liu_2023_CVPR}. Robust estimation has also been augmented by learning-guided hypothesis sampling (NG-RANSAC)\cite{Brachmann_2019} and differentiable robust selection (DSAC)\cite{brachmann2018dsacdifferentiableransac}. For direct two-view geometry estimation, Deep Fundamental Matrix Estimation\cite{Ranftl_2018} and its correspondence-free variant\cite{poursaeed2018deepfundamentalmatrixestimation} learn to predict or fit the fundamental matrix from image pairs.

Several works extend the feature toolkit beyond points. DeepLSD\cite{pautrat2023deeplsdlinesegmentdetection}, HAWP\cite{xue2020holisticallyattractedwireframeparsing}, and SOLD$^2$\cite{pautrat2021sold2selfsupervisedocclusionawareline} provide learned line or wireframe features, and GlueStick\cite{pautrat2023gluestickrobustimagematching} integrates point and line cues for robust matching. Multi-view matching and track construction can also be learned, for example with CoMatcher\cite{Zhang_2025_CVPR} and multi-view optimization of keypoint geometry\cite{Dusmanu_2020}. Learned optimization modules such as BA-Net\cite{tang2019banetdensebundleadjustment} act as bundle-adjustment-style refiners within otherwise classical pipelines.

Learned MVS modules such as MVSNet\cite{yao2018mvsnetdepthinferenceunstructured}, R-MVSNet\cite{yao2019recurrentmvsnethighresolutionmultiview}, CascadeMVSNet\cite{gu2020cascadecostvolumehighresolution}, and PatchmatchNet\cite{wang2020patchmatchnetlearnedmultiviewpatchmatch} provide dense depth and surface estimates after camera poses are estimated. Camera intrinsics can be estimated or regularized with DeepCalib\cite{bogdan2018deepcalib}, DeepFocal\cite{09525ae48a2041789462b93743c9e0d3}, Perspective Fields\cite{Jin_2023_CVPR}, and AnyCalib\cite{tiradogarín2025anycalibonmanifoldlearningmodelagnostic}. Finally, dynamic object masking is often introduced as a practical pre-processing step using DynaSLAM\cite{bescos2018dynaslamtrackingmappinginpainting} or generic segmentation models such as Mask R-CNN\cite{he2018maskrcnn} and SegFormer\cite{xie2021segformersimpleefficientdesign}.

\subsection{Fully learning-based SfM}
\label{sec:related:learning-sfm}
Recent fully learning-based SfM seeks to replace the entire classical pipeline with a single learned model.

Early works explore directly outputting camera poses from sparse-view images. RelPose~\cite{zhang2022relpose} formulates pairwise viewpoint estimation by predicting a distribution over relative rotations and then enforcing global rotation consistency across an image set. Building on this, RelPose++~\cite{lin2024relposepp} incorporates multi-view reasoning so additional observations help resolve ambiguities, and extends the framework toward full 6D pose recovery from sparse views. SparsePose~\cite{sinha2023sparsepose} targets extremely sparse and wide-baseline settings by regressing initial poses and iteratively refining them with learned features, improving stability when view overlap is limited. PoseDiffusion~\cite{Wang_2023_ICCV} casts pose estimation as a diffusion process analogous to iterative bundle adjustment and supports integrating geometric constraints during denoising. Finally, Cameras as Rays~\cite{zhang2024raydiffusion} represents cameras as collections of rays tied to images and uses diffusion models to improve sparse-view pose accuracy.

A second line focuses on feed-forward pointmap prediction, pioneered by DUSt3R\cite{wang2024dust3r}. MASt3R\cite{mast3r_eccv24} learns pointmap-based geometry-aware correspondences, which is later used in MASt3R-SfM\cite{duisterhof2025mastrsfm} for a full SfM pipeline. Speedy MASt3R\cite{11125757} targets faster matching building on top of MASt3R. Geometry-grounded transformers such as VGGSfM\cite{wang2024vggsfm} and VGGT\cite{Wang_2025_CVPR} predict correspondences, geometry, and poses in a unified architecture. DiffusionSfM\cite{zhao2025diffusionsfm} uses diffusion generative models instead of regressing a deterministic network. Other variants aim at stream processing (FlashVGGT\cite{wang2025flashvggt}, Streaming 4D VGGT\cite{streamVGGT}, InfiniteVGGT\cite{yuan2026infinitevggt}, CUT3R\cite{cut3r}), speed (Light3R-SfM\cite{zhu2025light3rsfm}, Fast3R\cite{yang2025fast3r}, FlashVGGT\cite{wang2025flashvggt}) or additional priors (G-CUT3R\cite{khafizov2025gcut3r}, Fin3R\cite{ren2025fin3r}).

A third line of work formulates SfM as learning scene coordinates or relocalizers over image collections. Scene Coordinate Reconstruction (ACE0)\cite{brachmann2024scene} learns a relocalizer incrementally to recover poses for large sets of images, building on the fast scene-coordinate encoder ACE\cite{brachmann2023ace}. R-SCoRe\cite{Jiang_2025_CVPR} revisits scene-coordinate regression for large-scale localization, and Reloc3r\cite{Dong_2025_CVPR} scales relative pose regression to improve generalization and speed.

\section{Synergy of Generation and Reconstruction}
\label{sec:related:synergy}
\noindent\textbf{Differentiable 3D representations as a shared substrate.} A key bridge between generation and reconstruction is the rise of differentiable 3D representations. NeRF\cite{mildenhall2020nerf} introduced a neural radiance field with differentiable volumetric rendering originally for reconstruction and novel-view synthesis, but it later proved to be important to generative pipelines as well. For instance, text-to-3D optimization methods such as DreamFusion\cite{poole2022dreamfusion} and Magic3D\cite{Lin_2023_CVPR} directly optimize NeRF-like representations under diffusion guidance, while 3D-aware image generators such as GRAF\cite{schwarz2020graf} and pi-GAN\cite{Chan_2021_CVPR} learn radiance-field-based generators for view-consistent synthesis. This shared representation makes it natural for ideas to transfer across tasks: improvements in rendering stability or representation efficiency in reconstruction can be reused in generation, and generative priors can regularize underconstrained reconstructions.

\noindent\textbf{Few-view reconstruction as a controllable generation strategy.} A popular trend in 3D generation is to generate a small set of consistent views and then reconstruct the 3D assets. Feed-forward pipelines such as One-2-3-45\cite{liu2023one2345}, LRM\cite{hong2023lrm}, and GS-LRM\cite{gslrm2024} use few-view image inputs to predict meshes or Gaussian splats in a single pass, while video-diffusion-based systems such as SV3D\cite{voleti2024sv3d} and V3D\cite{chen2024v3dvideodiffusionmodels} first synthesize multi-view orbit videos and then reconstruct NeRF or Gaussian representations. These examples underline that sparse-view reconstruction is becoming an important part for high-quality 3D generation pipelines because it can help fully utilize powerful text-to-image or text-to-video diffusion models.

\noindent\textbf{Large-scale learning priors strengthening both sides.} Large text-to-image diffusion models provide strong semantic priors that can be distilled into 3D generation, as exemplified by SDS-based methods and their extensions\cite{poole2022dreamfusion,swang2023score,podell2023sdxl,saharia2022photorealistic}. In parallel, transformer-based SfM and feed-forward reconstruction models (e.g., VGGSfM\cite{wang2024vggsfm}, VGGT\cite{Wang_2025_CVPR}, and DUSt3R\cite{wang2024dust3r}) show that large learned priors can mitigate classical SfM failure modes such as sparse viewpoints or texture-poor regions. Together, these developments indicate that scale and pretraining are now central to both generation and reconstruction.

\noindent\textbf{Classical SfM as a data engine for modern pipelines.} Traditional SfM systems remain essential for building large 3D datasets from raw RGB image collections. COLMAP\cite{schoenberger2016sfm} is widely used to construct large-scale reconstructions that serve as training data and benchmarks for modern learning-based systems, including VGGT\cite{Wang_2025_CVPR} as well as VGGSfM\cite{wang2024vggsfm}, DUSt3R\cite{wang2024dust3r}, and MASt3R-SfM\cite{duisterhof2025mastrsfm}. These reconstructions supply the supervision that underpins both learning-based reconstruction models and generation systems relying on large-scale 3D data.

\noindent\textbf{Toward unified multimodal models.} A related trend is the emergence of large multimodal language models that interleave multiple modalities in both input and output within a single model \cite{yu2023cm3leon,wu2023nextgpt,kondratyuk2023videopoet,ge2024seedx,xu2025qwen25omni}, suggesting that generation and reconstruction may eventually be unified rather than implemented as separate modules.

\chapter{3D Generation}
\label{chap:3d-generation}

\section{Instant3D}
\label{sec:instant3d}
\begingroup
\newcommand{\methodname}{Instant3D\xspace}
\newcommand{\imgset}{\mathcal{I}}
\newcommand{\img}{\textbf{I}}
\newcommand{\feat}{\boldsymbol{f}}
\label{sec:instant3d-intro}
Progress in 2D image generation has been driven by diffusion models~\citep{song2021denoising,ho2020denoising,ramesh2022hierarchical,rombach2021highresolution} and large-scale datasets such as Laion5B~\citep{schuhmann2022laion}. Extending these gains to 3D is constrained by data scarcity: while Laion5B has 5 billion text-image pairs, Objaverse-XL~\citep{deitke2023objaversexl}, the largest public 3D dataset, contains only 10 million 3D assets with lower diversity and weaker annotations. Directly training 3D diffusion models on existing 3D datasets~\citep{luo2021diffusion,nichol2022point,jun2023shap,gupta20233dgen,chen2023single} therefore tends to yield limited shape and appearance quality, diversity, and compositional complexity.

An alternative line of work~\citep{poole2022dreamfusion,swang2023score,lin2023magic3d,wang2023prolificdreamer,chen2023fantasia3d} uses pretrained 2D diffusion models to guide 3D optimization. These methods compute gradients on rendered images and optimize a 3D representation, typically a NeRF~\citep{mildenhall2020nerf}. They improve visual quality and text-3D alignment, but the per-prompt optimization is slow (often hours) and prone to artifacts such as over-saturated colors and the ``multi-face" problem. Diversity is also limited: varying the random seed often produces only minor geometric and texture changes.

This section introduces \methodname, a feed-forward method that generates high-quality and diverse 3D assets conditioned on text. 
\methodname builds on pretrained 2D diffusion models but splits 3D generation into two stages: 2D generation and 3D reconstruction. In the first stage, instead of generating images sequentially~\citep{liu2023zero1to3},
we fine-tune an existing text-to-image diffusion model~\citep{podell2023sdxl} to generate a sparse set of four-view images in the form of a $2\!\times\!2$ grid in a single denoising process. 
This design allows the multi-view images to attend to each other during generation, leading to more view-consistent results. In the second stage, instead of relying on a slow optimization-based reconstruction method, 
inspired by~\cite{hong2023lrm}, we introduce a novel sparse-view \emph{large reconstruction model} 
with a transformer-based architecture that can directly regress a triplane-based~\citep{chan2022eg3d} NeRF from a 
sparse set of multi-view images. 
The model projects sparse-view images into a set of pose-aware image tokens using pretrained 
vision transformers~\citep{caron2021emerging}, which are then fed to an image-to-triplane decoder 
that contains a sequence of transformer blocks with cross-attention and self-attention layers. 
The reconstruction model has more than $500$ million parameters 
and can robustly infer correct geometry and appearance of objects from just four images.

Both stages are fine-tuned/trained with multi-view rendered images of around $750$K 3D objects from Objaverse~\citep{deitke2023objaverse}, where the second stage uses the full dataset and the first stage can be fine-tuned with as little as $10$K data.
Although the 3D training data are much smaller than the pre-training datasets used in other modalities (e.g., C4~\cite{raffel2020exploring} for text and Laion5B for images), combining them with pretrained 2D diffusion models allows \methodname to generate high-quality and diverse 3D assets even for complex compositional prompts (see Figure~\ref{fig:instant3d-teaser}) that are absent from the 3D training set. 
Due to its feed-forward architecture, \methodname requires about 20 seconds per 3D asset, which is roughly $200\times$ faster than prior optimization-based methods~\citep{poole2022dreamfusion,wang2023prolificdreamer} while achieving comparable or better quality.

\begin{figure}[t]
    \centering
    \includegraphics[width=1\textwidth]{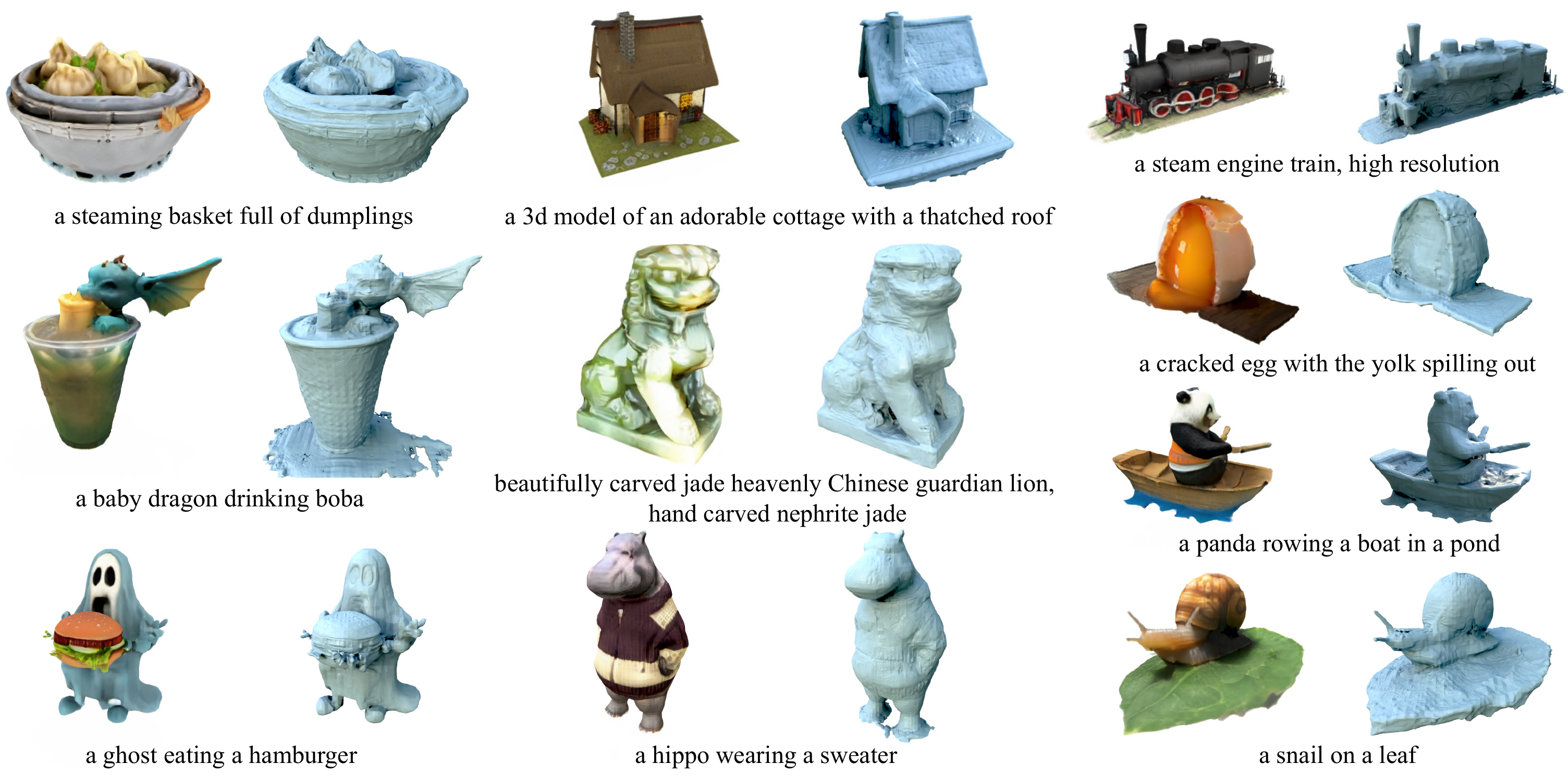} 
    \caption{
    Our method generates high-quality 3D NeRF assets from the given text prompts within 20 seconds. Here we show 
    novel view renderings from our generated NeRFs as well as the renderings of the extracted meshes from their density field. 
    }
    \label{fig:instant3d-teaser}

\end{figure}

\begin{figure}
    \centering
    \includegraphics[width=1\textwidth]{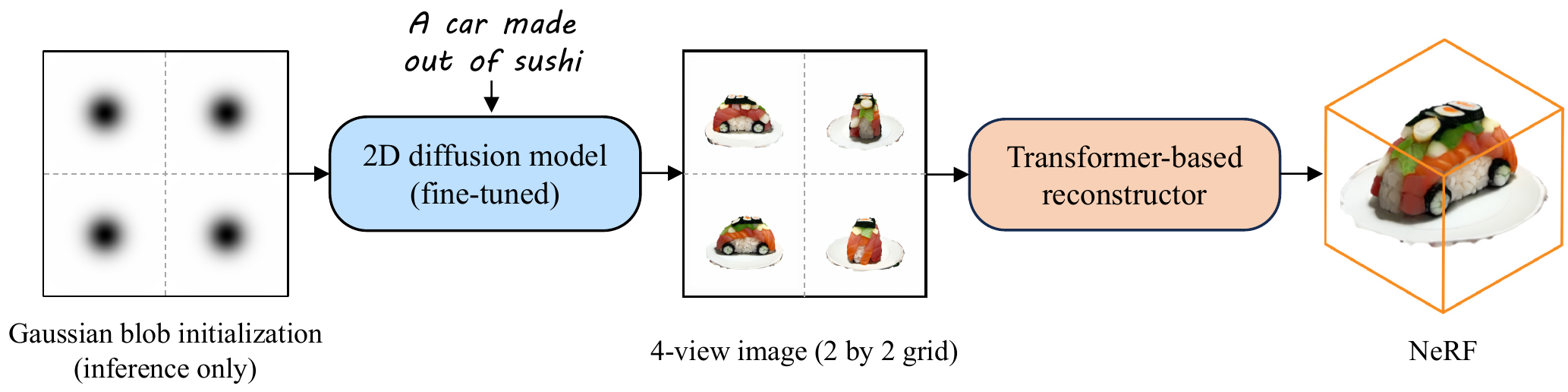}
    \caption{Overview of our method. Given a text prompt (`a car made out of sushi'), we perform multi-view generation 
    with Gaussian blobs as initialization using fine-tuned 2D diffusion model, 
    producing a 4-view image in the form of a $2\times 2$ grid. Then we apply a transformer-based 
    sparse-view 3D reconstructor on the 4-view image to generate 
    the final NeRF. 
    }
    \label{fig:instant3d-framework}

\end{figure}

\subsection{Method} \label{sec:instant3d-method}

Our method \methodname is composed of two stages: sparse-view generation and feed-forward NeRF reconstruction. In Section~\ref{sec:instant3d-sparse_view_gen}, we present our approach for generating sparse multi-view images conditioned on the text input. 
In Section~\ref{sec:instant3d-spase_view_recon}, we describe our transformer-based sparse-view large reconstruction model.

\subsubsection{Text-Conditioned Sparse View Generation}
\label{sec:instant3d-sparse_view_gen}
Given a text prompt, our goal is to generate a set of multi-view images that are aligned with the prompt and consistent with each other. 
We achieve this by fine-tuning a pretrained text-to-image diffusion model to generate a 
$2\times 2$ image grid as shown in Figure~\ref{fig:instant3d-framework}.

In the following paragraphs, we first illustrate that large text-to-image diffusion models (i.e., SDXL~\citep{podell2023sdxl}) have the capacity to generate view-consistent images thus a lightweight fine-tuning is possible.
We then introduce three essential techniques to achieve it: the image grid, the curation of the dataset, and also the Gaussian Blob noise initialization in inference.
As a result of these observations and technical improvements, we can fine-tune the 2D diffusion model for only 10K steps (on 10K data) to generate consistent sparse views.

\textbf{Multi-view generation with image grid.}
Previous methods~\citep{liu2023zero1to3, liu2023one2345} on novel-view synthesis show that image diffusion models are capable of understanding the multi-view consistency.
In light of this, we compile the images at different views into a single image in the form of an image grid, as depicted in Figure~\ref{fig:instant3d-framework}. 
This image-grid design can better match the original data format of the 2D diffusion model, and is suitable for simple direct fine-tuning protocol of 2D models.
We also observe that this simple protocol only works when the base 2D diffusion has enough capacity, as shown in the comparisons of Stable Diffusion v1.5~\citep{rombach2021highresolution} and SDXL~\citep{podell2023sdxl} in Section~\ref{sec:instant3d-ablation_study}.
The benefit from simplicity will also be illustrated later in unlocking the lightweight fine-tuning possibility.

Regarding the number of views in the image grid, there is a trade-off between the requirements of multi-view generation and 3D reconstruction.
More generated views make the problem of 3D reconstruction easier with more overlaps but increase possibility of view inconsistencies in generation and reduces the resolution of 
each generated view.
On the other hand, too few views  may cause insufficient coverage, requiring the reconstructor to hallucinate unseen parts, which is challenging for a deterministic 3D reconstruction model.
Our transformer-based reconstructor learns generic 3D priors from large-scale data,  
and greatly reduces the requirement for the number of views. We empirically found that 
using 4 views achieves a good balance in satisfying the two requirements above,
and they can be naturally arranged in a $2\times 2$ grid as shown in Figure~\ref{fig:instant3d-framework}. Next, we detail how the image grid data is created and curated.

\paragraph{Multi-view data creation and curation.} 
To fine-tune the text-to-image diffusion model, we create paired multi-view renderings and text prompts.
We adopt a large-scale synthetic 3D dataset Objaverse~\citep{deitke2023objaverse} and render four $512\times 512$ views of about $750$K objects with Blender. 
We distribute the four views at a fixed elevation (20 degrees) and four equidistant azimuths (0, 90, 180, 270 degrees) to achieve a better coverage of the object. 
We use Cap3D~\citep{luo2023scalable} to generate captions for each 3D object, which 
consolidates captions from multi-view renderings generated with pretrained image captioning
model BLIP-2~\citep{li2023blip} using a large language model (LLM).
Finally, the four views are assembled into a grid image in a fixed order and resized to the input resolution compatible with the 2D diffusion model.

We find that naively using all the data for fine-tuning reduces the 
photo-realism of the generated images and thus the quality of the 3D assets. 
Therefore, we train a simple scorer on a small amount (2000 samples) of manually labeled data to predict the quality of each 3D object. The model is a simple SVM on top of pretrained CLIP features extracted from multi-view renderings of the 3D object. 
During training, our model only takes the top 10K data ranked by our scorer.
We provide a quantitative study in Section~\ref{sec:instant3d-ablation_study} 
to validate the impact of different data curation strategies.
Although the difference is not very significant 
from the metric perspective, we found that our curated data is helpful in 
improving the visual quality. 
\paragraph{Inference with Gaussian blob initialization.} 
While our training data is multi-view images with a white background, we observe that 
during inference starting from standard Gaussian noise still results in 
images that have cluttered backgrounds (see Figure~\ref{fig:instant3d-ablate_gaussian}); 
this introduces extra difficulty for the  
feed-forward reconstructor in the second stage (Section~\ref{sec:instant3d-spase_view_recon}). 
To guide the model toward generating images with a clean white background, 
inspired by SDEdit~\citep{meng2022sdedit}, we first create an image of a $2\times 2$ grid 
with a solid white background that has the same resolution as the output image, 
and initialize each sub-grid with a 2D \emph{Gaussian blob} that is placed 
at the center of the image with a standard deviation of $0.1$. 
The visualization of this Gaussian Blob is shown in Figure~\ref{fig:instant3d-framework}.
The Gaussian blob image grid is fed to the auto-encoder to get its latent. 
We then add diffusion noise (e.g., use t=980/1000 for 50 DDIM denoising steps), and use 
it as the starting point for the denoising process. 
As seen in Figure~\ref{fig:instant3d-ablate_gaussian}, this technique 
effectively guides the model toward generating images with a clean background.

\paragraph{Lightweight fine-tuning.}
With all the above observations and techniques, we are able to adapt a text-to-image diffusion model to a text-to-multiview model with lightweight fine-tuning.
This lightweight fine-tuning shares a similar spirit to the `instruction fine-tuning'~\citep{mishra2022cross, wei2021finetuned} for LLM alignment. 
The assumption is that the base model is already capable of the task, and the fine-tuning 
is to unlock the base model's ability without introducing additional knowledge.

Since we utilize an image grid, the fine-tuning follows the exactly same protocol as the 2D diffusion model pre-training, except that we decrease the learning rate to $10^{-5}$.
We train the model with a batch size of 192 for only 10K iterations 
on the 10K curated multi-view data.
The training is done using 32 NVIDIA A100 GPUs for only 3 hours.
We study the impact of different training settings in Section~\ref{sec:instant3d-ablation_study}.

\begin{figure}
    \centering
    \includegraphics[width=1.\textwidth]{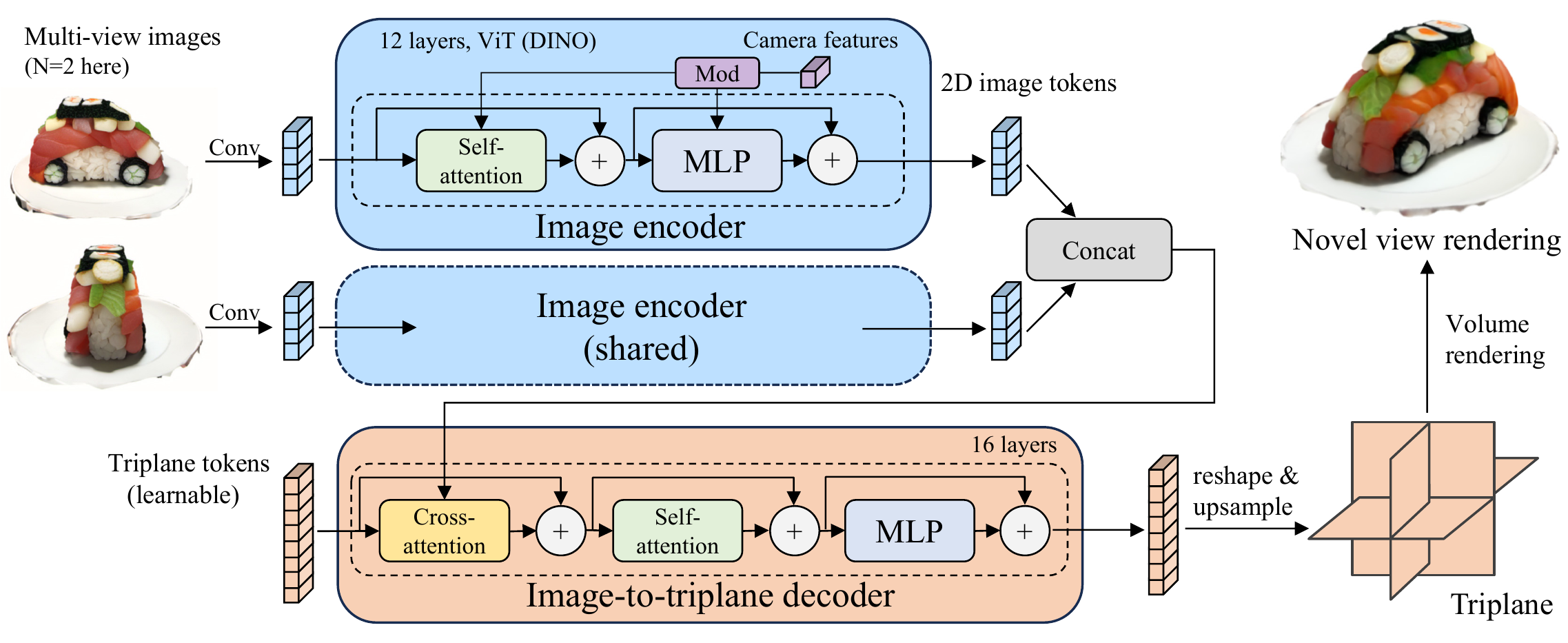}
    \caption{Architecture of our sparse-view reconstructor. The model applies a pretrained ViT 
    to encode multi-view images into pose-aware image tokens, from which we decode a triplane representation 
    of the scene using a transformer-based decoder. Finally we decode per-point triplane features 
    to its density and color and perform volume rendering to render novel views. 
    We illustrate here with 2 views and the actual implementation uses 4 views.
    }
    \label{fig:instant3d-second_stage}

\end{figure}

\subsubsection{Feed-Forward Sparse-View Large Reconstruction Model}\label{sec:instant3d-spase_view_recon}
In this stage, we aim to reconstruct a NeRF from the four-view images $\imgset = \{\img_i~|~i=1,...,4\}$ generated in the first stage.
3D reconstruction from sparse inputs with a large baseline is a challenging problem,
which requires strong model priors to resolve the inherent ambiguity. 
Inspired by a recent work LRM~\citep{hong2023lrm} that introduces a transformer-based model for single image 3D 
reconstruction, we propose a novel approach that enables us to predict a NeRF from a sparse set of 
input views with known poses.
Similar to~\cite{hong2023lrm}, our model consists of an image encoder, 
an image-to-triplane decoder, and a NeRF decoder. The image encoder encodes 
the multi-view images into a set of tokens. We feed the concatenated image tokens to the 
image-to-triplane decoder to output a triplane representation~\citep{chan2022eg3d} for the 3D object. Finally,
the triplane features are decoded into per-point density and colors via the NeRF MLP decoder.

In detail, we apply a pretrained Vision Transformer (ViT) DINO~\citep{caron2021emerging} as our image encoder. 
To support multi-view inputs, we inject camera information in the image encoder to make the output image tokens pose-aware.
This is different from \cite{hong2023lrm} that feeds the camera information in the image-to-triplane decoder 
because they take single image input.
The camera information injection is done by the AdaLN~\citep{huang2017arbitrary, peebles2022dit} camera 
modulation as described in \cite{hong2023lrm}.
The final output of the image encoder is a set of pose-aware image tokens $\feat_{\img_i}^*$,
and we concatenate the per-view tokens together as the feature descriptors for the 
multi-view images: $\feat_{\mathcal{I}} = \oplus(\feat_{\img_1}^*, ... \feat_{\img_4}^*)$

We use triplane as the scene representation.  
The triplane is flattened to a sequence of learnable tokens, and the image-to-triplane 
decoder connects these triplane tokens with the pose-aware image tokens $\feat_{\mathcal{I}}$ 
using cross-attention layers, followed by self-attention and MLP layers. 
The final output  tokens are reshaped and upsampled using a de-convolution layer to the final triplane representation.
During training, we ray march through the object bounding box and decode the triplane features  at each point to its density and color using a shared MLP, and finally get the  pixel color via volume rendering. 
We train the networks in an end-to-end manner with image reconstruction loss at novel views using a combination of MSE loss and LPIPS~\citep{zhang2018perceptual} loss.

\paragraph{Training details.} We train the model on multi-view renderings
of the Objaverse dataset~\citep{deitke2023objaverse}. 
Different from the first stage that performs data curation,
we use all the 3D objects in the dataset and scale them to $[-1, 1]^3$; then we generate 
multi-view renderings using  Blender under uniform lighting with a 
resolution of $512 \times 512$. 
While the output images from the first stage are generated in a structured setup with fixed camera poses, 
we train the model using random views as a data augmentation mechanism to increase the robustness. 
Particularly, we randomly sample $32$ views around each object. 
During training, we randomly select a subset of $4$ images as input and another random set of $4$ images as supervision.
For inference, we will reuse the fixed camera poses in the first stage as the camera input to the reconstructor.

\subsection{Experiments}\label{sec:instant3d-exps}

\begin{figure}
\includegraphics[width=\textwidth]{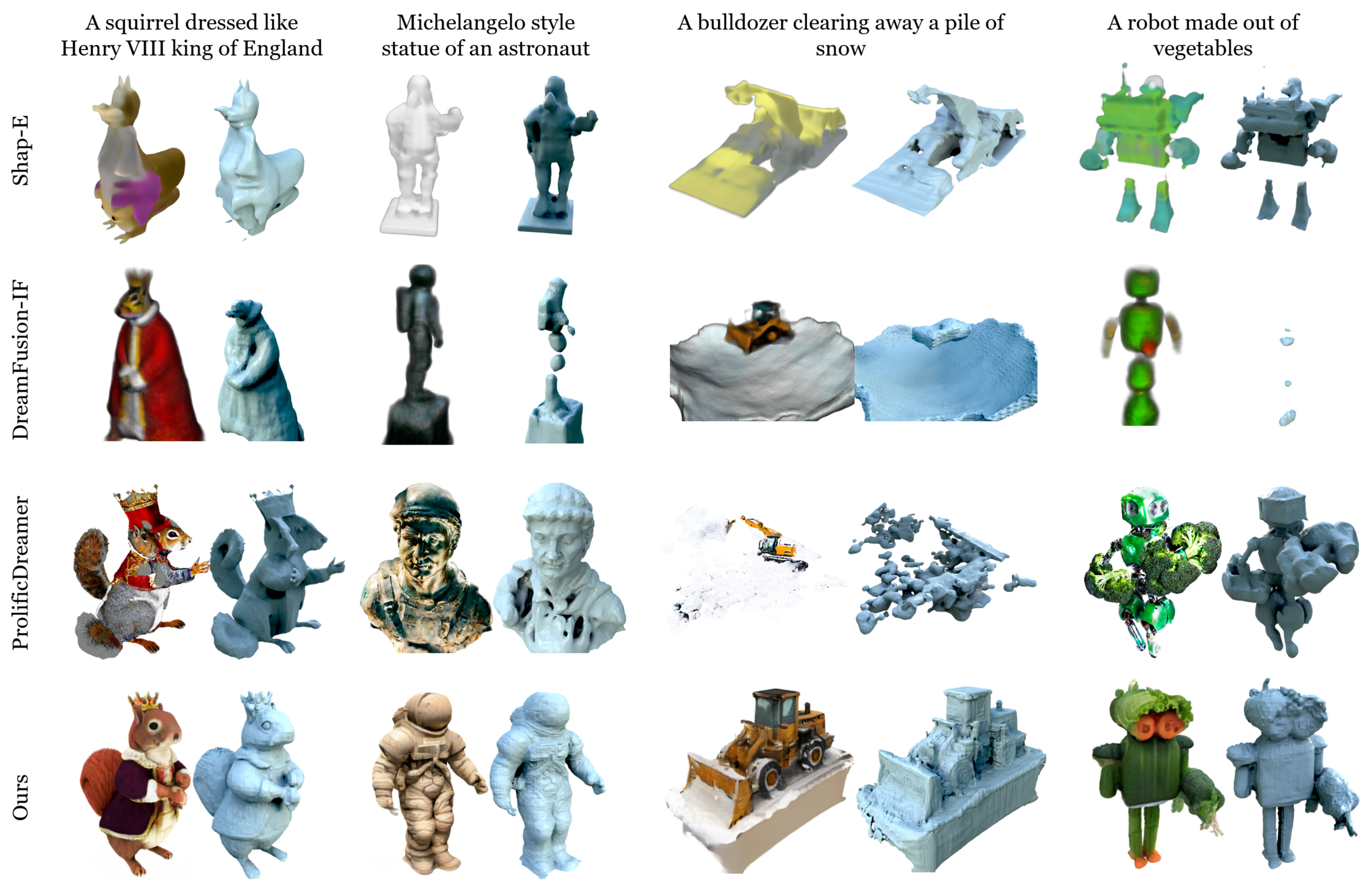}

\caption{
    Qualitative comparisons on text-to-3D against previous methods.
}
\label{fig:instant3d-comp_text_to_3d}

\end{figure}

In this section, we first do comparisons against previous methods on 
text-to-3D (Section~\ref{sec:instant3d-exp_tex_to_3d}), and then perform ablation studies
on different design choices of our method. 
By default, we report the results generated with fine-tuned SDXL models, unless otherwise noted.

\subsubsection{Text-to-3D}\label{sec:instant3d-exp_tex_to_3d}
We make comparisons to state-of-the-art methods on text-to-3D, including 
a feed-forward method Shap-E~\citep{jun2023shap}, and optimization-based 
methods including DreamFusion~\citep{poole2022dreamfusion} and 
ProlificDreamer~\citep{wang2023prolificdreamer}. We use the 
official code for Shap-E, and the implementation from 
three-studio~\citep{threestudio2023} for the other two as there 
is no official code. We use default hyper-parameters (number of optimization iterations, number of denoising steps) of these models. 
For our own model we use the SDXL base model fine-tuned on 10K data for 10K steps. During inference we take 100 DDIM steps.

\begin{table}[t]

\begin{minipage}[t]{0.5\linewidth}
\centering
\resizebox{\linewidth}{!}{

\begin{tabular}{ cccc} 
\specialrule{.15em}{.05em}{.05em}

  &  ViT-L/14 $\uparrow$  & ViT-bigG-14  $\uparrow$ & Time(s) $\downarrow$ \\ %
 \hline
 \hline
 Shap-E &  20.51 &  32.21 & 6 \\
 DreamFusion &  23.60  & 37.46  & 5400 \\
 ProlificDreamer & 27.39  & 42.98 & 36000 \\
 Ours  &  26.87  & 41.77 & 20 \\

\specialrule{.15em}{.05em}{.05em}
\end{tabular}
}
\caption{Quantitative comparisons on CLIP scores against baseline methods.}

\label{tab:instant3d-clip_comp_text_to_3d}

\end{minipage}
\hfill
\begin{minipage}[t]{0.4\linewidth}

\centering
\resizebox{\linewidth}{!}{
    \begin{tabular}{cccc}
        \specialrule{.15em}{.05em}{.05em}
            & PSNR $\uparrow$ & SSIM $\uparrow$  & LPIPS $\downarrow$ \\
            \hline
            \hline
           SparseNeus & 20.62 &  0.8360 & 0.1989\\ 
           Ours & 26.54 & 0.8934 & 0.0643 \\ 
        \specialrule{.15em}{.05em}{.05em}
    \end{tabular}
}
\caption{Quantitative comparisons against previous sparse-view reconstruction methods on GSO dataset.}
\label{fig:instant3d-sparseneus_comp}
\end{minipage}

\end{table}

\paragraph{Qualitative comparisons.} 
As shown in Figure~\ref{fig:instant3d-comp_text_to_3d}, our method generates visually better results than those of Shap-E, producing sharper textures, better geometry and substantially improved text-3D alignment.  
Shap-E applies a diffusion model that is exclusively trained on million-level 3D data, which might be evidence for 
the need of 2D data or models with 2D priors.
DreamFusion and ProlificDreamer 
achieve better text-3D alignment utilizing pretrained 
2D diffusion models. 
However, DreamFusion generates
results with over-saturated colors and over-smooth textures.
While ProlificDreamer results have better details, it still suffers from low-quality geometry (as in `A bulldozer clearing ...') and  the Janus problem (as in "a squirrel dressed like ...").
In comparison, our results have more photorealistic appearance with better geometric details.

\paragraph{Quantitative comparisons.} In Table~\ref{fig:instant3d-comp_text_to_3d}, 
we quantitatively assess the coherence between the generated models 
and text prompts using CLIP-based scores. 
We perform the evaluation on results with 400 text prompts 
from DreamFusion. 
For each model, we render 10 random views and calculate the average 
CLIP score between the rendered images and the input text.
We report the metric using multiple variants of CLIP models
with different model sizes and training data (i.e., ViT-L/14 from OpenAI and ViT-bigG-14 from OpenCLIP). 
From the results we can see that our model achieves higher CLIP scores than Shap-E, indicating better text-3D alignment. 
Our method even achieves consistently higher CLIP scores than optimization-based method DreamFusion  and competitive  scores to ProlificDreamer, from which we can see 
that our approach can effectively inherit the great text understanding capability from the pretrained SDXL model and preserve them in the generated 3D assets  via consistent sparse-view generation and robust 3D reconstruction. 

\paragraph{Inference time comparisons.} We present the time to generate a 3D asset in Table~\ref{tab:instant3d-clip_comp_text_to_3d}. 
The timing is measured using the default hyper-parameters of each method on an A100 GPU.  
Notably, our method is significantly faster than the optimization-based methods: while it takes 1.5 hours for DreamFusion and 10 hours for ProlificDreamer to generate a single asset, our method can finish the generation within 20 seconds, resulting in a $270\times$ and $1800 \times$ speed up respectively. 
We show that our inference time can be further reduced  
without obviously sacrificing the quality by decreasing the number of DDIM steps.

\subsubsection{Comparisons on Sparse View Reconstruction}
We make comparisons to previous sparse-view NeRF reconstruction works. Most of 
previous works~\citep{reizenstein2021common,trevithick2021grf,yu2020pixelnerf} are either 
trained on small-scale datasets such as ShapeNet, or trained in a category-specific manner.
Therefore, we make comparisons to a state-of-the-art method  
SparseNeus~\citep{long2022sparseneus}, which is also applied in One-2-3-45~\citep{liu2023one2345} 
where they train the model on the same Objaverse dataset for sparse-view reconstruction.
We do the comparisons on the Google Scan Object (GSO) dataset~\citep{gso}, which consists 
of 1019 objects. For each object, we render 4-view input following the 
structured setup and randomly select another 10 views for testing. 
We adopt the pretrained model from~\cite{liu2023one2345}.  Particularly, 
SparseNeus does not work  well for 4-view inputs with such a large baseline; 
therefore we add another set of $4$ input views in addition to our four input views
(our method still uses 4 views as input),
following the setup in~\cite{liu2023one2345}. We report the metrics 
on novel view renderings in  Table~\ref{fig:instant3d-sparseneus_comp}. From the table, we 
can see that our method outperforms the baseline method even 
with fewer input images, which demonstrates the superiority of our sparse-view 
reconstructor.

\subsubsection{Ablation Study for Sparse View Generation}\label{sec:instant3d-ablation_study}

We ablate several key decisions in our method design, including (1) the choice of the larger 2D base model SDXL, (2) the use of Gaussian Blob during inference, (3) the quality and size of the curated dataset, and lastly, (4) the need and requirements of lightweight fine-tuning. We gather the quantitative results in Table~\ref{tab:instant3d-ablation_data_amount}. We observe that qualitative results are more evident than quantitative results, thus we recommend a closer examination.

\paragraph{Scalability with 2D text-to-image models.}
One of the notable advantages of our method is that  its efficacy scales positively with the potency of the underlying 2D text-to-image model. 
We present qualitative comparisons between two distinct backbones (with their own tuned hyper-parameters): SD1.5~\citep{rombach2021highresolution} and SDXL~\citep{podell2023sdxl}.
It becomes readily apparent that SDXL, which boasts a model size $3\times$ larger than that of SD1.5, exhibits superior text comprehension and visual quality.
We also show a quantitative comparison on CLIP scores in Table~\ref{tab:instant3d-ablation_data_amount}. By comparing Exp(l, m) with Exp(d, g), we can see that the model with SD1.5 achieves consistently lower CLIP scores indicating worse text-3D alignment.

\paragraph{Gaussian blob initialization.}
In Figure~\ref{fig:instant3d-ablate_gaussian}, we show our results generated with and without 
Gaussian blob initialization. 
From the results we can see that while  our fine-tuned model can generate multi-view images without Gaussian blob initialization, they tend to have cluttered backgrounds, which challenges the second-stage feed-forward reconstructor. 
In contrast, our proposed Gaussian blob initialization enables the fine-tuned model to generate images with a clean white background, which better align with the requirements of the second stage.

\begin{figure}
\includegraphics[width=\textwidth]{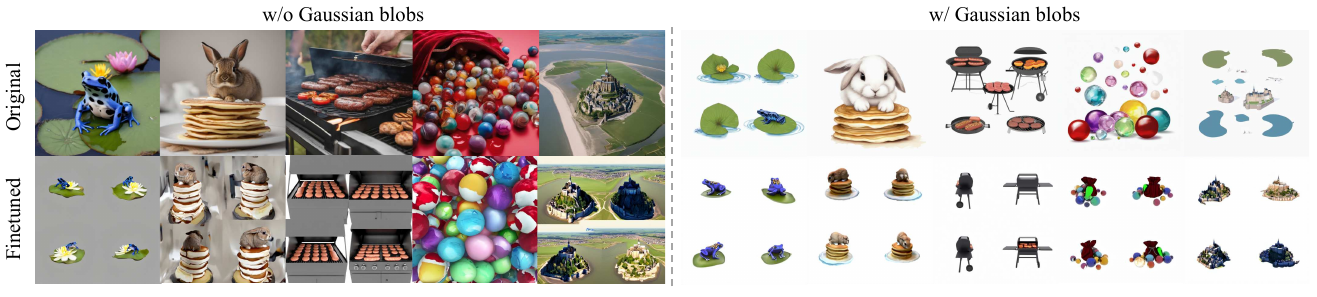}

\caption{
Qualitative comparisons on results generated with and without Gaussian blob initialization.
}
\label{fig:instant3d-ablate_gaussian}
\end{figure}

\begin{table}[t]
\centering
\resizebox{0.8\linewidth}{!}{
\begin{tabular}{llcccc|cc   } 
\specialrule{.15em}{.05em}{.05em}

                 Exp ID&Exp Name&  Base & \# Data & Curated & \# Steps & ViT-L/14  & ViT-bigG-14  \\ %
 \hline
 \hline
                 (a)&Curated-1K-s1k & SDXL & 1K & \cmark & 1K & 26.33 & 41.09 \\
                (b)&Curated-1K-s10k & SDXL & 1K & \cmark & 10k & 22.55 & 35.59 \\
 \hline
                (c)&Curated-10K-s4k & SDXL & 10K & \cmark & 4k & 26.55 & 41.08 \\
                (d)&Curated-10K-s10k & SDXL & 10K & \cmark & 10k & \textbf{26.87} & \textbf{41.77} \\
                (e)&Curated-10K-s20k & SDXL & 10K & \cmark & 20k & 25.96 & 40.56 \\
 \hline
                (f)&Curated-100K-s10k & SDXL & 100K & \cmark  & 10k & 25.79 & 40.32 \\
                (g)&Curated-100K-s40k & SDXL & 100K & \cmark  & 40k & 26.59 & 41.29 \\
                (h) &Curated-300K-s40k & SDXL & 300K & \cmark & 40K & 26.43 & 40.72 \\
  \hline
                (i)&Random-10K-s10k & SDXL & 10K & \xmark & 10k & 26.87 & 41.47\\
                (j) &Random-100K-s40k & SDXL & 100K & \xmark & 40k & 26.28 & 40.90\\
                (k) &AllData-s40k & SDXL & 700K & \xmark & 40k & 26.13 & 40.60 \\
   \hline
                (l) &Curated-10K-s10k (SD1.5) & SD1.5 & 10K & \cmark & 10k & 23.50 & 36.90 \\
                (m) &Curated-100K-s40k (SD1.5) & SD1.5 & 100K & \cmark & 40k & 25.48 & 39.07 \\

\specialrule{.15em}{.05em}{.05em}
\end{tabular}
}

\caption{Comparison on CLIP scores of NeRF renderings with different variants of fine-tuning settings.}
\label{tab:instant3d-ablation_data_amount}
\end{table}

\paragraph{Quality and size of fine-tuning dataset.}
We evaluate the impact of the quality and size of the dataset used for fine-tuning 2D text-to-image models.
We first make comparisons between curated and uncurated (randomly selected) data.
The CLIP score rises slightly as shown in Table~\ref{tab:instant3d-ablation_data_amount} (i.e., comparing Exp(d, i)), while there is a substantial quality improvement.
This aligns with the observation that the data quality can dramatically impact the results in the instruction fine-tuning stage 
of LLM~\citep{zhou2023lima}.

When it comes to data size, we observe a double descent from Table~\ref{tab:instant3d-ablation_data_amount} Exp(a, d, g) with 1K, 10K, and 100K data.
We pick Exp(a, d, g) here because they are the best results among different training steps for the same training data size.
The reason for this double descent can be spotlighted by qualitative comparisons, where training with 1K data can lead to inconsistent multi-view images, while training with 100K data can hurt the compositionality, photo-realism, and also text alignment.
\paragraph{Number of fine-tuning steps.}
We also quantitatively and qualitatively analyze the impact of fine-tuning steps.
For each block in Table~\ref{tab:instant3d-ablation_data_amount} 
we show the CLIP scores of different training steps.
Similar to the findings in instruction fine-tuning~\citep{ouyang2022training}, 
the results do not increase monotonically regarding the number of fine-tuning steps but have a peak in the middle.
For example, in our final setup with the SDXL base model and 10K curated data (i.e., Exp(c, d, e)), the results are peaked at 10K steps.
For other setups, the observations are similar.
There is an obvious degradation in the quality of the results for both 4K and 20K training steps. 

Another important observation is that the peak might move earlier when the model size becomes larger.
This can be observed by comparing between Exp(l,m) for SD1.5 and Exp(d,g) for SDXL.
Note that this comparison is not conclusive yet from the Table given that SD1.5 does not perform reasonably 
with our direct fine-tuning protocol.

We also found that Exp(a) with 1K steps on 1K data can achieve the best CLIP scores but the view consistency is actually disrupted.
A possible reason is that the CLIP score is insensitive to certain artifacts introduced by reconstruction from inconsistent images, which also calls for a more reliable evaluation metric for 3D generation.

\subsection{Conclusions}\label{sec:instant3d-conclusions}
We presented a novel feed-forward two-stage approach \methodname{} that can generate 
high-quality and diverse 3D assets from text prompts within 20 seconds. 
Our method finetunes a 2D text-to-image diffusion model
to generate consistent 4-view images, and lifts them to 3D with a
robust transformer-based large reconstruction model. The experiment results 
show that our method outperforms previous feed-forward methods in terms 
of quality while being equally fast, and achieves comparable or better performance
to previous optimization-based methods with a speed-up of more than $200$ times. 
\methodname{} allows novice users to easily create 3D assets 
and enables fast prototyping and iteration for various applications such as
3D design and modeling.

\paragraph{Ethics Statement.} 
The generation ability of our model is inherited from the public 2D diffusion model SDXL.
We only do lightweight fine-tuning over the SDXL model thus it is hard to introduce extra knowledge to it.
Also, our model can share similar ethical and legal considerations to SDXL.
The curation of the data for lightweight fine-tuning does not introduce outside annotators.
Thus the quality of the data might be biased towards the preference of the authors, 
which can lead to a potential bias on the generated results as well.
The text input to the model is not further checked by the model,
which means that the model will try to do the generation for every text prompt it gets
without the ability to acknowledge unknown knowledge.

\endgroup

\section{Carve3D}
\label{sec:carve3d}
\begingroup
\newcommand{\methodname}{Carve3D\xspace}
\newcommand{\metricname}{MRC\xspace}
\newcommand{\ignore}[1]{}
\label{sec:carve3d-intro}
Instant3D can generation high-quality 3D assets, but it relies heavily on a good multi-view diffusion model.
Most multi-view diffusion models~\cite{shi2023mvdream,li2023instant3d,liu2023syncdreamer,zhao2023efficientdreamer,liu2023zero1to3} rely on supervised finetuning (SFT) using multi-view renderings from 3D datasets~\cite{deitke2022objaverse, deitke2023objaversexl}. SFT improves consistency, but prolonged finetuning shifts the model toward the limited 3D data distribution, reducing diversity, texture detail, and realism~\cite{li2023instant3d}. A similar trade-off appears in LLM alignment: instruction SFT improves compliance but can introduce dataset bias and hallucination~\cite{schulman2023rlhf}, motivating reinforcement-learning finetuning (RLFT) in InstructGPT~\cite{ouyang2022InstructGPT}. By analogy, RLFT is a natural next step for multi-view diffusion to improve consistency without further distribution shift.

We therefore introduce \methodname, which couples an enhanced RLFT algorithm with a new Multi-view Reconstruction Consistency (MRC) metric to improve the consistency of multi-view diffusion models. \cref{fig:carve3d-teaser,fig:carve3d-overview} summarizes the method and its effects.

The MRC metric compares the generated multi-view images to images rendered from a reconstructed NeRF at the same camera viewpoints. We use the sparse-view Large Reconstruction Model (LRM)~\cite{hong2023lrm,li2023instant3d} for fast, feed-forward NeRF reconstruction from a few views. Image similarity is measured with LPIPS~\cite{zhang2018lpips}, and we normalize LPIPS by the foreground bounding boxes to prevent trivial reward hacking via object size reduction.
To validate the reliability of \metricname, we conduct extensive experiments with controlled inconsistency levels; 
starting from a set of perfectly consistent multi-view images rendered from a 3D asset~\cite{deitke2022objaverse}, we manually introduce distortion to one of the views to create inconsistency.
Our MRC metric provides robust evaluation of consistency of multi-view images, offers a valuable tool for assessing current multi-view generation methods and guiding future developments in the field.

With \metricname, we apply RLFT to multi-view diffusion models. The RLFT procedure repeatedly samples diverse multi-view images for curated creative prompts, computes MRC rewards, and updates the diffusion model (\cref{fig:carve3d-overview}). This diversity- and quality-preserving finetuning is not feasible with SFT alone, since collecting ground-truth multi-view images for such prompts is prohibitively expensive. We make three specific improvements to the RLFT algorithm~\cite{black2023DDPO}: we use a purely on-policy policy-gradient method~\cite{williams1992REINFORCE} instead of partially on-policy PPO~\cite{schulman2017ppo} to improve stability; we include KL regularization~\cite{fan2023dpok, ouyang2022InstructGPT} to stay close to the base model and avoid distribution shift; and we scale compute to reach optimal rewards using diffusion-model RLFT scaling laws identified empirically~\cite{black2023DDPO,fan2023dpok}.

Applying Carve3D RLFT to Instant3D-10K~\cite{li2023instant3d} (a multi-view diffusion model SFT from SDXL~\cite{podell2023sdxl}) yields the Carve3D Model (Carve3DM). Quantitative and qualitative experiments, along with a user study, show that Carve3DM (1) improves multi-view consistency and NeRF reconstruction quality over Instant3D-10K, -20K, and -100K, while (2) preserving prompt alignment, diversity, and realistic detail relative to the base Instant3D-10K, avoiding the degradation seen in longer SFT. These results indicate that pairing SFT with Carve3D RLFT is important for achieving multi-view consistency. We also apply MRC to existing models, revealing the prevalence of inconsistency when relying solely on SFT.
This work is the first application of RLFT to text-to-3D, including a 2.6B-parameter SDXL denoising UNet~\cite{podell2023sdxl}. We release code and data to support further research on RLFT and alignment in computer vision.

\begin{figure}
    \centering
    \includegraphics[width=0.49\textwidth]{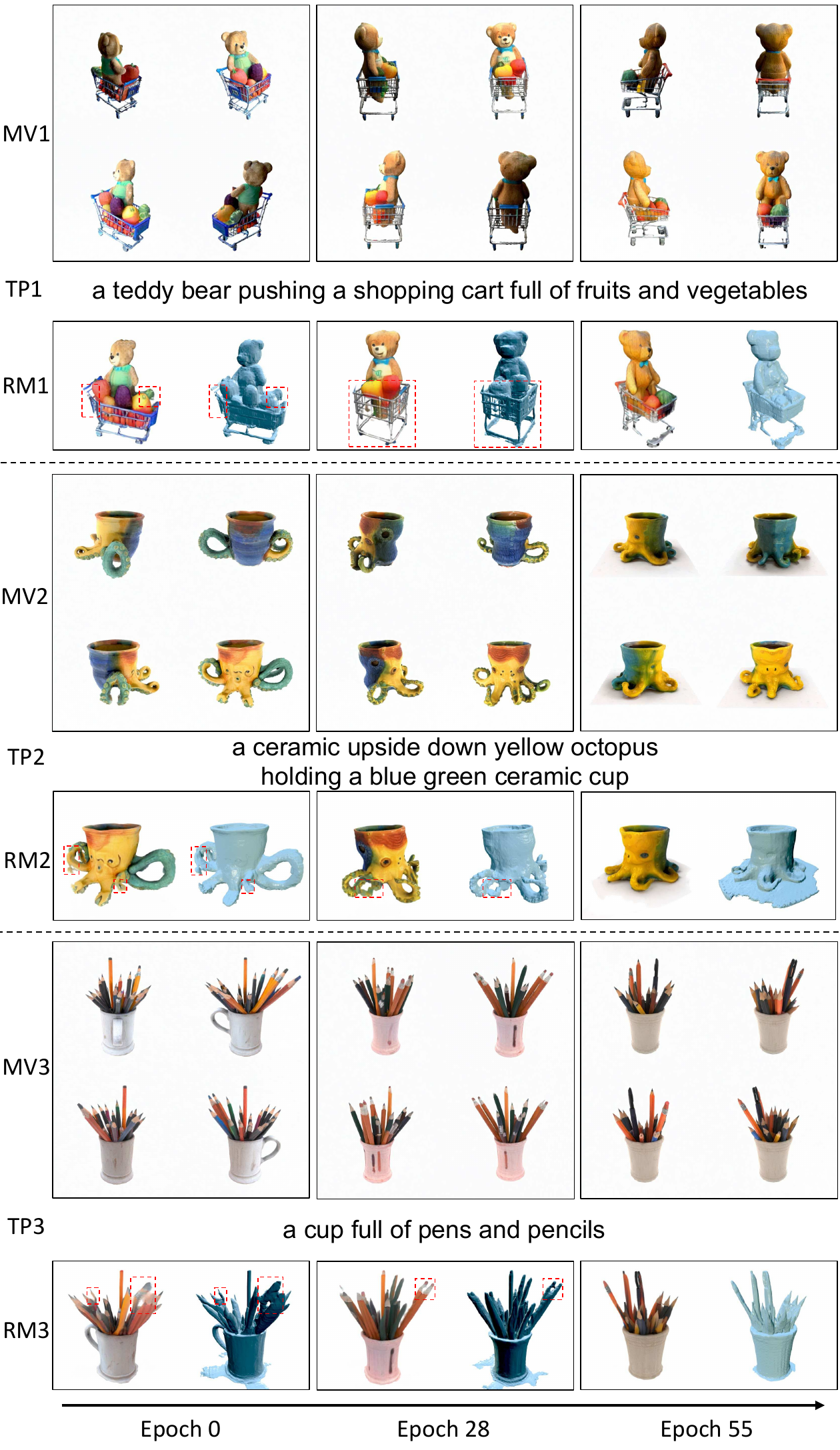}

    \caption{
        Our \methodname algorithm steadily improves the 3D consistency of a multi-view diffusion model and the resulting quality of the NeRF and the mesh, without sacrificing its image-prompt alignment, texture details, or realism. 
        Here, we show 3 testing-set results (in 3 rows, numbered as 1-3, separated by dotted lines) from the finetuning process (epoch 0, 28, and 55 in 3 columns). 
        Each row includes the generated multi-view images (denoted as MV), the reconstructed NeRF and extracted mesh (denoted as RM) and the text prompt (denoted as TP).
        The inconsistencies in the multi-view images, e.g. the facing direction of the shopping cart, the position of the octopus arms, and the position of the pencils, lead to artifacts in the NeRF and the mesh (highlighted in red).
    }

    \label{fig:carve3d-teaser}
\end{figure}

\begin{figure*}
    \centering
    \includegraphics[width=0.95\textwidth]{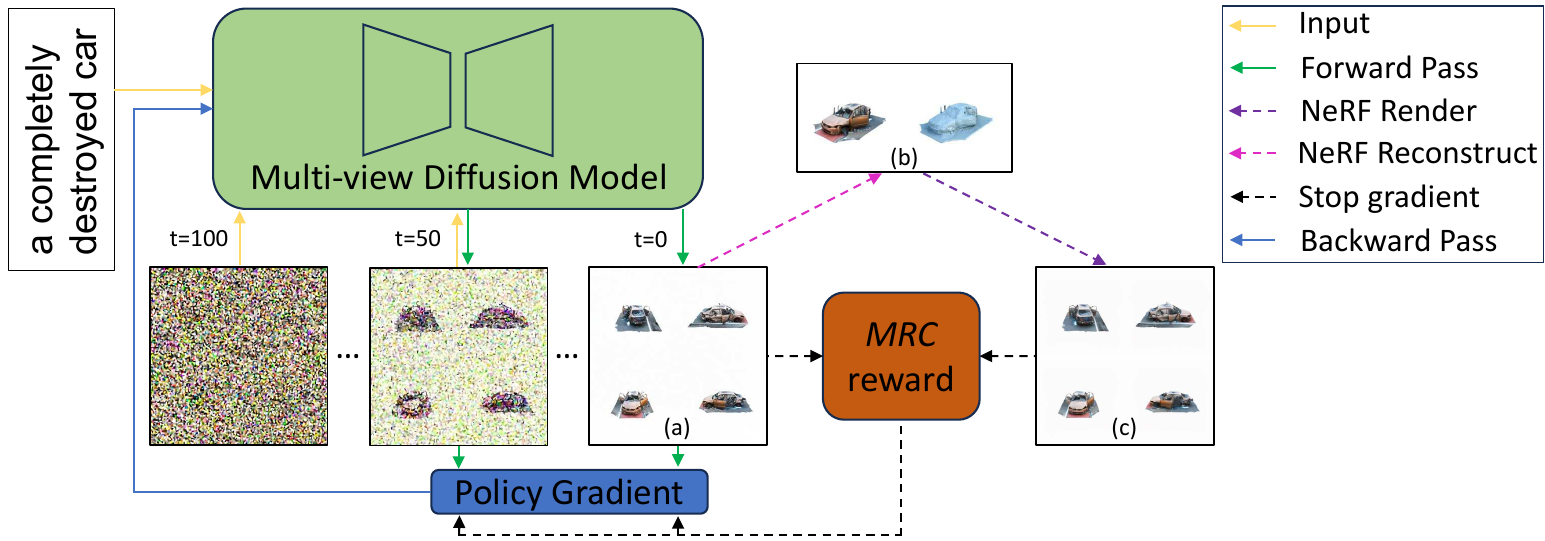}
    \caption{
        Overview of Carve3D. 
        Given a prompt sampled from our curated prompt set and a initial noisy image, we iteratively denoise the image using the UNet. 
        The final, clean image contains four multi-view images tiled in a 2-by-2 grid.
        \metricname reward is computed by comparing (a) the generated multi-view images with (c) the corresponding multi-view images rendered at the same camera viewpoints from (b) the reconstructed NeRF.
        Then, we train the model with policy gradient loss function, where the loss is derived from the reward and log probabilities of the UNet's predictions, accumulated over all denoising timesteps.
        By using only a set of training text prompts, our RLFT algorithm finetunes the diffusion model by evaluating its own generated outputs, without relying on ground truth multi-view images.
    }

    \label{fig:carve3d-overview}
\end{figure*}

\begin{figure*}
    \centering
    \includegraphics[width=0.85\linewidth]{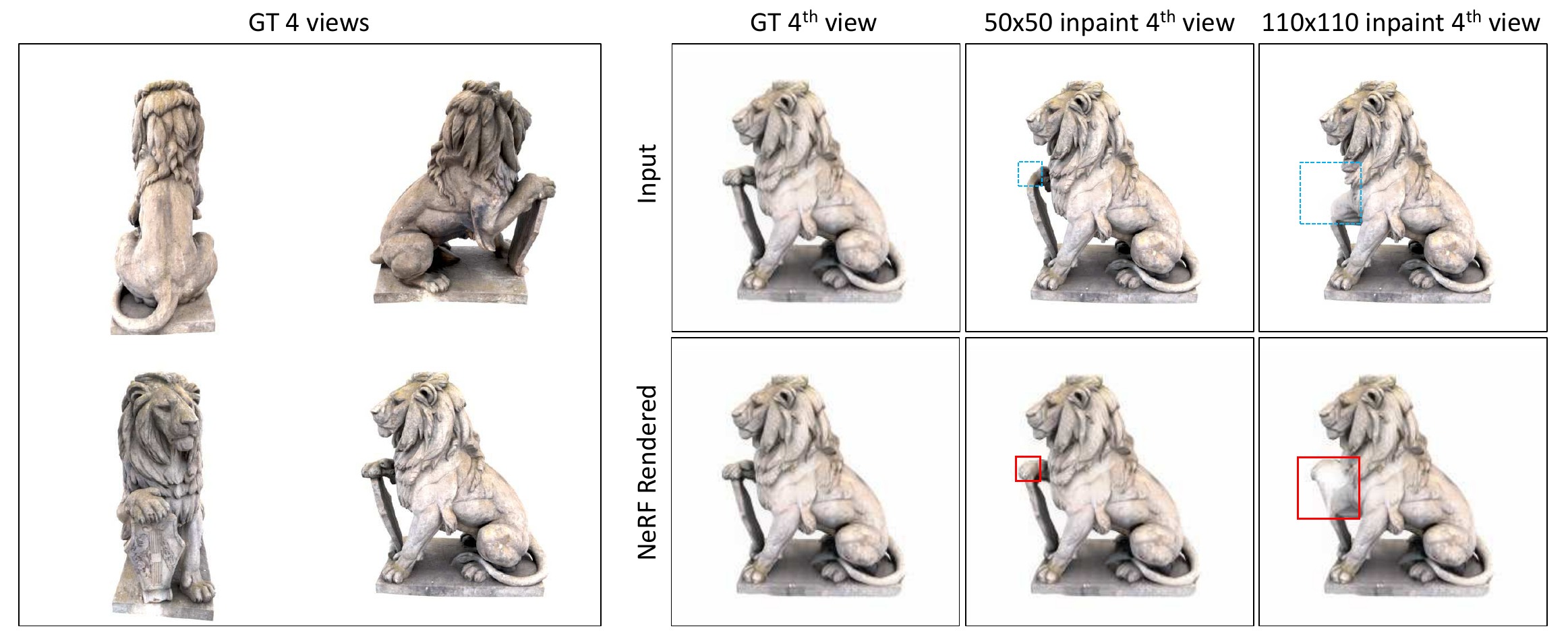}
    \caption{
        Qualitative correlation between \metricname and multi-view inconsistency with increasing intensity, introduced by inpainting with increasing mask sizes.
        Left: the four ground truth views.
        Right: the 4th view is inpainted with increasing area sizes, i.e. 0$\times$0, 50$\times$50 and 110$\times$110 pixels.
        The top row is the image after inpainting and the bottom row is the image rendered from the NeRF reconstructed with the top inpainted 4th view and the other 3 original GT views.
        We mark the inpainting area with blue and red boxes.
        Since the lion's right paw in the inpainted 4th views look different from the other three original views, its shape is broken in the NeRF and the rendered views. 
        This difference is captured in MRC's image dissimilarity metric.
}

    \label{fig:carve3d-method:consistency}

\end{figure*}

\begin{figure}
    \centering
    \includegraphics[width=0.6\textwidth]{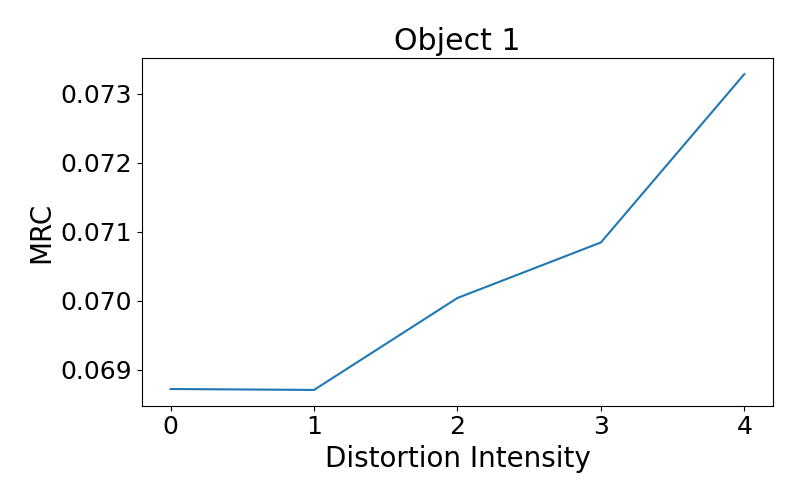}

    \caption{
    Quantitative correlation between \metricname and multi-view inconsistency with increasing intensity, for the object shown in~\Cref{fig:carve3d-method:consistency}.
    As inconsistency intensity rises, \metricname also monotonically increases.
    }

    \label{fig:carve3d-method:consistency-plot}
\end{figure}

\subsection{Multi-view Reconstruction Consistency}
\label{sec:carve3d-metric}

In this section, we propose the Multi-view Reconstruction Consistency (MRC) metric, for quantitative and robust evaluation of the consistency of multi-view images, which we define to be \textit{the degree of geometry and appearance uniformity of an object across the views}.

\subsubsection{Evaluate Consistency via NeRF Reconstruction}
\label{sec:carve3d-metric:viaNeRF}

A 3D model represented by Neural Radiance Field (NeRF) can be reconstructed from the view images of the object and their corresponding camera poses.
The quality of a NeRF notably depends on the consistency of the provided images images~\cite{mildenhall2020nerf, watson2022novel} -- inconsistent views lead to artifacts in the NeRF, which includes floaters, blurring, and broken geometry.
To address this challenge, we introduce a metric for assessing the consistency among multiple views.%

The intuition behind MRC comes from the relationship between multi-view consistency and the reconstructed NeRF.
As shown in~\cref{fig:carve3d-method:consistency}, when the multi-view images are consistent, they can produce a well reconstructed NeRF, preserving almost all the visual cues from the input images;
therefore, the views rendered from the NeRF at the same camera viewpoints will look the same as the original views;
conversely, when the multi-view images are inconsistent (e.g., intentionally introduced inconsistency in~\cref{fig:carve3d-method:consistency}), they will produce a NeRF with broken geometry and floater artifacts; 
thus, the NeRF rendered views will look different from the original views.
Building upon this observation, we propose the \metricname metric, defined as the image distances between the original multi-view images and the views of the reconstructed NeRF rendered at the same viewpoints, as illustrated in~\cref{fig:carve3d-overview}.

\subsubsection{Implementation}
\label{sec:carve3d-metric:implementation}
We formulate the implementation of \metricname as three parts: fast sparse-view NeRF reconstruction, measuring image distance between the input images and the rendered images, and a normalization technique for the image distance.

\paragraph{Fast Sparse-view Reconstruction}
We conduct NeRF reconstruction with sparse-view Large Reconstruction Model (LRM) proposed in ~\cite{li2023instant3d,hong2023lrm}.
Different from dense view NeRF reconstruction~\cite{mildenhall2020nerf,Chen2022tensorf,mueller2022instantngp}, sparse-view LRM reconstructs a NeRF with only $4$-$6$ view images.
Also, with its feed-forward reconstruction, it can achieve a speed two orders of magnitude faster than previous optimization-based reconstruction methods.
\metricname leverages all multi-view images for both NeRF reconstruction and 3D consistency evaluation.
Although the NeRF is reconstructed based on the visual prior of the input multi-views images, the rendering from the same views still exhibits notable differences if there is inconsistency inside the input, as shown in~\cref{fig:carve3d-method:consistency}.

\paragraph{Image Distance Metric}
In~\cref{sec:carve3d-metric:viaNeRF}, the consistency problem is reduced from 3D to a 2D image dissimilarity problem.
To measure the image dissimilarity between the input views and their corresponding NeRF rendered views, we utilize the perceptual image distance metric, LPIPS~\cite{zhang2018lpips}.
LPIPS exhibits smoother value changes with respect to the consistency of multi-view images compared to PSNR, SSIM, L1, and L2.
Such smoothness is derived from the non-pixel-aligned computation in LPIPS, as opposed to the other image distance metrics that are more pixel-aligned. 
Also, the smoothness is a crucial aspect for MRC to serve as the reward function in RLFT, because non-smooth, high-variance reward functions makes the RLFT training more challenging.

\paragraph{Bounding-box Normalization} 
Current multi-view diffusion models~\cite{shi2023mvdream,li2023instant3d,liu2023syncdreamer,zhao2023efficientdreamer} target single object generation with background.
Consequently, 
if computing LPIPS on the entire image, trivially reducing the object's relative size can exploit \metricname, as the majority of images will be the white background.
Therefore, we propose normalizing our metric with respect to the object's size.
Specifically, we identify the smallest square bounding box of the foreground object in the input view image.
Then we crop both the input images and the rendered images with that bounding box, resize them to a fixed resolution, and evaluate the LPIPS.
This normalization effectively prevents the reward hacking of \metricname by diminishing foreground object sizes.

\subsubsection{Metric Experiment}
\label{sec:carve3d-metric:exp}
The two key objectives for introducing the \metricname metric are (1) to assess the consistency of any multi-view generative model and (2) to enable  RLFT for improving the consistency of multi-view diffusion models.
Thus, the proposed consistency metric should ideally present two respective properties: 
(1) MRC should monotonically increase as inconsistency increases;
(2) the MRC vs. inconsistency curve should be smooth.

To validate the effectiveness and robustness of \metricname, i.e. whether it satisfies the two properties, we conduct evaluation on sets of multi-view images with controlled level of inconsistency.
Starting from a set of perfectly-consistent ground truth views rendered from a 3D asset from Objaverse~\cite{deitke2022objaverse}, we manually introduce inconsistency to one image.
We select a portion of this image and inpaint it with an image-to-image diffusion model\footnote{We use Adobe Photoshop's Generative Fill~\cite{adobe_firefly} without text prompt to add inpainting distortion, which is based on a diffusion model.}.
Therefore, we get different levels of distortion on one image, determined by the size of the inpainting area, that corresponds to different levels of inconsistency of the set of images.

\cref{fig:carve3d-method:consistency} shows the qualitative result on one object of our MRC metric experiment.
With increased inpainting area size, the NeRF rendered view also shows larger image difference, which is then captured by MRC's image distance metric, LPIPS.
\cref{fig:carve3d-method:consistency-plot} presents the quantitative curve of the same experiment. 
\metricname indeed shows a monotonically increasing pattern as the views become more inconsistent.
MRC constantly exhibits monotonically increasing pattern, and it is also smoother than the other MRC variants using PSNR, SSIM, L1, and L2.

\subsection{RLFT for Multi-view Consistency}
\label{sec:carve3d-RL}

In~\cref{sec:carve3d-metric}, we proposed a fast and reliable multi-view consistency metric named MRC, and in this section we describe how it can be used to finetune a multi-view diffusion model.
In this section, we present an improved Reinforcement Learning Finetuning (RLFT) algorithm for enhancing the consistency of 2D multi-view diffusion models, using the negative \metricname as the reward function. 
Building upon DDPO~\cite{black2023DDPO}, we opt for its pure on-policy version over the default partially on-policy version of the policy gradient algorithm for substantially improved training stability.
To maintain proximity to the base model, we incorporate KL divergence regularization similar to ~\cite{fan2023dpok,ouyang2022InstructGPT}.
In addition, we scale up the RLFT to achieve higher rewards by studying the scaling laws~\cite{kaplan2020scalinglaws} of diffusion model RLFT through extensive experiments.

\subsubsection{Preliminaries on DDPO}
\paragraph{Markov Decision Process}
To use RL for finetuning, we need to formulate the task as a Markov Decision Process (MDP).
In a MDP, an agent interacts with the environment at discrete timesteps;
at each timestep $t$, the agent is at a state $s_t$, takes an action $a_t$ according to its policy $\pi(a_t|s_t)$, receives a reward $r_t$, and transitions to the next state $s_{t+1}$.
Following denoising diffusion policy optimization (DDPO)~\cite{black2023DDPO}, the denoising process of a diffusion model is formulated as a multi-step MDP:
\begin{align}
    s_t&=(c, t, x_t),\\
    a_t&=x_{t-1},\\
    \pi(a_t|s_t)&=p_\theta(x_{t-1}|c, t, x_t),\\
    r(s_t, a_t)&=
    \begin{cases}
    r(x_0,c) & \text{if } t=0,\\
    0 & \text{otherwise},
    \end{cases}\\
    r(x_0,c) &= -\text{MRC}(x_0)
\end{align}
where each denoising step is a timestep, $c$ is the context, i.e. the text prompt, $x_t$ is the image being denoised at step $t$, $p_\theta$ is the diffusion model being finetuned, $x_T$ is the initial noisy image, $x_0$ is the fully denoised image, and $r(x_0,c)$ is the negative \metricname computed on the fully denoised image.

\paragraph{Policy Gradient}
In order to optimize the model with respect to the reward function, a family of RL algorithms, known as policy gradient methods, are commonly adopted, such as REINFORCE~\cite{williams1992REINFORCE} and Proximal Policy Optimization (PPO)~\cite{schulman2017ppo}.
$\text{DDPO}_{\text{SF}}$ is based on the vanilla policy gradient algorithm, REINFORCE~\cite{williams1992REINFORCE}, also known as the Score Function (SF) of diffusion models.
On the other hand, $\text{DDPO}_{\text{IS}}$ builds upon PPO~\cite{schulman2017ppo} and conducts multiple optimization steps per round of data using an importance sampling (IS) estimator and importance weight clipping.

As a common practice to reduce the variance of the policy gradients~\cite{mnih2016a3c}, DDPO~\cite{black2023DDPO} uses the advantages (\cref{eq:carve3d-A_r}), which are rewards normalized to have zero mean and unit variance, instead of directly using the rewards.
Specifically, the mean and standard deviation statistics of the rewards are tracked for each prompt $c$:
\begin{align}
    A_r(x_0,c) = \frac{r(x_0,c) - \mu_r(c)}{\sigma_r(c)}
    \label{eq:carve3d-A_r}
\end{align}
DDPO's~\cite{black2023DDPO} reward-normalizing advantage replaces the value model that is more widely adopted in PPO-based~\cite{schulman2017ppo} RLHF methods~\cite{ouyang2022InstructGPT, yao2023deepspeedchat, vonwerra2022trl}. 
This is similar to~\cite{li2023remax}, which shows that the value model creates unnecessary computation cost that can be replaced with a simpler advantage formulation. 

By using the advantage term $A_r$ (\cref{eq:carve3d-A_r}) in place of the reward $r$, the $\text{DDPO}_{\text{SF}}$ policy gradient function is:
\begin{align}
    \hat{g}_{\text{SF}} = \mathbb{E}\left[\sum_{t=0}^{T} \nabla_{\theta} \log p_{\theta}(x_{t-1} | c, t, x_t) A_r(x_0,c)\right]
    \label{eq:carve3d-DDPO_SF}
\end{align}
where the expectation is taken over data generated by the policy $\pi_{\theta}$ with the parameters $\theta$.
The log probability $\log p_\theta(x_{t-1} | c, t, x_t)$ can be easily obtained since the policy is an isotropic Gaussian distribution when using the DDIM sampler~\cite{black2023DDPO,song2022ddim}.
Black \etal~\cite{black2023DDPO} choose $\text{DDPO}_{\text{IS}}$ as the default policy gradient function, because it exhibits better sample efficiency than $\text{DDPO}_{\text{SF}}$ (Fig.~4 of~\cite{black2023DDPO}).
Such choice is consistent with the use of PPO~\cite{schulman2017ppo} in Large Language Model (LLM) Reinforcement Learning from Human Feedback (RLHF) literature~\cite{ouyang2022InstructGPT, yao2023deepspeedchat,vonwerra2022trl,bai2022helpfulharmlessRLHF,bai2022RLAIF}.

\begin{figure}
\centering
    \includegraphics[width=0.68\textwidth]{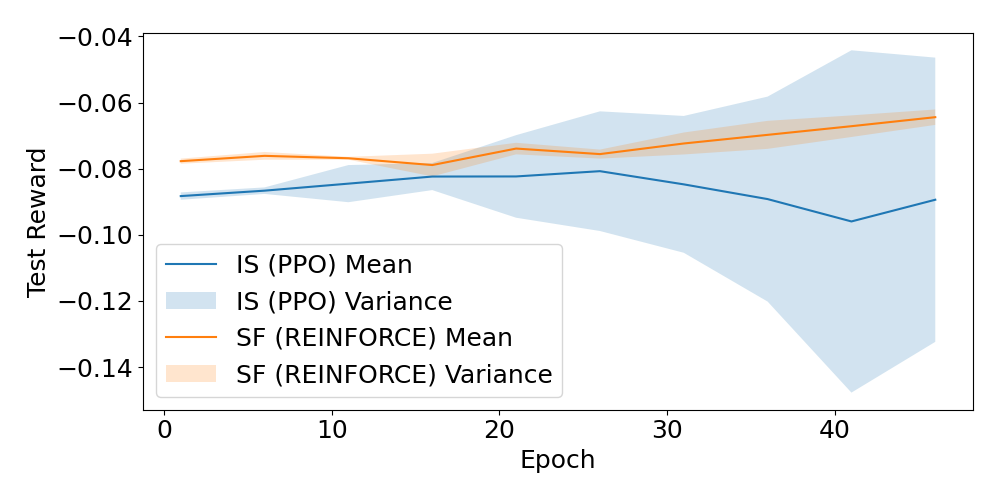}

    \caption{
    Comparing the IS and the SF versions of Carve3D reward curves on the testing set over 4 different random seeds.
    The IS version produces reward curves with high variance, including two runs that fails.
    In contrast, all runs of the SF version stably produces reward curves with low variance.
    }

    \label{fig:carve3d-IS_vs_SF_variance}
\end{figure}

\begin{figure}
    \centering
    \includegraphics[width=\textwidth]{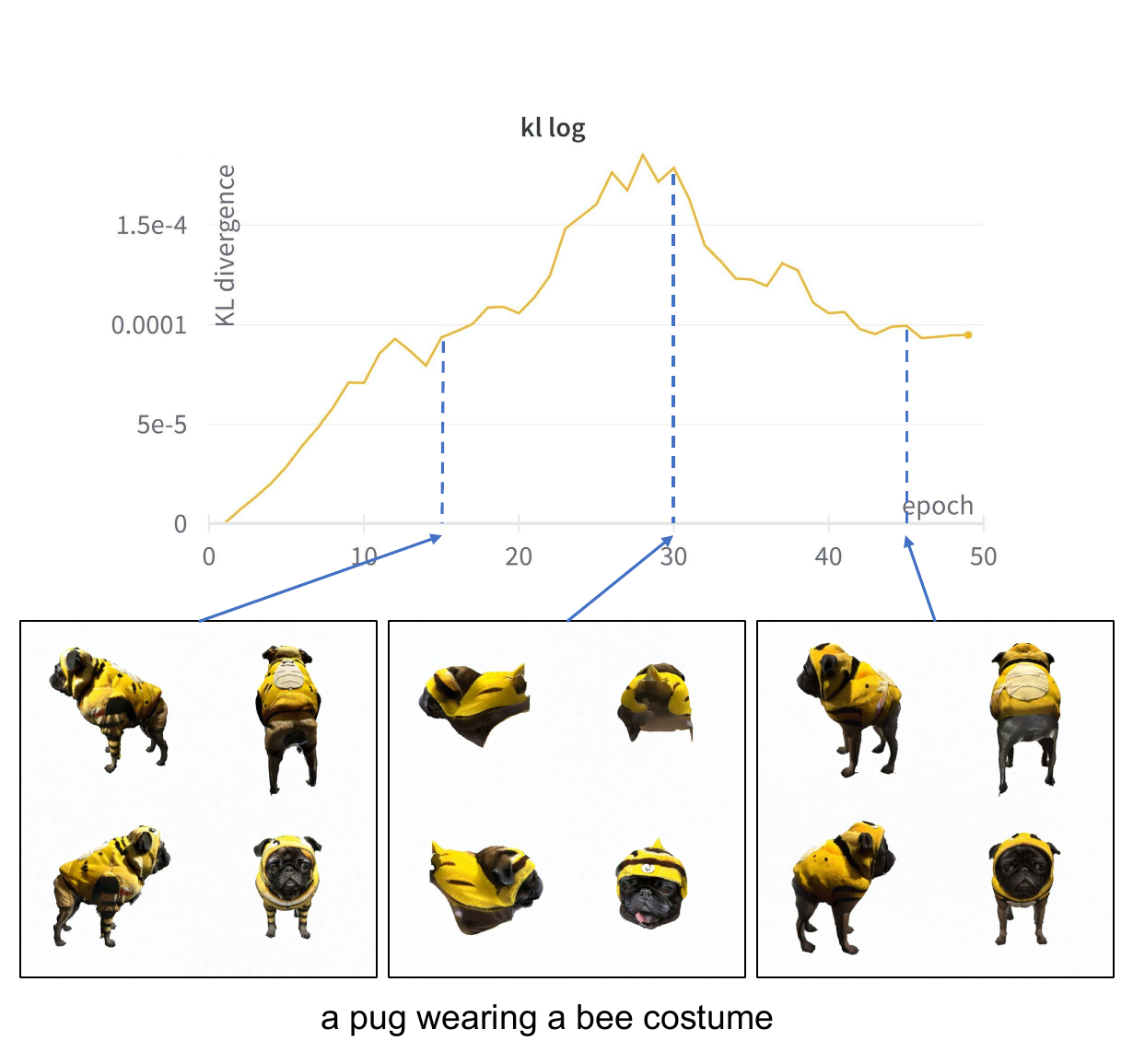}

    \caption{
    We observe qualitative correlation between the KL value and the prompt alignment degradation.
    Despite being distant in the finetuning process, epoch 15 and epoch 45, which have lower KL divergence to the base model, generates prompts that align better with the prompts.
    On the other hand, epoch 30, which has much higher KL divergence from the base model, generates results with broken identity, i.e. the body of the pug is missing.
    }

    \label{fig:carve3d-kl_degradation}
\end{figure}

\subsubsection{Improvements over DDPO}
\label{par:RL:improvements}

While RLFT with the default $\text{DDPO}_{\text{IS}}$ and our \metricname can enhance the 3D consistency of multi-view diffusion models, it still faces challenges regarding training stability, the shift of output distributions, and an unclear training scale setting to achieve optimal rewards with minimal distribution shift.
To address these issues, we propose three improvements over DDPO~\cite{black2023DDPO} in this section.
Given the universal nature of these challenges in RLFT, our enhancements may offer broader applicability across various tasks.

\paragraph{Pure On-policy Training}
Training stability is a major challenge in both RLFT~\cite{casper2023RLHFproblems, zheng2023RLHFsecrets} and traditional RL~\cite{eimer2023rlhyperparameters}.
With the default $\text{DDPO}_{\text{IS}}$, our training process is evidently unstable, as shown in~\cref{fig:carve3d-IS_vs_SF_variance}.
Training experiments with different random seeds or a slight change of hyperparameters can lead to different reward curves and qualitative results.
This complicates the training result evaluation as we cannot distinguish meaningful improvement or deterioration from the variance introduced by random seed.

We argue that such high variance is derived from the multi-step update in $\text{DDPO}_{\text{IS}}$~\cite{black2023DDPO}, originally proposed in PPO~\cite{schulman2017ppo}.
While it theoretically allows for better sample efficiency similar to off-policy methods~\cite{schulman2017ppo}, it also causes the training to be more unstable and the reward curves to be more variant, because it uses data collected with the older policy to update the newer policy.
Due to the undesirable consequences of training instability, we adopt the pure on-policy variant $\text{DDPO}_{\text{SF}}$, discarding the multi-step update from PPO.
As shown in~\cref{fig:carve3d-IS_vs_SF_variance}, $\text{DDPO}_{\text{SF}}$ significantly improves the training stability of our RLFT, while maintaining a comparable sample efficiency as the default $\text{DDPO}_{\text{IS}}$.

Diverging from DDPO~\cite{black2023DDPO} and most LLM RLHF literature~\cite{ouyang2022InstructGPT, yao2023deepspeedchat,vonwerra2022trl,bai2022helpfulharmlessRLHF,bai2022RLAIF}, we choose REINFORCE~\cite{williams1992REINFORCE} ($\text{DDPO}_{\text{SF}}$) over PPO~\cite{schulman2017ppo} ($\text{DDPO}_{\text{IS}}$) for its superior training stability.
We provide two hypotheses behind our surprising finding, including the difficulty of the task reward function and the size of the model being finetuned.
The favored use of REINFORCE~\cite{williams1992REINFORCE} over PPO~\cite{schulman2017ppo} could apply to broader scenarios that meet these two conditions. 
We leave the verification of our hypotheses as future work.

\paragraph{KL Divergence Regularization}
\label{sec:carve3d-RL:KL}
In RLFT methods, distribution shift (also known as reward overoptimization) can lead to low-quality results,
such as cartoon-like, less realistic style~\cite{black2023DDPO} or oversaturated colors and unnatural shape~\cite{fan2023dpok}, despite achieving high rewards.
In our case, we observe this as degradation of diversity, texture details and prompt alignment after prolonged RLFT with the \metricname reward. 
Previous methods~\cite{ouyang2022InstructGPT, fan2023dpok} suggest mitigating reward overoptimization by incorporating a penalty on the Kullback–Leibler (KL) divergence between the log probabilities of the outputs from the base and the finetuned models.
In our case, the base model is Instant3D-10K~\cite{li2023instant3d} without any additional finetuning.
By plotting the KL divergence values during finetuning, we also find that KL divergence correlates to the reward overoptimization problem (\cref{fig:carve3d-kl_degradation}), suggesting us to adopt KL divergence regularization.

Following the widely adopted implementation in LLM RLHF~\cite{yao2023deepspeedchat, ouyang2022InstructGPT}, we incorporate KL penalty into the reward function. %
Subtraction of the log probabilities is commonly used to approximate the full KL divergence~\cite{vonwerra2022trl, yao2023deepspeedchat}:
\begin{align}
    \text{KL}\left( \log{p_\theta(x_{0}|c,T,x_T)} || \log{p_{\theta_\text{base}}(x_{0}|c,T,x_T)}\right) \nonumber \\
    = \sum_{t=0}^T{\frac{\log{p_\theta(x_{t-1}|c,t,x_t)} - \log{p_{\theta_\text{base}}(x_{t-1}|c,t,x_t)}}{T+1}}
    \label{eq:carve3d-kl_computation}
\end{align}
where $p_{\theta_\text{base}}$ is the base model.
We will denote this approximated KL divergence term as $\text{KL}(x_0|c,x_T)$ for clarity in presentation. 

KL divergence values starts at $0$ and unavoidably increases as finetuning proceeds, making it hard to determine an optimal coefficient for the penalty term.
To enable a steady KL divergence regularization throughout the finetuning process, we propose to normalize the KL divergence penalty term.
This normalization ensures that the gradient consistently favors low-KL-divergence, high-reward samples, even in the early stages when KL divergence is still low compared to the later stages.
We extend DDPO's~\cite{black2023DDPO} per prompt stat tracking to also track the mean and standard deviation statistics of the KL divergence term in order to to normalize it:
\begin{align}
    A_\text{KL}(x_0,c) = \frac{\text{KL}(x_0|c,x_T) - \mu_\text{KL}(c)}{\sigma_\text{KL}(c)}.
    \label{eq:carve3d-kl_normalized}
\end{align}
Our advantage terms now consist of both the normalized reward (\cref{eq:carve3d-A_r}) and the normalized KL divergence (\cref{eq:carve3d-kl_normalized}).
Our final policy gradient function, used in our experiments, is a combination of~\cref{eq:carve3d-DDPO_SF,eq:carve3d-kl_normalized}
\begin{align}
    \hat{g}_{\text{SF,KL}} = \mathbb{E}\left[\sum_{t=0}^{T} \nabla_{\theta} \log p_{\theta}(x_{t-1} | c, t, x_t) \right. \nonumber \\
    \left. \vphantom{\sum_{t=0}^{T}} \cdot \left(\alpha A_r(x_0,c) - \beta A_\text{KL}(x_0,c)\right)\right]
    \label{eq:carve3d-DDPO_SF_KL}
\end{align}
where $\alpha$ and $\beta$ are the coefficients for the reward advantage and the KL advantage, respectively.

\begin{figure}
    \centering
    \includegraphics[width=0.68\textwidth]{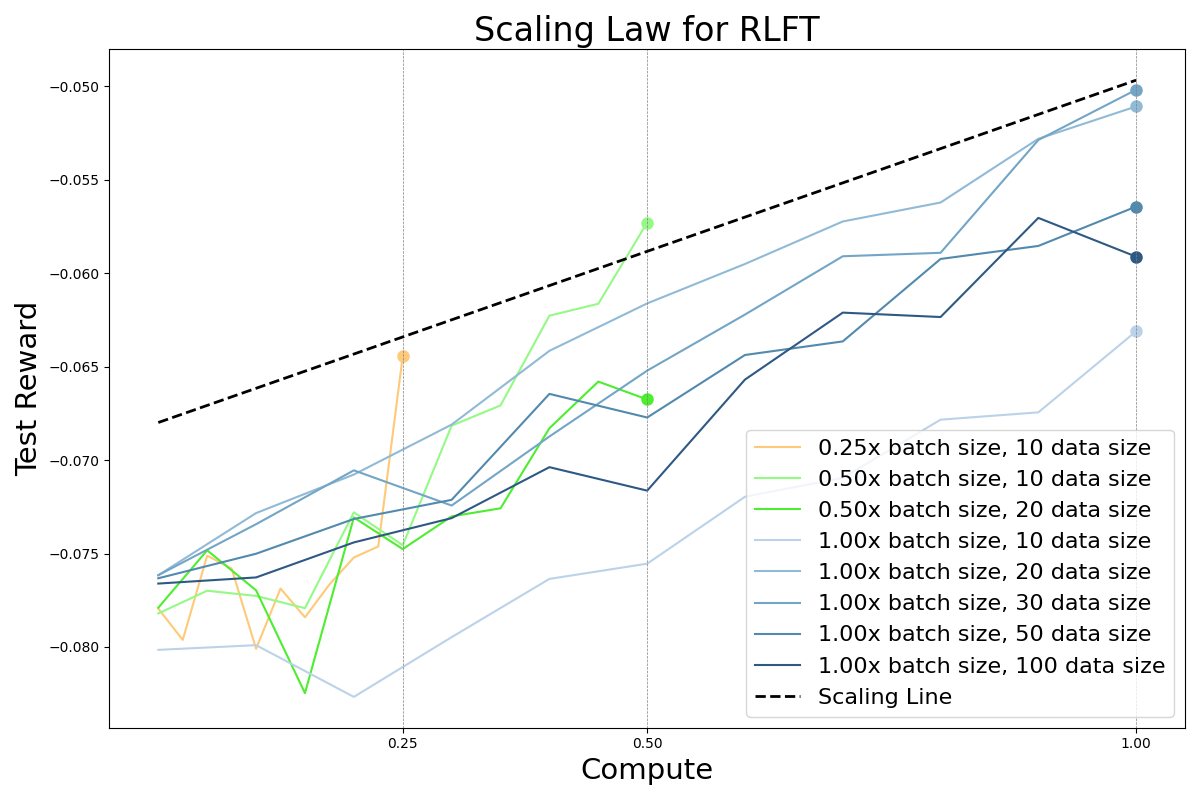}

    \caption{
    Scaling law for Carve3D's diffusion model RLFT algorithm.
    When we scale up the amount of compute, the model improves its reward smoothly under the optimal data size.
    The amount of compute scales linearly with respect to the batch size.
    The reward curves also become more stable (less variant) with a larger batch size.
    The reward curves are reported up to epoch $50$.
    }
    \label{fig:carve3d-data_vs_batch}

\end{figure}

\paragraph{Scaling Laws for RLFT}
\label{sec:carve3d-RL:scale}

The training of Reinforcement Learning (RL) is highly sensitive to the chosen scale setting~\cite{eimer2023rlhyperparameters}, impacting various results, including the final converged reward.
Through the scaling laws identified from extensive experiments, we scale up the amount of compute (equivalent to scaling up the batch size in our case) to achieve the optimal reward.
Although our scaling experiments are only conducted with the multi-view consistency task, our insights into the scaling laws of diffusion model RLFT are general and can be beneficial in broader scenarios.

\paragraph{Compute and Batch Size}
The reward curves from our experiments demonstrate a positive scaling law of the model's reward at epoch 50 with respect to the amount of compute (\cref{fig:carve3d-data_vs_batch}); 
the scaled up compute brings smooth improvement to the model's reward, under the optimal data sizes at each batch size.
Note that the amount of compute scales directly with respect to the batch size.

\paragraph{Data Size}
The model's reward does not directly scale with the data size but there exists a more complicated relationship between them. 
As shown in~\cref{fig:carve3d-data_vs_batch}, the optimal data size at each batch size grows as the batch size get larger, indicating that both factors need to be scaled up in tandem;
after the optimal data size, naively continuing to scale up the data size actually reduces the model's reward.
Surprisingly, even when trained on a prompt set as small as a size of $10$, the model still shows generalization to the testing prompts.
We choose data size of 30 with batch size 768 in our final experiments as it achieves the highest reward in our analysis.

\paragraph{Training Epochs}
With the pure on-policy $\text{DDPO}_{\text{SF}}$ (REINFORCE~\cite{williams1992REINFORCE}), the model steadily and smoothly improves its rewards throughout the finetuning process, meaning that more training epochs constantly lead to higher reward.
However, from our qualitative results, we also observe worse distribution shift, e.g. the degradation of prompt alignment and texture details, as training epoch increases. 
Due to the correlation between KL divergence and the quality degradation (\cref{fig:carve3d-kl_degradation}), we stop the finetuning early when a predefined KL divergence threshold is reached.
This threshold is empirically chosen based on qualitative results.
For fair comparisons, we report the reward curves up to epoch 50 in~\cref{fig:carve3d-data_vs_batch}.

\begin{figure}
    \centering
    \includegraphics[width=0.75\textwidth]{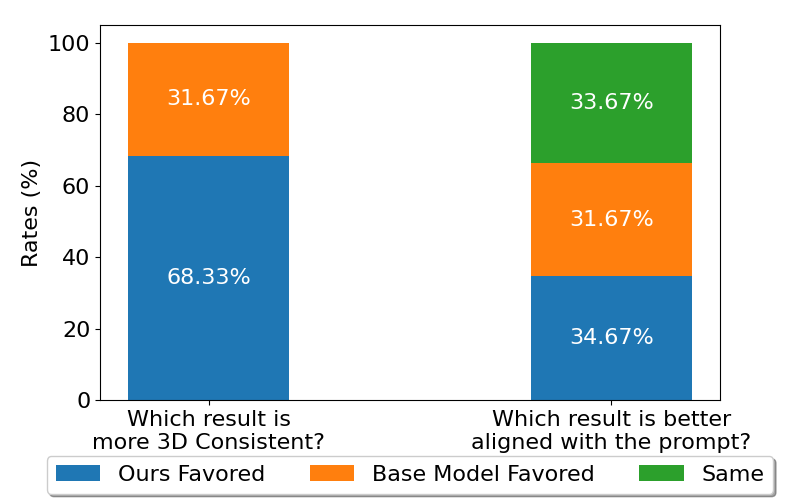}
    \caption{
    We conducted a user study with 20 randomly selected testing prompts and the corresponding outputs from both the base and fine-tuned model.
    $15$ participants took part in the study, with a majority favoring the 3D consistency of our fine-tuned model.
    Opinions are evenly split on which has better prompt alignment.}

    \label{fig:carve3d-results:user_study}
\end{figure}

\subsection{Experiments}

In this section, we evaluate the Carve3D Model (Carve3DM), obtained by applying the Carve3D Reinforcement Learning Finetuning (RLFT) algorithm on Instant3D~\cite{li2023instant3d}.

\subsubsection{Comparison with Base Model and Longer SFT}
\label{sec:carve3d-results:compare}
As shown in~\cref{fig:carve3d-IS_vs_SF_variance}, Carve3DM's Multi-view Reconstruction Consistency (MRC) reward steadily improves on the training set.
We aim to further answer the following questions:

\begin{enumerate}[leftmargin=0.8cm]
    \item Does Carve3DM's improved MRC generalize to the testing set?
    \item Quantitatively, is Carve3DM more consistent than the base model and the model with more Supervised Finetuning (SFT) steps?
    \item Qualitatively, does Carve3DM sacrifice the diversity, prompt alignment, and texture details of the base model?
\end{enumerate}

We quantitatively and qualitatively compare Carve3DM with Instant3D models with 10K, 20K, and 100K SFT steps, where Instant3D-10K is the default model in~\cite{li2023instant3d}.

\begin{table}[h]
\small
\centering
\begin{tabular*}{0.6\linewidth}{|@{\extracolsep{\fill}}l|c|}
\hline
\textbf{}               & \textbf{Avg \metricname on Testing Set} $\downarrow$ \\ \hline
MVDream     & $0.1222$        \\ \hline
Instant3D-10K (Base)     & $0.0892$        \\ \hline
Instant3D-20K     & $0.0795$        \\ \hline
Instant3D-100K     & $0.0685$        \\ \hline
\textbf{Carve3DM (Ours)} & \textbf{0.0606}  \\ \hline
Zero123++ & 0.0700  \\ \hline
SyncDreamer & 0.1018  \\ \hline
\end{tabular*}
\caption{
    Quantitative comparison of MVDream~\cite{shi2023mvdream}, Instant3D~\cite{li2023instant3d} with 10K (the base model), 20K, and 100K SFT steps, Carve3DM (ours, finetuned from Instant3D-10K), Zero123++~\cite{shi2023zero123plus}, and SyncDreamer~\cite{liu2023syncdreamer}.
    We evaluate them by generating 4 outputs for each of the 414 text prompts in the DreamFusion~\cite{poole2022dreamfusion} testing set.
    We let Zero123++ and SyncDreamer to use one of Carve3DM's output multi-view images as their input image conditioning.
    Carve3DM achieves substantially better MRC than all baselines, indicating its superior multi-view consistency.
}
\label{tab:carve3d-suppl:testing_mrc}

\end{table}

\paragraph{Quantitative Comparison and Generalization}
As shown in~\cref{tab:carve3d-suppl:testing_mrc}, 
when evaluated on the testing set, \methodname achieves substantially improved \metricname over the base model. 
More SFT steps indeed provides better multi-view consistency and achieves better MRC, i.e. Instant3D's 100K version is better than the 20K version, which is better than the 10K version. 
However, Carve3DM still outperforms even the most consistent 100K version of Instant3D by a notable gap.
This suggests that the explicit multi-view consistency objective in MRC, paired with our RLFT algorithm, can improve the model's consistency more efficiently than SFT.

Furthermore, although our RLFT only uses 30 training prompts, it brings multi-view consistency improvement that generalizes to the testing set containing 415 prompts.
Such generalization, also observed in~\cite{black2023DDPO,ouyang2022InstructGPT}, is likely derived from the strong knowledge from the base model.

\paragraph{Multi-view Consistency and NeRF Artifacts}
While the multi-view images generated by the base model may be inconsistent, causing artifacts such as floaters and broken geometry, Carve3D can fix such inconsistency in the multi-view images and produce NeRF with clean geometry, free of artifacts.

\paragraph{Prompt Alignment and Texture Details}
By virtue of our RLFT with KL-divergence regularization (\cref{sec:carve3d-RL:KL}), which prevents distribution shift,
Carve3DM preserves the prompt alignment and the texture details of the base model.
On the other hand, longer SFT causes additional distribution shift in Instant3D~\cite{li2023instant3d} from the base SDXL~\cite{podell2023sdxl} towards the SFT training set~\cite{deitke2022objaverse}.
Instant3D-20K and Instant3D-100K exhibits degradation in diversity, realism, and level of detailed textures.
This quality degradation with longer SFT is also observed in~\cite{li2023instant3d}.

\paragraph{Diversity}
Carve3DM preserves the generation diversity of the base model.
This owes to our RLFT process, which repeatedly samples different initial noises for the diffusion model to generate diverse results (\cref{fig:carve3d-overview}).

\subsubsection{Comparison with Existing Methods}
\label{sec:carve3d-suppl:mrc_mvdream}
We further compare Carve3DM with the text-conditioned multi-view diffusion model, MVDream~\cite{shi2023mvdream}, and two image-conditioned models, Zero123++~\cite{shi2023zero123plus} and SyncDreamer~\cite{liu2023syncdreamer}.
As shown in~\cref{tab:carve3d-suppl:testing_mrc}, Carve3DM's outputs have notably better multi-view consistency, realism, and level of details than the three baselines.

\subsubsection{User Study}
In addition to the quantitative and qualitative comparisons in~\cref{sec:carve3d-results:compare}, we conduct a user study to further understand the qualitative results of Carve3DM when perceived by human.
To run the user study, we \textit{randomly} selected 20 unseen testing prompts.
For each text prompt, we generated a pair of data from both the base and the finetuned models with the same initial noise. 
Then, we provided both the tiled 4-view image and the turntable video of the reconstructed NeRF to participants and asked them the following two questions:
(1) Which result is more 3D-consistent? and 
(2) Which result is better aligned with the prompt?
As shown in~\cref{fig:carve3d-results:user_study}, 68.33\% of participants believe that Carve3DM's generated results are more 3D consistent than that of the base model~\cite{li2023instant3d}.
Given that the multi-view consistency in the base model has already been much improved with SFT~\footnote{Please see \href{https://jiahao.ai/instant3d/}{https://jiahao.ai/instant3d/} for base model's 3D consistency},
the nearly $37\%$ gain in human preference introduced by \methodname on \textit{randomly} selected testing prompts is impressive.
Furthermore, participants find that Carve3DM exhibits similar prompt alignment as Instant3D.
The preservation of alignment can be attributed to the KL divergence regularization (\cref{sec:carve3d-RL:KL}) and early stopping the RLFT regarding KL divergence (\cref{sec:carve3d-RL:scale}).

\subsection{Conclusion}

We propose \methodname, a Reinforcement Learning Finetuning algorithm to improve the reconstruction consistency of 2D multi-view diffusion models.
\methodname relies on \metricname, a novel metric that measures the reconstruction consistency by comparing multi-view images with their corresponding NeRF renderings at the same viewpoints.
The resulting model, Carve3DM, demonstrates substantially improved multi-view consistency and NeRF quality without sacrificing the prompt alignment, texture details, or prompt alignment of the base model.
Our results conclude that pairing SFT with Carve3D's RLFT is essential for developing multi-view-consistent diffusion models. 

\ignore{
Broader Impacts: promote RL finetuning, alignment research, could be used to make safer, better aligned, reduced bias, harm diffusion models.

Metric could be used to evaluate/compare existing text-to-multi-view Diffusion Models. 
It could also be used to monitor the fineutning process and picking the best checkpoint.

Limitation: loss of high-frequency detail, due to limitation of Sparse-view LRM.
}

\endgroup

\section{DMV3D}
\label{sec:dmv3d}
\begingroup
\newcommand{\ourmethod}{{\it DMV3D}}
\newcommand{\para}[1]{\textbf{\noindent{#1}}}
\newcommand{\revise}[1]{#1}
\newcommand{\rmI}{\mathrm{I}}
\newcommand{\vc}{\mathbf{c}}
\newcommand{\vx}{\mathbf{x}}
\newcommand{\vr}{\mathbf{r}}
\newcommand{\vo}{\mathbf{o}}
\newcommand{\vd}{\mathbf{d}}
\newcommand{\FID}{FID $\downarrow$ \xspace}
\newcommand{\CLIP}{CLIP $\uparrow$ \xspace}
\newcommand{\PSNR}{PSNR $\uparrow$ \xspace}
\newcommand{\SSIM}{SSIM $\uparrow$ \xspace}
\newcommand{\LPIPS}{LPIPS $\downarrow$ \xspace}
\newcommand{\CD}{CD $\downarrow$ \xspace}
\newcommand{\FS}{F-Score $\uparrow$ \xspace}
\newcommand{\VIOU}{Volume IOU $\uparrow$ \xspace}
\definecolor{LightGray}{gray}{0.9}
\label{sec:dmv3d-intro}

Both Instant3D and Carve3D are two-stage generation methods, which first generate multi-view images and then reconstruct the NeRFs.
We also explored a single-stage direction, which results in DMV3D, a single-stage diffusion model that generates triplane NeRFs from text or a single image via direct diffusion inference. It can produce diverse, high-fidelity 3D objects in under a minute per asset (Fig.~\ref{fig:dmv3d-teaser}). DMV3D is a multi-view 2D image diffusion model whose denoiser integrates 3D NeRF reconstruction and rendering, trained end-to-end without direct 3D supervision. This avoids both pre-training per-asset NeRFs (two-stage methods) and per-asset optimization (SDS methods).

Conceptually, the model jointly solves 2D diffusion denoising and 3D reconstruction. It is inspired by RenderDiffusion~\citep{renderdiff}, which performs single-view diffusion for 3D generation, but avoids the category-specific priors and canonical poses that limit scalability. Instead, DMV3D uses a sparse set of four views that surround an object, which is sufficient to represent a full 3D asset but requires solving the challenging sparse-view 3D reconstruction problem, especially when inputs are noisy.

\revise{To address this, we leverage large transformer models that have shown strong scalability on challenging problems~\citep{hong2023lrm, gpt, shen2023gina}. Building on the 3D Large Reconstruction Model (LRM)~\citep{hong2023lrm}, we introduce a model for joint reconstruction and denoising. From a sparse set of noisy multi-view images, the transformer reconstructs a clean NeRF that supports rendering at arbitrary viewpoints. The model is conditioned on diffusion time steps to handle varying noise levels, enabling it to serve as the multi-view denoiser in a diffusion framework.}

\revise{DMV3D supports both single-image and text conditioning. For image conditioning, one sparse view is fixed as a noise-free input and the remaining views are denoised, analogous to multi-view inpainting~\citep{xie2023smartbrush}. For text, we apply attention-based conditioning and classifier-free guidance, as in 2D diffusion. Training uses only image-space supervision on large-scale datasets of synthetic renderings from Objaverse~\citep{objaverse} and real captures from MVImgNet~\citep{mvimgnet}.}
The resulting model achieves state-of-the-art single-image 3D reconstruction, outperforming prior SDS-based methods and 3D diffusion models, and also delivers strong text-to-3D results.

In sum, the main contributions are:
\begin{itemize}

    \item A single-stage diffusion framework that leverages multi-view 2D image diffusion to achieve 3D generation;

\item \revise{An LRM-based multi-view denoiser that reconstructs noise-free triplane NeRFs from noisy multi-view images;}

\item A general approach for high-quality text-to-3D generation and single-image reconstruction.

\end{itemize}

\begin{figure}

    \includegraphics[width=1.\textwidth]{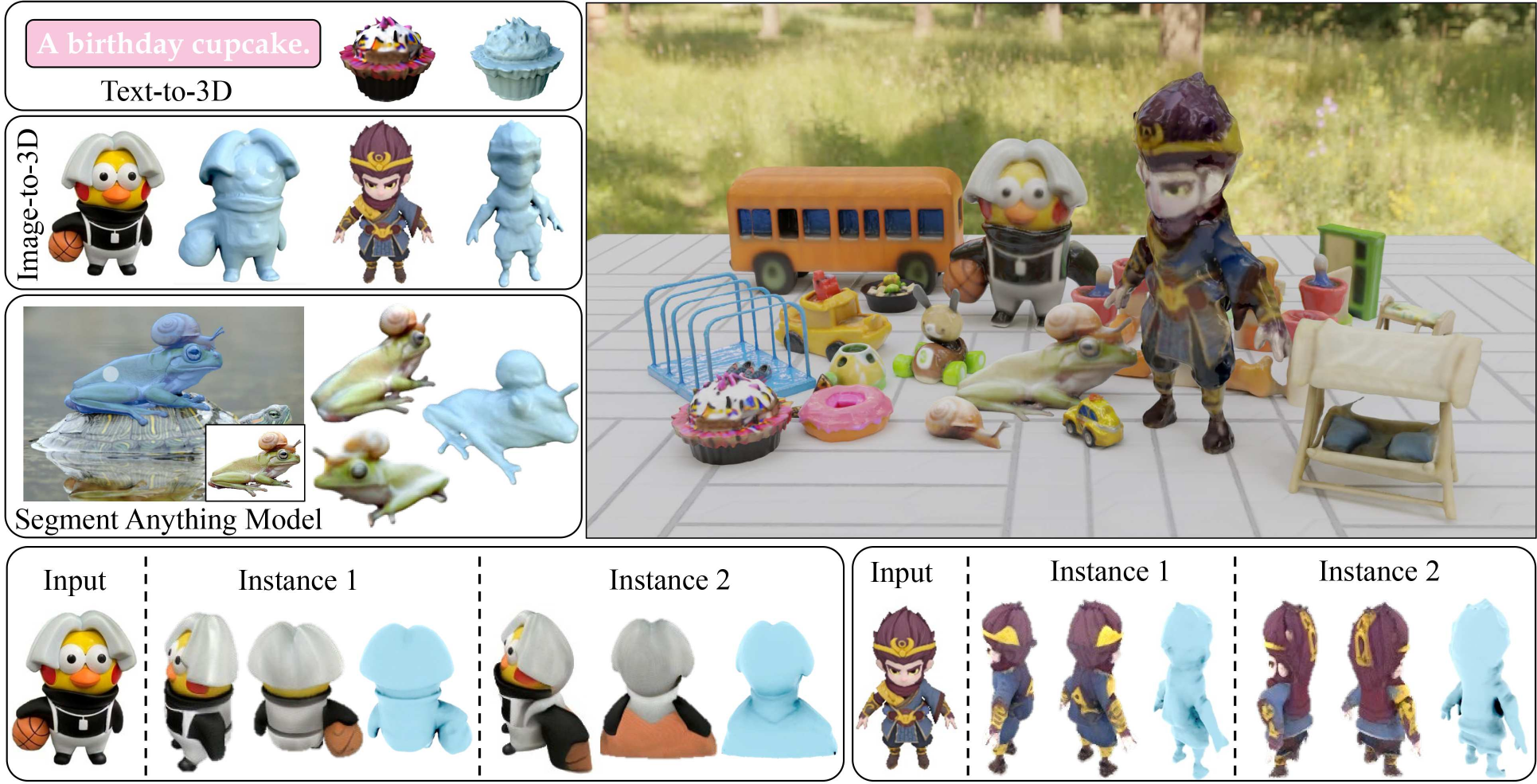}

   \captionof{figure}{Top left: our approach achieves fast 3D generation from text or single-image input; the latter one, combined with 2D segmentation methods (like SAM \citep{kirillov2023segment}), allows us to reconstruct objects segmented from natural images. Bottom: as a probabilistic generative model, our model can produce multiple reasonable 3D assets from the same image. Top right: we demonstrate a scene comprising diverse 3D objects generated by our models, each within one minute.}

    \label{fig:dmv3d-teaser}
\end{figure}

This perspective bridges 2D and 3D generative modeling by unifying reconstruction and generation, and it suggests a path toward foundation models for a broad range of 3D vision and graphics tasks.

\subsection{Method}
\label{sec:dmv3d-method}

\begin{figure}\label{fig:dmv3d-method}
\centering
\includegraphics[width=0.9\textwidth]{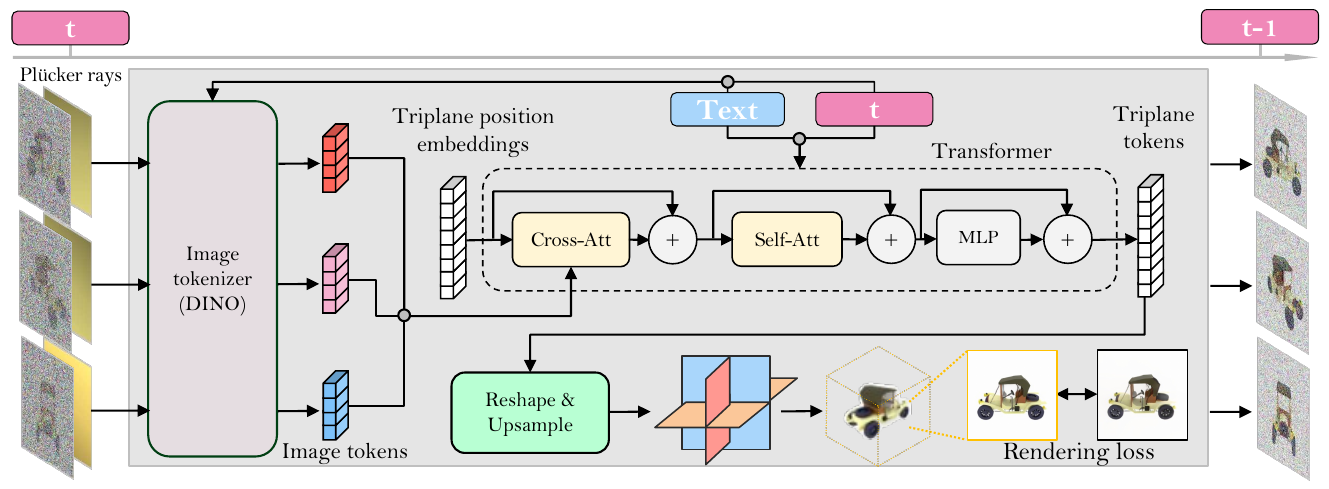}
\caption{\textbf{Overview of our method.} 
We denoise multiple views (three shown in the figure; four used in experiments) for 3D generation. Our multi-view denoiser is a large transformer model that reconstructs a noise-free triplane NeRF from input noisy images with camera poses (parameterized by Plucker rays). During training, we supervise the triplane NeRF with a rendering loss at input and novel viewpoints.
During inference, we render denoised images at input viewpoints and combine them with noise to obtain less noisy input for the next denoising step. Once the multi-view images are fully denoised, our model offers a clean triplane NeRF, enabling 3D generation. Refer to Sec.~\ref{sec:dmv3d-method:conditional} for how to extend this model to condition on single image.
}

\end{figure}

We now present our single-stage diffusion model. 
In particular, we introduce a novel diffusion framework that uses a reconstruction-based denoiser to denoise multi-view noisy images for 3D generation (Sec.~\ref{sec:dmv3d-method:mvdiffusion}).
Based on this, we propose a novel large 3D reconstruction model conditioned on diffusion time step, functioning as the multi-view denoiser, to denoise multi-view images via 3D NeRF reconstruction and rendering (Sec.~\ref{sec:dmv3d-method:reconstructor}).
We further extend our model to support text and image conditioning, enabling practical and controllable generation (Sec.~\ref{sec:dmv3d-method:conditional}).

\subsubsection{Multi-view Diffusion and Denoising}
\label{sec:dmv3d-method:mvdiffusion}

\para{Diffusion.} Denoising Diffusion Problistic Models (DDPM) extends the data distribution $\vx_0 \sim q(\vx)$ with a $T$-step Markov Chain using a Gaussian noise schedule.
The generation process is the reverse of a forward diffusion process. 
The diffusion data $\vx_{t}$ at timestep $t$ can be derived by $\vx_{t} = \sqrt{\bar{\alpha}_{t}} \vx_0 + \sqrt{1 - \bar{\alpha}_{t}} \epsilon$,
where $\epsilon \sim \mathcal{N}(0, \rmI)$ represents Gaussian noise and $\bar{\alpha}_t$ is a monotonically decreasing noise schedule.

\para{Multi-view diffusion.} The original $\vx_0$ distribution addressed in 2D DMs is the (single) image distribution in a dataset. We instead consider the (joint) distribution of multi-view images $\mathcal{I} = \{\rmI_1, ..., \rmI_N\}$, where each set of $\mathcal{I}$ are image observations of the same 3D scene (asset) from viewpoints $\mathcal{C} = \{\vc_1, ..., \vc_N\}$. The diffusion process is equivalent to diffusing each image independently but with the same noise schedule: 
\begin{align}
    \mathcal{I}_t &= \{\sqrt{\bar{\alpha}_{t}} \rmI + \sqrt{1 - \bar{\alpha}_{t}} \epsilon_{\rmI} | \rmI \in \mathcal{I}  \}
\end{align}
Note that this diffusion process is identical to the original one in DDPM, despite that we consider a specific type of data distribution $\vx = \mathcal{I}$ of per-asset 2D multi-view images. 

\para{Reconstruction-based denoising.} The reverse of the 2D diffusion process is essentially denoising.
In this work, we propose to leverage 3D reconstruction and rendering to achieve 2D multi-view image denoising, while outputting a clean 3D model for 3D generation. %
In particular, we leverage a 3D reconstruction module $\mathrm{E}(\cdot)$ to reconstruct a 3D representation $\mathrm{S}$ from the noisy multi-view images $\mathcal{I}_t$ (at time step $t$), and render denoised images with a differentiable rendering module $\mathrm{R}(\cdot)$:
\begin{align}
    \rmI_{r,t} = \mathrm{R}(\mathrm{S}_t, \vc), \quad \mathrm{S}_t = \mathrm{E}(\mathcal{I}_t, t, \mathcal{C})  
    \label{eqn:reconrender}
\end{align}
where $\rmI_{r,t}$ represents a rendered image from $\mathrm{S}_t$ at a specific viewpoint $\vc$.

Denoising the multi-view input $\mathcal{I}_t$ is done by rendering $\mathrm{S}_t$ at the viewpoints $\mathcal{C}$, leading to the prediction of noise-free $\mathcal{I}_0$.
This is equivalent to $\vx_0$ prediction in 2D DMs \citep{ddim}, which can be used to predict $\vx_{t-1}$, enabling progressive denoising inference.
However, unlike pure 2D generation, we find merely supervising $\mathcal{I}_0$ prediction at input viewpoints cannot guarantee high-quality 3D generation (see Tab.~\ref{tab:dmv3d-ablation}), often leading to rendering artifacts at novel viewpoints.
Therefore, we propose to also supervise images rendered at novel viewpoints from the 3D model $\mathrm{S}_t$.
In essence, we reposition the original 2D image $\vx_0$ ($\mathcal{I}_0$) prediction to a (hidden) 3D $\mathrm{S}_0$ prediction task, ensuring consistent high-quality rendering across arbitrary viewpoints. 
The denoising objective is written as 
\revise{
\begin{align}
    \mathrm{L_\mathit{recon}}(t) &= \mathbb{E}_{\rmI,\vc \sim {\mathcal{I}_\mathit{full}, \mathcal{C}_\mathit{full}}}\ \ \ell \big(\rmI, \mathrm{R}(\mathrm{E}(\mathcal{I}_t, t, \mathcal{C}), \vc)\big)
\end{align}
where $\mathcal{I}_\mathit{full}$ and $\mathcal{C}_\mathit{full}$ represent the full set of images and poses (from both randomly selected input and novel views), and $\ell(\cdot, \cdot)$ is an image reconstruction loss penalizing the difference between groundtruth $\rmI$ and rendering $\mathrm{R}(\mathrm{E}(\mathcal{I}_t, t, \mathcal{C}), \vc)$.}
Note that our framework is general -- potentially any 3D representations ($\mathrm{S}$) can be applied. In this work, we consider a (triplane) NeRF representation (where $\mathrm{R}(\cdot)$ becomes neural volumetric rendering) and propose a transformer-based reconstructor $\mathrm{E(\cdot)}$.

\subsubsection{Reconstructor-based Multi-view Denoiser}
\label{sec:dmv3d-method:reconstructor}

\revise{We build our multi-view denoiser upon LRM~\citep{hong2023lrm} and uses large transformer model to reconstruct a clean triplane NeRF~\citep{eg3d} from noisy sparse-view posed images. Renderings from the reconstructed triplane NeRF are then used as denoising outputs.}

\para{Reconstruction and rendering.} 
\revise{As shown in Fig.~\ref{fig:dmv3d-method}, we use a Vision Transformer (DINO~\citep{caron2021emerging}) to convert input images $\mathcal{I} = \{\rmI_1, ..., \rmI_N\}$  to 2D tokens, and then use a transformer to map a learnable triplane positional embedding to the final triplane representing the 3D shape and appearance of  an asset; the predicted triplane is then used to decode volume density and color with an MLP (not shown in Fig.~\ref{fig:dmv3d-method} to avoid clutterness) for differentiable volume rendering. The transformer model consists of  a series of triplane-to-images cross-attention and triplane-to-triplane self-attention layers as in the LRM work~\citep{hong2023lrm}. We further enable time conditioning for diffusion-based progressive denoising and introduce a new technique for camera conditioning.}

\para{Time Conditioning}.
Our transformer-based model requires different designs for time-conditioning, compared to DDPM and its variants that are based on CNN UNets.
Inspired by DiT~\citep{peebles2022dit}, we apply time condition through the \textit{adaLN-Zero block} in our self- and cross- attention layers in our model, allowing our model to effectively handle input with different diffusion noise levels.

\para{Camera Conditioning}.
\revise{Training our model on datasets with highly diverse camera intrinsics and extrinsics, e.g., MVImgNet~\citep{mvimgnet}, requires an effective design of input camera conditioning to facilitate the model's understanding of cameras for 3D reasoning.}
A basic strategy is, as in the case of time conditioning, to use \textit{adaLN-Zero block} on the camera parameters (as done in \cite{li2023instant3d}). However,  we find that conditioning on camera and time simultaneously with the same strategy tends to weaken the effects of these two conditions and often leads to an unstable training process and slow convergence.
Instead, we propose a novel approach -- parameterizing cameras with sets of pixel-aligned rays. 
In particular, following LFN~\citep{sitzmann2021light}, we parameterize rays using Plucker coordinates as $\vr = (\vo \times \vd, \vd)$, where $\vo$ and $\vd$ are the origin and direction of a pixel ray and can be computed from the camera parameters. %
We concatenate the Plucker coordinates with image pixels, and send them to the ViT transformer for 2D image tokenization, achieving effective camera conditioning.

\subsubsection{Conditioning on single image or text}
\label{sec:dmv3d-method:conditional}
The methods described thus far enable our model to function as an unconditional generative model.
We now introduce how to model the conditional probabilistic distribution with a conditional denoiser $\mathrm{E}(\mathcal{I}_t, t, \mathcal{C}, y)$, where $y$ is text and image conditioning, enabling controllable 3D generation. 

\para{Image Conditioning}.
Unlike previous methods \citep{zero123} that design new modules to inject image conditioning to a DM, we propose a simple but effective view-inpainting strategy for our multi-view model.
In particular, we keep the first view $\rmI_1$ (in the denoiser input) noise-free as the image condition, while applying diffusion and denoising on other views. 
In this case, the denoiser essentially learns to fill in the missing pixels within the noisy views using cues extracted from the first input view, similar to the task of image inpainting which has been shown to be addressable by 2D DMs \citep{ldm}.
In addition, to improve the generalizability of our image-conditioned model, we generate tri-planes in a coordinate frame aligned with the conditional view and render other images using poses relative to the conditional one. 

\para{Text Conditioning}.
To add text conditioning into our model, we adopt a strategy similar to that presented in Stable Diffusion~\citep{ldm}. 
We use the text encoder from CLIP~\citep{radford2021learning} to generate text embeddings and inject them into our denoiser using cross-attention.
Specifically, we include an additional cross-attention layer after each self-attention block in the ViT and each cross-attention blocak in the triplane transformer, enabling text-driven 3D generation. 

\subsubsection{Training and Inference}
\label{sec:dmv3d-method:training}

\para{Training}. 
During the training phase, we uniformly sample time steps $t$ within the range $[1, T]$, and add noise according to a cosine schedule. 
We sample input images with random camera poses, instead of fixing ones, enhancing the robustness of our system. We also randomly sample additional novel viewpoints to supervise the renderings (as discussed in Sec.~\ref{sec:dmv3d-method:mvdiffusion}) for better quality.
We minimize the following training objective with conditional signal $y$:
\revise{
\begin{align}
    \mathrm{L} = \mathbb{E}_{t\sim U[1, T], (\rmI,\vc) \sim (\mathcal{I}_{full}, \mathcal{C}_{full})}\ \ \ell \big( \rmI, \mathrm{R}(\mathrm{E}(\mathcal{I}_t, t, \mathcal{D}, y), \vc)\big)
\end{align}
For the image reconstruction loss $\ell(\cdot, \cdot)$, we use a combination of L2 loss and LPIPS loss~\citep{zhang2018perceptual}, with loss weights being 1 and 2, respectively.}

\para{Inference}.
For inference, we select four viewpoints that uniformly surround the object in a circle with the same pitch, to ensure the reconstruction model (denoiser) can capture the full 3D shape and appearance. 
We utilize DDIM~\citep{ddim} to improve the inference speed in the progressive multi-view denoising.
Once the 2D multi-view images are fully denoised at the final step, we can directly obtain a clean triplane NeRF model from the denoiser, achieving fast 3D generation without requiring any extra optimization to fit the multi-view denoised images.

\subsection{Experiments}
\label{sec:dmv3d-exp}

In this section, we present an extensive evaluation of our method.
In particular, we briefly describe our experiment settings (Sec.~\ref{sec:dmv3d-exp:setting}), compare our results with previous works (Sec.~\ref{sec:dmv3d-exp:img}), and show additional analysis and ablation experiments (Sec.~\ref{sec:dmv3d-exp:ablation}). 

\subsubsection{Settings}
\label{sec:dmv3d-exp:setting}
\para{Implementation details}.
We use Adam optimizer to train our model with an initial learning rate of 4$e^{-4}$. We also apply a warm-up stage for 3$K$ steps and a cosine decay on the learning rate.
We train our denoiser with 256 $\times$ 256 input images and render 128 $\times$ 128 image crops for supervision. 
Our final model is a large transformer with 48 attention layers and $64^{3}$ triplane tokens with 32 channels. 
We use 128 NVIDIA A100 GPUs to train this model with a batch size of 8 per GPU for 100$K$ steps, taking about 7 days.
Since the final model takes a lot of resources, it is impractical for us to evaluate the design choices with this large model for our ablation study.
Therefore, we also train a small model that consists of 36 attention layers to conduct our ablation study. 
The small model is trained with 32 NVIDIA A100 GPUs for 200$K$ steps (4 days).

\para{Datasets}.
Our model requires only 2D image supervision.
We use rendered multi-view images from $\sim$700k scenes in the Objaverse \citep{objaverse} dataset to train our text-to-3D model, for which we use Cap3D \citep{cap3d} to generate the text prompts.
For each scene, we render 32 images under uniform lighting at random viewpoints with a fixed 50$^\circ$ FOV. 
For image-conditioned (single-view reconstruction) model, we combine the Objaverse data with additional real captures of $\sim$200k scenes from the MVImgNet \citep{mvimgnet} dataset, enhancing the generalization to out-of-domain input.
In general, these datasets contain a large variety of synthetic and real assets from numerous categories, allowing us to train a generic and scalable 3D generative model. 

\begin{table}[t]
\centering
\resizebox{\textwidth}{!}{
\begin{tabular}{@{}lccccc|ccccc@{}}
\toprule
& \multicolumn{5}{c}{ABO dataset} & \multicolumn{5}{c}{GSO dataset} \\
                        & \FID & \CLIP & \PSNR  & \LPIPS & \CD   & \FID & \CLIP & \PSNR  & \LPIPS & \CD   \\
\midrule
Point-E                 &  112.29   &  0.806    & 17.03      &  0.363     &  0.127   &   123.70  &   0.741    &  15.60     & 0.308   &  0.099                    \\
Shap-E                        &  79.80   & 0.864     &    15.29   &   0.331    &  0.097   & 97.05   & 0.805  &    14.36   & 0.289    & 0.085                   \\
Zero123                        &  31.59   & 0.927      & 17.33      & 0.194      &  $-$  & 32.44    & 0.896     &  17.36     &  0.182     &   $-$                  \\
One2345                      &  190.81   &  0.748    &   12.00    &   0.514    & 0.163 & 139.24    & 0.713     &  12.42     & 0.448      &   0.123                       \\
Magic123                      &   34.93  & 0.928 & 18.47    & 0.180        & 0.136   &   34.06   &  0.901    &    18.68     &  0.159  &0.113                      \\
\midrule
Ours (S) &  36.77 & 0.915 & 22.62 & 0.194 & 0.059  & 35.16 & 0.888 & 21.80 & 0.150 & 0.046 \\
Ours  & 27.88 & 0.949 & 24.15  & 0.127 & 0.046  & 30.01 & 0.928 & 22.57 & 0.126 & 0.040 \\
\bottomrule
\end{tabular}
}
\caption{Evaluation Metrics of single-image 3D reconstruction on ABO and GSO datasets.}
\label{tab:dmv3d-evalabo}
\end{table}

We evaluate our image-conditioned model with novel synthetic datasets, including 100 scenes from the Google Scanned Object (GSO) \citep{gso} and 100 scenes from the Amazon Berkeley Object (ABO) \citep{abo} datasets. 
This allows for direct comparison of single-view reconstruction with the groundtruth.
Note that accurate quantitative evaluation of 3D generation remains a challenge in the field, we use the most applicable metrics from earlier works to assess our and baseline models.

\begin{figure}[t]
\begin{center}

\includegraphics[width=0.95\textwidth]{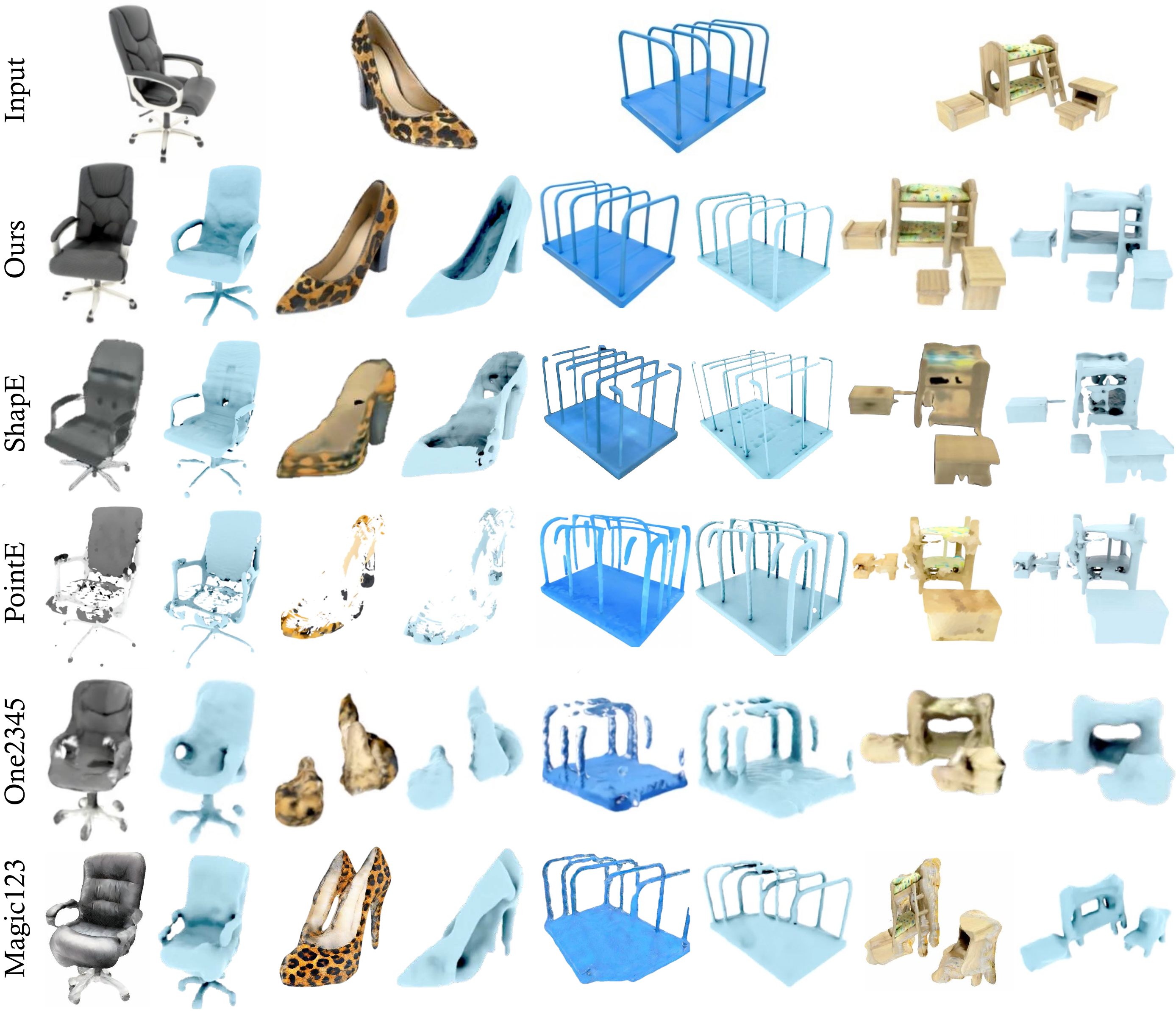}
\end{center}

\caption{ \textbf{Qualitative comparisons on single-image reconstruction.}}

\label{fig:dmv3d-singleimg}
\end{figure}

\subsubsection{Results and comparisons}
\label{sec:dmv3d-exp:img}

\para{Single-image reconstruction.} We compare our image-conditioned model with previous methods, including Point-E \citep{pointe}, Shap-E \citep{shape}, Zero123 \citep{zero123}, One2345 \citep{one2345}, and Magic123 \citep{Magic123}, on single-image reconstruction.
We evaluate the novel-view rendering quality from all methods using PSNR, LPIPS, CLIP precision (including top-1 R-precision and averaged precision), and FID, computed between the rendered  and GT images. In addition, we also compute the Chamfer distance (CD) for geometry evaluation, for which we use marching cubes to extract meshes from NeRFs.

\setlength{\tabcolsep}{4pt}
\begin{wraptable}[10]{R}{6cm}

\centering
\resizebox{\linewidth}{!}{%
\begin{tabular}{@{}lcccc@{}}
\toprule
\multirow{2}{*}{Method} & \multicolumn{2}{c}{VIT-B/32}    & \multicolumn{2}{c}{ViT-L/14} \\ \cmidrule(lr){2-3} \cmidrule(lr){4-5} %
                        & R-Prec & AP & R-Prec & AP \\
\midrule
Point-E    &  33.33          &   40.06           &  46.4   & 54.13 \\
Shap-E    & 38.39 & 46.02 & 51.40 & 58.03 \\
\midrule
\rowcolor{LightGray}
Ours  & 39.72 & 47.96 & 55.14 & 61.32\\
\bottomrule
\end{tabular}
}
\caption{Evaluation Metrics on Text-to-3D.}
\label{tab:dmv3d-text23d}

\end{wraptable} 

Table~\ref{tab:dmv3d-evalabo} report the quantitative results on the GSO and ABO testing sets respectively.
Note that our models (even ours-small ) can outperforms all baseline methods, achieving the best scores across all metrics for both datasets.
Our high generation quality is reflected by the qualitative results shown in Fig.~\ref{fig:dmv3d-singleimg}. Our model generates realistic results with more complete geometry and much sharper appearance details, compared to all baselines. 

In particular, the two-stage 3D DMs, ShapE and Point-E, lead to lower quality, often with incomplete shapes and blurry textures; this suggests the inherent difficulties in denoising pretrained 3D latent spaces, a problem our model avoids.
On the other hand, Zero123 leads to better quantitative results than ShapE and Point-E on appearnce, because it is a 2D diffusion model and trained to generate high-quality images.
However, Zero123 alone cannot output a 3D model required by many 3D applications and their rendered images suffer from severe inconsistency across viewpoints. 
This inconsistency also leads to the low reconstruction and rendering quality from One2345, which attempts to reconstruct meshes from Zero123's image outputs.
On the other hand, the per-asset optimization-based method Magic123 can achieve rendering quality comparable to Zero123 while offering a 3D mdoel. However, these methods require long (hours of) optimization time and also often suffer from unrealistic Janus artifacts (as shown in the second object in Fig.~\ref{fig:dmv3d-singleimg}).
In contrast, our approach is a single-stage model with 2D image training objectives and directly generates a 3D NeRF model (without per-asset optimization) while denoising multi-view diffusion. Our scalable model learns strong data priors from massive training data and produces realistic 3D assets without Janus artifacts.  In general, our approach leads to fast 3D generation and state-of-the-art single-image 3D reconstruction results. 

\setlength{\tabcolsep}{4pt}
\begin{wraptable}[11]{R}{8cm}

\centering
\resizebox{8cm}{!}{
\begin{tabular}{lccccccc@{}}
\toprule
\#Views & \FID & \CLIP & \PSNR & \SSIM & \LPIPS & \CD     \\
\midrule
4 (Ours) & 35.16 & 0.888 & 21.798 & 0.852 & 0.150 & 0.0459 \\
\midrule
1 & 70.59 & 0.788 & 17.560 & 0.832 & 0.304 & 0.0775 \\
2 & 47.69 & 0.896 & 20.965 & 0.851 & 0.167 & 0.0544 \\
6 & 39.11 & 0.899 & 21.545 & 0.861 & 0.148 & 0.0454 \\
\midrule
\textit{w.o} Novel & 102.00 & 0.801 & 17.772 & 0.838 & 0.289 & 0.185 \\
\midrule
\textit{w.o} Plucker & 43.31 & 0.883 & 20.930 & 0.842& 0.185 & 0.505\\
\bottomrule
\end{tabular}
}
\caption{Ablation on GSO dataset (DMV3D-S).}
\label{tab:dmv3d-ablation}
\end{wraptable}

\para{Text-to-3D.}
We also evaluate our text-to-3D generation results and compare with 3D diffusion models Shap-E \citep{shape} and Point-E \citep{pointe}, that are also category-agnostic and support fast direct inference.
For this experiment, we use Shap-E's 50 text prompts for the generation, and evaluate the results with CLIP precisions using two different ViT models, shown in Table.~\ref{tab:dmv3d-text23d}.
From the table, we can see that our model achieves the best precision.
We also show qualitative results in Fig.~\ref{fig:dmv3d-text23d}, in which our results clearly contain more geometry and appearance details and look more realistic than the compared ones.

\begin{figure}[t]
\centering
\includegraphics[width=0.92\textwidth]{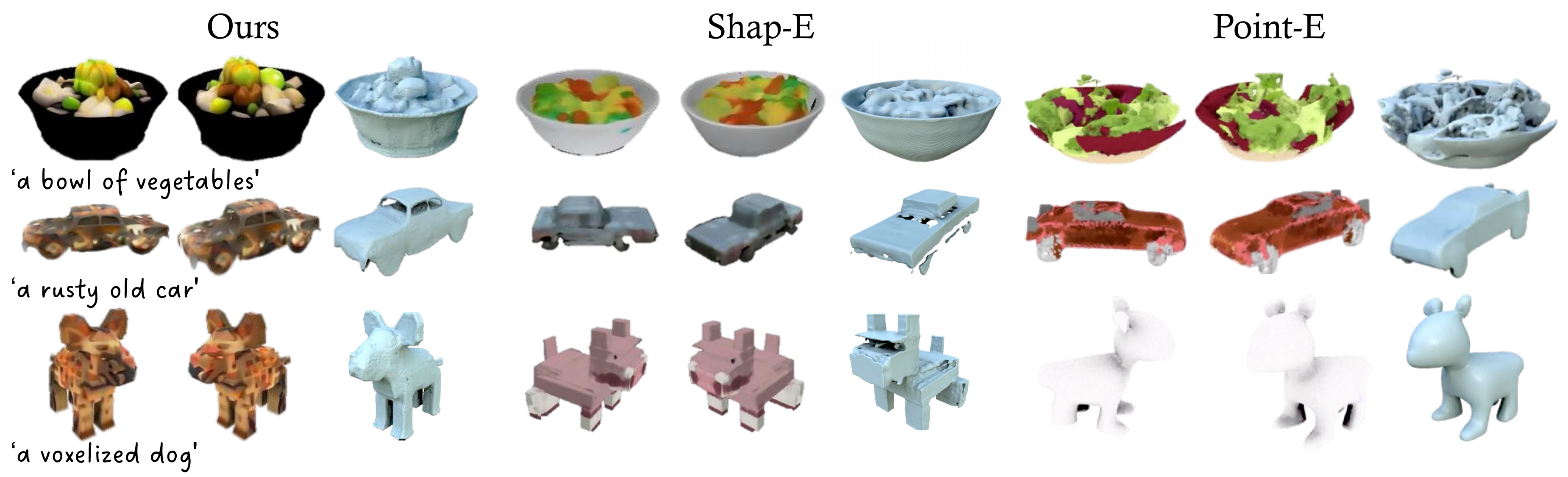}
\caption{\textbf{Qualitative comparison on Text-to-3D .} }
\label{fig:dmv3d-text23d}

\end{figure}

\subsubsection{Analysis, ablation, and application}
\label{sec:dmv3d-exp:ablation}
We analyze our image-conditioned model and verify our design choices using our small model architecture for better energy-efficiency.

\para{\#Views.} We show quantitative and qualitative comparisons of our models trained with different numbers (1, 2, 4, 6) of input views in Tab.~\ref{tab:dmv3d-ablation}.
We can see that our model consistently achieves better quality when using more images, benefiting from capturing more shape and appearance information. 
However, the performance improvement of 6 views over four views is marginal, where some metrics (like PSNR) from the 4-view model is even better.
We therefore use four views as the default setting to generate all of our main results.

\para{Multiple inference generation.} Similar to other DMs, our model can generate various instances from the same input image with different random seeds as shown in Fig.~\ref{fig:dmv3d-teaser}, demonstrating the diversity of our generation results.
In general, we find the multiple inference results can all reproduce the frontal input view while containing varying shape and appearance in the unseen back side.

\para{Input sources.} Our model is category-agnostic and generally works on various input sources as shown in many previous figures.
We show additional results in Fig.~\ref{fig:dmv3d-abl_more} with various inputs, out of our training domains, including synthetic rendering, real capture, and generated images. Our method can robustly reconstruct the geometry and appearance of all cases.

\para{Training data.}
We compare our models trained w/ and w.o the real MVImgNet dataset on two challenging examples. We can see that the model without MVImgNet can lead to unrealistic flat shapes, showcasing the importance of having diverse training data.

\para{More ablation.} We compare with our ablated models including one trained without the novel-view rendering supervision, and one without the Plucker coordinate view conditioning (using the \textit{adaLN-Zero block} conditioning instead).
We can also see that the novel view rendering supervision is critical for our model. Without it, all quantitative scores drop by a large margin. In general, the novel view supervision is crucial for our model to achieve meaningful 3D generation, avoiding the model to learn a local minima that merely recovers the sparse multi-view images. 
In addition, our design of Plucker coordinate-based camera conditioning is also effective, leading to better quantitative results than the ablated model.

\para{Application.}
The flexibility and generality of our method can potentially enable broad 3D applications. One useful image editing application is to lift any objects in a 2D photo to 3D by segment them (using methods like SAM~\citep{kirillov2023segment}) and reconstruct the 3D model with our method.

\begin{figure}[t]
\centering
\includegraphics[width=1\textwidth]{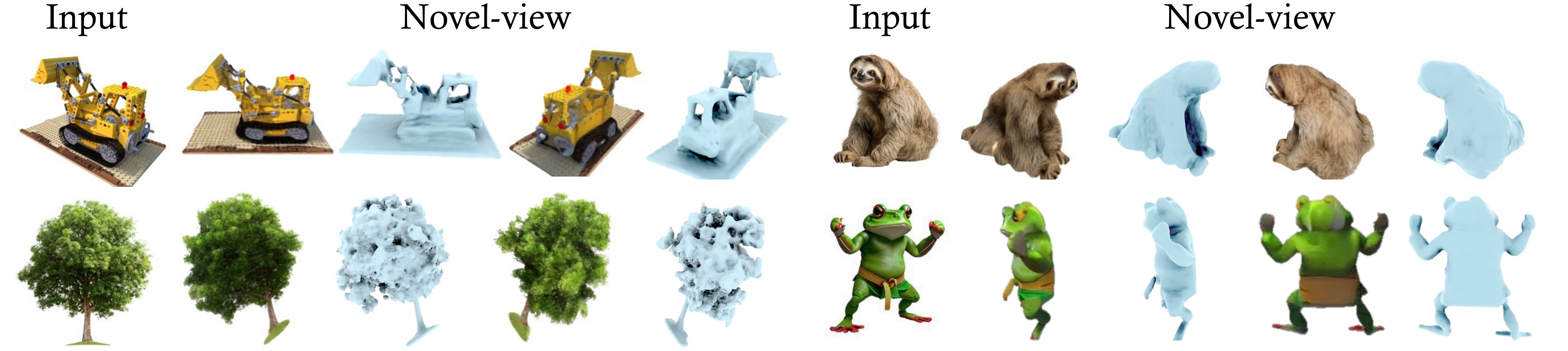}
\caption{\textbf{Robustness on out-of-domain inputs of synthetic, real, and generated images.}} 
\label{fig:dmv3d-abl_more}
\end{figure}

\subsection{Conclusion}
\label{sec:dmv3d-con}
We presented a novel single-stage diffusion model for 3D generation, generating 3D assets by denoising multi-view image diffusion. Our multi-view denoiser is based on a large transformer model, which takes multi-view noisy images to reconstruct a clean triplane NeRF, outputting denoised images through neural rendering.
Our framework generally supports text- and image-conditioning inputs, achieving fast 3D generation via direct diffusion inference without per-asset optimization. Our method outperforms previous 3D diffusion models for text-to-3D generation and achieves state-of-the-art quality on single-view reconstruction on various testing datasets.
Our approach combines 2D diffusion and 3D reconstruction, bridging the gap between 2D and 3D generation and paving the way for future directions on extending 2D diffusion applications for 3D generation.

\paragraph{Ethics Statement.} 
Our generative model is trained on the Objaverse data and MvImgNet data.
The dataset (about 1M) is smaller than the dataset in training 2D diffusion models (about 100M to 1000M).
The lack of data can raise two considerations.
First, it can possibly bias towards the training data distribution.
Secondly, it might not be powerful enough to cover all the diversity of testing images and testing texts.
Our model has certain generalization ability but might not cover as much modes as the 2D diffusion model can.
Given that our model does not have the ability to identify the content that is out of its knowledge, it might introduce to unsatisfying user experience.
Also, our model can possibly leak the training data if the text prompt or image input highly align with some data sample.
This potential leakage raises legal and security considerations, and is shared among all generative data (such as LLM and 2D diffusion models).

\paragraph{Reproducibility Statement.} 
We provide detailed implementation of our training method in the main text.
We will help to resolve any uncertainty of our implementation during review discussions.

\endgroup

\chapter{3D Reconstruction}
\label{chap:3d-reconstruction}

\section{FastMap}
\label{sec:fastmap}
\begingroup
\newcommand{\alg}{\textsc{FastMap}\xspace}
\newcommand{\minihead}[1]{\noindent\textbf{#1}}
\newcommand{\tablestyle}[2]{%
  \centering
  \setlength{\tabcolsep}{#1}
  \renewcommand{\arraystretch}{#2}
  \footnotesize
}
\newcommand{\coloredCell}[3]{\definecolor{mycolor}{HTML}{#2}\tikz[baseline=(char.base)]{\node[fill=mycolor,inner ysep=2pt, inner xsep=5pt, minimum width=#1, align=right] (char) {#3};}}
\newcommand{\cellCD}[3][24pt]{\coloredCell{#1}{#2}{#3}}
\newcommand{\mE}{\mathbf{E}}
\newcommand{\mF}{\mathbf{F}}
\newcommand{\mH}{\mathbf{H}}
\newcommand{\mA}{\mathbf{A}}
\newcommand{\mR}{\mathbf{R}}
\newcommand{\mW}{\mathbf{W}}
\newcommand{\ve}{\bm{e}}
\newcommand{\vg}{\bm{g}}
\newcommand{\vo}{\bm{o}}
\newcommand{\vp}{\bm{p}}
\newcommand{\vs}{\bm{s}}
\newcommand{\vt}{\bm{t}}
\newcommand{\vw}{\bm{w}}
\newcommand{\vx}{\bm{x}}

\label{sec:fastmap-intro}

Large-scale data underpins modern computer vision. Recently, synthetic 3D datasets~\cite{tartanair,dsetspring,pointodyssey,greff2022kubric,urbansyn,deitke2023objaverse}
have been scaled up to provide supervision for diverse tasks such as
Visual-SLAM~\cite{teed2021droid}, 3D point tracking~\cite{karaev2024cotracker,harley2025alltracker}, 
3D asset generation~\cite{shi2023mvdream,li2023instant3d,trellis}, etc. 
Scaling real-world 3D data remains difficult because ground-truth camera poses are rarely available. Many applications such as monocular depth estimation~\cite{ranftl2020towards, bhat2023zoedepth, ke2024repurposing} and learning-based 3D reconstruction~\cite{wang2024dust3r, duisterhof2024mast3r, Wang_2025_CVPR} therefore still rely on pseudo-ground-truth produced by geometry-based structure-from-motion
systems (SfM)
such as COLMAP~\cite{schoenberger2016sfm}. However, COLMAP is slow---processing a scene consisting of thousands of images can take multiple days. Global SfM methods such as GLOMAP~\cite{pan2024glomap} improve upon COLMAP's speed, but still take many hours to converge on large scenes.  Efficiently scaling learning-based systems to more training data requires a fast and high-quality ground-truth annotator.

\begin{figure}[!t] %
\centering
\includegraphics[width=0.7\columnwidth]{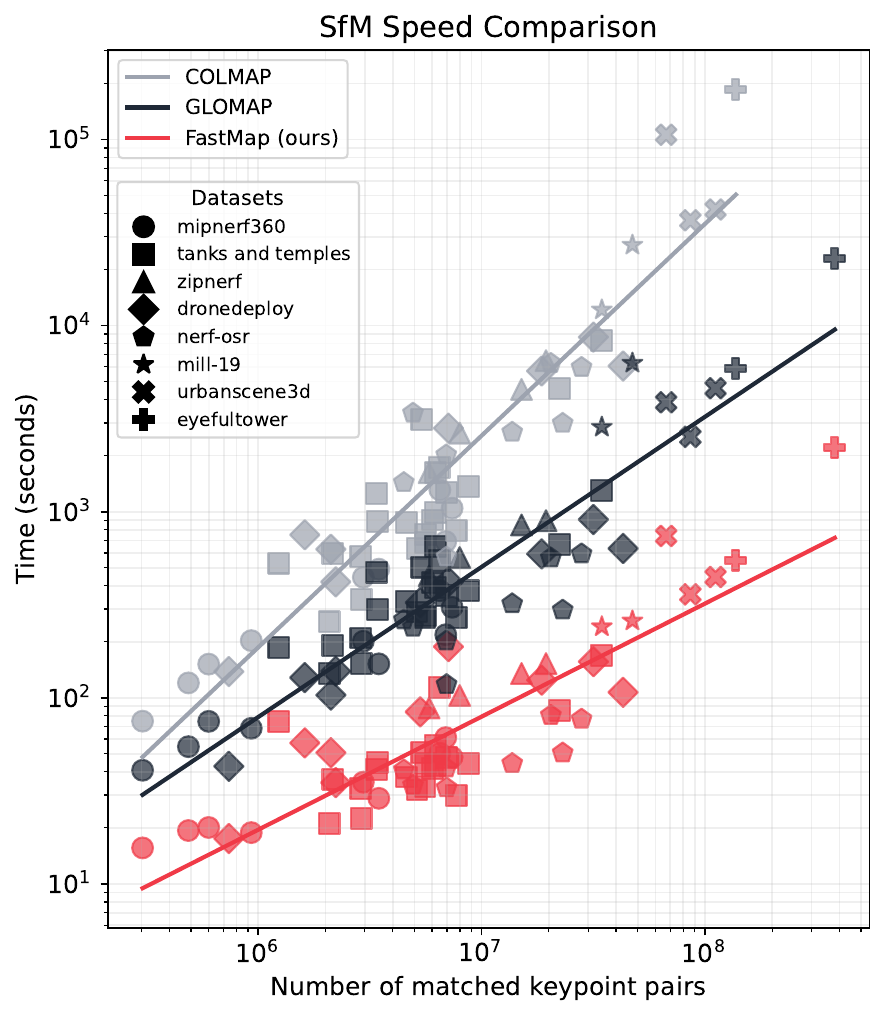}
\caption{
    Timing of \textcolor{red}{\alg} compared to \textcolor{gray}{COLMAP} %
    and GLOMAP~%
    (all with GPU acceleration on a single A6000) on scenes from eight datasets, %
    excluding the matching stage for all methods. Note the \textbf{logarithmic} time scale. Lines represent a least squares power function fit to timing across multiple datasets, as a function of the number of matched keypoint pairs.
}
\label{fig:fastmap-timing}

\end{figure}
A typical SfM pipeline spends most of its time on optimization. COLMAP~\cite{schoenberger2016sfm} registers one image at a time and runs global bundle adjustment every few images to reduce drift. GLOMAP~\cite{pan2024glomap} estimates global translation by optimizing camera centers and 3D points from random initializations, followed by a final bundle adjustment. These optimization problems are nonlinear~\cite{triggs2000bundle} and are commonly solved with quasi second-order methods such as Levenberg-Marquardt (LM)~\cite{Agarwal_Ceres_Solver_2022}. 
LM is a trust-region variant of the Gauss-Newton method, which estimates the Hessian with the residual Jacobian. While several techniques have been adopted to speed up LM optimization, such as the Schur complement trick and sparse Cholesky decomposition, each iteration still takes long wall-clock time to compute. %
Adaptive first-order methods~\cite{kingma2014adam} could eliminate the scalability bottleneck and simplify algorithm design, but they remain underexplored in the SfM literature.

This section presents \alg, a global SfM method that relies only on first-order optimization. Two speed bottlenecks arise when switching from second-order to first-order optimizers. First, many SfM sub-problems, such as bundle adjustment~\cite{triggs2000bundle} and global positioning~\cite{pan2024glomap}, jointly optimize camera poses and 3D points. The number of 3D points is typically orders of magnitude larger than the number of image pairs. Second-order methods spend most of their time solving the reduced linear system from the Schur complement, whereas first-order methods require per-step gradient computation that can scale poorly with the number of points. We therefore design \alg so that all optimization steps have per-iteration complexity independent of the number of 3D points.

The second bottleneck is implementation. While it is straightforward to implement gradient descent 
with modern deep learning Autograd engines~\cite{paszke2019pytorch}, we find that it leads to sub-optimal utilization of GPU resources. This is mainly due to the large overhead of kernel launching, unnecessary data movement between global and shared memory,
and improper kernel choice by the library. Instead, we use kernel fusion to perform 
forward and backward steps
in one CUDA kernel, which significantly speeds up each optimization step. 

Experiments on eight datasets show that \alg is up to $10\times$ faster than COLMAP and GLOMAP with GPU-accelerated Ceres~\cite{Agarwal_Ceres_Solver_2022} (\cref{fig:fastmap-timing}), while achieving comparable pose accuracy and novel view synthesis quality. 

In summary, \alg provides the following contributions:
\begin{itemize}
    \item We show that first-order optimization can be used to make a scalable and accurate SfM system.
    \item We design a fully 3D point-free pipeline that is friendly to first-order optimizers.
    \item We show that kernel fusion can significantly speed up gradient computation by eliminating overhead.
\end{itemize}

\subsection{Method}
\label{sec:fastmap-method}

\begin{figure}[!t] %
    \centering
    \includegraphics[width=1.0\columnwidth]{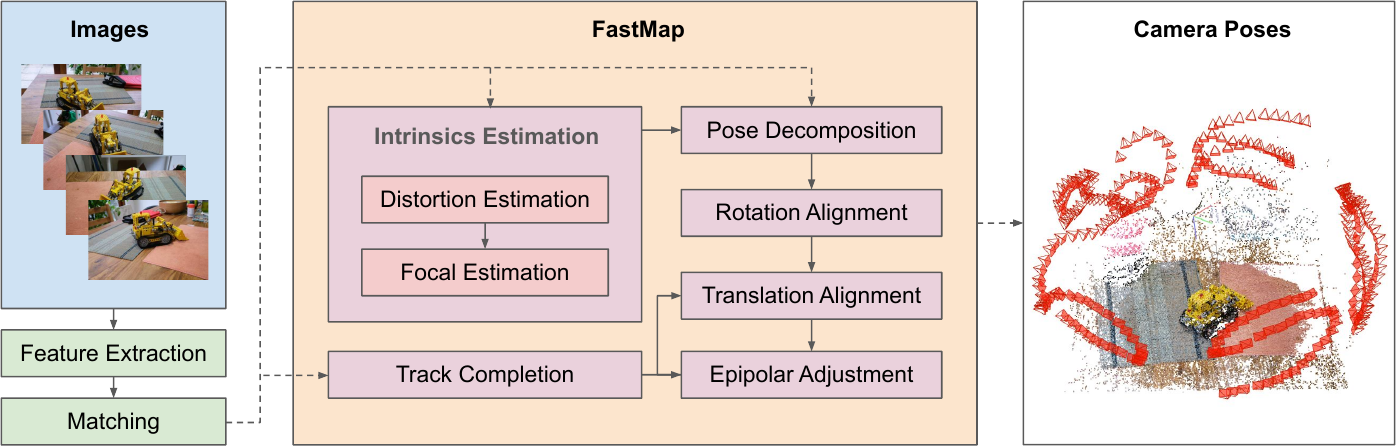}
    \caption{An overview of \alg. Input images are processed using feature extraction and matching. Given the matching results, \alg estimates the intrinsics and extrinsics of the cameras. Finally a sparse point cloud is generated by triangulation.} %
    \label{fig:fastmap-pipeline}

\end{figure}

\minihead{Overview} 
Our proposed \alg framework (\cref{fig:fastmap-pipeline}) consists of multiple stages roughly in sequential order. In this section we introduce the algorithmic details. \cref{sec:fastmap-method-intrinsics} describes how we estimate the distortion parameters and focal lengths for each camera after extracting and matching keypoints. Then, \cref{sec:fastmap-method-first-order-optimization} analyzes the pros and cons of first-order vs second-order optimization, and \cref{sec:fastmap-method-optimization-problems} describes all the optimization problems for global pose estimation and refinement. Finally, we discuss kernel fusion (\cref{sec:fastmap-method-kernel-fusion}) for further speed-up.

\minihead{Matching} \alg's matching stage is identical to that of both COLMAP and GLOMAP: it involves first extracting and matching keypoints from the input images, followed by geometric verification of the resulting image pairs~\cite{schoenberger2016sfm}. The output of the matching stage consists of the 
set of inlier keypoint pairs and either an estimated fundamental matrix $\mF_{ij}$ or a homography matrix $\mH_{ij}$ (the latter if it is consistent with sufficiently many inlier matches) for each image pair $(i,j)$ with enough correspondences.

\subsubsection{Intrinsics Estimation}
\label{sec:fastmap-method-intrinsics}

Accurate intrinsics estimation has a direct impact on the precision of relative pose decomposition, which is critical for later stages. In this section, we describe the algorithms that \alg employs to estimate the focal lengths and distortion from the matching results. 

\minihead{Camera Assumptions} We use a one-parameter radial distortion model. The principal point is fixed to be the center, therefore the only intrinsics parameters to estimate are the distortion parameter and the focal length. We also assume that all the images are taken with a small number of distinct cameras, and that we know which images are from the same camera. In practice, this can be inferred from image resolutions, EXIF tags, file and directory names, etc. 

\minihead{Distortion Estimation} 
\label{sec:fastmap-method-distortion}
We formulate distortion estimation as the problem of finding the distortion parameters that result in the most consistent two-view geometric model for each image pair (e.g., the fundamental matrix estimated from undistorted keypoints has the lowest epipolar error). We do so using the one-parameter division undistortion model~\cite{fitzgibbon2001simultaneous, barreto2005fundamental, erdnuss2021review, henrique2013radial}
\begin{equation}
    x_u = \frac{x_d}{1+\alpha r_d^2} \qquad
    y_u = \frac{y_d}{1+\alpha r_d^2},
\end{equation}
where $(x_d, y_d)$ and $(x_u, y_u)$ are the distorted and undistorted coordinates, respectively, $r_d = \sqrt{x_d^2+y_d^2}$ and $r_u = \sqrt{x_u^2+y_u^2}$, and $\alpha$ is the distortion model parameter. The model can be inverted in closed form to apply distortion to keypoints~\cite{erdnuss2021review}. %
We found this model to be more convenient than the commonly used, but difficult to invert Brown-Conrady distortion model~\cite{duane1971close}.

We use brute-force interval search to estimate the distortion parameter $\alpha$. Given a set of image pairs that share the same $\alpha$, we sample set of candidate values from an interval $[\alpha_{\text{min}}, \alpha_{\text{max}}]$, and evaluate the average epipolar errors for each candidate after undistorting and re-estimating the fundamental matrices based on the sampled $\alpha$ (we ignore all the homography matrices at this stage). This method directly minimizes our objective (epipolar error) and takes into account information from multiple different image pairs, improving robustness to noise. Moreover, each candidate can be evaluated independently, making it highly parallelizable on a GPU.

\paragraph{Hierarchical search} To accelerate the interval search, which scales linearly with the number of candidates, we employ a hierarchical strategy that iteratively shrinks the interval. At each level, we take the left and right candidates adjacent to the current solution as the endpoints of the next interval. The solution at the final level is the distortion estimate.

\paragraph{Multiple cameras} If two images in a pair do not share intrinsics but the distortion parameter of one image is known, we can use the same 1D search to estimate the other image's distortion. We extend the algorithm to multiple cameras by estimating one camera at a time. An image pair is \textit{ready} for a camera if either (1) both images correspond to that camera, or (2) only one image corresponds to that camera and the distortion parameter of the other image is already estimated. We pick the camera with the largest number of ready image pairs, estimate its distortion by averaging epipolar errors across all its ready pairs, and repeat until all cameras are processed.

\paragraph{Importance of undistortion} Incorrect distortion has a direct impact on focal length estimation. Figure~\ref{fig:fastmap-ablation-distortion} visualizes the focal length validity score (smoothed over discrete samples) on two scenes with and without distortion estimation. After keypoints are undistorted (green), the validity score peaks at an accurate FoV estimation. Without distortion estimation (red), the score decreases drastically and the peak shifts away from the correct FoV.

\begin{figure}[!t]
    \centering
    \includegraphics[width=1.0\columnwidth]{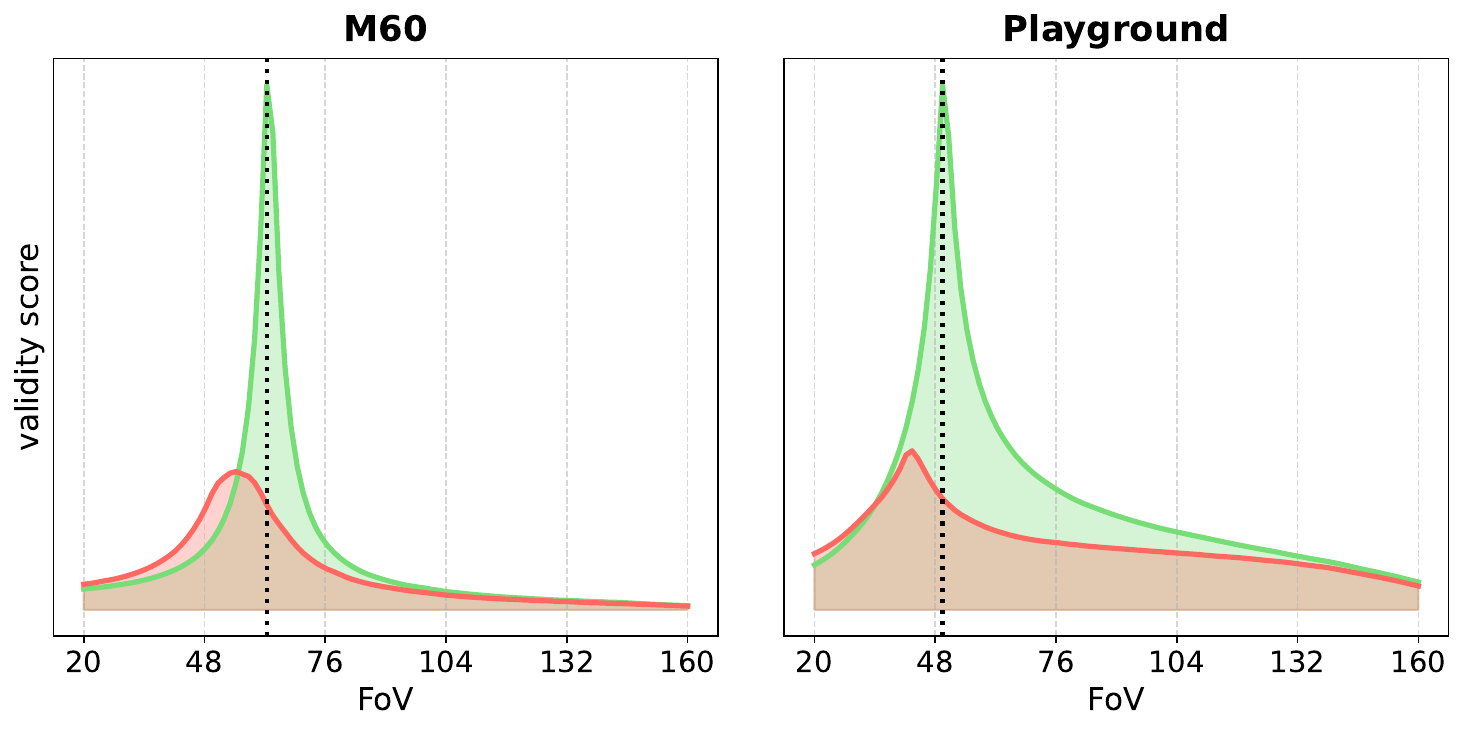}
    \caption{Effect of distortion on focal length estimation. Curves in {\color{green}green} are with undistortion, and curves in {\color{red}red} without. The dotted lines indicate the ground-truth FoVs.}
    \label{fig:fastmap-ablation-distortion}
\end{figure}

\minihead{Focal Length Estimation} 
\label{sec:fastmap-method-focal}
We use the estimated distortion model to undistort all the keypoints, and re-estimate the fundamental and homography matrices. At this point, the only remaining unknown intrinsic parameter for each camera is the focal length.
We estimate the focal length based on the re-estimated fundamental matrices from undistorted keypoints. While this is a well studied problem~\cite{hartley1993extraction, kanatani2000closed, sweeney2015optimizing, barath2017minimal, kocur2024robust}, existing methods are
susceptible to noise or require nonlinear optimization. Instead, we employ an interval search strategy similar to that used for distortion estimation, but with a different objective.

Our method is based on the well-known property that a $3 \times 3$ matrix is an essential matrix if and only if its singular values are such that $\lambda_1=\lambda_2$ and $\lambda_3=0$ (where $\lambda_1 \geq \lambda_2 \geq \lambda_3$)~\cite{faugeras1993three, hartley2003multiple}. %
Given the correct fundamental matrix $\mathbf{F}$ and intrinsics matrix $\mathbf{K}$, the essential matrix $\mathbf{E}=\mathbf{K}^\top \mathbf{F} \mathbf{K}$ should satisfy $\frac{\lambda_1}{\lambda_2}=1$. If all images share the same intrinsics, with a set of fundamental matrices $\{\mathbf{F}_i\}$, we can evaluate the accuracy of a candidate focal length $f$ using the singular value ratio above. Letting $\lambda_1^{(i)} \geq \lambda_2^{(i)} \geq \lambda_3^{(i)}$ be the singular values of $\mathbf{K}^\top \mathbf{F}_i \mathbf{K}$, where $\mathbf{K}$ is a function of $f$, we can measure the validity of $f$ as
\begin{align} \label{eq:fastmap-focal-vote}
    v = \sum_{i} \exp\left(\frac{1-\lambda_1^{(i)}/\lambda_2^{(i)}}{\tau}\right),
\end{align}
where $\tau$ is a temperature hyper-parameter. %
Intuitively, the above formula is close to one when $\lambda_1^{(i)}/\lambda_2^{(i)}$ is close to one, and decreases exponentially as $\lambda_1^{(i)}/\lambda_2^{(i)}$ increases.

We sample a set of candidate focal lengths and evaluate them using Eqn.~\ref{eq:fastmap-focal-vote}. We choose the candidate with the highest value~\eqref{eq:fastmap-focal-vote} as the final estimate. This method can be easily generalized to images with different focal lengths.
After estimating the focal length, we transform the keypoints, fundamental matrices, and homography matrices using the estimated intrinsic matrices so all components are calibrated.

When multiple cameras are present, we apply the same strategy as distortion estimation to estimate the focal length for each camera, but do not use hierarchical sampling since each candidate evaluation is relatively cheap.

\subsubsection{First-order Optimization}
\label{sec:fastmap-method-first-order-optimization}

\minihead{Levenberg-Marquardt} Most of the previous SfM methods use quasi second-order methods such as Levenberg-Marquardt (LM) for optimization. Almost all methods use LM for bundle adjustment~\cite{triggs2000bundle}, and GLOMAP~\cite{pan2024glomap} uses it for global translation alignment. Levenberg-Marquardt is a Gauss-Newton method that first approximates the Hessian with Jacobians of the residuals, and then solves a large linear system to compute the update direction. Techniques like the Schur's complement and sparse Cholesky decomposition are used to improve computational efficiency by exploiting the sparsity of the system~\cite{triggs2000bundle, Agarwal_Ceres_Solver_2022}.

While second-order methods can converge quickly near the optimum, they suffer from poor scalability. Even with the Schur complement method, each step requires solving a reduced linear system whose size grows quadratically with the number of images. In practice, this results in a cubic-time cost in the number of images, which dominates the computation of the update direction. In very large and densely-connected problems, this becomes prohibitively expensive, even with GPU acceleration. To address this, many frameworks employ the preconditioned conjugate gradient (PCG) method, which approximately solves the reduced linear system at a per-iteration cost that scales only quadratically in the number of images. However, PCG slows convergence and introduces implementation complexity.

\minihead{First-order Optimization} On the other hand, first-order optimization methods are prevalent in other fields of computer vision, thanks to the success of deep learning. Optimizing a neural network with a large number of parameters is only tractable with first-order methods, and many adaptive gradient methods exist that  accelerate the naive gradient descent.
In this paper, we investigate the use of first-order optimization methods in SfM.

\minihead{Efficiency} Unfortunately, first-order methods usually converge much more slowly than Gauss-Newton methods in terms of the decrease in loss at each iteration (i.e., they require more iterations). The key to the success of using first-order optimization in SfM is to make the computation of each step as fast as possible. We identify the two most important speed bottlenecks:
\begin{enumerate}
    \item \textit{3D points}:
    One of the most important components of a typical SfM pipeline is bundle adjustment~\cite{triggs2000bundle}, which jointly optimizes camera poses and 3D points. In practice, the number of 3D points is usually orders-of-magnitude larger than the number of image pairs. To make Gauss-Newton tractable in this setting, methods employ the Schur complement method~\cite{triggs2000bundle} to first eliminate the 3D point variables to form a reduced system independent of the number of points, and then reciver the 3D points via back-substitution. In this stage, solving the reduced system usually dominates the compute time. However, if we switch to gradient descent, the main bottleneck becomes computing the forward and backward passes for the 3D point parameters. To address this, we design all the optimization problems in our method (\cref{sec:fastmap-method-optimization-problems}) so that at each iteration, the computation complexity is independent of the number of 3D points.
    \item \textit{Kernel implementation}:
    The optimization problems that \alg solves can be easily implemented using modern Autograd frameworks such as PyTorch. These libraries are highly optimized for large-scale deep learning applications that involve a lot of linear operations on large tensors. In our case, most of the operations are relatively small (e.g., $3\times 3$ matrix multiplication), and naively implementing everything with high-level PyTorch optimizations induces significant kernel launching and data movement overhead. We solve this problem through kernel fusion (\cref{sec:fastmap-method-kernel-fusion}), which eliminates most of the overhead and increases GPU utilization.
\end{enumerate}
\noindent  Section~\ref{sec:fastmap-method-optimization-problems} introduces all the optimization problems present in our method. They are chosen such that the computational complexity of each step is independent of the number of 3D points. In \cref{sec:fastmap-method-kernel-fusion}, we describe the kernel fusion technique to fully exploit the power of GPUs for even more speedup.

\subsubsection{Optimization Formulations}
\label{sec:fastmap-method-optimization-problems}
Here, we introduce the optimization-based formulations of estimating and finetuning the global poses.

\minihead{Global Rotation} 
With the estimated intrinsics, we can decompose the fundamental and homography matrices into relative rotations and translations~\cite{hartley2003multiple, malis2007deeper}. Given the set of image pairs $\mathcal{P}=\{(i, j)\}$ and the corresponding relative rotation matrices $\{\mR^{i\rightarrow j}\}_{(i, j)\in \mathcal{P}}$, \alg next estimates the world-to-camera global rotation $\mathbf{R}^{(i)}$ matrix for each image $i$. We formulate this as an optimization of a loss defined over all image pairs $\mathcal{P}=\{(i, j)\}$ %
\begin{equation} \label{eq:fastmap-rotation-objective}
    \mathcal{L}_{\text{R}} = \frac{1}{|\mathcal{P}|}\sum_{(i, j)\in\mathcal{P}} d( \mR^{(j)}, \mR^{i\rightarrow j} \mR^{(i)}),
\end{equation}
where $d(\cdot, \cdot)$ is the geodesic distance between rotations %
\begin{equation}
    d(\mR, \mR^\prime) = \cos^{-1} \left( \frac{\text{Tr}(\mR^T \mR^\prime) - 1}{2} \right).
\end{equation}
For simplicity, we parameterize the global rotation matrices $\mathbf{R}_i$ using a differentiable 6D representation~\cite{zhou2019continuity}.

Unfortunately, directly optimizing the above objective from random initialization of $\mathbf{R}_i$ is prone to local minima. We use a slightly modified version of the method proposed by \citet{martinec2007robust} to obtain a good initialization. The basic idea of the method is that although the column vectors in a rotation matrix are constrained by orthogonality, each column vector alone is only subject to a unit length constraint. If we consider one column at a time, we can formulate the optimization as as a least squares problem and solve it using SVD.

\paragraph{Initialization details}
Denote the image set as $\mathcal{I}$ and image pairs as $\mathcal{P}=\{(i, j)\}_{(i,j)\in\mathcal{I}\times\mathcal{I}}$ with relative rotations $\mR^{i\rightarrow j}$. We seek global rotations $\{\mR_i\}_{i\in\mathcal{I}}$ by minimizing the L2 objective
\begin{equation}
    \mathcal{L} = \sum_{(i, j)\in\mathcal{P}}\left\Vert \mR^{(j)} - \mR^{i\rightarrow j} \mR^{(i)} \right\Vert^2.
\end{equation}
Because $\mR^{(i)}\in\mathrm{SO}(3)$, we solve for columns sequentially. Let $\mA_{*,k}$ denote column $k$ of $\mA$. Splitting the sum yields
\begin{align}
    \mathcal{L} = \sum_{(i, j)\in\mathcal{P}} \sum_{k=1,2,3} \left\Vert \mR^{(j)}_{*,k} - \mR^{i\rightarrow j} \mR^{(i)}_{*,k} \right\Vert^2.
\end{align}
For the first column, we ignore unit-length constraints and solve a least-squares problem
\begin{align}
    \mathcal{L}^{(1)} = \sum_{(i, j)\in\mathcal{P}}  \left\Vert \mR^{(j)}_{*,1} - \mR^{i\rightarrow j} \mR^{(i)}_{*,1} \right\Vert^2,
\end{align}
then normalize each $\mR^{(i)}_{*,1}$ to unit length. The second column is estimated similarly with an orthogonality penalty,
\begin{equation}
    \begin{split}
        \mathcal{L}^{(2)} = &\frac{1}{|\mathcal{P}|} \sum_{(i, j)\in\mathcal{P}}  \left\Vert \mR^{(j)}_{*,2} - \mR^{i\rightarrow j} \mR^{(i)}_{*,2} \right\Vert^2   \\
        &+ \frac{1}{|\mathcal{I}|} \sum_{i\in\mathcal{I}} \left\Vert {\mR^{(i)}_{*,1}}^\top  \mR^{(i)}_{*,2} \right\Vert^2,
    \end{split}
\end{equation}
followed by normalization and a Gram-Schmidt step. The third column is given by $\mR^{(i)}_{*,3} = \mR^{(i)}_{*,1} \times \mR^{(i)}_{*,2}$.

\paragraph{Filtering}
To improve robustness, we filter image pairs whose number of inlier point pairs falls below a threshold. We start from a large threshold, halve it if it disconnects the image graph, and iterate until all images are connected or a minimal threshold is reached. Images that remain disconnected at the minimal threshold are treated as outliers and ignored in rotation alignment and later stages.

\subsubsection{Tracks and Track Completion}
\label{sec:fastmap-method-tracks}
Consider the graph where 2D keypoints are nodes and pairwise edges denote matches. A 3D scene point observed in $m$ images should yield a complete subgraph, but in practice low recall makes it far from complete. We make up for low matching recall by \emph{track completion}. A \emph{track} is a connected component in the keypoint connectivity graph and implies a shared 3D point. Tracks are heavily used in SfM for imposing constraints (e.g., bundle adjustment). \alg avoids bundle adjustment and explicit 3D point data structures; instead we convert tracks into additional matches by augmenting the original matching results with all pairwise combinations of keypoints in each track. These additional point pairs are introduced after global rotation alignment and used in translation alignment and epipolar adjustment.

\minihead{Global Translation} 
After global rotation alignment, we re-estimate the relative translations between image pairs. The next step is to utilize these relative translations to estimate the 3D coordinates of the camera centers in a common (world) coordinate frame, up to a similarity transformation. This step is usually called \textit{translation averaging}. It is notoriously susceptible to noise, and making it robust to all kinds of scenarios is the focus of many global SfM papers~\cite{jiang2013global, wilson2014robust, ozyesil2015robust, zhuang2018baseline, pan2024glomap}. However, this is not the focus of our paper and so we choose a relatively simple method to tackle this problem, which we find to be sufficient for most of the scenes we evaluate on. A more robust design for this stage is left for future work.

\paragraph{Relative translation}
We re-estimate unit relative translation vectors using the improved global rotations. We sample candidates on the unit sphere, evaluate the mean epipolar error of each candidate, and select the one with the lowest error. This also leverages new point pairs introduced by track completion, which can enable translation estimation for image pairs that previously had no matches.

Given %
world-to-camera rotations $\{\mR_i\}_{1\leq i\leq N}$ for $N$ images and unit-length relative translations $\{\vt^{i\rightarrow j}\}_{(i,j)\in\mathcal{P}}$ for a set of image pairs $\mathcal{P}$, we compute the normalized vector from the camera centers of image $i$ to $j$ in world coordinates
\begin{equation}
\vo^{i\rightarrow j} = -\mR_j^\top \vt^{i\rightarrow j}.
\end{equation}

We estimate the camera locations $\{\vo_i\}_{1\leq i\leq N}$ in the world frame by minimizing the error between the normalized relative translation $\frac{\vo_j - \vo_i}{\|\vo_j - \vo_i\|_2}$ and the target $\vo^{i\rightarrow j}$ above with gradient descent 
\begin{equation}\label{eq:fastmap-global-translation-objective}
    \mathcal{L}_{\text{t}} = \frac{1}{|\mathcal{P}|} \sum_{(i,j)\in\mathcal{P}} \left\| \frac{\vo_j - \vo_i}{\|\vo_j - \vo_i\|_2} - \vo^{i\rightarrow j} \right\|_1.
\end{equation}
Unlike global rotation optimization, this objective can often be effectively optimized from a random initialization. %
GLOMAP~\cite{pan2024glomap} makes a similar observation, but it optimizes poses and 3D points jointly, and is much more computationally expensive.

Although random initialization works surprisingly well for the objective in Eqn.~\ref{eq:fastmap-global-translation-objective}, it occasionally produces a small number of outliers. To deal with this, we perform multiple independent runs from different random initializations and merge the solutions as the initialization for the final optimization loop.

\paragraph{Multiple initializations}
Different random initializations can produce different outliers. We align the resulting solutions by centering them at the origin and rescaling to unit average norm, then pick for each image the solution with the lowest average loss. The merged solution serves as the initialization for the final optimization loop.

\minihead{Epipolar Adjustment} 
A typical SfM pipeline relies on \emph{bundle adjustment} (BA)~\cite{triggs2000bundle} to jointly refine the camera poses and inferred 3D points. Directly implementing BA with first-order optimizers is computationally expensive when the number of points is large. 
Instead, we refine the poses from previous stages using \emph{re-weighting epipolar adjustment}, an optimization method~\citep{rodriguez2011reduced} for which the computational complexity in each iteration is linear only in the number of image pairs, not in the number of points.

\paragraph{L2 objective derivation}
For the L2 epipolar objective, each error term is linear in the essential matrix, which allows a compact quadratic form. Let $\vw_{nm}=\text{flatten}(\tilde{\vx}^{(2)}_{nm}\tilde{\vx}^{(1)\top}_{nm})\in\mathbb{R}^9$ and $\ve_{n}=\text{flatten}(\mE_{n})\in\mathbb{R}^9$. Then
\begin{subequations}
    \begin{align}
        \mathcal{L} &= \frac{1}{Z}\sum_{n=1}^{|\mathcal{P}|}\sum_{m=1}^{|\tilde{\mathcal{Q}}_n|} ( \tilde{\vx}^{(2)\top}_{nm}\mE_{n}\tilde{\vx}^{(1)}_{nm} )^2 \\
        &= \frac{1}{Z}\sum_{n=1}^{|\mathcal{P}|}\sum_{m=1}^{|\tilde{\mathcal{Q}}_n|} ( \vw^{\top}_{nm}\ve_{n})^2 \\
        &= \frac{2}{Z}\sum_{n=1}^{|\mathcal{P}|} \ve_{n}^\top \mW_n \ve_{n},
    \end{align}
\end{subequations}
where $\mW_n=\sum_{m=1}^{|\tilde{\mathcal{Q}}_n|}  \vw_{nm}\vw^{\top}_{nm}\in\mathbb{R}^{9\times9}$.

In relative translation re-estimation, we obtain a set of image pairs with number of inliers above some threshold. We denote the set of such image pairs as $\mathcal{P}=\{(i_n, j_n)\}_{1\leq n\leq |\mathcal{P}|}$ (abusing the notation for the original set of images above), where $i_n$ and $j_n$ are the indices of the first and second images in the pair. %
For an image pair $(i_n, j_n)\in\mathcal{P}$, we represent the set of point pairs as $\mathcal{Q}_n=\{(\vx^{(1)}_{nm}, \vx^{(2)}_{nm})\in\mathbb{R}^2\times\mathbb{R}^2\}_{1\leq m\leq |\mathcal{Q}_n|}$, and let  $\tilde{\mathcal{Q}}_n=\{(\tilde{\vx}^{(1)}_{nm}, \tilde{\vx}^{(2)}_{nm})\in\mathbb{P}^2\times\mathbb{P}^2\}_{1\leq m\leq |\tilde{\mathcal{Q}}_n|}$ be the set of point pairs in normalized homogeneous coordinates.

\minihead{Sparse Reconstruction}
After pose estimation, we triangulate matched keypoint pairs from track completion to obtain a sparse point cloud. 3D points corresponding to the same track are merged by averaging. We remove outliers by computing the re-projection error for each 2D keypoint, discarding those with large errors. A 3D point is dropped if the number of inlier keypoints in the track is smaller than three, and we also filter out points with small maximal triangulation angles.

Using estimated initializations from the previous stages, we optimize over the world-to-camera global rotations and translations to minimize the absolute epipolar error
\begin{equation} \label{eq:fastmap-epipolar-objective}
    \mathcal{L}_{\text{e}} = \frac{1}{Z} \sum_{n=1}^{|\mathcal{P}|}\sum_{m=1}^{|\tilde{\mathcal{Q}}_n|} \lvert \tilde{\vx}^{(2)\top}_{nm}\mE_n\tilde{\vx}^{(1)}_{nm} \rvert,
\end{equation}
where $Z=\sum_{n=1}^{|\mathcal{P}|}|\tilde{\mathcal{Q}}_n|$ is the total number of point pairs, and $\mE_n$ is the essential matrix computed from the global rotations and translations for images $i_n$ and $j_n$.

Evaluating Eqn.~\ref{eq:fastmap-epipolar-objective} for every iteration is expensive because it involves every point pair. However, if we replace the cost terms with the L2 loss (as in \citet{rodriguez2011gea}), the overall objective can be re-organized to aggregate terms that involve point pairs shared by the same image pair into a compact quadratic form%
\begin{equation} \label{eq:fastmap-epipolar-objective-l2}
    \mathcal{L}_{\text{e}}= \frac{1}{Z}\sum_{n=1}^{|\mathcal{P}|}\sum_{m=1}^{|\tilde{\mathcal{Q}}_n|} ( \tilde{\vx}^{(2)\top}_{nm}\mE_n\tilde{\vx}^{(1)}_{nm} )^2=\frac{2}{Z}\sum_{n=1}^{|\mathcal{P}|} \ve_n^\top \mW_n \ve_n,
\end{equation}
where  $\ve_n=\text{flatten}(\mE_n)\in\mathbb{R}^9$, and $\mW_n\in\mathbb{R}^{9\times 9}$ is a matrix computed from all the point pairs in $\tilde{\mathcal{Q}}_n$. Note that only $\ve_n$ is a function of the parameters to be optimized. The matrices $\mW_n$ can be pre-computed for each image pair before optimization. With the precomputed $\mW_n$, the cost of each optimization step is linear in the number of image pairs. 

The expedience of Eqn.~\ref{eq:fastmap-epipolar-objective-l2} comes with the side effect that the L2 loss is sensitive to outliers. We propose to robustify the loss function
but still preserve the benefit of pre-computation with \emph{iterative re-weighted least squares (IRLS)}. The intuition is that if we have an initialization close to the optimum, the L1 loss, which is more robust to outliers, can be approximated by a weighted L2 loss. In other words, for some differentiable computed scalar value $z$, we have $ z^2\approx z^2 / |\hat{z}|$, where $\hat{z}$ is the value of $z$ at initialization. In our case, after global translation alignment, we already have a good initialization of global poses, so we can compute the absolute epipolar error $|\hat{\epsilon}_{nm}|$ for each point pair, and use it to weight the L2 loss used above to get an approximate robust L1 loss
\begin{equation}
        \hat{\mathcal{L}}_{\text{e}}
        = \frac{1}{Z}\sum_{n=1}^{|\mathcal{P}|}\sum_{m=1}^{|\tilde{\mathcal{Q}}_n|} \frac{( \tilde{\vx}^{(2)\top}_{nm}\mE_n\tilde{\vx}^{(1)}_{nm} )^2}{|\hat{\epsilon}_{nm}|}
        = \frac{2}{Z}\sum_{n=1}^{|\mathcal{P}|} \ve_{n}^\top \hat{\mW}_n \ve_{n},
\end{equation}
where $\hat{\mW}_n$ is similar to $\mW_n$ in Eqn.~\ref{eq:fastmap-epipolar-objective-l2}, but computed from $\tilde{\mathcal{Q}}_n$ weighted by $|\hat{\epsilon}_{nm}|$.

After a round of optimization, we can better approximate the L1 loss by re-computing the weights and then doing another optimization loop.
To further reduce the impact of outliers, we periodically filter out point pairs with large epipolar error. We start from a relatively large initial threshold and gradually decrease it to a pre-determined minimum.

The above optimization problem only involves camera poses. We can also optimize the focal lengths by incorporating them into the computation of the essential matrices (in this case, the results are actually fundamental matrices).

\subsubsection{Kernel Fusion}
\label{sec:fastmap-method-kernel-fusion}

A standard way to implement gradient descent in the above algorithms is to use the Autograd  feature in modern deep learning libraries such as PyTorch~\cite{paszke2019pytorch}. However, the tensors in our setting are mostly batches of small matrices and vectors of shapes $B\times 3\times 3$ and $B\times 3$, where $B$ is the number of image pairs, and a naive PyTorch implementation introduces some significant bottlenecks:
\begin{enumerate}
    \item \textit{Kernel Launching Overhead:} When the scene is relatively small (i.e., the number of image pairs is small), the kernel launching overhead dominates the running time and is limited by the CPU speed.
    \item \textit{Data Movement:} Computing the objective and gradients involves a series of small operations (e.g., $3\times 3$ matrix multiplication and cross product). Each operation involves reading the data from the high-latency global memory to the fast on-chip shared memory and writing back when finished. his leads to substantial inefficiency and makes the computation predominantly memory-bound.
    \item \textit{Kernel Design:} PyTorch kernels are optimized for deep learning workload, which usually consists of linear operations on large tensors. These kernels can lead to sub-optimal performance when applied to tensors without the assumed shapes.
\end{enumerate}

\noindent To address this problem, we fuse all the operations for computing the gradients, including forward and backward passes, into one single custom CUDA kernel. This introduces challenges in shared-memory management when the computation involves many small operations. However, since almost all input and intermediate tensors have one of the three shapes ($B\times 3\times 3$, $B\times 3$, or simply $B$), we can efficiently reuse shared-memory slots to limit the reduction in thread occupancy. In \cref{tab:fastmap-kernel_fusion} we show that the fused kernel can be more than two orders-of-magnitude faster than a naive PyTorch implementation under different scene sizes and hardware settings. Please refer to the ablation study (\cref{sec:fastmap-ablation}) for a more detailed analysis.

\subsection{Experiments}
\label{sec:fastmap-exp}

\subsubsection{Setup}
We compare \alg with two state-of-the-art methods:
COLMAP~\cite{schoenberger2016sfm} 
(commit \href{https://github.com/colmap/colmap/commit/c4a3b308bddf391b4e7c62e835720f83c13aea8b}{\small \texttt{c4a3b30}})
and GLOMAP~\cite{pan2024glomap} 
(commit \href{https://github.com/colmap/glomap/commit/01060b4be509902ea9aaefaae982a2cb941ae5c3}{\small \texttt{01060b4}}), both with GPU-accelerated Ceres~\cite{Agarwal_Ceres_Solver_2022} solver enabled.
For all three methods, we use the COLMAP image matching system. We 
use shared intrinsics if all images in a scene are from the same camera, and leave all other hyper-parameters at their default values in COLMAP and GLOMAP. We run all three methods on a machine with a single A6000 (Ampere) GPU and an AMD EPYC 9274F CPU (Zen 4) with 24 cores / 48 threads. By default, \alg uses 2 CPU threads, whereas COLMAP and GLOMAP use all 48 threads.

\begin{table*}[t]
    \centering
    \resizebox{1.0\linewidth}{!}{
        
\newcommand{\ratio}[1]{\textcolor{teal}{\textsubscript{$\times$#1}}}

\tablestyle{4pt}{1.1}
\begin{tabular}{l r r r | r r r | r r }
\toprule
  &  &
\multicolumn{2}{c}{\scriptsize \alg (\textbf{1}G+\textbf{2}C) } & \multicolumn{3}{c}{\scriptsize GLOMAP} & \multicolumn{2}{c}{\scriptsize COLMAP} \\
\cmidrule(lr){3-4} \cmidrule(lr){5-7} \cmidrule(lr){8-9}
  & n\_imgs &
\multicolumn{1}{c}{\scriptsize w/ cuda} &
\multicolumn{1}{c}{\scriptsize w/o cuda } &
\multicolumn{1}{c}{\scriptsize \textbf{1}G+\textbf{48}C} &
\multicolumn{1}{c}{\scriptsize \textbf{1}G+\textbf{12}C} &
\multicolumn{1}{c}{\scriptsize \textbf{48}C} &
\multicolumn{1}{c}{\scriptsize \textbf{1}G+\textbf{48}C} &
\multicolumn{1}{c}{\scriptsize \textbf{48}C}\\
\midrule
     z\_alameda  & 1734 & 134 : \ratio{1.0} & 917 : \ratio{6.8} & 848 : \ratio{6.3} & 934 : \ratio{6.9} & 3805 : \ratio{28.2} & 4541 : \ratio{33.6} & 24641 : \ratio{182.5} \\
      z\_berlin  & 1511 & 152 : \ratio{1.0} & 545 : \ratio{3.6} & 893 : \ratio{5.9} & 1015 : \ratio{6.7} & 1802 : \ratio{11.8} & 6478 : \ratio{42.4} & 24648 : \ratio{161.4} \\
      z\_london  & 1874 & 102 : \ratio{1.0} & 556 : \ratio{5.4} & 566 : \ratio{5.5} & 669 : \ratio{6.5} & 2092 : \ratio{20.3} & 2643 : \ratio{25.7} & 19238 : \ratio{187.1} \\
         z\_nyc  &  990 & 88 : \ratio{1.0} & 338 : \ratio{3.8} & 451 : \ratio{5.1} & 487 : \ratio{5.5} & 921 : \ratio{10.5} & 1618 : \ratio{18.4} & 1988 : \ratio{22.6} \\
mill19\_building  & 1920 & 258 : \ratio{1.0} & 1366 : \ratio{5.3} & 6289 : \ratio{24.3} & 7792 : \ratio{30.1} & 38428 : \ratio{148.4} & 27080 : \ratio{104.6} & 152839 : \ratio{590.1} \\
 mill19\_rubble  & 1657 & 240 : \ratio{1.0} & 789 : \ratio{3.3} & 2849 : \ratio{11.8} & 2466 : \ratio{10.2} & 11571 : \ratio{48.0} & 12153 : \ratio{50.4} & 64987 : \ratio{269.8} \\
   urbn\_Campus  & 5871 & 740 : \ratio{1.0} & 3009 : \ratio{4.1} & 3869 : \ratio{5.2} & 4175 : \ratio{5.6} & 21916 : \ratio{29.6} & 106055 : \ratio{143.2} & 349490 : \ratio{472.0} \\
  urbn\_Sci-Art  & 3019 & 445 : \ratio{1.0} & 1760 : \ratio{4.0} & 4601 : \ratio{10.3} & 5712 : \ratio{12.8} & 28824 : \ratio{64.7} & 42032 : \ratio{94.4} & 286454 : \ratio{643.4} \\
 eft\_apartment  & 3804 & 549 : \ratio{1.0} & 1003 : \ratio{1.8} & 5905 : \ratio{10.8} & 8341 : \ratio{15.2} & 124310 : \ratio{226.3} & 185361 : \ratio{337.5} & \multicolumn{1}{c}{timeout} \\
   eft\_kitchen  & 6042 & 2202 : \ratio{1.0} & 6796 : \ratio{3.1} & 22884 : \ratio{10.4} & 34287 : \ratio{15.6} & \multicolumn{1}{c}{timeout} & \multicolumn{1}{c}{timeout} & \multicolumn{1}{c}{timeout} \\
\bottomrule
\end{tabular}

    }
    \caption{Detailed system runtime comparisons (seconds:\textcolor{teal}{speed\_ratio}) with different GPU (G) and CPU threads (C) configurations. Despite the CUDA-accelerated Ceres solver for bundle adjustment, a significant part of the GLOMAP and COLMAP pipeline workload is still CPU-bound, and having at least 12 threads is necessary for higher speed. \alg performs data structure marshaling on GPU with non-trivial tensor indexing and consumes less CPU resource.}
    \label{tab:fastmap-detailed_timing}
\end{table*}

\begin{table*}[t]
    \centering
    \resizebox{\textwidth}{!}{
        \tablestyle{2pt}{1.1}
\resizebox{\textwidth}{!}{%
\begin{tabular}{r r  r@{~}r@{~}r | r@{~}r@{~}r | r@{~}r@{~}r | r@{~}r@{~}r | r@{~}r@{~}r | r@{~}r@{~}r}

\toprule

  &  &  \multicolumn{3}{c}{time (sec)} &
\multicolumn{3}{c}{ATE$\downarrow$} &
\multicolumn{3}{c}{RTA@3$\uparrow$} &
\multicolumn{3}{c}{AUC-R\&T @ 3 $\uparrow$} &
\multicolumn{3}{c}{RTA@1$\uparrow$} &
\multicolumn{3}{c}{AUC-R\&T @ 1 $\uparrow$} \\

\cmidrule(lr){3-5} \cmidrule(lr){6-8} \cmidrule(lr){9-11}
\cmidrule(lr){12-14} \cmidrule(lr){15-17} \cmidrule(lr){18-20}

  & n\_imgs &
\multicolumn{1}{c}{\scriptsize \alg} & \multicolumn{1}{c}{\scriptsize GLOMAP} & \multicolumn{1}{c}{\scriptsize COLMAP} &
\multicolumn{1}{c}{\scriptsize \alg} & \multicolumn{1}{c}{\scriptsize GLOMAP} & \multicolumn{1}{c}{\scriptsize COLMAP} &
\multicolumn{1}{c}{\scriptsize \alg} & \multicolumn{1}{c}{\scriptsize GLOMAP} & \multicolumn{1}{c}{\scriptsize COLMAP} &
\multicolumn{1}{c}{\scriptsize \alg} & \multicolumn{1}{c}{\scriptsize GLOMAP} & \multicolumn{1}{c}{\scriptsize COLMAP} &
\multicolumn{1}{c}{\scriptsize \alg} & \multicolumn{1}{c}{\scriptsize GLOMAP} & \multicolumn{1}{c}{\scriptsize COLMAP} &
\multicolumn{1}{c}{\scriptsize \alg} & \multicolumn{1}{c}{\scriptsize GLOMAP} & \multicolumn{1}{c}{\scriptsize COLMAP} \\

\midrule
     mipnerf360  (9) & 215.6 &  \cellCD[24pt]{69a84f}{33} &  \cellCD[24pt]{b7d7a8}{165} &  \cellCD[24pt]{ffffff}{503} &  \cellCD[22pt]{ffffff}{4.2e-4} &  \cellCD[22pt]{69a84f}{3.3e-5} &  \cellCD[22pt]{b7d7a8}{5.8e-5} &  \cellCD[22pt]{b7d7a8}{99.9} &  \cellCD[22pt]{b7d7a8}{100.0} &  \cellCD[22pt]{b7d7a8}{100.0} &  \cellCD[22pt]{b7d7a8}{97.4} &  \cellCD[22pt]{b7d7a8}{98.2} &  \cellCD[22pt]{b7d7a8}{97.2} &  \cellCD[22pt]{b7d7a8}{99.8} &  \cellCD[22pt]{b7d7a8}{100.0} &  \cellCD[22pt]{b7d7a8}{99.7} &  \cellCD[22pt]{b7d7a8}{92.4} &  \cellCD[22pt]{69a84f}{94.6} &  \cellCD[22pt]{b7d7a8}{91.9} \\
  tnt\_advanced  (6) & 337.8 &  \cellCD[24pt]{69a84f}{61} &  \cellCD[24pt]{b7d7a8}{357} &  \cellCD[24pt]{ffffff}{1016} &  \cellCD[22pt]{b7d7a8}{6.4e-3} &  \cellCD[22pt]{ffffff}{1.2e-2} &  \cellCD[22pt]{69a84f}{1.2e-3} &  \cellCD[22pt]{ffffff}{71.4} &  \cellCD[22pt]{b7d7a8}{79.1} &  \cellCD[22pt]{69a84f}{98.5} &  \cellCD[22pt]{ffffff}{42.6} &  \cellCD[22pt]{b7d7a8}{75.3} &  \cellCD[22pt]{69a84f}{94.8} &  \cellCD[22pt]{ffffff}{42.3} &  \cellCD[22pt]{b7d7a8}{77.5} &  \cellCD[22pt]{69a84f}{97.0} &  \cellCD[22pt]{ffffff}{16.7} &  \cellCD[22pt]{b7d7a8}{69.8} &  \cellCD[22pt]{69a84f}{90.0} \\
tnt\_intermediate  (8) & 268.6 &  \cellCD[24pt]{69a84f}{35} &  \cellCD[24pt]{b7d7a8}{314} &  \cellCD[24pt]{ffffff}{833} &  \cellCD[22pt]{b7d7a8}{7.8e-5} &  \cellCD[22pt]{69a84f}{1.9e-5} &  \cellCD[22pt]{ffffff}{2.6e-4} &  \cellCD[22pt]{b7d7a8}{99.9} &  \cellCD[22pt]{b7d7a8}{100.0} &  \cellCD[22pt]{b7d7a8}{99.8} &  \cellCD[22pt]{ffffff}{94.1} &  \cellCD[22pt]{b7d7a8}{99.0} &  \cellCD[22pt]{b7d7a8}{98.9} &  \cellCD[22pt]{b7d7a8}{99.3} &  \cellCD[22pt]{b7d7a8}{99.9} &  \cellCD[22pt]{b7d7a8}{99.5} &  \cellCD[22pt]{ffffff}{83.1} &  \cellCD[22pt]{b7d7a8}{96.9} &  \cellCD[22pt]{b7d7a8}{97.3} \\
  tnt\_training  (7) & 470.1 &  \cellCD[24pt]{69a84f}{63} &  \cellCD[24pt]{b7d7a8}{515} &  \cellCD[24pt]{ffffff}{2751} &  \cellCD[22pt]{b7d7a8}{3.0e-3} &  \cellCD[22pt]{ffffff}{1.1e-2} &  \cellCD[22pt]{69a84f}{3.0e-4} &  \cellCD[22pt]{b7d7a8}{87.8} &  \cellCD[22pt]{b7d7a8}{88.7} &  \cellCD[22pt]{69a84f}{99.9} &  \cellCD[22pt]{ffffff}{77.2} &  \cellCD[22pt]{b7d7a8}{87.9} &  \cellCD[22pt]{69a84f}{99.5} &  \cellCD[22pt]{ffffff}{82.1} &  \cellCD[22pt]{b7d7a8}{88.6} &  \cellCD[22pt]{69a84f}{99.9} &  \cellCD[22pt]{ffffff}{60.5} &  \cellCD[22pt]{b7d7a8}{86.3} &  \cellCD[22pt]{69a84f}{98.7} \\
      nerf\_osr  (8) & 402.8 &  \cellCD[24pt]{69a84f}{50} &  \cellCD[24pt]{b7d7a8}{324} &  \cellCD[24pt]{ffffff}{3163} &  \cellCD[22pt]{b7d7a8}{1.6e-3} &  \cellCD[22pt]{b7d7a8}{1.1e-3} &  \cellCD[22pt]{b7d7a8}{1.3e-3} &  \cellCD[22pt]{b7d7a8}{91.7} &  \cellCD[22pt]{b7d7a8}{92.0} &  \cellCD[22pt]{b7d7a8}{92.1} &  \cellCD[22pt]{b7d7a8}{70.9} &  \cellCD[22pt]{b7d7a8}{71.9} &  \cellCD[22pt]{b7d7a8}{71.7} &  \cellCD[22pt]{b7d7a8}{71.1} &  \cellCD[22pt]{b7d7a8}{71.9} &  \cellCD[22pt]{b7d7a8}{71.7} &  \cellCD[22pt]{b7d7a8}{43.2} &  \cellCD[22pt]{b7d7a8}{45.2} &  \cellCD[22pt]{b7d7a8}{44.7} \\
  drone\_deploy  (9) & 524.7 &  \cellCD[24pt]{69a84f}{91} &  \cellCD[24pt]{b7d7a8}{365} &  \cellCD[24pt]{ffffff}{3352} &  \cellCD[22pt]{b7d7a8}{4.9e-3} &  \cellCD[22pt]{b7d7a8}{4.3e-3} &  \cellCD[22pt]{69a84f}{2.0e-3} &  \cellCD[22pt]{b7d7a8}{97.9} &  \cellCD[22pt]{b7d7a8}{98.2} &  \cellCD[22pt]{ffffff}{91.3} &  \cellCD[22pt]{b7d7a8}{79.2} &  \cellCD[22pt]{b7d7a8}{81.1} &  \cellCD[22pt]{ffffff}{65.2} &  \cellCD[22pt]{b7d7a8}{89.6} &  \cellCD[22pt]{b7d7a8}{91.5} &  \cellCD[22pt]{ffffff}{73.5} &  \cellCD[22pt]{b7d7a8}{50.4} &  \cellCD[22pt]{69a84f}{53.5} &  \cellCD[22pt]{ffffff}{40.2} \\
        zipnerf  (4) & 1527.2 &  \cellCD[24pt]{69a84f}{119} &  \cellCD[24pt]{b7d7a8}{690} &  \cellCD[24pt]{ffffff}{3820} &  \cellCD[22pt]{b7d7a8}{3.0e-3} &  \cellCD[22pt]{ffffff}{7.1e-3} &  \cellCD[22pt]{69a84f}{3.4e-4} &  \cellCD[22pt]{b7d7a8}{99.0} &  \cellCD[22pt]{b7d7a8}{98.1} &  \cellCD[22pt]{b7d7a8}{99.7} &  \cellCD[22pt]{ffffff}{92.6} &  \cellCD[22pt]{b7d7a8}{96.6} &  \cellCD[22pt]{b7d7a8}{98.1} &  \cellCD[22pt]{b7d7a8}{97.4} &  \cellCD[22pt]{b7d7a8}{98.0} &  \cellCD[22pt]{b7d7a8}{99.6} &  \cellCD[22pt]{ffffff}{81.4} &  \cellCD[22pt]{b7d7a8}{93.6} &  \cellCD[22pt]{b7d7a8}{95.2} \\
   urban\_scene  (3) & 3824 &  \cellCD[24pt]{69a84f}{515} &  \cellCD[24pt]{b7d7a8}{3664} &  \cellCD[24pt]{ffffff}{61622} &  \cellCD[22pt]{b7d7a8}{1.7e-5} &  \cellCD[22pt]{b7d7a8}{1.4e-5} &  \cellCD[22pt]{b7d7a8}{1.4e-5} &  \cellCD[22pt]{b7d7a8}{99.9} &  \cellCD[22pt]{b7d7a8}{99.9} &  \cellCD[22pt]{b7d7a8}{100.0} &  \cellCD[22pt]{b7d7a8}{95.3} &  \cellCD[22pt]{b7d7a8}{97.0} &  \cellCD[22pt]{b7d7a8}{97.0} &  \cellCD[22pt]{b7d7a8}{99.5} &  \cellCD[22pt]{b7d7a8}{99.6} &  \cellCD[22pt]{b7d7a8}{99.6} &  \cellCD[22pt]{ffffff}{86.3} &  \cellCD[22pt]{b7d7a8}{91.2} &  \cellCD[22pt]{b7d7a8}{91.3} \\
mill19\_building  & 1920 &  \cellCD[24pt]{69a84f}{258} &  \cellCD[24pt]{b7d7a8}{6289} &  \cellCD[24pt]{ffffff}{27080} &  \cellCD[22pt]{b7d7a8}{3.0e-4} &  \cellCD[22pt]{ffffff}{1.3e-2} &  \cellCD[22pt]{69a84f}{1.9e-5} &  \cellCD[22pt]{b7d7a8}{99.9} &  \cellCD[22pt]{f4cccc}{0.1} &  \cellCD[22pt]{b7d7a8}{99.9} &  \cellCD[22pt]{b7d7a8}{95.5} &  \cellCD[22pt]{f4cccc}{0.0} &  \cellCD[22pt]{b7d7a8}{95.6} &  \cellCD[22pt]{b7d7a8}{99.3} &  \cellCD[22pt]{f4cccc}{0.0} &  \cellCD[22pt]{b7d7a8}{99.3} &  \cellCD[22pt]{b7d7a8}{87.0} &  \cellCD[22pt]{f4cccc}{0.0} &  \cellCD[22pt]{b7d7a8}{87.4} \\
 mill19\_rubble  & 1657 &  \cellCD[24pt]{69a84f}{240} &  \cellCD[24pt]{b7d7a8}{2849} &  \cellCD[24pt]{ffffff}{12153} &  \cellCD[22pt]{b7d7a8}{3.6e-5} &  \cellCD[22pt]{ffffff}{6.4e-5} &  \cellCD[22pt]{b7d7a8}{3.4e-5} &  \cellCD[22pt]{b7d7a8}{99.9} &  \cellCD[22pt]{b7d7a8}{99.8} &  \cellCD[22pt]{b7d7a8}{99.9} &  \cellCD[22pt]{b7d7a8}{93.6} &  \cellCD[22pt]{b7d7a8}{94.5} &  \cellCD[22pt]{b7d7a8}{94.6} &  \cellCD[22pt]{b7d7a8}{98.6} &  \cellCD[22pt]{b7d7a8}{98.6} &  \cellCD[22pt]{b7d7a8}{98.7} &  \cellCD[22pt]{ffffff}{81.6} &  \cellCD[22pt]{b7d7a8}{84.7} &  \cellCD[22pt]{b7d7a8}{84.8} \\
eyeful\_apartment  & 3804 &  \cellCD[24pt]{69a84f}{549} &  \cellCD[24pt]{b7d7a8}{5905} &  \cellCD[24pt]{ffffff}{185361} &  \cellCD[22pt]{b7d7a8}{2.8e-3} &  \cellCD[22pt]{ffffff}{9.4e-3} &  \cellCD[22pt]{b7d7a8}{2.2e-3} &  \cellCD[22pt]{b7d7a8}{86.8} &  \cellCD[22pt]{ffffff}{75.0} &  \cellCD[22pt]{69a84f}{90.2} &  \cellCD[22pt]{ffffff}{45.5} &  \cellCD[22pt]{b7d7a8}{50.5} &  \cellCD[22pt]{69a84f}{62.0} &  \cellCD[22pt]{ffffff}{51.1} &  \cellCD[22pt]{b7d7a8}{61.3} &  \cellCD[22pt]{69a84f}{71.7} &  \cellCD[22pt]{ffffff}{6.4} &  \cellCD[22pt]{b7d7a8}{18.2} &  \cellCD[22pt]{69a84f}{21.9} \\
eyeful\_kitchen  & 6042 &  \cellCD[24pt]{69a84f}{2202} &  \cellCD[24pt]{b7d7a8}{22884} &  \cellCD[24pt]{c0c0c0}{timeout} &  \cellCD[22pt]{69a84f}{3.1e-3} &  \cellCD[22pt]{b7d7a8}{7.4e-3} &  \cellCD[22pt]{c0c0c0}{\vphantom{0}-} &  \cellCD[22pt]{69a84f}{85.0} &  \cellCD[22pt]{b7d7a8}{59.9} &  \cellCD[22pt]{c0c0c0}{\vphantom{0}-} &  \cellCD[22pt]{b7d7a8}{38.1} &  \cellCD[22pt]{69a84f}{41.2} &  \cellCD[22pt]{c0c0c0}{\vphantom{0}-} &  \cellCD[22pt]{b7d7a8}{46.7} &  \cellCD[22pt]{69a84f}{51.7} &  \cellCD[22pt]{c0c0c0}{\vphantom{0}-} &  \cellCD[22pt]{b7d7a8}{4.6} &  \cellCD[22pt]{69a84f}{14.4} &  \cellCD[22pt]{c0c0c0}{\vphantom{0}-} \\
\bottomrule
\end{tabular}
}

    }
    \caption{Speed and pose accuracy of \alg, GLOMAP, and COLMAP. All three methods are accelerated by GPU. For datasets with more than two scenes, we denote the average metrics as \texttt{dataset-name(\#scenes)}. In particular, Tanks and Temples~\cite{knapitsch2017tanks} has three official splits, and we do the averaging separately for them. Mill-19~\cite{turki2022mega} and Eyeful Tower~\cite{VRNeRF} scenes are listed separately. Metrics are color-coded in {\color{teal}green}, with color changes if the percentage gap $>$$2\%$ or ATE ratio $>$$1.5$. {\color{red}Red} denotes complete failures and {\color{gray}gray} means the method did not finish in a week. Note the significant speedup of \alg vs. previous work, especially on larger scenes.}
    \label{tab:fastmap-pose_overall}

\end{table*}
\begin{table}[t]
\centering
\resizebox{0.8\linewidth}{!}{%
    \tablestyle{4pt}{1.1}
\begin{tabular}{l l | ccc | cc}
\toprule
  &  & \multicolumn{3}{c}{Absolute PSNR $\uparrow$} & \multicolumn{2}{c}{Relative to COLMAP} \\
\cmidrule(lr){3-5}  \cmidrule(lr){6-7}
  &  & \alg & GLOMAP & COLMAP & \alg & GLOMAP \\
\midrule
m360\_bicycle & { Zip-NeRF} &  \cellCD[18pt]{b7d7a8}{25.60} &  \cellCD[18pt]{b7d7a8}{25.78} &  \cellCD[18pt]{b7d7a8}{25.86} &  \cellCD[18pt]{f4cccc}{-0.26} &  \cellCD[18pt]{ffffff}{-0.08} \\
 & { + CamP} &  \cellCD[18pt]{b7d7a8}{26.21} &  \cellCD[18pt]{b7d7a8}{26.36} &  \cellCD[18pt]{b7d7a8}{26.41} &  \cellCD[18pt]{ffffff}{-0.21} &  \cellCD[18pt]{ffffff}{-0.05} \\
\cmidrule(lr){2-7}
 & { GSplat} &  \cellCD[18pt]{b7d7a8}{25.51} &  \cellCD[18pt]{b7d7a8}{25.59} &  \cellCD[18pt]{b7d7a8}{25.62} &  \cellCD[18pt]{ffffff}{-0.11} &  \cellCD[18pt]{ffffff}{-0.03} \\
\midrule
m360\_bonsai & { Zip-NeRF} &  \cellCD[18pt]{b7d7a8}{34.78} &  \cellCD[18pt]{b7d7a8}{34.91} &  \cellCD[18pt]{ffffff}{34.47} &  \cellCD[18pt]{fff2cc}{0.31} &  \cellCD[18pt]{fff2cc}{0.44} \\
 & { + CamP} &  \cellCD[18pt]{b7d7a8}{35.26} &  \cellCD[18pt]{b7d7a8}{35.32} &  \cellCD[18pt]{b7d7a8}{35.37} &  \cellCD[18pt]{ffffff}{-0.12} &  \cellCD[18pt]{ffffff}{-0.05} \\
\cmidrule(lr){2-7}
 & { GSplat} &  \cellCD[18pt]{b7d7a8}{32.32} &  \cellCD[18pt]{b7d7a8}{32.29} &  \cellCD[18pt]{ffffff}{31.49} &  \cellCD[18pt]{fff2cc}{0.84} &  \cellCD[18pt]{fff2cc}{0.81} \\
\midrule
m360\_counter & { Zip-NeRF} &  \cellCD[18pt]{b7d7a8}{28.97} &  \cellCD[18pt]{b7d7a8}{28.95} &  \cellCD[18pt]{b7d7a8}{29.18} &  \cellCD[18pt]{ffffff}{-0.21} &  \cellCD[18pt]{ffffff}{-0.23} \\
 & { + CamP} &  \cellCD[18pt]{b7d7a8}{29.09} &  \cellCD[18pt]{b7d7a8}{29.18} &  \cellCD[18pt]{b7d7a8}{29.29} &  \cellCD[18pt]{ffffff}{-0.20} &  \cellCD[18pt]{ffffff}{-0.12} \\
\cmidrule(lr){2-7}
 & { GSplat} &  \cellCD[18pt]{b7d7a8}{28.99} &  \cellCD[18pt]{b7d7a8}{29.06} &  \cellCD[18pt]{b7d7a8}{29.02} &  \cellCD[18pt]{ffffff}{-0.02} &  \cellCD[18pt]{ffffff}{0.04} \\
\midrule
m360\_flowers & { Zip-NeRF} &  \cellCD[18pt]{b7d7a8}{22.05} &  \cellCD[18pt]{b7d7a8}{22.29} &  \cellCD[18pt]{b7d7a8}{21.89} &  \cellCD[18pt]{ffffff}{0.15} &  \cellCD[18pt]{fff2cc}{0.40} \\
 & { + CamP} &  \cellCD[18pt]{b7d7a8}{23.53} &  \cellCD[18pt]{b7d7a8}{23.47} &  \cellCD[18pt]{b7d7a8}{23.27} &  \cellCD[18pt]{fff2cc}{0.25} &  \cellCD[18pt]{ffffff}{0.20} \\
\cmidrule(lr){2-7}
 & { GSplat} &  \cellCD[18pt]{b7d7a8}{21.74} &  \cellCD[18pt]{b7d7a8}{21.79} &  \cellCD[18pt]{b7d7a8}{21.59} &  \cellCD[18pt]{ffffff}{0.15} &  \cellCD[18pt]{ffffff}{0.20} \\
\midrule
m360\_garden & { Zip-NeRF} &  \cellCD[18pt]{b7d7a8}{28.10} &  \cellCD[18pt]{b7d7a8}{28.20} &  \cellCD[18pt]{b7d7a8}{28.20} &  \cellCD[18pt]{ffffff}{-0.11} &  \cellCD[18pt]{ffffff}{0.00} \\
 & { + CamP} &  \cellCD[18pt]{b7d7a8}{28.54} &  \cellCD[18pt]{b7d7a8}{28.49} &  \cellCD[18pt]{b7d7a8}{28.54} &  \cellCD[18pt]{ffffff}{0.00} &  \cellCD[18pt]{ffffff}{-0.05} \\
\cmidrule(lr){2-7}
 & { GSplat} &  \cellCD[18pt]{b7d7a8}{27.61} &  \cellCD[18pt]{b7d7a8}{27.67} &  \cellCD[18pt]{b7d7a8}{27.72} &  \cellCD[18pt]{ffffff}{-0.11} &  \cellCD[18pt]{ffffff}{-0.05} \\
\midrule
m360\_kitchen & { Zip-NeRF} &  \cellCD[18pt]{b7d7a8}{32.29} &  \cellCD[18pt]{b7d7a8}{32.43} &  \cellCD[18pt]{b7d7a8}{32.31} &  \cellCD[18pt]{ffffff}{-0.02} &  \cellCD[18pt]{ffffff}{0.12} \\
 & { + CamP} &  \cellCD[18pt]{69a84f}{32.47} &  \cellCD[18pt]{b7d7a8}{32.19} &  \cellCD[18pt]{b7d7a8}{32.21} &  \cellCD[18pt]{fff2cc}{0.27} &  \cellCD[18pt]{ffffff}{-0.02} \\
\cmidrule(lr){2-7}
 & { GSplat} &  \cellCD[18pt]{b7d7a8}{31.36} &  \cellCD[18pt]{b7d7a8}{31.62} &  \cellCD[18pt]{b7d7a8}{31.58} &  \cellCD[18pt]{ffffff}{-0.22} &  \cellCD[18pt]{ffffff}{0.05} \\
\midrule
m360\_room & { Zip-NeRF} &  \cellCD[18pt]{b7d7a8}{32.81} &  \cellCD[18pt]{b7d7a8}{32.94} &  \cellCD[18pt]{b7d7a8}{32.93} &  \cellCD[18pt]{ffffff}{-0.12} &  \cellCD[18pt]{ffffff}{0.01} \\
 & { + CamP} &  \cellCD[18pt]{b7d7a8}{32.51} &  \cellCD[18pt]{b7d7a8}{32.48} &  \cellCD[18pt]{b7d7a8}{32.44} &  \cellCD[18pt]{ffffff}{0.07} &  \cellCD[18pt]{ffffff}{0.04} \\
\cmidrule(lr){2-7}
 & { GSplat} &  \cellCD[18pt]{b7d7a8}{31.77} &  \cellCD[18pt]{b7d7a8}{31.71} &  \cellCD[18pt]{b7d7a8}{31.67} &  \cellCD[18pt]{ffffff}{0.11} &  \cellCD[18pt]{ffffff}{0.04} \\
\midrule
m360\_stump & { Zip-NeRF} &  \cellCD[18pt]{b7d7a8}{27.34} &  \cellCD[18pt]{b7d7a8}{27.41} &  \cellCD[18pt]{b7d7a8}{27.43} &  \cellCD[18pt]{ffffff}{-0.09} &  \cellCD[18pt]{ffffff}{-0.02} \\
 & { + CamP} &  \cellCD[18pt]{b7d7a8}{28.10} &  \cellCD[18pt]{b7d7a8}{28.03} &  \cellCD[18pt]{b7d7a8}{28.03} &  \cellCD[18pt]{ffffff}{0.07} &  \cellCD[18pt]{ffffff}{0.00} \\
\cmidrule(lr){2-7}
 & { GSplat} &  \cellCD[18pt]{b7d7a8}{26.97} &  \cellCD[18pt]{b7d7a8}{26.89} &  \cellCD[18pt]{b7d7a8}{26.84} &  \cellCD[18pt]{ffffff}{0.13} &  \cellCD[18pt]{ffffff}{0.05} \\
\midrule
m360\_treehill & { Zip-NeRF} &  \cellCD[18pt]{ffffff}{23.73} &  \cellCD[18pt]{b7d7a8}{24.05} &  \cellCD[18pt]{b7d7a8}{24.04} &  \cellCD[18pt]{f4cccc}{-0.31} &  \cellCD[18pt]{ffffff}{0.01} \\
 & { + CamP} &  \cellCD[18pt]{b7d7a8}{25.73} &  \cellCD[18pt]{b7d7a8}{25.74} &  \cellCD[18pt]{b7d7a8}{25.99} &  \cellCD[18pt]{f4cccc}{-0.26} &  \cellCD[18pt]{ffffff}{-0.25} \\
\cmidrule(lr){2-7}
 & { GSplat} &  \cellCD[18pt]{b7d7a8}{22.71} &  \cellCD[18pt]{b7d7a8}{22.88} &  \cellCD[18pt]{b7d7a8}{22.80} &  \cellCD[18pt]{ffffff}{-0.08} &  \cellCD[18pt]{ffffff}{0.08} \\
\midrule
tnt\_training (7) & { InstantNGP} &  \cellCD[18pt]{b7d7a8}{20.73} &  \cellCD[18pt]{ffffff}{19.37} &  \cellCD[18pt]{69a84f}{21.05} &  \cellCD[18pt]{f4cccc}{-0.32} &  \cellCD[18pt]{f4cccc}{-1.68} \\
\cmidrule(lr){2-7}
 & { GSplat} &  \cellCD[18pt]{b7d7a8}{23.22} &  \cellCD[18pt]{ffffff}{21.54} &  \cellCD[18pt]{69a84f}{24.19} &  \cellCD[18pt]{f4cccc}{-0.97} &  \cellCD[18pt]{f4cccc}{-2.65} \\
\midrule
tnt\_intermediate (8) & { InstantNGP} &  \cellCD[18pt]{b7d7a8}{22.29} &  \cellCD[18pt]{b7d7a8}{22.51} &  \cellCD[18pt]{b7d7a8}{22.38} &  \cellCD[18pt]{ffffff}{-0.09} &  \cellCD[18pt]{ffffff}{0.13} \\
\cmidrule(lr){2-7}
 & { GSplat} &  \cellCD[18pt]{ffffff}{24.24} &  \cellCD[18pt]{b7d7a8}{25.28} &  \cellCD[18pt]{b7d7a8}{25.24} &  \cellCD[18pt]{f4cccc}{-1.00} &  \cellCD[18pt]{ffffff}{0.03} \\
\midrule
tnt\_advanced (6) & { InstantNGP} &  \cellCD[18pt]{ffffff}{16.59} &  \cellCD[18pt]{b7d7a8}{16.94} &  \cellCD[18pt]{69a84f}{17.55} &  \cellCD[18pt]{f4cccc}{-0.95} &  \cellCD[18pt]{f4cccc}{-0.60} \\
\cmidrule(lr){2-7}
 & { GSplat} &  \cellCD[18pt]{b7d7a8}{18.82} &  \cellCD[18pt]{b7d7a8}{18.74} &  \cellCD[18pt]{69a84f}{22.06} &  \cellCD[18pt]{f4cccc}{-3.24} &  \cellCD[18pt]{f4cccc}{-3.32} \\
\bottomrule
\end{tabular}

}
\caption{
Novel view synthesis evaluation on MipNeRF360~\cite{barron2022mipnerf360} and Tanks and Temples~\cite{knapitsch2017tanks}.  Results for MipNeRF360 are listed separately for each scene, and those for Tanks and Temples are averaged over all scenes in each of the three splits. The color changes only if the PSNR difference $>$$0.25$. We report results for Zip-NeRF, Zip-NeRF + CamP optimization, and Gaussian Splatting. 
}
\label{tab:fastmap-nerf_m360}
\end{table}

\begin{table}[t]
\centering
\resizebox{0.8\linewidth}{!}{
    \tablestyle{4pt}{1.1}
\begin{tabular}{l l ccc | ccc }
\toprule
& & \multicolumn{3}{c}{Instant-NGP} & \multicolumn{3}{c}{Gaussian Splatting}  \\
\cmidrule(lr){3-5} \cmidrule(lr){6-8}
& &
\scriptsize \alg & \scriptsize GLOMAP & \scriptsize COLMAP &
\scriptsize \alg & \scriptsize GLOMAP & \scriptsize COLMAP \\
\midrule
\multirow{7}{*}{\rotatebox{90}{training}} & Barn &  \cellCD[18pt]{ffffff}{23.37} &  \cellCD[18pt]{b7d7a8}{23.68} &  \cellCD[18pt]{b7d7a8}{23.69} &  \cellCD[18pt]{ffffff}{26.17} &  \cellCD[18pt]{b7d7a8}{27.81} &  \cellCD[18pt]{b7d7a8}{27.79} \\
 & Caterpillar &  \cellCD[18pt]{b7d7a8}{20.17} &  \cellCD[18pt]{b7d7a8}{20.20} &  \cellCD[18pt]{b7d7a8}{20.23} &  \cellCD[18pt]{b7d7a8}{23.30} &  \cellCD[18pt]{b7d7a8}{23.47} &  \cellCD[18pt]{b7d7a8}{23.59} \\
 & Courthouse &  \cellCD[18pt]{b7d7a8}{19.96} &  \cellCD[18pt]{ffffff}{14.73} &  \cellCD[18pt]{69a84f}{20.27} &  \cellCD[18pt]{b7d7a8}{21.12} &  \cellCD[18pt]{ffffff}{12.23} &  \cellCD[18pt]{69a84f}{22.25} \\
 & Ignatius &  \cellCD[18pt]{b7d7a8}{18.11} &  \cellCD[18pt]{b7d7a8}{18.14} &  \cellCD[18pt]{b7d7a8}{18.26} &  \cellCD[18pt]{ffffff}{21.42} &  \cellCD[18pt]{b7d7a8}{22.04} &  \cellCD[18pt]{b7d7a8}{21.86} \\
 & Meetingroom &  \cellCD[18pt]{ffffff}{21.61} &  \cellCD[18pt]{b7d7a8}{22.59} &  \cellCD[18pt]{b7d7a8}{22.41} &  \cellCD[18pt]{ffffff}{23.71} &  \cellCD[18pt]{b7d7a8}{25.35} &  \cellCD[18pt]{b7d7a8}{25.17} \\
 & Truck &  \cellCD[18pt]{b7d7a8}{21.19} &  \cellCD[18pt]{ffffff}{16.86} &  \cellCD[18pt]{69a84f}{21.45} &  \cellCD[18pt]{b7d7a8}{23.58} &  \cellCD[18pt]{ffffff}{18.34} &  \cellCD[18pt]{69a84f}{24.50} \\
\midrule
\multirow{8}{*}{\rotatebox{90}{intermediate}} & Family &  \cellCD[18pt]{b7d7a8}{22.10} &  \cellCD[18pt]{b7d7a8}{21.99} &  \cellCD[18pt]{ffffff}{21.45} &  \cellCD[18pt]{ffffff}{23.67} &  \cellCD[18pt]{b7d7a8}{24.54} &  \cellCD[18pt]{b7d7a8}{24.75} \\
 & Francis &  \cellCD[18pt]{b7d7a8}{23.68} &  \cellCD[18pt]{b7d7a8}{23.73} &  \cellCD[18pt]{b7d7a8}{23.48} &  \cellCD[18pt]{ffffff}{26.94} &  \cellCD[18pt]{b7d7a8}{27.30} &  \cellCD[18pt]{69a84f}{27.59} \\
 & Horse &  \cellCD[18pt]{b7d7a8}{21.07} &  \cellCD[18pt]{b7d7a8}{21.04} &  \cellCD[18pt]{b7d7a8}{21.13} &  \cellCD[18pt]{ffffff}{22.96} &  \cellCD[18pt]{b7d7a8}{24.05} &  \cellCD[18pt]{b7d7a8}{23.89} \\
 & Lighthouse &  \cellCD[18pt]{b7d7a8}{20.84} &  \cellCD[18pt]{b7d7a8}{20.99} &  \cellCD[18pt]{b7d7a8}{20.91} &  \cellCD[18pt]{b7d7a8}{22.00} &  \cellCD[18pt]{b7d7a8}{22.19} &  \cellCD[18pt]{b7d7a8}{22.12} \\
 & M60 &  \cellCD[18pt]{ffffff}{24.80} &  \cellCD[18pt]{b7d7a8}{25.11} &  \cellCD[18pt]{b7d7a8}{25.11} &  \cellCD[18pt]{ffffff}{26.44} &  \cellCD[18pt]{b7d7a8}{28.07} &  \cellCD[18pt]{b7d7a8}{27.95} \\
 & Panther &  \cellCD[18pt]{ffffff}{25.25} &  \cellCD[18pt]{69a84f}{26.01} &  \cellCD[18pt]{b7d7a8}{25.74} &  \cellCD[18pt]{ffffff}{27.13} &  \cellCD[18pt]{69a84f}{28.27} &  \cellCD[18pt]{b7d7a8}{27.97} \\
 & Playground &  \cellCD[18pt]{ffffff}{21.48} &  \cellCD[18pt]{b7d7a8}{21.96} &  \cellCD[18pt]{b7d7a8}{21.99} &  \cellCD[18pt]{ffffff}{24.12} &  \cellCD[18pt]{b7d7a8}{26.00} &  \cellCD[18pt]{b7d7a8}{26.03} \\
 & Train &  \cellCD[18pt]{b7d7a8}{19.14} &  \cellCD[18pt]{b7d7a8}{19.26} &  \cellCD[18pt]{b7d7a8}{19.23} &  \cellCD[18pt]{ffffff}{20.66} &  \cellCD[18pt]{b7d7a8}{21.79} &  \cellCD[18pt]{b7d7a8}{21.65} \\
\midrule
\multirow{6}{*}{\rotatebox{90}{advanced}} & Auditorium &  \cellCD[18pt]{ffffff}{17.05} &  \cellCD[18pt]{b7d7a8}{19.41} &  \cellCD[18pt]{69a84f}{19.76} &  \cellCD[18pt]{b7d7a8}{17.38} &  \cellCD[18pt]{ffffff}{16.67} &  \cellCD[18pt]{69a84f}{24.20} \\
 & Ballroom &  \cellCD[18pt]{b7d7a8}{17.94} &  \cellCD[18pt]{ffffff}{13.92} &  \cellCD[18pt]{69a84f}{19.12} &  \cellCD[18pt]{b7d7a8}{19.17} &  \cellCD[18pt]{ffffff}{11.91} &  \cellCD[18pt]{69a84f}{23.64} \\
 & Courtroom &  \cellCD[18pt]{ffffff}{17.39} &  \cellCD[18pt]{69a84f}{18.95} &  \cellCD[18pt]{b7d7a8}{17.80} &  \cellCD[18pt]{ffffff}{20.87} &  \cellCD[18pt]{69a84f}{23.11} &  \cellCD[18pt]{b7d7a8}{22.77} \\
 & Museum &  \cellCD[18pt]{b7d7a8}{14.58} &  \cellCD[18pt]{b7d7a8}{14.73} &  \cellCD[18pt]{b7d7a8}{14.53} &  \cellCD[18pt]{ffffff}{20.00} &  \cellCD[18pt]{b7d7a8}{20.97} &  \cellCD[18pt]{b7d7a8}{20.96} \\
 & Palace &  \cellCD[18pt]{b7d7a8}{16.94} &  \cellCD[18pt]{69a84f}{17.57} &  \cellCD[18pt]{b7d7a8}{17.19} &  \cellCD[18pt]{ffffff}{16.41} &  \cellCD[18pt]{b7d7a8}{18.99} &  \cellCD[18pt]{69a84f}{20.05} \\
 & Temple &  \cellCD[18pt]{ffffff}{15.65} &  \cellCD[18pt]{b7d7a8}{17.10} &  \cellCD[18pt]{b7d7a8}{16.87} &  \cellCD[18pt]{ffffff}{19.08} &  \cellCD[18pt]{b7d7a8}{20.80} &  \cellCD[18pt]{b7d7a8}{20.73} \\
\bottomrule
\end{tabular}

}
\caption{Per-scene novel view synthesis results on Tanks and Temples.}
\label{tab:fastmap-nerf_tnt_full}
\end{table}

\begin{table}[t]
\centering
\resizebox{1.0\linewidth}{!}{
\tablestyle{1pt}{1.1}
\begin{tabular}{l ccc@{\,} | ccc@{\,} | ccc}
\toprule
& \multicolumn{3}{c}{tnt\_training (7)} & \multicolumn{3}{c}{tnt\_intermediate (8)} & \multicolumn{3}{c}{tnt\_advanced (6)} \\
\cmidrule(lr){2-4} \cmidrule(lr){5-7} \cmidrule(lr){8-10}
& \scriptsize ATE & \scriptsize RTA@5 & \scriptsize RRA@5
& \scriptsize ATE & \scriptsize RTA@5 & \scriptsize RRA@5
& \scriptsize ATE & \scriptsize RTA@5 & \scriptsize RRA@5 \\
\midrule
ACE-Zero~\cite{brachmann2024scene}      &  \cellCD[22pt]{ffffff}{1.2e-2} &  \cellCD[18pt]{ffffff}{72.9} &  \cellCD[18pt]{ffffff}{73.9} &  \cellCD[22pt]{ffffff}{8.0e-3} &  \cellCD[18pt]{ffffff}{74.0} &  \cellCD[18pt]{ffffff}{67.5} &  \cellCD[22pt]{ffffff}{2.8e-2} &  \cellCD[18pt]{ffffff}{19.1} &  \cellCD[18pt]{ffffff}{22.9}  \\
MAST3R-SfM~\cite{duisterhof2024mast3r}  &  \cellCD[22pt]{b7d7a8}{6.2e-3} &  \cellCD[18pt]{ffffff}{64.9} &  \cellCD[18pt]{ffffff}{56.2} &  \cellCD[22pt]{ffffff}{7.2e-3} &  \cellCD[18pt]{ffffff}{57.5} &  \cellCD[18pt]{ffffff}{50.8} &  \cellCD[22pt]{ffffff}{2.0e-2} &  \cellCD[18pt]{ffffff}{36.5} &  \cellCD[18pt]{ffffff}{38.8}  \\
GLOMAP                                  &  \cellCD[22pt]{ffffff}{1.1e-2} &  \cellCD[18pt]{b7d7a8}{88.8} &  \cellCD[18pt]{b7d7a8}{89.3} &  \cellCD[22pt]{69a84f}{1.9e-5} &  \cellCD[18pt]{b7d7a8}{100.0} &  \cellCD[18pt]{b7d7a8}{100.0} &  \cellCD[22pt]{b7d7a8}{1.2e-2} &  \cellCD[18pt]{69a84f}{79.3} &  \cellCD[18pt]{b7d7a8}{80.5}  \\
\alg                                    &  \cellCD[22pt]{69a84f}{3.2e-3} &  \cellCD[18pt]{b7d7a8}{88.8} &  \cellCD[18pt]{69a84f}{95.8} &  \cellCD[22pt]{b7d7a8}{9.2e-5} &  \cellCD[18pt]{b7d7a8}{100.0} &  \cellCD[18pt]{b7d7a8}{100.0} &  \cellCD[22pt]{69a84f}{6.8e-3} &  \cellCD[18pt]{b7d7a8}{70.5} &  \cellCD[18pt]{b7d7a8}{82.1}  \\
\bottomrule
\end{tabular}

}
\caption{
Comparison to learning-based SfM on Tanks and Temples~\cite{knapitsch2017tanks}. We use COLMAP poses from \citet{kulhanek2024nerfbaselines} as reference. We average numbers for scenes in each split. 
}
\label{tab:fastmap-compare_learning}
\end{table}

\minihead{Datasets}
We focus on the case of high-overlap images densely connected by feature matching, and evaluate the three methods on eight datasets: 
MipNeRF360~\cite{barron2022mipnerf360}, Tanks and Temples~\cite{knapitsch2017tanks}, ZipNeRF~\cite{barron2023zipnerf}, NeRF-OSR~\cite{rudnev2022nerf}, DroneDeploy~\cite{Pilkington2022}, Mill-19~\cite{turki2022mega}, Urbanscene3D~\cite{lin2022capturing}, and Eyeful Tower~\cite{VRNeRF}. They cover a wide range of real-world scenarios and camera trajectory patterns. The number of images per scene ranges from around 200 to 6000.

\paragraph{Data ground truth}
With the exception of Tanks and Temples, each dataset includes author-provided reference camera poses. These reference poses are obtained through different means, including COLMAP (MipNeRF360, ZipNeRF, NeRF-OSR), PixSfM~\cite{lindenberger2021pixel} (Mill-19 and Urbanscene3D), and commercial software (DroneDeploy and Eyeful Tower). For Urbanscene3D, we use the poses provided by \citet{turki2022mega}. For Tanks and Temples, we use COLMAP poses provided by \citet{kulhanek2024nerfbaselines}. On one scene from Tanks and Temples (Courthouse), we found the reference poses to be inconsistent with the images, but still treat them as ground-truth.

\minihead{Metrics}
We report wall-clock time in seconds, excluding the time required for feature extraction and matching (identical for all three methods and dominated by the SfM backend time for large scenes). We evaluate pose accuracy using the standard metrics~\cite{schoenberger2016sfm,pan2024glomap, duisterhof2024mast3r}: ATE, RRA@$\delta$, RTA@$\delta$, and AUC@$\delta$. For some of the scenes, we also evaluate the novel view synthesis quality of NeRF~\cite{mildenhall2021nerf} and Gaussian Splatting~\cite{kerbl20233d} trained on the output poses, intrinsics, and triangulated point clouds.

\subsubsection{Analysis}

\noindent\textbf{Pose accuracy} Table~\ref{tab:fastmap-pose_overall} compares the three methods in average camera pose metrics on all the datasets. In general, our method is much faster than both GLOMAP and COLMAP. The speedup over GLOMAP is less dramatic when there are only a few hundred images (e.g., MipNeRF360), but it can be about $10\times$ faster on scenes with several thousand images (e.g., Urbanscene3D, Mill-19, and Eyeful Tower). On most datasets, \alg is on par with GLOMAP and COLMAP in terms of RTA@$3$. There is a more prominent difference for stricter metrics (RTA@$1$, AUC@$3$, AUC@$1$). This shows that while \alg succeeds in recovering the overall structures of camera trajectories, it does achieve the highest level of precision when the error is reduced to one or two degrees.

None of the methods are perfect. \alg performs particularly bad on the Advanced split of Tanks and Temples, probably because there are many erroneous matches due to repetitive patterns and symmetric structures in the scenes. This is a well-known problem of global SfM (i.e., GLOMAP also suffers a significant drop in performance compared to COLMAP), and incremental SfM methods like COLMAP are more robust in these settings. On the building scene of the Mill-19 dataset, GLOMAP fail catastrophically, however \alg and COLMAP remain highly accurate. On DroneDeploy, none of the three methods is very good in terms of AUC@$1$ and RTA@$1$.

In \cref{tab:fastmap-compare_learning}, we compare \alg to two representative learning-based methods, ACE-Zero~\cite{brachmann2024scene} and MAST3R-SfM~\cite{duisterhof2024mast3r}, where we include the results from \citet[Tab.~10]{duisterhof2024mast3r}. 
Both methods perform significantly worse than \alg and GLOMAP. This indicates that while learning-based methods are promising, they still lag far behind traditional methods in terms of pose accuracy.

\noindent\textbf{Novel view synthesis} Table~\ref{tab:fastmap-nerf_m360} evaluates the quality of novel view synthesis on MipNeRF360 and Tanks and Temples when using \alg, COLMAP, and GLOMAP to estimate the camera poses. We use ZipNeRF~\cite{barron2023zipnerf}, a very high-quality NeRF method, for MipNeRF360, and use Instant-NGP~\cite{muller2022instant} for Tank and Temples, which offers a better trade-off between quality and speed. We also evaluate the performance of Gaussian Splatting~\cite{kerbl20233d} on both datasets.

While \alg lags behind GLOMAP and COLMAP on most MipNeRF360 scenes, the PSNR difference is within $0.5$. 
On Tanks and Temples, \alg performs on par with GLOMAP, but both are worse than COLMAP. Here, again, the lower pose accuracy of \alg under the strictest metrics does not prevent \alg poses from yielding competitive PSNR. %
These results suggest that pose accuracy under a strict metric could be a misleading proxy for downstream view synthesis quality, and vice versa.

We also investigate the impact of different SfM poses on rendering with CamP~\cite{park2023camp}, which simultaneously optimizes the radiance field and refines the camera poses. We include the results in \cref{tab:fastmap-nerf_m360} for comparison. 
In general, CamP improves the PSNR for all the three methods, and for some scenes (e.g., flowers, garden, kitchen, etc.) the gap in rendering quality is closed and sometimes even reversed.

\subsubsection{More Results}
Additional speed benchmarking with different hardware configurations is reported in Table~\ref{tab:fastmap-detailed_timing}. We show per-scene pose accuracy metrics for all datasets in Tables~\ref{tab:fastmap-supp_pose_m360} to~\ref{tab:fastmap-supp_pose_large}. The per-scene NeRF and Gaussian Splatting evaluation on Tanks and Temples is shown in Table~\ref{tab:fastmap-nerf_tnt_full}.

\begin{table*}[t]
    \centering
    \resizebox{1.0\linewidth}{!}{
        \tablestyle{2pt}{1.1}
\begin{tabular}{l r  r@{~}r@{~}r | r@{~}r@{~}r | r@{~}r@{~}r | r@{~}r@{~}r | r@{~}r@{~}r | r@{~}r@{~}r | r@{~}r@{~}r | r@{~}r@{~}r }

\toprule

  &  & \multicolumn{3}{c}{time (sec)} &
\multicolumn{3}{c}{ATE$\downarrow$} &
\multicolumn{3}{c}{RTA@3$\uparrow$} &
\multicolumn{3}{c}{RRA@3$\uparrow$} &
\multicolumn{3}{c}{RTA@1$\uparrow$} &
\multicolumn{3}{c}{RRA@1$\uparrow$} &
\multicolumn{3}{c}{AUC-R\&T @ 3 $\uparrow$} &
\multicolumn{3}{c}{AUC-R\&T @ 1 $\uparrow$} \\

\cmidrule(lr){3-5} \cmidrule(lr){6-8} \cmidrule(lr){9-11}
\cmidrule(lr){12-14} \cmidrule(lr){15-17} \cmidrule(lr){18-20}
\cmidrule(lr){21-23} \cmidrule(lr){24-26}

   & n\_imgs &
\multicolumn{1}{c}{\scriptsize \alg} & \multicolumn{1}{c}{\scriptsize GLOMAP} & \multicolumn{1}{c}{\scriptsize COLMAP} &
\multicolumn{1}{c}{\scriptsize \alg} & \multicolumn{1}{c}{\scriptsize GLOMAP} & \multicolumn{1}{c}{\scriptsize COLMAP} &
\multicolumn{1}{c}{\scriptsize \alg} & \multicolumn{1}{c}{\scriptsize GLOMAP} & \multicolumn{1}{c}{\scriptsize COLMAP} &
\multicolumn{1}{c}{\scriptsize \alg} & \multicolumn{1}{c}{\scriptsize GLOMAP} & \multicolumn{1}{c}{\scriptsize COLMAP} &
\multicolumn{1}{c}{\scriptsize \alg} & \multicolumn{1}{c}{\scriptsize GLOMAP} & \multicolumn{1}{c}{\scriptsize COLMAP} &
\multicolumn{1}{c}{\scriptsize \alg} & \multicolumn{1}{c}{\scriptsize GLOMAP} & \multicolumn{1}{c}{\scriptsize COLMAP} &
\multicolumn{1}{c}{\scriptsize \alg} & \multicolumn{1}{c}{\scriptsize GLOMAP} & \multicolumn{1}{c}{\scriptsize COLMAP} &
\multicolumn{1}{c}{\scriptsize \alg} & \multicolumn{1}{c}{\scriptsize GLOMAP} & \multicolumn{1}{c}{\scriptsize COLMAP} \\

\midrule
  m360\_bicycle &  194 &  \cellCD[24pt]{69a84f}{20} &  \cellCD[24pt]{b7d7a8}{74} &  \cellCD[24pt]{ffffff}{151} &  \cellCD[22pt]{b7d7a8}{7.5e-5} &  \cellCD[22pt]{b7d7a8}{5.4e-5} &  \cellCD[22pt]{b7d7a8}{5.7e-5} &  \cellCD[22pt]{b7d7a8}{100.0} &  \cellCD[22pt]{b7d7a8}{100.0} &  \cellCD[22pt]{b7d7a8}{100.0} &  \cellCD[22pt]{b7d7a8}{100.0} &  \cellCD[22pt]{b7d7a8}{100.0} &  \cellCD[22pt]{b7d7a8}{100.0} &  \cellCD[22pt]{b7d7a8}{99.7} &  \cellCD[22pt]{b7d7a8}{100.0} &  \cellCD[22pt]{b7d7a8}{99.9} &  \cellCD[22pt]{b7d7a8}{100.0} &  \cellCD[22pt]{b7d7a8}{100.0} &  \cellCD[22pt]{b7d7a8}{100.0} &  \cellCD[22pt]{b7d7a8}{96.9} &  \cellCD[22pt]{b7d7a8}{97.6} &  \cellCD[22pt]{b7d7a8}{97.5} &  \cellCD[22pt]{b7d7a8}{90.8} &  \cellCD[22pt]{b7d7a8}{92.7} &  \cellCD[22pt]{b7d7a8}{92.6} \\
   m360\_bonsai &  292 &  \cellCD[24pt]{69a84f}{47} &  \cellCD[24pt]{b7d7a8}{306} &  \cellCD[24pt]{ffffff}{1043} &  \cellCD[22pt]{b7d7a8}{3.9e-5} &  \cellCD[22pt]{69a84f}{2.1e-5} &  \cellCD[22pt]{ffffff}{1.7e-4} &  \cellCD[22pt]{b7d7a8}{100.0} &  \cellCD[22pt]{b7d7a8}{100.0} &  \cellCD[22pt]{b7d7a8}{99.9} &  \cellCD[22pt]{b7d7a8}{100.0} &  \cellCD[22pt]{b7d7a8}{100.0} &  \cellCD[22pt]{b7d7a8}{100.0} &  \cellCD[22pt]{b7d7a8}{100.0} &  \cellCD[22pt]{b7d7a8}{100.0} &  \cellCD[22pt]{ffffff}{97.9} &  \cellCD[22pt]{b7d7a8}{100.0} &  \cellCD[22pt]{b7d7a8}{100.0} &  \cellCD[22pt]{b7d7a8}{100.0} &  \cellCD[22pt]{b7d7a8}{96.4} &  \cellCD[22pt]{69a84f}{98.6} &  \cellCD[22pt]{ffffff}{92.8} &  \cellCD[22pt]{b7d7a8}{89.2} &  \cellCD[22pt]{69a84f}{95.7} &  \cellCD[22pt]{ffffff}{79.7} \\
  m360\_counter &  240 &  \cellCD[24pt]{69a84f}{35} &  \cellCD[24pt]{b7d7a8}{201} &  \cellCD[24pt]{ffffff}{443} &  \cellCD[22pt]{ffffff}{1.3e-5} &  \cellCD[22pt]{b7d7a8}{2.5e-6} &  \cellCD[22pt]{b7d7a8}{2.3e-6} &  \cellCD[22pt]{b7d7a8}{100.0} &  \cellCD[22pt]{b7d7a8}{100.0} &  \cellCD[22pt]{b7d7a8}{100.0} &  \cellCD[22pt]{b7d7a8}{100.0} &  \cellCD[22pt]{b7d7a8}{100.0} &  \cellCD[22pt]{b7d7a8}{100.0} &  \cellCD[22pt]{b7d7a8}{100.0} &  \cellCD[22pt]{b7d7a8}{100.0} &  \cellCD[22pt]{b7d7a8}{100.0} &  \cellCD[22pt]{b7d7a8}{100.0} &  \cellCD[22pt]{b7d7a8}{100.0} &  \cellCD[22pt]{b7d7a8}{100.0} &  \cellCD[22pt]{b7d7a8}{99.1} &  \cellCD[22pt]{b7d7a8}{99.8} &  \cellCD[22pt]{b7d7a8}{99.8} &  \cellCD[22pt]{ffffff}{97.2} &  \cellCD[22pt]{b7d7a8}{99.5} &  \cellCD[22pt]{b7d7a8}{99.5} \\
  m360\_flowers &  173 &  \cellCD[24pt]{69a84f}{19} &  \cellCD[24pt]{b7d7a8}{54} &  \cellCD[24pt]{ffffff}{120} &  \cellCD[22pt]{b7d7a8}{6.5e-5} &  \cellCD[22pt]{b7d7a8}{6.0e-5} &  \cellCD[22pt]{ffffff}{1.6e-4} &  \cellCD[22pt]{b7d7a8}{100.0} &  \cellCD[22pt]{b7d7a8}{100.0} &  \cellCD[22pt]{b7d7a8}{100.0} &  \cellCD[22pt]{b7d7a8}{100.0} &  \cellCD[22pt]{b7d7a8}{100.0} &  \cellCD[22pt]{b7d7a8}{100.0} &  \cellCD[22pt]{b7d7a8}{99.8} &  \cellCD[22pt]{b7d7a8}{99.9} &  \cellCD[22pt]{b7d7a8}{99.6} &  \cellCD[22pt]{b7d7a8}{100.0} &  \cellCD[22pt]{b7d7a8}{100.0} &  \cellCD[22pt]{b7d7a8}{100.0} &  \cellCD[22pt]{b7d7a8}{96.1} &  \cellCD[22pt]{b7d7a8}{96.4} &  \cellCD[22pt]{ffffff}{92.0} &  \cellCD[22pt]{b7d7a8}{88.5} &  \cellCD[22pt]{b7d7a8}{89.3} &  \cellCD[22pt]{ffffff}{76.3} \\
   m360\_garden &  185 &  \cellCD[24pt]{69a84f}{28} &  \cellCD[24pt]{b7d7a8}{152} &  \cellCD[24pt]{ffffff}{490} &  \cellCD[22pt]{69a84f}{7.7e-6} &  \cellCD[22pt]{b7d7a8}{1.5e-5} &  \cellCD[22pt]{b7d7a8}{1.5e-5} &  \cellCD[22pt]{b7d7a8}{100.0} &  \cellCD[22pt]{b7d7a8}{100.0} &  \cellCD[22pt]{b7d7a8}{100.0} &  \cellCD[22pt]{b7d7a8}{100.0} &  \cellCD[22pt]{b7d7a8}{100.0} &  \cellCD[22pt]{b7d7a8}{100.0} &  \cellCD[22pt]{b7d7a8}{100.0} &  \cellCD[22pt]{b7d7a8}{100.0} &  \cellCD[22pt]{b7d7a8}{100.0} &  \cellCD[22pt]{b7d7a8}{100.0} &  \cellCD[22pt]{b7d7a8}{100.0} &  \cellCD[22pt]{b7d7a8}{100.0} &  \cellCD[22pt]{b7d7a8}{99.6} &  \cellCD[22pt]{b7d7a8}{99.0} &  \cellCD[22pt]{b7d7a8}{99.1} &  \cellCD[22pt]{b7d7a8}{98.8} &  \cellCD[22pt]{b7d7a8}{97.1} &  \cellCD[22pt]{b7d7a8}{97.3} \\
  m360\_kitchen &  279 &  \cellCD[24pt]{69a84f}{50} &  \cellCD[24pt]{b7d7a8}{376} &  \cellCD[24pt]{ffffff}{1308} &  \cellCD[22pt]{b7d7a8}{4.4e-5} &  \cellCD[22pt]{b7d7a8}{4.0e-5} &  \cellCD[22pt]{b7d7a8}{3.9e-5} &  \cellCD[22pt]{b7d7a8}{100.0} &  \cellCD[22pt]{b7d7a8}{100.0} &  \cellCD[22pt]{b7d7a8}{100.0} &  \cellCD[22pt]{b7d7a8}{100.0} &  \cellCD[22pt]{b7d7a8}{100.0} &  \cellCD[22pt]{b7d7a8}{100.0} &  \cellCD[22pt]{b7d7a8}{100.0} &  \cellCD[22pt]{b7d7a8}{100.0} &  \cellCD[22pt]{b7d7a8}{100.0} &  \cellCD[22pt]{b7d7a8}{100.0} &  \cellCD[22pt]{b7d7a8}{100.0} &  \cellCD[22pt]{b7d7a8}{100.0} &  \cellCD[22pt]{b7d7a8}{97.2} &  \cellCD[22pt]{b7d7a8}{97.2} &  \cellCD[22pt]{b7d7a8}{97.2} &  \cellCD[22pt]{b7d7a8}{91.5} &  \cellCD[22pt]{b7d7a8}{91.5} &  \cellCD[22pt]{b7d7a8}{91.7} \\
     m360\_room &  311 &  \cellCD[24pt]{69a84f}{61} &  \cellCD[24pt]{b7d7a8}{218} &  \cellCD[24pt]{ffffff}{691} &  \cellCD[22pt]{ffffff}{3.4e-3} &  \cellCD[22pt]{b7d7a8}{1.1e-5} &  \cellCD[22pt]{b7d7a8}{9.3e-6} &  \cellCD[22pt]{b7d7a8}{99.3} &  \cellCD[22pt]{b7d7a8}{100.0} &  \cellCD[22pt]{b7d7a8}{100.0} &  \cellCD[22pt]{b7d7a8}{99.4} &  \cellCD[22pt]{b7d7a8}{100.0} &  \cellCD[22pt]{b7d7a8}{100.0} &  \cellCD[22pt]{b7d7a8}{98.8} &  \cellCD[22pt]{b7d7a8}{100.0} &  \cellCD[22pt]{b7d7a8}{100.0} &  \cellCD[22pt]{b7d7a8}{99.4} &  \cellCD[22pt]{b7d7a8}{100.0} &  \cellCD[22pt]{b7d7a8}{100.0} &  \cellCD[22pt]{ffffff}{96.6} &  \cellCD[22pt]{b7d7a8}{98.9} &  \cellCD[22pt]{b7d7a8}{99.0} &  \cellCD[22pt]{ffffff}{91.5} &  \cellCD[22pt]{b7d7a8}{96.7} &  \cellCD[22pt]{b7d7a8}{97.0} \\
    m360\_stump &  125 &  \cellCD[24pt]{69a84f}{15} &  \cellCD[24pt]{b7d7a8}{40} &  \cellCD[24pt]{ffffff}{74} &  \cellCD[22pt]{ffffff}{3.5e-5} &  \cellCD[22pt]{b7d7a8}{1.9e-5} &  \cellCD[22pt]{b7d7a8}{2.1e-5} &  \cellCD[22pt]{b7d7a8}{100.0} &  \cellCD[22pt]{b7d7a8}{100.0} &  \cellCD[22pt]{b7d7a8}{100.0} &  \cellCD[22pt]{b7d7a8}{100.0} &  \cellCD[22pt]{b7d7a8}{100.0} &  \cellCD[22pt]{b7d7a8}{100.0} &  \cellCD[22pt]{b7d7a8}{99.9} &  \cellCD[22pt]{b7d7a8}{100.0} &  \cellCD[22pt]{b7d7a8}{100.0} &  \cellCD[22pt]{b7d7a8}{100.0} &  \cellCD[22pt]{b7d7a8}{100.0} &  \cellCD[22pt]{b7d7a8}{100.0} &  \cellCD[22pt]{b7d7a8}{98.8} &  \cellCD[22pt]{b7d7a8}{99.4} &  \cellCD[22pt]{b7d7a8}{99.3} &  \cellCD[22pt]{b7d7a8}{96.5} &  \cellCD[22pt]{b7d7a8}{98.1} &  \cellCD[22pt]{b7d7a8}{98.0} \\
 m360\_treehill &  141 &  \cellCD[24pt]{69a84f}{18} &  \cellCD[24pt]{b7d7a8}{68} &  \cellCD[24pt]{ffffff}{202} &  \cellCD[22pt]{b7d7a8}{8.7e-5} &  \cellCD[22pt]{b7d7a8}{7.7e-5} &  \cellCD[22pt]{69a84f}{4.6e-5} &  \cellCD[22pt]{b7d7a8}{100.0} &  \cellCD[22pt]{b7d7a8}{100.0} &  \cellCD[22pt]{b7d7a8}{100.0} &  \cellCD[22pt]{b7d7a8}{100.0} &  \cellCD[22pt]{b7d7a8}{100.0} &  \cellCD[22pt]{b7d7a8}{100.0} &  \cellCD[22pt]{b7d7a8}{99.7} &  \cellCD[22pt]{b7d7a8}{99.8} &  \cellCD[22pt]{b7d7a8}{99.9} &  \cellCD[22pt]{b7d7a8}{100.0} &  \cellCD[22pt]{b7d7a8}{100.0} &  \cellCD[22pt]{b7d7a8}{100.0} &  \cellCD[22pt]{b7d7a8}{95.8} &  \cellCD[22pt]{b7d7a8}{96.8} &  \cellCD[22pt]{b7d7a8}{98.3} &  \cellCD[22pt]{ffffff}{87.6} &  \cellCD[22pt]{b7d7a8}{90.6} &  \cellCD[22pt]{69a84f}{94.9} \\
\bottomrule
\end{tabular}

    }
    \caption{Per-scene camera pose metrics on the MipNeRF360 dataset.}
    \label{tab:fastmap-supp_pose_m360}
\end{table*}

\begin{table*}[t]
    \centering
    \resizebox{1.0\linewidth}{!}{
        \tablestyle{2pt}{1.1}
\begin{tabular}{l r  r@{~}r@{~}r | r@{~}r@{~}r | r@{~}r@{~}r | r@{~}r@{~}r | r@{~}r@{~}r | r@{~}r@{~}r | r@{~}r@{~}r | r@{~}r@{~}r }

\toprule

  &  & \multicolumn{3}{c}{time (sec)} &
\multicolumn{3}{c}{ATE$\downarrow$} &
\multicolumn{3}{c}{RTA@3$\uparrow$} &
\multicolumn{3}{c}{RRA@3$\uparrow$} &
\multicolumn{3}{c}{RTA@1$\uparrow$} &
\multicolumn{3}{c}{RRA@1$\uparrow$} &
\multicolumn{3}{c}{AUC-R\&T @ 3 $\uparrow$} &
\multicolumn{3}{c}{AUC-R\&T @ 1 $\uparrow$} \\

\cmidrule(lr){3-5} \cmidrule(lr){6-8} \cmidrule(lr){9-11}
\cmidrule(lr){12-14} \cmidrule(lr){15-17} \cmidrule(lr){18-20}
\cmidrule(lr){21-23} \cmidrule(lr){24-26}

   & n\_imgs &
\multicolumn{1}{c}{\scriptsize \alg} & \multicolumn{1}{c}{\scriptsize GLOMAP} & \multicolumn{1}{c}{\scriptsize COLMAP} &
\multicolumn{1}{c}{\scriptsize \alg} & \multicolumn{1}{c}{\scriptsize GLOMAP} & \multicolumn{1}{c}{\scriptsize COLMAP} &
\multicolumn{1}{c}{\scriptsize \alg} & \multicolumn{1}{c}{\scriptsize GLOMAP} & \multicolumn{1}{c}{\scriptsize COLMAP} &
\multicolumn{1}{c}{\scriptsize \alg} & \multicolumn{1}{c}{\scriptsize GLOMAP} & \multicolumn{1}{c}{\scriptsize COLMAP} &
\multicolumn{1}{c}{\scriptsize \alg} & \multicolumn{1}{c}{\scriptsize GLOMAP} & \multicolumn{1}{c}{\scriptsize COLMAP} &
\multicolumn{1}{c}{\scriptsize \alg} & \multicolumn{1}{c}{\scriptsize GLOMAP} & \multicolumn{1}{c}{\scriptsize COLMAP} &
\multicolumn{1}{c}{\scriptsize \alg} & \multicolumn{1}{c}{\scriptsize GLOMAP} & \multicolumn{1}{c}{\scriptsize COLMAP} &
\multicolumn{1}{c}{\scriptsize \alg} & \multicolumn{1}{c}{\scriptsize GLOMAP} & \multicolumn{1}{c}{\scriptsize COLMAP} \\

\midrule
tnt\_advn\_Auditorium &  298 &  \cellCD[24pt]{69a84f}{74} &  \cellCD[24pt]{b7d7a8}{185} &  \cellCD[24pt]{ffffff}{529} &  \cellCD[22pt]{b7d7a8}{1.4e-2} &  \cellCD[22pt]{ffffff}{3.3e-2} &  \cellCD[22pt]{69a84f}{1.8e-3} &  \cellCD[22pt]{ffffff}{29.6} &  \cellCD[22pt]{b7d7a8}{94.3} &  \cellCD[22pt]{b7d7a8}{95.8} &  \cellCD[22pt]{ffffff}{55.2} &  \cellCD[22pt]{b7d7a8}{94.0} &  \cellCD[22pt]{b7d7a8}{93.4} &  \cellCD[22pt]{ffffff}{11.0} &  \cellCD[22pt]{b7d7a8}{93.7} &  \cellCD[22pt]{b7d7a8}{92.8} &  \cellCD[22pt]{ffffff}{17.0} &  \cellCD[22pt]{69a84f}{94.0} &  \cellCD[22pt]{b7d7a8}{89.6} &  \cellCD[22pt]{ffffff}{12.0} &  \cellCD[22pt]{69a84f}{90.8} &  \cellCD[22pt]{b7d7a8}{88.0} &  \cellCD[22pt]{f4cccc}{1.8} &  \cellCD[22pt]{b7d7a8}{84.8} &  \cellCD[22pt]{b7d7a8}{83.8} \\
tnt\_advn\_Ballroom &  324 &  \cellCD[24pt]{69a84f}{55} &  \cellCD[24pt]{b7d7a8}{652} &  \cellCD[24pt]{ffffff}{1615} &  \cellCD[22pt]{b7d7a8}{1.8e-2} &  \cellCD[22pt]{ffffff}{3.2e-2} &  \cellCD[22pt]{69a84f}{5.1e-3} &  \cellCD[22pt]{b7d7a8}{49.1} &  \cellCD[22pt]{ffffff}{32.5} &  \cellCD[22pt]{69a84f}{98.1} &  \cellCD[22pt]{b7d7a8}{63.7} &  \cellCD[22pt]{ffffff}{36.3} &  \cellCD[22pt]{69a84f}{98.2} &  \cellCD[22pt]{b7d7a8}{28.3} &  \cellCD[22pt]{b7d7a8}{28.3} &  \cellCD[22pt]{69a84f}{97.7} &  \cellCD[22pt]{ffffff}{28.6} &  \cellCD[22pt]{b7d7a8}{32.2} &  \cellCD[22pt]{69a84f}{98.2} &  \cellCD[22pt]{b7d7a8}{25.0} &  \cellCD[22pt]{b7d7a8}{26.5} &  \cellCD[22pt]{69a84f}{93.5} &  \cellCD[22pt]{ffffff}{7.7} &  \cellCD[22pt]{b7d7a8}{17.8} &  \cellCD[22pt]{69a84f}{84.6} \\
tnt\_advn\_Courtroom &  301 &  \cellCD[24pt]{69a84f}{45} &  \cellCD[24pt]{b7d7a8}{296} &  \cellCD[24pt]{ffffff}{882} &  \cellCD[22pt]{ffffff}{9.3e-4} &  \cellCD[22pt]{b7d7a8}{4.3e-5} &  \cellCD[22pt]{69a84f}{2.3e-5} &  \cellCD[22pt]{ffffff}{85.1} &  \cellCD[22pt]{b7d7a8}{99.9} &  \cellCD[22pt]{b7d7a8}{99.9} &  \cellCD[22pt]{b7d7a8}{98.3} &  \cellCD[22pt]{b7d7a8}{100.0} &  \cellCD[22pt]{b7d7a8}{100.0} &  \cellCD[22pt]{ffffff}{27.9} &  \cellCD[22pt]{b7d7a8}{99.3} &  \cellCD[22pt]{b7d7a8}{99.8} &  \cellCD[22pt]{ffffff}{47.4} &  \cellCD[22pt]{b7d7a8}{100.0} &  \cellCD[22pt]{b7d7a8}{100.0} &  \cellCD[22pt]{ffffff}{37.7} &  \cellCD[22pt]{b7d7a8}{96.8} &  \cellCD[22pt]{b7d7a8}{98.6} &  \cellCD[22pt]{ffffff}{3.8} &  \cellCD[22pt]{b7d7a8}{91.0} &  \cellCD[22pt]{69a84f}{96.1} \\
tnt\_advn\_Museum &  301 &  \cellCD[24pt]{69a84f}{44} &  \cellCD[24pt]{b7d7a8}{275} &  \cellCD[24pt]{ffffff}{752} &  \cellCD[22pt]{ffffff}{8.5e-4} &  \cellCD[22pt]{b7d7a8}{5.5e-5} &  \cellCD[22pt]{69a84f}{3.3e-5} &  \cellCD[22pt]{ffffff}{91.4} &  \cellCD[22pt]{b7d7a8}{100.0} &  \cellCD[22pt]{b7d7a8}{100.0} &  \cellCD[22pt]{b7d7a8}{100.0} &  \cellCD[22pt]{b7d7a8}{100.0} &  \cellCD[22pt]{b7d7a8}{100.0} &  \cellCD[22pt]{ffffff}{39.9} &  \cellCD[22pt]{b7d7a8}{99.7} &  \cellCD[22pt]{b7d7a8}{99.9} &  \cellCD[22pt]{ffffff}{85.2} &  \cellCD[22pt]{b7d7a8}{100.0} &  \cellCD[22pt]{b7d7a8}{100.0} &  \cellCD[22pt]{ffffff}{53.8} &  \cellCD[22pt]{b7d7a8}{97.2} &  \cellCD[22pt]{b7d7a8}{98.7} &  \cellCD[22pt]{ffffff}{12.1} &  \cellCD[22pt]{b7d7a8}{91.9} &  \cellCD[22pt]{69a84f}{96.1} \\
tnt\_advn\_Palace &  501 &  \cellCD[24pt]{69a84f}{113} &  \cellCD[24pt]{b7d7a8}{547} &  \cellCD[24pt]{ffffff}{1722} &  \cellCD[22pt]{b7d7a8}{3.3e-3} &  \cellCD[22pt]{ffffff}{6.4e-3} &  \cellCD[22pt]{69a84f}{1.5e-4} &  \cellCD[22pt]{b7d7a8}{74.5} &  \cellCD[22pt]{ffffff}{47.9} &  \cellCD[22pt]{69a84f}{97.1} &  \cellCD[22pt]{b7d7a8}{84.3} &  \cellCD[22pt]{ffffff}{46.4} &  \cellCD[22pt]{69a84f}{98.4} &  \cellCD[22pt]{b7d7a8}{51.6} &  \cellCD[22pt]{ffffff}{44.5} &  \cellCD[22pt]{69a84f}{92.6} &  \cellCD[22pt]{ffffff}{36.1} &  \cellCD[22pt]{b7d7a8}{45.9} &  \cellCD[22pt]{69a84f}{96.5} &  \cellCD[22pt]{b7d7a8}{41.2} &  \cellCD[22pt]{b7d7a8}{42.2} &  \cellCD[22pt]{69a84f}{91.6} &  \cellCD[22pt]{ffffff}{13.3} &  \cellCD[22pt]{b7d7a8}{37.9} &  \cellCD[22pt]{69a84f}{84.4} \\
tnt\_advn\_Temple &  302 &  \cellCD[24pt]{69a84f}{36} &  \cellCD[24pt]{b7d7a8}{190} &  \cellCD[24pt]{ffffff}{596} &  \cellCD[22pt]{ffffff}{9.8e-4} &  \cellCD[22pt]{69a84f}{5.0e-5} &  \cellCD[22pt]{b7d7a8}{1.5e-4} &  \cellCD[22pt]{b7d7a8}{98.8} &  \cellCD[22pt]{b7d7a8}{99.9} &  \cellCD[22pt]{b7d7a8}{99.8} &  \cellCD[22pt]{b7d7a8}{100.0} &  \cellCD[22pt]{b7d7a8}{100.0} &  \cellCD[22pt]{b7d7a8}{100.0} &  \cellCD[22pt]{ffffff}{95.1} &  \cellCD[22pt]{b7d7a8}{99.6} &  \cellCD[22pt]{b7d7a8}{99.4} &  \cellCD[22pt]{b7d7a8}{99.1} &  \cellCD[22pt]{b7d7a8}{100.0} &  \cellCD[22pt]{b7d7a8}{100.0} &  \cellCD[22pt]{ffffff}{85.8} &  \cellCD[22pt]{b7d7a8}{98.4} &  \cellCD[22pt]{b7d7a8}{98.1} &  \cellCD[22pt]{ffffff}{61.3} &  \cellCD[22pt]{b7d7a8}{95.6} &  \cellCD[22pt]{b7d7a8}{95.0} \\
tnt\_intrmdt\_Family &  152 &  \cellCD[24pt]{69a84f}{22} &  \cellCD[24pt]{b7d7a8}{152} &  \cellCD[24pt]{ffffff}{335} &  \cellCD[22pt]{ffffff}{8.3e-5} &  \cellCD[22pt]{b7d7a8}{1.1e-5} &  \cellCD[22pt]{69a84f}{2.8e-6} &  \cellCD[22pt]{b7d7a8}{100.0} &  \cellCD[22pt]{b7d7a8}{100.0} &  \cellCD[22pt]{b7d7a8}{100.0} &  \cellCD[22pt]{b7d7a8}{100.0} &  \cellCD[22pt]{b7d7a8}{100.0} &  \cellCD[22pt]{b7d7a8}{100.0} &  \cellCD[22pt]{b7d7a8}{99.8} &  \cellCD[22pt]{b7d7a8}{100.0} &  \cellCD[22pt]{b7d7a8}{100.0} &  \cellCD[22pt]{b7d7a8}{100.0} &  \cellCD[22pt]{b7d7a8}{100.0} &  \cellCD[22pt]{b7d7a8}{100.0} &  \cellCD[22pt]{ffffff}{95.4} &  \cellCD[22pt]{b7d7a8}{99.4} &  \cellCD[22pt]{b7d7a8}{99.9} &  \cellCD[22pt]{ffffff}{86.5} &  \cellCD[22pt]{b7d7a8}{98.3} &  \cellCD[22pt]{b7d7a8}{99.7} \\
tnt\_intrmdt\_Francis &  302 &  \cellCD[24pt]{69a84f}{29} &  \cellCD[24pt]{b7d7a8}{269} &  \cellCD[24pt]{ffffff}{789} &  \cellCD[22pt]{ffffff}{4.2e-5} &  \cellCD[22pt]{b7d7a8}{6.2e-6} &  \cellCD[22pt]{69a84f}{2.4e-6} &  \cellCD[22pt]{b7d7a8}{99.9} &  \cellCD[22pt]{b7d7a8}{100.0} &  \cellCD[22pt]{b7d7a8}{100.0} &  \cellCD[22pt]{b7d7a8}{100.0} &  \cellCD[22pt]{b7d7a8}{100.0} &  \cellCD[22pt]{b7d7a8}{100.0} &  \cellCD[22pt]{b7d7a8}{99.5} &  \cellCD[22pt]{b7d7a8}{99.9} &  \cellCD[22pt]{b7d7a8}{100.0} &  \cellCD[22pt]{b7d7a8}{100.0} &  \cellCD[22pt]{b7d7a8}{100.0} &  \cellCD[22pt]{b7d7a8}{100.0} &  \cellCD[22pt]{ffffff}{96.2} &  \cellCD[22pt]{b7d7a8}{99.5} &  \cellCD[22pt]{b7d7a8}{99.9} &  \cellCD[22pt]{ffffff}{89.2} &  \cellCD[22pt]{b7d7a8}{98.5} &  \cellCD[22pt]{b7d7a8}{99.7} \\
tnt\_intrmdt\_Horse &  151 &  \cellCD[24pt]{69a84f}{21} &  \cellCD[24pt]{b7d7a8}{134} &  \cellCD[24pt]{ffffff}{255} &  \cellCD[22pt]{ffffff}{6.8e-5} &  \cellCD[22pt]{b7d7a8}{1.4e-5} &  \cellCD[22pt]{69a84f}{4.6e-6} &  \cellCD[22pt]{b7d7a8}{100.0} &  \cellCD[22pt]{b7d7a8}{100.0} &  \cellCD[22pt]{b7d7a8}{100.0} &  \cellCD[22pt]{b7d7a8}{100.0} &  \cellCD[22pt]{b7d7a8}{100.0} &  \cellCD[22pt]{b7d7a8}{100.0} &  \cellCD[22pt]{b7d7a8}{99.9} &  \cellCD[22pt]{b7d7a8}{100.0} &  \cellCD[22pt]{b7d7a8}{100.0} &  \cellCD[22pt]{b7d7a8}{100.0} &  \cellCD[22pt]{b7d7a8}{100.0} &  \cellCD[22pt]{b7d7a8}{100.0} &  \cellCD[22pt]{ffffff}{97.2} &  \cellCD[22pt]{b7d7a8}{99.4} &  \cellCD[22pt]{b7d7a8}{99.8} &  \cellCD[22pt]{ffffff}{91.5} &  \cellCD[22pt]{b7d7a8}{98.3} &  \cellCD[22pt]{b7d7a8}{99.4} \\
tnt\_intrmdt\_Lighthouse &  309 &  \cellCD[24pt]{69a84f}{47} &  \cellCD[24pt]{b7d7a8}{355} &  \cellCD[24pt]{ffffff}{1270} &  \cellCD[22pt]{ffffff}{1.0e-4} &  \cellCD[22pt]{b7d7a8}{2.3e-5} &  \cellCD[22pt]{69a84f}{9.4e-6} &  \cellCD[22pt]{b7d7a8}{99.6} &  \cellCD[22pt]{b7d7a8}{100.0} &  \cellCD[22pt]{b7d7a8}{100.0} &  \cellCD[22pt]{b7d7a8}{100.0} &  \cellCD[22pt]{b7d7a8}{100.0} &  \cellCD[22pt]{b7d7a8}{100.0} &  \cellCD[22pt]{ffffff}{97.4} &  \cellCD[22pt]{b7d7a8}{99.8} &  \cellCD[22pt]{b7d7a8}{100.0} &  \cellCD[22pt]{b7d7a8}{99.2} &  \cellCD[22pt]{b7d7a8}{100.0} &  \cellCD[22pt]{b7d7a8}{100.0} &  \cellCD[22pt]{ffffff}{92.0} &  \cellCD[22pt]{b7d7a8}{98.3} &  \cellCD[22pt]{b7d7a8}{99.4} &  \cellCD[22pt]{ffffff}{78.3} &  \cellCD[22pt]{b7d7a8}{95.1} &  \cellCD[22pt]{69a84f}{98.2} \\
tnt\_intrmdt\_M60 &  313 &  \cellCD[24pt]{69a84f}{37} &  \cellCD[24pt]{b7d7a8}{329} &  \cellCD[24pt]{ffffff}{873} &  \cellCD[22pt]{ffffff}{4.4e-5} &  \cellCD[22pt]{b7d7a8}{1.2e-5} &  \cellCD[22pt]{69a84f}{6.2e-6} &  \cellCD[22pt]{b7d7a8}{99.9} &  \cellCD[22pt]{b7d7a8}{100.0} &  \cellCD[22pt]{b7d7a8}{100.0} &  \cellCD[22pt]{b7d7a8}{100.0} &  \cellCD[22pt]{b7d7a8}{100.0} &  \cellCD[22pt]{b7d7a8}{100.0} &  \cellCD[22pt]{b7d7a8}{99.5} &  \cellCD[22pt]{b7d7a8}{100.0} &  \cellCD[22pt]{b7d7a8}{100.0} &  \cellCD[22pt]{b7d7a8}{100.0} &  \cellCD[22pt]{b7d7a8}{100.0} &  \cellCD[22pt]{b7d7a8}{100.0} &  \cellCD[22pt]{ffffff}{96.4} &  \cellCD[22pt]{b7d7a8}{99.1} &  \cellCD[22pt]{b7d7a8}{99.7} &  \cellCD[22pt]{ffffff}{89.5} &  \cellCD[22pt]{b7d7a8}{97.3} &  \cellCD[22pt]{b7d7a8}{99.0} \\
tnt\_intrmdt\_Panther &  314 &  \cellCD[24pt]{69a84f}{43} &  \cellCD[24pt]{b7d7a8}{390} &  \cellCD[24pt]{ffffff}{989} &  \cellCD[22pt]{b7d7a8}{6.4e-5} &  \cellCD[22pt]{b7d7a8}{6.4e-5} &  \cellCD[22pt]{ffffff}{1.5e-4} &  \cellCD[22pt]{b7d7a8}{99.9} &  \cellCD[22pt]{b7d7a8}{100.0} &  \cellCD[22pt]{b7d7a8}{99.9} &  \cellCD[22pt]{b7d7a8}{100.0} &  \cellCD[22pt]{b7d7a8}{100.0} &  \cellCD[22pt]{b7d7a8}{100.0} &  \cellCD[22pt]{b7d7a8}{99.0} &  \cellCD[22pt]{b7d7a8}{99.7} &  \cellCD[22pt]{b7d7a8}{97.4} &  \cellCD[22pt]{b7d7a8}{100.0} &  \cellCD[22pt]{b7d7a8}{100.0} &  \cellCD[22pt]{b7d7a8}{100.0} &  \cellCD[22pt]{b7d7a8}{94.1} &  \cellCD[22pt]{69a84f}{97.3} &  \cellCD[22pt]{b7d7a8}{94.8} &  \cellCD[22pt]{ffffff}{82.9} &  \cellCD[22pt]{69a84f}{92.1} &  \cellCD[22pt]{b7d7a8}{85.8} \\
tnt\_intrmdt\_Playground &  307 &  \cellCD[24pt]{69a84f}{41} &  \cellCD[24pt]{b7d7a8}{473} &  \cellCD[24pt]{ffffff}{1255} &  \cellCD[22pt]{b7d7a8}{1.4e-4} &  \cellCD[22pt]{69a84f}{1.8e-5} &  \cellCD[22pt]{ffffff}{1.9e-3} &  \cellCD[22pt]{b7d7a8}{100.0} &  \cellCD[22pt]{b7d7a8}{100.0} &  \cellCD[22pt]{b7d7a8}{98.7} &  \cellCD[22pt]{b7d7a8}{100.0} &  \cellCD[22pt]{b7d7a8}{100.0} &  \cellCD[22pt]{b7d7a8}{98.7} &  \cellCD[22pt]{b7d7a8}{99.6} &  \cellCD[22pt]{b7d7a8}{99.9} &  \cellCD[22pt]{b7d7a8}{98.7} &  \cellCD[22pt]{b7d7a8}{97.3} &  \cellCD[22pt]{b7d7a8}{100.0} &  \cellCD[22pt]{b7d7a8}{98.7} &  \cellCD[22pt]{ffffff}{87.8} &  \cellCD[22pt]{b7d7a8}{99.2} &  \cellCD[22pt]{b7d7a8}{98.3} &  \cellCD[22pt]{ffffff}{63.9} &  \cellCD[22pt]{b7d7a8}{97.5} &  \cellCD[22pt]{b7d7a8}{97.5} \\
tnt\_intrmdt\_Train &  301 &  \cellCD[24pt]{69a84f}{41} &  \cellCD[24pt]{b7d7a8}{414} &  \cellCD[24pt]{ffffff}{901} &  \cellCD[22pt]{ffffff}{8.0e-5} &  \cellCD[22pt]{b7d7a8}{7.8e-6} &  \cellCD[22pt]{69a84f}{4.3e-6} &  \cellCD[22pt]{b7d7a8}{99.9} &  \cellCD[22pt]{b7d7a8}{100.0} &  \cellCD[22pt]{b7d7a8}{100.0} &  \cellCD[22pt]{b7d7a8}{100.0} &  \cellCD[22pt]{b7d7a8}{100.0} &  \cellCD[22pt]{b7d7a8}{100.0} &  \cellCD[22pt]{b7d7a8}{99.5} &  \cellCD[22pt]{b7d7a8}{99.9} &  \cellCD[22pt]{b7d7a8}{100.0} &  \cellCD[22pt]{b7d7a8}{100.0} &  \cellCD[22pt]{b7d7a8}{100.0} &  \cellCD[22pt]{b7d7a8}{100.0} &  \cellCD[22pt]{ffffff}{94.1} &  \cellCD[22pt]{b7d7a8}{99.4} &  \cellCD[22pt]{b7d7a8}{99.8} &  \cellCD[22pt]{ffffff}{82.7} &  \cellCD[22pt]{b7d7a8}{98.2} &  \cellCD[22pt]{b7d7a8}{99.3} \\
tnt\_trng\_Barn &  410 &  \cellCD[24pt]{69a84f}{50} &  \cellCD[24pt]{b7d7a8}{503} &  \cellCD[24pt]{ffffff}{3126} &  \cellCD[22pt]{ffffff}{1.0e-4} &  \cellCD[22pt]{b7d7a8}{6.7e-6} &  \cellCD[22pt]{b7d7a8}{4.6e-6} &  \cellCD[22pt]{b7d7a8}{100.0} &  \cellCD[22pt]{b7d7a8}{100.0} &  \cellCD[22pt]{b7d7a8}{100.0} &  \cellCD[22pt]{b7d7a8}{100.0} &  \cellCD[22pt]{b7d7a8}{100.0} &  \cellCD[22pt]{b7d7a8}{100.0} &  \cellCD[22pt]{b7d7a8}{99.7} &  \cellCD[22pt]{b7d7a8}{100.0} &  \cellCD[22pt]{b7d7a8}{100.0} &  \cellCD[22pt]{b7d7a8}{100.0} &  \cellCD[22pt]{b7d7a8}{100.0} &  \cellCD[22pt]{b7d7a8}{100.0} &  \cellCD[22pt]{ffffff}{94.2} &  \cellCD[22pt]{b7d7a8}{99.4} &  \cellCD[22pt]{b7d7a8}{99.7} &  \cellCD[22pt]{ffffff}{82.9} &  \cellCD[22pt]{b7d7a8}{98.2} &  \cellCD[22pt]{b7d7a8}{99.2} \\
tnt\_trng\_Caterpillar &  383 &  \cellCD[24pt]{69a84f}{44} &  \cellCD[24pt]{b7d7a8}{374} &  \cellCD[24pt]{ffffff}{1367} &  \cellCD[22pt]{ffffff}{5.1e-5} &  \cellCD[22pt]{b7d7a8}{5.6e-6} &  \cellCD[22pt]{b7d7a8}{4.1e-6} &  \cellCD[22pt]{b7d7a8}{99.8} &  \cellCD[22pt]{b7d7a8}{100.0} &  \cellCD[22pt]{b7d7a8}{100.0} &  \cellCD[22pt]{b7d7a8}{100.0} &  \cellCD[22pt]{b7d7a8}{100.0} &  \cellCD[22pt]{b7d7a8}{100.0} &  \cellCD[22pt]{b7d7a8}{99.5} &  \cellCD[22pt]{b7d7a8}{99.9} &  \cellCD[22pt]{b7d7a8}{99.9} &  \cellCD[22pt]{b7d7a8}{100.0} &  \cellCD[22pt]{b7d7a8}{100.0} &  \cellCD[22pt]{b7d7a8}{100.0} &  \cellCD[22pt]{ffffff}{96.2} &  \cellCD[22pt]{b7d7a8}{99.4} &  \cellCD[22pt]{b7d7a8}{99.7} &  \cellCD[22pt]{ffffff}{89.0} &  \cellCD[22pt]{b7d7a8}{98.4} &  \cellCD[22pt]{b7d7a8}{99.3} \\
tnt\_trng\_Church &  507 &  \cellCD[24pt]{69a84f}{85} &  \cellCD[24pt]{b7d7a8}{666} &  \cellCD[24pt]{ffffff}{4589} &  \cellCD[22pt]{b7d7a8}{8.8e-3} &  \cellCD[22pt]{ffffff}{2.6e-2} &  \cellCD[22pt]{69a84f}{8.0e-4} &  \cellCD[22pt]{b7d7a8}{75.6} &  \cellCD[22pt]{ffffff}{71.1} &  \cellCD[22pt]{69a84f}{99.6} &  \cellCD[22pt]{b7d7a8}{94.8} &  \cellCD[22pt]{ffffff}{75.4} &  \cellCD[22pt]{69a84f}{100.0} &  \cellCD[22pt]{ffffff}{57.4} &  \cellCD[22pt]{b7d7a8}{70.3} &  \cellCD[22pt]{69a84f}{99.3} &  \cellCD[22pt]{b7d7a8}{72.4} &  \cellCD[22pt]{b7d7a8}{70.8} &  \cellCD[22pt]{69a84f}{100.0} &  \cellCD[22pt]{ffffff}{52.3} &  \cellCD[22pt]{b7d7a8}{69.3} &  \cellCD[22pt]{69a84f}{98.4} &  \cellCD[22pt]{ffffff}{19.5} &  \cellCD[22pt]{b7d7a8}{66.6} &  \cellCD[22pt]{69a84f}{96.1} \\
tnt\_trng\_Courthouse & 1106 &  \cellCD[24pt]{69a84f}{169} &  \cellCD[24pt]{b7d7a8}{1297} &  \cellCD[24pt]{ffffff}{8285} &  \cellCD[22pt]{b7d7a8}{1.2e-2} &  \cellCD[22pt]{b7d7a8}{1.7e-2} &  \cellCD[22pt]{69a84f}{1.3e-3} &  \cellCD[22pt]{ffffff}{40.9} &  \cellCD[22pt]{b7d7a8}{97.3} &  \cellCD[22pt]{69a84f}{99.8} &  \cellCD[22pt]{ffffff}{71.4} &  \cellCD[22pt]{b7d7a8}{97.3} &  \cellCD[22pt]{69a84f}{99.8} &  \cellCD[22pt]{ffffff}{35.9} &  \cellCD[22pt]{b7d7a8}{97.3} &  \cellCD[22pt]{69a84f}{99.8} &  \cellCD[22pt]{ffffff}{67.3} &  \cellCD[22pt]{b7d7a8}{97.3} &  \cellCD[22pt]{69a84f}{99.8} &  \cellCD[22pt]{ffffff}{33.7} &  \cellCD[22pt]{b7d7a8}{96.5} &  \cellCD[22pt]{69a84f}{99.4} &  \cellCD[22pt]{ffffff}{24.0} &  \cellCD[22pt]{b7d7a8}{95.0} &  \cellCD[22pt]{69a84f}{98.6} \\
tnt\_trng\_Ignatius &  263 &  \cellCD[24pt]{69a84f}{33} &  \cellCD[24pt]{b7d7a8}{269} &  \cellCD[24pt]{ffffff}{682} &  \cellCD[22pt]{ffffff}{5.2e-5} &  \cellCD[22pt]{b7d7a8}{7.6e-6} &  \cellCD[22pt]{69a84f}{2.1e-6} &  \cellCD[22pt]{b7d7a8}{100.0} &  \cellCD[22pt]{b7d7a8}{100.0} &  \cellCD[22pt]{b7d7a8}{100.0} &  \cellCD[22pt]{b7d7a8}{100.0} &  \cellCD[22pt]{b7d7a8}{100.0} &  \cellCD[22pt]{b7d7a8}{100.0} &  \cellCD[22pt]{b7d7a8}{99.8} &  \cellCD[22pt]{b7d7a8}{100.0} &  \cellCD[22pt]{b7d7a8}{100.0} &  \cellCD[22pt]{b7d7a8}{100.0} &  \cellCD[22pt]{b7d7a8}{100.0} &  \cellCD[22pt]{b7d7a8}{100.0} &  \cellCD[22pt]{ffffff}{96.9} &  \cellCD[22pt]{b7d7a8}{99.5} &  \cellCD[22pt]{b7d7a8}{99.9} &  \cellCD[22pt]{ffffff}{90.9} &  \cellCD[22pt]{b7d7a8}{98.6} &  \cellCD[22pt]{b7d7a8}{99.8} \\
tnt\_trng\_Meetingroom &  371 &  \cellCD[24pt]{69a84f}{32} &  \cellCD[24pt]{b7d7a8}{209} &  \cellCD[24pt]{ffffff}{575} &  \cellCD[22pt]{ffffff}{3.1e-4} &  \cellCD[22pt]{b7d7a8}{1.4e-5} &  \cellCD[22pt]{69a84f}{8.3e-6} &  \cellCD[22pt]{b7d7a8}{98.5} &  \cellCD[22pt]{b7d7a8}{100.0} &  \cellCD[22pt]{b7d7a8}{100.0} &  \cellCD[22pt]{b7d7a8}{100.0} &  \cellCD[22pt]{b7d7a8}{100.0} &  \cellCD[22pt]{b7d7a8}{100.0} &  \cellCD[22pt]{ffffff}{82.5} &  \cellCD[22pt]{b7d7a8}{99.9} &  \cellCD[22pt]{b7d7a8}{100.0} &  \cellCD[22pt]{ffffff}{84.4} &  \cellCD[22pt]{b7d7a8}{100.0} &  \cellCD[22pt]{b7d7a8}{100.0} &  \cellCD[22pt]{ffffff}{72.0} &  \cellCD[22pt]{b7d7a8}{99.0} &  \cellCD[22pt]{b7d7a8}{99.4} &  \cellCD[22pt]{ffffff}{30.5} &  \cellCD[22pt]{b7d7a8}{97.0} &  \cellCD[22pt]{b7d7a8}{98.2} \\
tnt\_trng\_Truck &  251 &  \cellCD[24pt]{69a84f}{31} &  \cellCD[24pt]{b7d7a8}{289} &  \cellCD[24pt]{ffffff}{635} &  \cellCD[22pt]{b7d7a8}{6.9e-5} &  \cellCD[22pt]{ffffff}{3.4e-2} &  \cellCD[22pt]{69a84f}{2.5e-6} &  \cellCD[22pt]{b7d7a8}{100.0} &  \cellCD[22pt]{ffffff}{52.7} &  \cellCD[22pt]{b7d7a8}{100.0} &  \cellCD[22pt]{b7d7a8}{100.0} &  \cellCD[22pt]{ffffff}{52.6} &  \cellCD[22pt]{b7d7a8}{100.0} &  \cellCD[22pt]{b7d7a8}{99.8} &  \cellCD[22pt]{ffffff}{52.6} &  \cellCD[22pt]{b7d7a8}{100.0} &  \cellCD[22pt]{b7d7a8}{100.0} &  \cellCD[22pt]{ffffff}{52.6} &  \cellCD[22pt]{b7d7a8}{100.0} &  \cellCD[22pt]{b7d7a8}{95.4} &  \cellCD[22pt]{ffffff}{51.8} &  \cellCD[22pt]{69a84f}{99.9} &  \cellCD[22pt]{b7d7a8}{86.4} &  \cellCD[22pt]{ffffff}{50.3} &  \cellCD[22pt]{69a84f}{99.6} \\
\bottomrule
\end{tabular}

    }
    \caption{Per-scene camera pose metrics on the Tanks and Temples dataset.}
    \label{tab:fastmap-supp_pose_tnt}
\end{table*}

\begin{table*}[t]
    \centering
    \resizebox{1.0\linewidth}{!}{
        \tablestyle{2pt}{1.1}
\resizebox{\textwidth}{!}{%
\begin{tabular}{l r  r@{~}r@{~}r | r@{~}r@{~}r | r@{~}r@{~}r | r@{~}r@{~}r | r@{~}r@{~}r | r@{~}r@{~}r | r@{~}r@{~}r | r@{~}r@{~}r }

\toprule

  &  & \multicolumn{3}{c}{time (sec)} &
\multicolumn{3}{c}{ATE$\downarrow$} &
\multicolumn{3}{c}{RTA@3$\uparrow$} &
\multicolumn{3}{c}{RRA@3$\uparrow$} &
\multicolumn{3}{c}{RTA@1$\uparrow$} &
\multicolumn{3}{c}{RRA@1$\uparrow$} &
\multicolumn{3}{c}{AUC-R\&T @ 3 $\uparrow$} &
\multicolumn{3}{c}{AUC-R\&T @ 1 $\uparrow$} \\

\cmidrule(lr){3-5} \cmidrule(lr){6-8} \cmidrule(lr){9-11}
\cmidrule(lr){12-14} \cmidrule(lr){15-17} \cmidrule(lr){18-20}
\cmidrule(lr){21-23} \cmidrule(lr){24-26}

   & n\_imgs &
\multicolumn{1}{c}{\scriptsize \alg} & \multicolumn{1}{c}{\scriptsize GLOMAP} & \multicolumn{1}{c}{\scriptsize COLMAP} &
\multicolumn{1}{c}{\scriptsize \alg} & \multicolumn{1}{c}{\scriptsize GLOMAP} & \multicolumn{1}{c}{\scriptsize COLMAP} &
\multicolumn{1}{c}{\scriptsize \alg} & \multicolumn{1}{c}{\scriptsize GLOMAP} & \multicolumn{1}{c}{\scriptsize COLMAP} &
\multicolumn{1}{c}{\scriptsize \alg} & \multicolumn{1}{c}{\scriptsize GLOMAP} & \multicolumn{1}{c}{\scriptsize COLMAP} &
\multicolumn{1}{c}{\scriptsize \alg} & \multicolumn{1}{c}{\scriptsize GLOMAP} & \multicolumn{1}{c}{\scriptsize COLMAP} &
\multicolumn{1}{c}{\scriptsize \alg} & \multicolumn{1}{c}{\scriptsize GLOMAP} & \multicolumn{1}{c}{\scriptsize COLMAP} &
\multicolumn{1}{c}{\scriptsize \alg} & \multicolumn{1}{c}{\scriptsize GLOMAP} & \multicolumn{1}{c}{\scriptsize COLMAP} &
\multicolumn{1}{c}{\scriptsize \alg} & \multicolumn{1}{c}{\scriptsize GLOMAP} & \multicolumn{1}{c}{\scriptsize COLMAP} \\

\midrule
   nosr\_europa &  309 &  \cellCD[24pt]{69a84f}{33} &  \cellCD[24pt]{b7d7a8}{239} &  \cellCD[24pt]{ffffff}{3387} &  \cellCD[22pt]{b7d7a8}{9.3e-4} &  \cellCD[22pt]{b7d7a8}{9.0e-4} &  \cellCD[22pt]{b7d7a8}{8.8e-4} &  \cellCD[22pt]{b7d7a8}{88.7} &  \cellCD[22pt]{b7d7a8}{88.8} &  \cellCD[22pt]{b7d7a8}{89.4} &  \cellCD[22pt]{b7d7a8}{100.0} &  \cellCD[22pt]{b7d7a8}{100.0} &  \cellCD[22pt]{b7d7a8}{100.0} &  \cellCD[22pt]{b7d7a8}{59.2} &  \cellCD[22pt]{b7d7a8}{60.1} &  \cellCD[22pt]{b7d7a8}{60.2} &  \cellCD[22pt]{b7d7a8}{98.5} &  \cellCD[22pt]{b7d7a8}{97.2} &  \cellCD[22pt]{b7d7a8}{98.3} &  \cellCD[22pt]{b7d7a8}{63.2} &  \cellCD[22pt]{b7d7a8}{63.5} &  \cellCD[22pt]{b7d7a8}{63.9} &  \cellCD[22pt]{b7d7a8}{33.7} &  \cellCD[22pt]{b7d7a8}{33.7} &  \cellCD[22pt]{b7d7a8}{34.1} \\
      nosr\_lk2 &  199 &  \cellCD[24pt]{69a84f}{32} &  \cellCD[24pt]{b7d7a8}{117} &  \cellCD[24pt]{ffffff}{570} &  \cellCD[22pt]{ffffff}{1.7e-3} &  \cellCD[22pt]{b7d7a8}{5.0e-4} &  \cellCD[22pt]{b7d7a8}{4.8e-4} &  \cellCD[22pt]{b7d7a8}{96.3} &  \cellCD[22pt]{b7d7a8}{97.2} &  \cellCD[22pt]{b7d7a8}{97.2} &  \cellCD[22pt]{b7d7a8}{100.0} &  \cellCD[22pt]{b7d7a8}{100.0} &  \cellCD[22pt]{b7d7a8}{100.0} &  \cellCD[22pt]{b7d7a8}{87.4} &  \cellCD[22pt]{b7d7a8}{88.5} &  \cellCD[22pt]{b7d7a8}{88.2} &  \cellCD[22pt]{ffffff}{96.7} &  \cellCD[22pt]{b7d7a8}{100.0} &  \cellCD[22pt]{b7d7a8}{99.0} &  \cellCD[22pt]{b7d7a8}{83.4} &  \cellCD[22pt]{b7d7a8}{84.8} &  \cellCD[22pt]{b7d7a8}{84.3} &  \cellCD[22pt]{b7d7a8}{63.0} &  \cellCD[22pt]{b7d7a8}{65.1} &  \cellCD[22pt]{b7d7a8}{64.0} \\
      nosr\_lwp &  354 &  \cellCD[24pt]{69a84f}{41} &  \cellCD[24pt]{b7d7a8}{202} &  \cellCD[24pt]{ffffff}{2020} &  \cellCD[22pt]{b7d7a8}{5.9e-4} &  \cellCD[22pt]{b7d7a8}{6.0e-4} &  \cellCD[22pt]{b7d7a8}{5.9e-4} &  \cellCD[22pt]{b7d7a8}{96.0} &  \cellCD[22pt]{b7d7a8}{96.1} &  \cellCD[22pt]{b7d7a8}{96.2} &  \cellCD[22pt]{b7d7a8}{100.0} &  \cellCD[22pt]{b7d7a8}{100.0} &  \cellCD[22pt]{b7d7a8}{100.0} &  \cellCD[22pt]{b7d7a8}{74.5} &  \cellCD[22pt]{b7d7a8}{75.7} &  \cellCD[22pt]{b7d7a8}{76.2} &  \cellCD[22pt]{b7d7a8}{100.0} &  \cellCD[22pt]{b7d7a8}{100.0} &  \cellCD[22pt]{b7d7a8}{100.0} &  \cellCD[22pt]{b7d7a8}{74.2} &  \cellCD[22pt]{b7d7a8}{75.9} &  \cellCD[22pt]{b7d7a8}{76.2} &  \cellCD[22pt]{ffffff}{43.5} &  \cellCD[22pt]{b7d7a8}{48.3} &  \cellCD[22pt]{b7d7a8}{48.4} \\
  nosr\_rathaus &  515 &  \cellCD[24pt]{69a84f}{76} &  \cellCD[24pt]{b7d7a8}{593} &  \cellCD[24pt]{ffffff}{5965} &  \cellCD[22pt]{b7d7a8}{5.4e-4} &  \cellCD[22pt]{b7d7a8}{5.1e-4} &  \cellCD[22pt]{b7d7a8}{5.1e-4} &  \cellCD[22pt]{b7d7a8}{88.3} &  \cellCD[22pt]{b7d7a8}{88.1} &  \cellCD[22pt]{b7d7a8}{88.6} &  \cellCD[22pt]{b7d7a8}{99.9} &  \cellCD[22pt]{b7d7a8}{100.0} &  \cellCD[22pt]{b7d7a8}{100.0} &  \cellCD[22pt]{b7d7a8}{54.9} &  \cellCD[22pt]{b7d7a8}{55.1} &  \cellCD[22pt]{b7d7a8}{56.5} &  \cellCD[22pt]{b7d7a8}{99.5} &  \cellCD[22pt]{b7d7a8}{100.0} &  \cellCD[22pt]{b7d7a8}{100.0} &  \cellCD[22pt]{b7d7a8}{59.9} &  \cellCD[22pt]{b7d7a8}{60.5} &  \cellCD[22pt]{b7d7a8}{61.2} &  \cellCD[22pt]{b7d7a8}{26.9} &  \cellCD[22pt]{b7d7a8}{28.4} &  \cellCD[22pt]{b7d7a8}{28.8} \\
  nosr\_schloss &  379 &  \cellCD[24pt]{69a84f}{44} &  \cellCD[24pt]{b7d7a8}{320} &  \cellCD[24pt]{ffffff}{2675} &  \cellCD[22pt]{b7d7a8}{6.1e-4} &  \cellCD[22pt]{b7d7a8}{6.1e-4} &  \cellCD[22pt]{b7d7a8}{5.8e-4} &  \cellCD[22pt]{b7d7a8}{92.2} &  \cellCD[22pt]{b7d7a8}{92.1} &  \cellCD[22pt]{b7d7a8}{92.2} &  \cellCD[22pt]{b7d7a8}{100.0} &  \cellCD[22pt]{b7d7a8}{100.0} &  \cellCD[22pt]{b7d7a8}{100.0} &  \cellCD[22pt]{b7d7a8}{72.7} &  \cellCD[22pt]{b7d7a8}{73.2} &  \cellCD[22pt]{b7d7a8}{73.8} &  \cellCD[22pt]{b7d7a8}{99.9} &  \cellCD[22pt]{b7d7a8}{100.0} &  \cellCD[22pt]{b7d7a8}{100.0} &  \cellCD[22pt]{b7d7a8}{71.7} &  \cellCD[22pt]{b7d7a8}{72.3} &  \cellCD[22pt]{b7d7a8}{72.6} &  \cellCD[22pt]{b7d7a8}{43.3} &  \cellCD[22pt]{b7d7a8}{44.8} &  \cellCD[22pt]{b7d7a8}{45.1} \\
       nosr\_st &  397 &  \cellCD[24pt]{69a84f}{40} &  \cellCD[24pt]{b7d7a8}{262} &  \cellCD[24pt]{ffffff}{1435} &  \cellCD[22pt]{b7d7a8}{2.9e-3} &  \cellCD[22pt]{b7d7a8}{3.0e-3} &  \cellCD[22pt]{b7d7a8}{2.9e-3} &  \cellCD[22pt]{b7d7a8}{96.5} &  \cellCD[22pt]{b7d7a8}{96.2} &  \cellCD[22pt]{b7d7a8}{96.0} &  \cellCD[22pt]{b7d7a8}{98.0} &  \cellCD[22pt]{b7d7a8}{98.5} &  \cellCD[22pt]{b7d7a8}{98.0} &  \cellCD[22pt]{b7d7a8}{86.9} &  \cellCD[22pt]{b7d7a8}{86.4} &  \cellCD[22pt]{ffffff}{83.4} &  \cellCD[22pt]{b7d7a8}{96.8} &  \cellCD[22pt]{b7d7a8}{95.6} &  \cellCD[22pt]{b7d7a8}{96.7} &  \cellCD[22pt]{b7d7a8}{80.1} &  \cellCD[22pt]{b7d7a8}{80.3} &  \cellCD[22pt]{b7d7a8}{78.6} &  \cellCD[22pt]{b7d7a8}{53.3} &  \cellCD[22pt]{b7d7a8}{54.8} &  \cellCD[22pt]{b7d7a8}{51.5} \\
  nosr\_stjacob &  722 &  \cellCD[24pt]{69a84f}{80} &  \cellCD[24pt]{b7d7a8}{564} &  \cellCD[24pt]{ffffff}{6267} &  \cellCD[22pt]{b7d7a8}{4.9e-3} &  \cellCD[22pt]{69a84f}{1.7e-3} &  \cellCD[22pt]{b7d7a8}{3.3e-3} &  \cellCD[22pt]{b7d7a8}{90.4} &  \cellCD[22pt]{b7d7a8}{93.0} &  \cellCD[22pt]{b7d7a8}{92.2} &  \cellCD[22pt]{b7d7a8}{97.0} &  \cellCD[22pt]{b7d7a8}{99.7} &  \cellCD[22pt]{b7d7a8}{98.9} &  \cellCD[22pt]{b7d7a8}{75.7} &  \cellCD[22pt]{b7d7a8}{78.0} &  \cellCD[22pt]{b7d7a8}{76.8} &  \cellCD[22pt]{ffffff}{96.8} &  \cellCD[22pt]{b7d7a8}{99.4} &  \cellCD[22pt]{b7d7a8}{98.9} &  \cellCD[22pt]{b7d7a8}{73.1} &  \cellCD[22pt]{b7d7a8}{75.9} &  \cellCD[22pt]{b7d7a8}{74.8} &  \cellCD[22pt]{ffffff}{47.3} &  \cellCD[22pt]{b7d7a8}{50.6} &  \cellCD[22pt]{b7d7a8}{49.5} \\
 nosr\_stjohann &  347 &  \cellCD[24pt]{69a84f}{50} &  \cellCD[24pt]{b7d7a8}{296} &  \cellCD[24pt]{ffffff}{2986} &  \cellCD[22pt]{b7d7a8}{7.8e-4} &  \cellCD[22pt]{b7d7a8}{7.9e-4} &  \cellCD[22pt]{b7d7a8}{7.9e-4} &  \cellCD[22pt]{b7d7a8}{84.9} &  \cellCD[22pt]{b7d7a8}{84.7} &  \cellCD[22pt]{b7d7a8}{85.1} &  \cellCD[22pt]{b7d7a8}{100.0} &  \cellCD[22pt]{b7d7a8}{100.0} &  \cellCD[22pt]{b7d7a8}{100.0} &  \cellCD[22pt]{b7d7a8}{57.8} &  \cellCD[22pt]{b7d7a8}{57.9} &  \cellCD[22pt]{b7d7a8}{58.3} &  \cellCD[22pt]{b7d7a8}{99.3} &  \cellCD[22pt]{b7d7a8}{99.4} &  \cellCD[22pt]{b7d7a8}{99.4} &  \cellCD[22pt]{b7d7a8}{61.6} &  \cellCD[22pt]{b7d7a8}{61.9} &  \cellCD[22pt]{b7d7a8}{62.1} &  \cellCD[22pt]{b7d7a8}{34.7} &  \cellCD[22pt]{b7d7a8}{36.1} &  \cellCD[22pt]{b7d7a8}{35.8} \\
\bottomrule
\end{tabular}
}

    }
    \caption{Per-scene camera pose metrics on the NeRF-OSR dataset.}
    \label{tab:fastmap-supp_pose_nosr}
\end{table*}

\begin{table*}[t]
    \centering
    \resizebox{1.0\linewidth}{!}{
        \tablestyle{2pt}{1.1}
\resizebox{\textwidth}{!}{%
\begin{tabular}{l r  r@{~}r@{~}r | r@{~}r@{~}r | r@{~}r@{~}r | r@{~}r@{~}r | r@{~}r@{~}r | r@{~}r@{~}r | r@{~}r@{~}r | r@{~}r@{~}r }

\toprule

  &  & \multicolumn{3}{c}{time (sec)} &
\multicolumn{3}{c}{ATE$\downarrow$} &
\multicolumn{3}{c}{RTA@3$\uparrow$} &
\multicolumn{3}{c}{RRA@3$\uparrow$} &
\multicolumn{3}{c}{RTA@1$\uparrow$} &
\multicolumn{3}{c}{RRA@1$\uparrow$} &
\multicolumn{3}{c}{AUC-R\&T @ 3 $\uparrow$} &
\multicolumn{3}{c}{AUC-R\&T @ 1 $\uparrow$} \\

\cmidrule(lr){3-5} \cmidrule(lr){6-8} \cmidrule(lr){9-11}
\cmidrule(lr){12-14} \cmidrule(lr){15-17} \cmidrule(lr){18-20}
\cmidrule(lr){21-23} \cmidrule(lr){24-26}

   & n\_imgs &
\multicolumn{1}{c}{\scriptsize \alg} & \multicolumn{1}{c}{\scriptsize GLOMAP} & \multicolumn{1}{c}{\scriptsize COLMAP} &
\multicolumn{1}{c}{\scriptsize \alg} & \multicolumn{1}{c}{\scriptsize GLOMAP} & \multicolumn{1}{c}{\scriptsize COLMAP} &
\multicolumn{1}{c}{\scriptsize \alg} & \multicolumn{1}{c}{\scriptsize GLOMAP} & \multicolumn{1}{c}{\scriptsize COLMAP} &
\multicolumn{1}{c}{\scriptsize \alg} & \multicolumn{1}{c}{\scriptsize GLOMAP} & \multicolumn{1}{c}{\scriptsize COLMAP} &
\multicolumn{1}{c}{\scriptsize \alg} & \multicolumn{1}{c}{\scriptsize GLOMAP} & \multicolumn{1}{c}{\scriptsize COLMAP} &
\multicolumn{1}{c}{\scriptsize \alg} & \multicolumn{1}{c}{\scriptsize GLOMAP} & \multicolumn{1}{c}{\scriptsize COLMAP} &
\multicolumn{1}{c}{\scriptsize \alg} & \multicolumn{1}{c}{\scriptsize GLOMAP} & \multicolumn{1}{c}{\scriptsize COLMAP} &
\multicolumn{1}{c}{\scriptsize \alg} & \multicolumn{1}{c}{\scriptsize GLOMAP} & \multicolumn{1}{c}{\scriptsize COLMAP} \\

\midrule
  dploy\_house1 &  220 &  \cellCD[24pt]{69a84f}{34} &  \cellCD[24pt]{b7d7a8}{138} &  \cellCD[24pt]{ffffff}{419} &  \cellCD[22pt]{b7d7a8}{7.5e-5} &  \cellCD[22pt]{b7d7a8}{1.1e-4} &  \cellCD[22pt]{b7d7a8}{1.1e-4} &  \cellCD[22pt]{b7d7a8}{100.0} &  \cellCD[22pt]{b7d7a8}{99.9} &  \cellCD[22pt]{b7d7a8}{99.9} &  \cellCD[22pt]{b7d7a8}{100.0} &  \cellCD[22pt]{b7d7a8}{100.0} &  \cellCD[22pt]{b7d7a8}{100.0} &  \cellCD[22pt]{b7d7a8}{99.5} &  \cellCD[22pt]{b7d7a8}{99.1} &  \cellCD[22pt]{b7d7a8}{99.0} &  \cellCD[22pt]{b7d7a8}{100.0} &  \cellCD[22pt]{b7d7a8}{100.0} &  \cellCD[22pt]{b7d7a8}{100.0} &  \cellCD[22pt]{b7d7a8}{93.9} &  \cellCD[22pt]{b7d7a8}{93.2} &  \cellCD[22pt]{b7d7a8}{92.9} &  \cellCD[22pt]{b7d7a8}{82.1} &  \cellCD[22pt]{b7d7a8}{80.2} &  \cellCD[22pt]{b7d7a8}{79.6} \\
  dploy\_house2 &  725 &  \cellCD[24pt]{69a84f}{124} &  \cellCD[24pt]{b7d7a8}{592} &  \cellCD[24pt]{ffffff}{5702} &  \cellCD[22pt]{b7d7a8}{6.9e-5} &  \cellCD[22pt]{b7d7a8}{7.2e-5} &  \cellCD[22pt]{ffffff}{1.3e-4} &  \cellCD[22pt]{b7d7a8}{99.9} &  \cellCD[22pt]{b7d7a8}{99.9} &  \cellCD[22pt]{b7d7a8}{99.8} &  \cellCD[22pt]{b7d7a8}{100.0} &  \cellCD[22pt]{b7d7a8}{100.0} &  \cellCD[22pt]{b7d7a8}{100.0} &  \cellCD[22pt]{b7d7a8}{99.2} &  \cellCD[22pt]{b7d7a8}{99.3} &  \cellCD[22pt]{ffffff}{95.6} &  \cellCD[22pt]{b7d7a8}{99.6} &  \cellCD[22pt]{b7d7a8}{99.9} &  \cellCD[22pt]{ffffff}{87.3} &  \cellCD[22pt]{b7d7a8}{91.6} &  \cellCD[22pt]{b7d7a8}{91.7} &  \cellCD[22pt]{ffffff}{82.0} &  \cellCD[22pt]{b7d7a8}{75.4} &  \cellCD[22pt]{b7d7a8}{75.6} &  \cellCD[22pt]{ffffff}{51.1} \\
  dploy\_house3 &  180 &  \cellCD[24pt]{69a84f}{50} &  \cellCD[24pt]{b7d7a8}{103} &  \cellCD[24pt]{ffffff}{627} &  \cellCD[22pt]{b7d7a8}{4.6e-3} &  \cellCD[22pt]{ffffff}{1.2e-2} &  \cellCD[22pt]{69a84f}{1.1e-3} &  \cellCD[22pt]{b7d7a8}{95.1} &  \cellCD[22pt]{b7d7a8}{95.9} &  \cellCD[22pt]{b7d7a8}{97.2} &  \cellCD[22pt]{b7d7a8}{94.5} &  \cellCD[22pt]{b7d7a8}{95.6} &  \cellCD[22pt]{69a84f}{97.9} &  \cellCD[22pt]{b7d7a8}{80.7} &  \cellCD[22pt]{69a84f}{85.8} &  \cellCD[22pt]{b7d7a8}{78.7} &  \cellCD[22pt]{b7d7a8}{70.5} &  \cellCD[22pt]{b7d7a8}{71.2} &  \cellCD[22pt]{ffffff}{59.1} &  \cellCD[22pt]{b7d7a8}{66.7} &  \cellCD[22pt]{b7d7a8}{68.0} &  \cellCD[22pt]{ffffff}{58.6} &  \cellCD[22pt]{b7d7a8}{24.7} &  \cellCD[22pt]{b7d7a8}{26.4} &  \cellCD[22pt]{ffffff}{21.3} \\
  dploy\_house4 &  349 &  \cellCD[24pt]{69a84f}{57} &  \cellCD[24pt]{b7d7a8}{128} &  \cellCD[24pt]{ffffff}{751} &  \cellCD[22pt]{b7d7a8}{1.7e-2} &  \cellCD[22pt]{b7d7a8}{1.9e-2} &  \cellCD[22pt]{69a84f}{9.4e-3} &  \cellCD[22pt]{ffffff}{94.8} &  \cellCD[22pt]{b7d7a8}{98.3} &  \cellCD[22pt]{b7d7a8}{98.8} &  \cellCD[22pt]{ffffff}{95.6} &  \cellCD[22pt]{b7d7a8}{97.7} &  \cellCD[22pt]{b7d7a8}{98.9} &  \cellCD[22pt]{ffffff}{90.8} &  \cellCD[22pt]{b7d7a8}{97.8} &  \cellCD[22pt]{b7d7a8}{97.9} &  \cellCD[22pt]{ffffff}{86.2} &  \cellCD[22pt]{b7d7a8}{97.7} &  \cellCD[22pt]{b7d7a8}{98.3} &  \cellCD[22pt]{ffffff}{81.1} &  \cellCD[22pt]{b7d7a8}{89.9} &  \cellCD[22pt]{b7d7a8}{91.0} &  \cellCD[22pt]{ffffff}{62.0} &  \cellCD[22pt]{b7d7a8}{74.5} &  \cellCD[22pt]{69a84f}{76.7} \\
  dploy\_pipes1 &   97 &  \cellCD[24pt]{69a84f}{17} &  \cellCD[24pt]{b7d7a8}{42} &  \cellCD[24pt]{ffffff}{138} &  \cellCD[22pt]{b7d7a8}{6.8e-5} &  \cellCD[22pt]{b7d7a8}{6.2e-5} &  \cellCD[22pt]{b7d7a8}{6.1e-5} &  \cellCD[22pt]{b7d7a8}{100.0} &  \cellCD[22pt]{b7d7a8}{100.0} &  \cellCD[22pt]{b7d7a8}{100.0} &  \cellCD[22pt]{b7d7a8}{100.0} &  \cellCD[22pt]{b7d7a8}{100.0} &  \cellCD[22pt]{b7d7a8}{100.0} &  \cellCD[22pt]{b7d7a8}{99.9} &  \cellCD[22pt]{b7d7a8}{99.9} &  \cellCD[22pt]{b7d7a8}{99.9} &  \cellCD[22pt]{b7d7a8}{100.0} &  \cellCD[22pt]{b7d7a8}{100.0} &  \cellCD[22pt]{b7d7a8}{100.0} &  \cellCD[22pt]{b7d7a8}{95.8} &  \cellCD[22pt]{b7d7a8}{95.8} &  \cellCD[22pt]{b7d7a8}{95.8} &  \cellCD[22pt]{b7d7a8}{87.5} &  \cellCD[22pt]{b7d7a8}{87.3} &  \cellCD[22pt]{b7d7a8}{87.4} \\
  dploy\_ruins2 & 1171 &  \cellCD[24pt]{69a84f}{156} &  \cellCD[24pt]{b7d7a8}{907} &  \cellCD[24pt]{ffffff}{8672} &  \cellCD[22pt]{b7d7a8}{5.2e-5} &  \cellCD[22pt]{b7d7a8}{4.1e-5} &  \cellCD[22pt]{ffffff}{2.0e-4} &  \cellCD[22pt]{b7d7a8}{100.0} &  \cellCD[22pt]{b7d7a8}{100.0} &  \cellCD[22pt]{b7d7a8}{98.3} &  \cellCD[22pt]{b7d7a8}{100.0} &  \cellCD[22pt]{b7d7a8}{100.0} &  \cellCD[22pt]{b7d7a8}{100.0} &  \cellCD[22pt]{b7d7a8}{99.0} &  \cellCD[22pt]{b7d7a8}{99.4} &  \cellCD[22pt]{ffffff}{82.3} &  \cellCD[22pt]{b7d7a8}{98.0} &  \cellCD[22pt]{69a84f}{100.0} &  \cellCD[22pt]{ffffff}{81.6} &  \cellCD[22pt]{b7d7a8}{88.8} &  \cellCD[22pt]{69a84f}{92.1} &  \cellCD[22pt]{ffffff}{71.9} &  \cellCD[22pt]{b7d7a8}{67.2} &  \cellCD[22pt]{69a84f}{76.5} &  \cellCD[22pt]{ffffff}{30.6} \\
  dploy\_ruins3 &  523 &  \cellCD[24pt]{69a84f}{84} &  \cellCD[24pt]{b7d7a8}{324} &  \cellCD[24pt]{ffffff}{4986} &  \cellCD[22pt]{69a84f}{1.2e-3} &  \cellCD[22pt]{ffffff}{5.6e-3} &  \cellCD[22pt]{b7d7a8}{3.5e-3} &  \cellCD[22pt]{69a84f}{98.3} &  \cellCD[22pt]{b7d7a8}{94.9} &  \cellCD[22pt]{b7d7a8}{95.7} &  \cellCD[22pt]{b7d7a8}{98.4} &  \cellCD[22pt]{b7d7a8}{99.3} &  \cellCD[22pt]{ffffff}{83.4} &  \cellCD[22pt]{b7d7a8}{75.4} &  \cellCD[22pt]{69a84f}{79.8} &  \cellCD[22pt]{ffffff}{45.8} &  \cellCD[22pt]{b7d7a8}{39.4} &  \cellCD[22pt]{69a84f}{46.0} &  \cellCD[22pt]{ffffff}{24.1} &  \cellCD[22pt]{b7d7a8}{56.4} &  \cellCD[22pt]{69a84f}{59.0} &  \cellCD[22pt]{ffffff}{39.1} &  \cellCD[22pt]{b7d7a8}{8.9} &  \cellCD[22pt]{69a84f}{10.9} &  \cellCD[22pt]{ffffff}{3.8} \\
  dploy\_tower1 &  775 &  \cellCD[24pt]{69a84f}{188} &  \cellCD[24pt]{b7d7a8}{419} &  \cellCD[24pt]{ffffff}{2809} &  \cellCD[22pt]{ffffff}{2.0e-2} &  \cellCD[22pt]{b7d7a8}{2.0e-3} &  \cellCD[22pt]{b7d7a8}{2.5e-3} &  \cellCD[22pt]{b7d7a8}{93.7} &  \cellCD[22pt]{b7d7a8}{95.2} &  \cellCD[22pt]{ffffff}{87.7} &  \cellCD[22pt]{b7d7a8}{99.2} &  \cellCD[22pt]{b7d7a8}{99.5} &  \cellCD[22pt]{b7d7a8}{98.3} &  \cellCD[22pt]{b7d7a8}{86.4} &  \cellCD[22pt]{b7d7a8}{84.5} &  \cellCD[22pt]{ffffff}{51.5} &  \cellCD[22pt]{b7d7a8}{63.7} &  \cellCD[22pt]{69a84f}{74.4} &  \cellCD[22pt]{ffffff}{37.9} &  \cellCD[22pt]{b7d7a8}{65.6} &  \cellCD[22pt]{b7d7a8}{67.5} &  \cellCD[22pt]{ffffff}{47.2} &  \cellCD[22pt]{b7d7a8}{20.9} &  \cellCD[22pt]{69a84f}{23.7} &  \cellCD[22pt]{ffffff}{10.1} \\
  dploy\_tower2 &  682 &  \cellCD[24pt]{69a84f}{106} &  \cellCD[24pt]{b7d7a8}{634} &  \cellCD[24pt]{ffffff}{6060} &  \cellCD[22pt]{b7d7a8}{1.7e-4} &  \cellCD[22pt]{b7d7a8}{1.6e-4} &  \cellCD[22pt]{ffffff}{1.3e-3} &  \cellCD[22pt]{b7d7a8}{99.9} &  \cellCD[22pt]{b7d7a8}{99.8} &  \cellCD[22pt]{ffffff}{44.6} &  \cellCD[22pt]{b7d7a8}{100.0} &  \cellCD[22pt]{b7d7a8}{100.0} &  \cellCD[22pt]{ffffff}{25.9} &  \cellCD[22pt]{b7d7a8}{75.5} &  \cellCD[22pt]{69a84f}{77.6} &  \cellCD[22pt]{ffffff}{10.6} &  \cellCD[22pt]{b7d7a8}{100.0} &  \cellCD[22pt]{b7d7a8}{100.0} &  \cellCD[22pt]{ffffff}{9.6} &  \cellCD[22pt]{b7d7a8}{72.7} &  \cellCD[22pt]{b7d7a8}{72.9} &  \cellCD[22pt]{ffffff}{8.4} &  \cellCD[22pt]{b7d7a8}{24.5} &  \cellCD[22pt]{b7d7a8}{26.2} &  \cellCD[22pt]{f4cccc}{0.8} \\
\bottomrule
\end{tabular}
}

    }
    \caption{Per-scene camera pose metrics on the DroneDeploy dataset.}
    \label{tab:fastmap-supp_pose_dploy}
\end{table*}

\begin{table*}[t]
    \centering
    \resizebox{1.0\linewidth}{!}{
        \tablestyle{2pt}{1.1}
\resizebox{\textwidth}{!}{%
\begin{tabular}{l r  r@{~}r@{~}r | r@{~}r@{~}r | r@{~}r@{~}r | r@{~}r@{~}r | r@{~}r@{~}r | r@{~}r@{~}r | r@{~}r@{~}r | r@{~}r@{~}r }

\toprule

  &  & \multicolumn{3}{c}{time (sec)} &
\multicolumn{3}{c}{ATE$\downarrow$} &
\multicolumn{3}{c}{RTA@3$\uparrow$} &
\multicolumn{3}{c}{RRA@3$\uparrow$} &
\multicolumn{3}{c}{RTA@1$\uparrow$} &
\multicolumn{3}{c}{RRA@1$\uparrow$} &
\multicolumn{3}{c}{AUC-R\&T @ 3 $\uparrow$} &
\multicolumn{3}{c}{AUC-R\&T @ 1 $\uparrow$} \\

\cmidrule(lr){3-5} \cmidrule(lr){6-8} \cmidrule(lr){9-11}
\cmidrule(lr){12-14} \cmidrule(lr){15-17} \cmidrule(lr){18-20}
\cmidrule(lr){21-23} \cmidrule(lr){24-26}

   & n\_imgs &
\multicolumn{1}{c}{\scriptsize \alg} & \multicolumn{1}{c}{\scriptsize GLOMAP} & \multicolumn{1}{c}{\scriptsize COLMAP} &
\multicolumn{1}{c}{\scriptsize \alg} & \multicolumn{1}{c}{\scriptsize GLOMAP} & \multicolumn{1}{c}{\scriptsize COLMAP} &
\multicolumn{1}{c}{\scriptsize \alg} & \multicolumn{1}{c}{\scriptsize GLOMAP} & \multicolumn{1}{c}{\scriptsize COLMAP} &
\multicolumn{1}{c}{\scriptsize \alg} & \multicolumn{1}{c}{\scriptsize GLOMAP} & \multicolumn{1}{c}{\scriptsize COLMAP} &
\multicolumn{1}{c}{\scriptsize \alg} & \multicolumn{1}{c}{\scriptsize GLOMAP} & \multicolumn{1}{c}{\scriptsize COLMAP} &
\multicolumn{1}{c}{\scriptsize \alg} & \multicolumn{1}{c}{\scriptsize GLOMAP} & \multicolumn{1}{c}{\scriptsize COLMAP} &
\multicolumn{1}{c}{\scriptsize \alg} & \multicolumn{1}{c}{\scriptsize GLOMAP} & \multicolumn{1}{c}{\scriptsize COLMAP} &
\multicolumn{1}{c}{\scriptsize \alg} & \multicolumn{1}{c}{\scriptsize GLOMAP} & \multicolumn{1}{c}{\scriptsize COLMAP} \\

\midrule
     z\_alameda & 1734 &  \cellCD[24pt]{69a84f}{134} &  \cellCD[24pt]{b7d7a8}{848} &  \cellCD[24pt]{ffffff}{4541} &  \cellCD[22pt]{ffffff}{1.1e-3} &  \cellCD[22pt]{b7d7a8}{1.9e-4} &  \cellCD[22pt]{69a84f}{6.7e-6} &  \cellCD[22pt]{b7d7a8}{99.1} &  \cellCD[22pt]{b7d7a8}{99.4} &  \cellCD[22pt]{b7d7a8}{100.0} &  \cellCD[22pt]{b7d7a8}{99.7} &  \cellCD[22pt]{b7d7a8}{100.0} &  \cellCD[22pt]{b7d7a8}{100.0} &  \cellCD[22pt]{b7d7a8}{98.9} &  \cellCD[22pt]{b7d7a8}{99.3} &  \cellCD[22pt]{b7d7a8}{99.9} &  \cellCD[22pt]{b7d7a8}{99.1} &  \cellCD[22pt]{b7d7a8}{99.7} &  \cellCD[22pt]{b7d7a8}{99.9} &  \cellCD[22pt]{ffffff}{95.4} &  \cellCD[22pt]{b7d7a8}{97.8} &  \cellCD[22pt]{b7d7a8}{98.5} &  \cellCD[22pt]{ffffff}{88.1} &  \cellCD[22pt]{b7d7a8}{94.8} &  \cellCD[22pt]{b7d7a8}{95.6} \\
      z\_berlin & 1511 &  \cellCD[24pt]{69a84f}{152} &  \cellCD[24pt]{b7d7a8}{893} &  \cellCD[24pt]{ffffff}{6478} &  \cellCD[22pt]{b7d7a8}{1.0e-3} &  \cellCD[22pt]{ffffff}{1.5e-2} &  \cellCD[22pt]{b7d7a8}{1.1e-3} &  \cellCD[22pt]{b7d7a8}{97.7} &  \cellCD[22pt]{ffffff}{93.0} &  \cellCD[22pt]{b7d7a8}{98.9} &  \cellCD[22pt]{b7d7a8}{96.8} &  \cellCD[22pt]{ffffff}{92.9} &  \cellCD[22pt]{69a84f}{98.9} &  \cellCD[22pt]{b7d7a8}{91.8} &  \cellCD[22pt]{b7d7a8}{92.8} &  \cellCD[22pt]{69a84f}{98.8} &  \cellCD[22pt]{b7d7a8}{94.7} &  \cellCD[22pt]{b7d7a8}{92.8} &  \cellCD[22pt]{69a84f}{98.3} &  \cellCD[22pt]{ffffff}{83.6} &  \cellCD[22pt]{b7d7a8}{90.9} &  \cellCD[22pt]{69a84f}{96.6} &  \cellCD[22pt]{ffffff}{61.0} &  \cellCD[22pt]{b7d7a8}{86.9} &  \cellCD[22pt]{69a84f}{92.4} \\
      z\_london & 1874 &  \cellCD[24pt]{69a84f}{102} &  \cellCD[24pt]{b7d7a8}{566} &  \cellCD[24pt]{ffffff}{2643} &  \cellCD[22pt]{69a84f}{6.6e-5} &  \cellCD[22pt]{ffffff}{1.3e-2} &  \cellCD[22pt]{b7d7a8}{2.8e-4} &  \cellCD[22pt]{b7d7a8}{99.8} &  \cellCD[22pt]{b7d7a8}{99.9} &  \cellCD[22pt]{b7d7a8}{99.9} &  \cellCD[22pt]{b7d7a8}{99.8} &  \cellCD[22pt]{b7d7a8}{99.9} &  \cellCD[22pt]{b7d7a8}{99.9} &  \cellCD[22pt]{b7d7a8}{99.6} &  \cellCD[22pt]{b7d7a8}{99.9} &  \cellCD[22pt]{b7d7a8}{99.9} &  \cellCD[22pt]{b7d7a8}{99.5} &  \cellCD[22pt]{b7d7a8}{99.9} &  \cellCD[22pt]{b7d7a8}{99.9} &  \cellCD[22pt]{ffffff}{96.4} &  \cellCD[22pt]{b7d7a8}{98.6} &  \cellCD[22pt]{b7d7a8}{98.5} &  \cellCD[22pt]{ffffff}{89.9} &  \cellCD[22pt]{b7d7a8}{96.0} &  \cellCD[22pt]{b7d7a8}{95.8} \\
         z\_nyc &  990 &  \cellCD[24pt]{69a84f}{88} &  \cellCD[24pt]{b7d7a8}{451} &  \cellCD[24pt]{ffffff}{1618} &  \cellCD[22pt]{ffffff}{9.7e-3} &  \cellCD[22pt]{b7d7a8}{6.8e-6} &  \cellCD[22pt]{b7d7a8}{5.3e-6} &  \cellCD[22pt]{b7d7a8}{99.4} &  \cellCD[22pt]{b7d7a8}{100.0} &  \cellCD[22pt]{b7d7a8}{100.0} &  \cellCD[22pt]{b7d7a8}{99.4} &  \cellCD[22pt]{b7d7a8}{100.0} &  \cellCD[22pt]{b7d7a8}{100.0} &  \cellCD[22pt]{b7d7a8}{99.1} &  \cellCD[22pt]{b7d7a8}{100.0} &  \cellCD[22pt]{b7d7a8}{100.0} &  \cellCD[22pt]{b7d7a8}{99.4} &  \cellCD[22pt]{b7d7a8}{100.0} &  \cellCD[22pt]{b7d7a8}{100.0} &  \cellCD[22pt]{ffffff}{95.0} &  \cellCD[22pt]{b7d7a8}{98.9} &  \cellCD[22pt]{b7d7a8}{98.9} &  \cellCD[22pt]{ffffff}{86.5} &  \cellCD[22pt]{b7d7a8}{96.8} &  \cellCD[22pt]{b7d7a8}{96.8} \\
mill19\_building & 1920 &  \cellCD[24pt]{69a84f}{258} &  \cellCD[24pt]{b7d7a8}{6289} &  \cellCD[24pt]{ffffff}{27080} &  \cellCD[22pt]{b7d7a8}{3.0e-4} &  \cellCD[22pt]{ffffff}{1.3e-2} &  \cellCD[22pt]{69a84f}{1.9e-5} &  \cellCD[22pt]{b7d7a8}{99.9} &  \cellCD[22pt]{f4cccc}{0.1} &  \cellCD[22pt]{b7d7a8}{99.9} &  \cellCD[22pt]{b7d7a8}{100.0} &  \cellCD[22pt]{ffffff}{7.4} &  \cellCD[22pt]{b7d7a8}{100.0} &  \cellCD[22pt]{b7d7a8}{99.3} &  \cellCD[22pt]{f4cccc}{0.0} &  \cellCD[22pt]{b7d7a8}{99.3} &  \cellCD[22pt]{b7d7a8}{100.0} &  \cellCD[22pt]{f4cccc}{1.9} &  \cellCD[22pt]{b7d7a8}{99.9} &  \cellCD[22pt]{b7d7a8}{95.5} &  \cellCD[22pt]{f4cccc}{0.0} &  \cellCD[22pt]{b7d7a8}{95.6} &  \cellCD[22pt]{b7d7a8}{87.0} &  \cellCD[22pt]{f4cccc}{0.0} &  \cellCD[22pt]{b7d7a8}{87.4} \\
 mill19\_rubble & 1657 &  \cellCD[24pt]{69a84f}{240} &  \cellCD[24pt]{b7d7a8}{2849} &  \cellCD[24pt]{ffffff}{12153} &  \cellCD[22pt]{b7d7a8}{3.6e-5} &  \cellCD[22pt]{ffffff}{6.4e-5} &  \cellCD[22pt]{b7d7a8}{3.4e-5} &  \cellCD[22pt]{b7d7a8}{99.9} &  \cellCD[22pt]{b7d7a8}{99.8} &  \cellCD[22pt]{b7d7a8}{99.9} &  \cellCD[22pt]{b7d7a8}{100.0} &  \cellCD[22pt]{b7d7a8}{99.9} &  \cellCD[22pt]{b7d7a8}{100.0} &  \cellCD[22pt]{b7d7a8}{98.6} &  \cellCD[22pt]{b7d7a8}{98.6} &  \cellCD[22pt]{b7d7a8}{98.7} &  \cellCD[22pt]{b7d7a8}{100.0} &  \cellCD[22pt]{b7d7a8}{99.9} &  \cellCD[22pt]{b7d7a8}{100.0} &  \cellCD[22pt]{b7d7a8}{93.6} &  \cellCD[22pt]{b7d7a8}{94.5} &  \cellCD[22pt]{b7d7a8}{94.6} &  \cellCD[22pt]{ffffff}{81.6} &  \cellCD[22pt]{b7d7a8}{84.7} &  \cellCD[22pt]{b7d7a8}{84.8} \\
   urbn\_Campus & 5871 &  \cellCD[24pt]{69a84f}{740} &  \cellCD[24pt]{b7d7a8}{3869} &  \cellCD[24pt]{ffffff}{106055} &  \cellCD[22pt]{ffffff}{1.1e-5} &  \cellCD[22pt]{b7d7a8}{4.7e-6} &  \cellCD[22pt]{b7d7a8}{5.0e-6} &  \cellCD[22pt]{b7d7a8}{100.0} &  \cellCD[22pt]{b7d7a8}{100.0} &  \cellCD[22pt]{b7d7a8}{100.0} &  \cellCD[22pt]{b7d7a8}{100.0} &  \cellCD[22pt]{b7d7a8}{100.0} &  \cellCD[22pt]{b7d7a8}{100.0} &  \cellCD[22pt]{b7d7a8}{99.9} &  \cellCD[22pt]{b7d7a8}{99.9} &  \cellCD[22pt]{b7d7a8}{99.9} &  \cellCD[22pt]{b7d7a8}{100.0} &  \cellCD[22pt]{b7d7a8}{100.0} &  \cellCD[22pt]{b7d7a8}{100.0} &  \cellCD[22pt]{ffffff}{94.0} &  \cellCD[22pt]{b7d7a8}{98.0} &  \cellCD[22pt]{b7d7a8}{97.9} &  \cellCD[22pt]{ffffff}{81.9} &  \cellCD[22pt]{b7d7a8}{94.1} &  \cellCD[22pt]{b7d7a8}{93.7} \\
urbn\_Residence & 2582 &  \cellCD[24pt]{69a84f}{359} &  \cellCD[24pt]{b7d7a8}{2523} &  \cellCD[24pt]{ffffff}{36778} &  \cellCD[22pt]{b7d7a8}{2.8e-5} &  \cellCD[22pt]{b7d7a8}{2.7e-5} &  \cellCD[22pt]{b7d7a8}{2.6e-5} &  \cellCD[22pt]{b7d7a8}{99.8} &  \cellCD[22pt]{b7d7a8}{99.9} &  \cellCD[22pt]{b7d7a8}{99.9} &  \cellCD[22pt]{b7d7a8}{100.0} &  \cellCD[22pt]{b7d7a8}{100.0} &  \cellCD[22pt]{b7d7a8}{100.0} &  \cellCD[22pt]{b7d7a8}{98.8} &  \cellCD[22pt]{b7d7a8}{98.9} &  \cellCD[22pt]{b7d7a8}{99.0} &  \cellCD[22pt]{b7d7a8}{100.0} &  \cellCD[22pt]{b7d7a8}{100.0} &  \cellCD[22pt]{b7d7a8}{100.0} &  \cellCD[22pt]{b7d7a8}{94.6} &  \cellCD[22pt]{b7d7a8}{95.2} &  \cellCD[22pt]{b7d7a8}{95.4} &  \cellCD[22pt]{b7d7a8}{84.6} &  \cellCD[22pt]{b7d7a8}{86.3} &  \cellCD[22pt]{b7d7a8}{86.8} \\
  urbn\_Sci-Art & 3019 &  \cellCD[24pt]{69a84f}{445} &  \cellCD[24pt]{b7d7a8}{4601} &  \cellCD[24pt]{ffffff}{42032} &  \cellCD[22pt]{b7d7a8}{1.4e-5} &  \cellCD[22pt]{b7d7a8}{1.0e-5} &  \cellCD[22pt]{b7d7a8}{1.1e-5} &  \cellCD[22pt]{b7d7a8}{100.0} &  \cellCD[22pt]{b7d7a8}{100.0} &  \cellCD[22pt]{b7d7a8}{100.0} &  \cellCD[22pt]{b7d7a8}{100.0} &  \cellCD[22pt]{b7d7a8}{100.0} &  \cellCD[22pt]{b7d7a8}{100.0} &  \cellCD[22pt]{b7d7a8}{99.9} &  \cellCD[22pt]{b7d7a8}{99.9} &  \cellCD[22pt]{b7d7a8}{99.9} &  \cellCD[22pt]{b7d7a8}{100.0} &  \cellCD[22pt]{b7d7a8}{100.0} &  \cellCD[22pt]{b7d7a8}{100.0} &  \cellCD[22pt]{b7d7a8}{97.4} &  \cellCD[22pt]{b7d7a8}{97.7} &  \cellCD[22pt]{b7d7a8}{97.8} &  \cellCD[22pt]{b7d7a8}{92.2} &  \cellCD[22pt]{b7d7a8}{93.1} &  \cellCD[22pt]{b7d7a8}{93.5} \\
 eft\_apartment & 3804 &  \cellCD[24pt]{69a84f}{549} &  \cellCD[24pt]{b7d7a8}{5905} &  \cellCD[24pt]{ffffff}{185361} &  \cellCD[22pt]{b7d7a8}{2.8e-3} &  \cellCD[22pt]{ffffff}{9.4e-3} &  \cellCD[22pt]{b7d7a8}{2.2e-3} &  \cellCD[22pt]{b7d7a8}{86.8} &  \cellCD[22pt]{ffffff}{75.0} &  \cellCD[22pt]{69a84f}{90.2} &  \cellCD[22pt]{b7d7a8}{89.1} &  \cellCD[22pt]{ffffff}{75.6} &  \cellCD[22pt]{69a84f}{92.4} &  \cellCD[22pt]{ffffff}{51.1} &  \cellCD[22pt]{b7d7a8}{61.3} &  \cellCD[22pt]{69a84f}{71.7} &  \cellCD[22pt]{ffffff}{38.1} &  \cellCD[22pt]{b7d7a8}{56.6} &  \cellCD[22pt]{69a84f}{70.6} &  \cellCD[22pt]{ffffff}{45.5} &  \cellCD[22pt]{b7d7a8}{50.5} &  \cellCD[22pt]{69a84f}{62.0} &  \cellCD[22pt]{ffffff}{6.4} &  \cellCD[22pt]{b7d7a8}{18.2} &  \cellCD[22pt]{69a84f}{21.9} \\
   eft\_kitchen & 6042 &  \cellCD[24pt]{69a84f}{2202} &  \cellCD[24pt]{b7d7a8}{22884} &  \cellCD[24pt]{c0c0c0}{timeout} &  \cellCD[22pt]{69a84f}{3.1e-3} &  \cellCD[22pt]{b7d7a8}{7.4e-3} &  \cellCD[22pt]{c0c0c0}{\vphantom{0}-} &  \cellCD[22pt]{69a84f}{85.0} &  \cellCD[22pt]{b7d7a8}{59.9} &  \cellCD[22pt]{c0c0c0}{\vphantom{0}-} &  \cellCD[22pt]{69a84f}{85.1} &  \cellCD[22pt]{b7d7a8}{62.3} &  \cellCD[22pt]{c0c0c0}{\vphantom{0}-} &  \cellCD[22pt]{b7d7a8}{46.7} &  \cellCD[22pt]{69a84f}{51.7} &  \cellCD[22pt]{c0c0c0}{\vphantom{0}-} &  \cellCD[22pt]{b7d7a8}{26.4} &  \cellCD[22pt]{69a84f}{44.5} &  \cellCD[22pt]{c0c0c0}{\vphantom{0}-} &  \cellCD[22pt]{b7d7a8}{38.1} &  \cellCD[22pt]{69a84f}{41.2} &  \cellCD[22pt]{c0c0c0}{\vphantom{0}-} &  \cellCD[22pt]{b7d7a8}{4.6} &  \cellCD[22pt]{69a84f}{14.4} &  \cellCD[22pt]{c0c0c0}{\vphantom{0}-} \\
\bottomrule
\end{tabular}
}

    }
    \caption{Per-scene camera pose metrics on large-scale datasets including ZipNeRF, Mill-19, Urbanscene3D, and Eyeful Tower.}
    \label{tab:fastmap-supp_pose_large}
\end{table*}

\subsubsection{Ablations}
\label{sec:fastmap-ablation}

\minihead{Kernel fusion} In \cref{tab:fastmap-kernel_fusion}, we show the timing comparison of naive PyTorch and kernel fusion approaches to implementing the first-order optimization of epipolar adjustment. Profiling and comparing the CPU and GPU times for these two approaches is challenging due to the various forms of execution overlap. Instead, we directly compare the wall-clock time on different hardware setups. On small-to-medium scale scenes (i.e., 5k and 50k image pairs), the running time is severely bottlenecked by CPU overhead, and using a slower CPU can significantly impact the speed. Interestingly, the PyTorch version is faster on the less-powerful 2080 Ti than A6000, reflecting that its kernel implementation cannot fully utilize the power of high-end GPUs and is not suitable in our case. Across all the three hardware settings and scene sizes, our fused kernel implementation is around $20\times$ to $90\times$ faster than the naive PyTorch version.
\begin{table}[t]
    \centering
    \resizebox{1.0\linewidth}{!}{
    \tablestyle{8pt}{1.1}
\renewcommand{\arraystretch}{1.0}
\begin{tabular}{l l l rrr}
\toprule
 $\#$ pairs & CPU & GPU &
torch (ms) &
fused (ms) &
speedup \\
\midrule
\multirow{3}{4em}{5k} 
 & 4.05GHz & A6000 & 2.83 & \textbf{0.05} & $56\times$ \\
 & 2.2GHz & A6000 & 9.82 & \textbf{0.11} & $89\times$ \\
 & 2.2GHz & 2080 Ti & 9.41 & \textbf{0.11} & $85\times$ \\
\midrule
\multirow{3}{4em}{50k} 
 & 4.05GHz & A6000 & 8.20 & \textbf{0.14} & $58\times$ \\
 & 2.2GHz & A6000 & 12.47 & \textbf{0.20} & $62\times$ \\
 & 2.2GHz & 2080 Ti & 11.93 & \textbf{0.27} & $44\times$ \\
\midrule
\multirow{3}{4em}{500k} 
 & 4.05GHz & A6000 & 65.94 & \textbf{1.16} & $56\times$ \\
 & 2.2GHz & A6000 & 69.31 & \textbf{1.21} & $53\times$ \\
 & 2.2GHz & 2080 Ti & 44.32 & \textbf{1.92} & $23\times$ \\
\bottomrule

\end{tabular}

    }
    \caption{Effect of kernel fusion for epipolar adjustment under different hardware settings and scene sizes ($\#$pairs refers to the number of image pairs). Interestingly, the naive PyTorch implementation is faster on 2080 Ti than A6000 with 500k image pairs, showing that the native PyTorch kernel implementation cannot fully utilize the GPU for our problems. Note that the performance of the same CPU or GPU can be slightly different on different machines.}
    \label{tab:fastmap-kernel_fusion}
\end{table}

\minihead{Distortion estimation} is one of the first steps of \alg, and its accuracy is critical to the final performance. 
\begin{table}[t!]
    \centering
    \resizebox{1.0\linewidth}{!}{
        \tablestyle{5pt}{1.1}
\begin{tabular}{l  rr @{~~~~} | rr @{~~~~} | rr @{~~~~} | rr}
\toprule
 &
\multicolumn{2}{c}{\scriptsize AUC@3} &
\multicolumn{2}{c}{\scriptsize AUC@10} &
\multicolumn{2}{c}{\scriptsize RTA@3} &
\multicolumn{2}{c}{\scriptsize RTA@10} \\
\cmidrule(lr){2-3} \cmidrule(lr){4-5} \cmidrule(lr){6-7} \cmidrule(lr){8-9}

 & \multicolumn{1}{c}{w/} & \multicolumn{1}{c}{w/o}
& \multicolumn{1}{c}{w/} & \multicolumn{1}{c}{w/o}
& \multicolumn{1}{c}{w/} & \multicolumn{1}{c}{w/o}
& \multicolumn{1}{c}{w/} & \multicolumn{1}{c}{w/o} \\
\midrule
Family & \textbf{95.1} & 72.8 & \textbf{98.5} & 91.8 & \textbf{100.0} & 99.9 & \textbf{100.0} & 100.0 \\
Francis & \textbf{95.5} & 71.1 & \textbf{98.6} & 91.2 & \textbf{99.9} & 99.6 & \textbf{100.0} & 99.9 \\
Horse & \textbf{96.8} & 76.8 & \textbf{99.0} & 93.0 & \textbf{100.0} & 100.0 & \textbf{100.0} & 100.0 \\
Lighthouse & \textbf{90.7} & \cellcolor[HTML]{f4cccc} 4.6 & \textbf{97.1} & \cellcolor[HTML]{f4cccc} 42.2 & \textbf{99.6} & \cellcolor[HTML]{f4cccc} 46.5 & \textbf{100.0} & 98.5 \\
M60 & \textbf{95.6} & \cellcolor[HTML]{f4cccc} 28.3 & \textbf{98.7} & 72.9 & \textbf{99.9} & 85.9 & \textbf{100.0} & 99.7 \\
Panther & \textbf{93.0} & \cellcolor[HTML]{f4cccc} 12.1 & \textbf{97.9} & 64.2 & \textbf{99.9} & 78.3 & \textbf{100.0} & 99.9 \\
Playground & \textbf{84.6} & \cellcolor[HTML]{f4cccc} 2.2 & \textbf{95.4} & \cellcolor[HTML]{f4cccc} 14.4 & \textbf{100.0} & \cellcolor[HTML]{f4cccc} 15.2 & \textbf{100.0} & \cellcolor[HTML]{f4cccc} 51.9 \\
Train & \textbf{92.4} & \cellcolor[HTML]{f4cccc} 54.2 & \textbf{97.7} & 86.0 & \textbf{99.9} & 99.6 & \textbf{100.0} & 99.9 \\
\bottomrule
\end{tabular}

    }
    \caption{Effect of camera distortion estimation on pose accuracy.}
    \label{tab:fastmap-ablate_distortion}
\end{table}
Table~\ref{tab:fastmap-ablate_distortion} presents the performance of \alg with and without distortion estimation on the Intermediate split of Tanks and Temples. Without distortion estimation, results drop, sometimes catastrophically.

\minihead{Additional ablation study}
\begin{table}[t]
    \centering
    \resizebox{1.0\linewidth}{!}{
        \tablestyle{8pt}{1.1}
\renewcommand{\arraystretch}{1.0}
\resizebox{\textwidth}{!}{%
\begin{tabular}{l l rrrrr}
\toprule
 &  &
\multicolumn{1}{c}{$m=1$} &
\multicolumn{1}{c}{$m=2$} &
\multicolumn{1}{c}{$m=3$} &
\multicolumn{1}{c}{$m=4$} &
\multicolumn{1}{c}{$m=8$} \\
\midrule
\multirow{2}{7.5em}{zipnerf\_nyc} & RTE@30 & \textbf{0.59} & \textbf{0.09} & 0.09 & 0.09 & 0.10 \\
                  & RTA@3 & 98.59 & 99.45 & 99.45 & 99.45 & 99.45 \\
\midrule
\multirow{2}{7.5em}{zipnerf\_alameda}  & RTE@30 & \textbf{0.32} & \textbf{0.14} & 0.13 & 0.09 & 0.09 \\
                  & RTA@3 & 97.90 & 98.28 & 98.29 & 98.40 & 98.39 \\
\midrule
\multirow{2}{7.5em}{tnt\_Train}        & RTE@30 & \textbf{1.40} & \textbf{0.02} & 0.02 & 0.02 & 0.29 \\
                  & RTA@3 & 97.15 & 99.62 & 99.64 & 99.61 & 99.08 \\
\midrule
\multirow{2}{7.5em}{tnt\_Lighthouse}   & RTE@30 & \textbf{1.98} & \textbf{0.01} & 0.01 & 0.02 & 0.01 \\
                  & RTA@3 & 96.82 & 99.26 & 99.23 & 99.34 & 99.33 \\
\midrule
\multirow{2}{7.5em}{dploy\_ruins3}     & RTE@30 & \textbf{1.23} & \textbf{0.66} & 0.67 & 0.92 & 0.69 \\
                  & RTA@3 & 92.70 & 94.56 & 94.44 & 94.06 & 93.77 \\
\midrule
\multirow{2}{7.5em}{dploy\_house4}     & RTE@30 & \textbf{3.82} & \textbf{0.83} & 1.00 & 0.83 & 0.83 \\
                  & RTA@3 & 93.00 & 96.92 & 95.69 & 97.48 & 95.48 \\
\bottomrule

\end{tabular}
}

    }
    \caption{Ablation of multiple translation initializations on selected scenes. Results are obtained right after translation alignment and before epipolar adjustment. We bold the RTE@$30$ entries for $m=1$ and $m=2$ initializations to highlight the effect.}
    \label{tab:fastmap-ablate_multiinit}
\end{table}

\begin{table}[t]
    \centering
    \resizebox{1.0\linewidth}{!}{
        \tablestyle{2pt}{1.1}
\renewcommand{\arraystretch}{1.0}
\scriptsize
\begin{tabular}{l @{~~~} l rrrrr }
\toprule
 & &
\multicolumn{1}{c}{AUC@3} &
\multicolumn{1}{c}{AUC@10} &
\multicolumn{1}{c}{RTA@1} &
\multicolumn{1}{c}{RTA@5} &
\multicolumn{1}{c}{RRA@3} \\
\midrule
\multirow{3}{4em}{m360 (9)} &  \alg & \textbf{97.2} & \textbf{99.1} & \textbf{99.8} & \textbf{100.0} & \textbf{100.0} \\
         &  w/o epipolar adjustment & 75.0 & 90.8 & 85.5 & 99.5 & 94.5 \\
         &  w/o track completion & 80.4 & 86.4 & 83.3 & 91.4 & 83.6 \\
\midrule
 \multirow{3}{4em}{alameda}  &  \alg & \textbf{95.2} & \textbf{98.1} & \textbf{99.0} & \textbf{99.3} & \textbf{99.9} \\
         &  w/o epipolar adjustment & 86.7 & 95.5 & 94.9 & 99.2 & 99.8 \\
         &  w/o track completion & 94.8 & \textbf{98.1} & \textbf{99.0} & 99.4 & \textbf{99.9} \\
\midrule
\multirow{3}{4em}{berlin}   &  \alg & \textbf{81.6} & \textbf{93.2} & \textbf{92.8} & \textbf{99.2} & \textbf{97.5} \\
         &  w/o epipolar adjustment & 70.4 & 89.5 & 82.4 & 98.8 & 95.7 \\
         &  w/o track completion & 60.4 & 81.3 & 70.4 & 90.8 & 90.3 \\
\midrule
\multirow{3}{4em}{london}   &  \alg & \textbf{96.6} & \textbf{98.8} & \textbf{99.6} & \textbf{99.9} & 99.7 \\
         &  w/o epipolar adjustment & 90.1 & 96.6 & 97.8 & 99.7 & 99.3 \\
         &  w/o track completion & 96.1 & 98.7 & \textbf{99.6} & \textbf{99.9} & \textbf{99.8} \\
\midrule
\multirow{3}{4em}{nyc}      &  \alg & \textbf{94.6} & \textbf{98.1} & 99.2 & 99.6 & 99.6 \\
         &  w/o epipolar adjustment & 89.7 & 96.7 & 97.0 & 99.7 & 99.6 \\
         &  w/o track completion & 93.9 & 98.0 & \textbf{99.4} & \textbf{99.8} &  \textbf{99.8} \\
\bottomrule
\end{tabular}

    }
    \caption{Epipolar adjustment and track completion ablation on the MipNeRF360~\cite{barron2022mipnerf360} and ZipNeRF~\cite{barron2023zipnerf} datasets. Results for MipNeRF360 are averaged over all scenes, and for ZipNeRF each scene is listed separately.}
    \label{tab:fastmap-ablate_epioptim_and_tracks}
\end{table}

\paragraph{Track completion}
Table~\ref{tab:fastmap-ablate_epioptim_and_tracks} shows the performance of \alg with and without augmented point pairs from track completion (\cref{sec:fastmap-method-tracks}). Track completion improves performance for MipNeRF360 scenes and some ZipNeRF scenes.

\paragraph{Multiple translation initializations}
As shown in Table~\ref{tab:fastmap-ablate_multiinit}, a single initialization is prone to large-error outliers (RTE@$30$ defined as $100.0$ - RTA@$30$). Increasing the number of initializations improves performance, but the effect plateaus and does not fully fix outliers.

\paragraph{Epipolar adjustment}
Table~\ref{tab:fastmap-ablate_epioptim_and_tracks} also presents the performance of \alg with and without the final epipolar adjustment step. Epipolar adjustment consistently improves over the poses from global translation alignment, especially under stricter metrics (RTA@1).

\subsection{Limitations and Conclusions}
\label{sec:fastmap-conclusions}

\subsubsection{Limitations}
\noindent\textbf{Sparse views.} Our method assumes that the input images densely cover the 3D scene. Many components in the pipeline implicitly assume dense coverage so that outlier image or point pairs are averaged out. If the coverage is sparse, the pipeline is sensitive to outliers and likely breaks down. We evaluated on ETH3D MVS (DSLR)~\cite{schops2017multi}, where each scene contains only a small number of images, and report results in Table~\ref{tab:fastmap-eth3d}. While \alg succeeds on many scenes, it is less robust than GLOMAP.

\begin{table}[t]
    \centering
    \resizebox{1.0\linewidth}{!}{
        
\tablestyle{2pt}{1.1}
\resizebox{\textwidth}{!}{%
\begin{tabular}{l r  r@{~}r | r@{~}r | r@{~}r | r@{~}r r }
\toprule
  &  &
\multicolumn{2}{c}{ATE$\downarrow$} &
\multicolumn{2}{c}{RTA@3$\uparrow$} &
\multicolumn{2}{c}{AUC-R\&T @ 3 $\uparrow$} &
\multicolumn{2}{c}{AUC-R\&T @ 1 $\uparrow$} \\
\cmidrule(lr){3-4} \cmidrule(lr){5-6} \cmidrule(lr){7-8} \cmidrule(lr){9-10}
   & n\_imgs &
\multicolumn{1}{c}{\scriptsize \alg} & \multicolumn{1}{c}{\scriptsize GLOMAP} &
\multicolumn{1}{c}{\scriptsize \alg} & \multicolumn{1}{c}{\scriptsize GLOMAP} &
\multicolumn{1}{c}{\scriptsize \alg} & \multicolumn{1}{c}{\scriptsize GLOMAP} &
\multicolumn{1}{c}{\scriptsize \alg} & \multicolumn{1}{c}{\scriptsize GLOMAP} \\
\midrule
botanical\_garden &   30 &  \cellCD[22pt]{b7d7a8}{8.2e-3} &  \cellCD[22pt]{69a84f}{4.3e-4} &  \cellCD[22pt]{b7d7a8}{86.9} &  \cellCD[22pt]{69a84f}{100.0} &  \cellCD[22pt]{b7d7a8}{68.3} &  \cellCD[22pt]{69a84f}{94.3} &  \cellCD[22pt]{b7d7a8}{52.0} &  \cellCD[22pt]{69a84f}{83.8} \\
       boulders &   26 &  \cellCD[22pt]{b7d7a8}{6.7e-4} &  \cellCD[22pt]{69a84f}{1.4e-4} &  \cellCD[22pt]{b7d7a8}{99.1} &  \cellCD[22pt]{b7d7a8}{100.0} &  \cellCD[22pt]{b7d7a8}{91.2} &  \cellCD[22pt]{69a84f}{97.0} &  \cellCD[22pt]{b7d7a8}{76.2} &  \cellCD[22pt]{69a84f}{91.0} \\
         bridge &  110 &  \cellCD[22pt]{b7d7a8}{1.3e-2} &  \cellCD[22pt]{69a84f}{2.0e-5} &  \cellCD[22pt]{b7d7a8}{92.9} &  \cellCD[22pt]{69a84f}{100.0} &  \cellCD[22pt]{b7d7a8}{85.3} &  \cellCD[22pt]{69a84f}{97.7} &  \cellCD[22pt]{b7d7a8}{73.3} &  \cellCD[22pt]{69a84f}{93.1} \\
      courtyard &   38 &  \cellCD[22pt]{b7d7a8}{3.8e-2} &  \cellCD[22pt]{69a84f}{1.8e-4} &  \cellCD[22pt]{f4cccc}{18.9} &  \cellCD[22pt]{69a84f}{100.0} &  \cellCD[22pt]{f4cccc}{6.9} &  \cellCD[22pt]{69a84f}{96.0} &  \cellCD[22pt]{f4cccc}{2.2} &  \cellCD[22pt]{69a84f}{88.2} \\
 delivery\_area &   44 &  \cellCD[22pt]{b7d7a8}{8.4e-2} &  \cellCD[22pt]{69a84f}{8.1e-5} &  \cellCD[22pt]{f4cccc}{23.6} &  \cellCD[22pt]{69a84f}{100.0} &  \cellCD[22pt]{f4cccc}{13.9} &  \cellCD[22pt]{69a84f}{97.8} &  \cellCD[22pt]{f4cccc}{6.1} &  \cellCD[22pt]{69a84f}{93.3} \\
           door &    7 &  \cellCD[22pt]{f4cccc}{-\vphantom{1.0}} &  \cellCD[22pt]{69a84f}{1.2e-4} &  \cellCD[22pt]{f4cccc}{-\vphantom{1.0}} &  \cellCD[22pt]{69a84f}{100.0} &  \cellCD[22pt]{f4cccc}{-\vphantom{1.0}} &  \cellCD[22pt]{69a84f}{98.0} &  \cellCD[22pt]{f4cccc}{-\vphantom{1.0}} &  \cellCD[22pt]{69a84f}{94.1} \\
        electro &   45 &  \cellCD[22pt]{b7d7a8}{7.5e-2} &  \cellCD[22pt]{69a84f}{3.0e-2} &  \cellCD[22pt]{b7d7a8}{86.3} &  \cellCD[22pt]{69a84f}{95.2} &  \cellCD[22pt]{b7d7a8}{76.9} &  \cellCD[22pt]{69a84f}{91.1} &  \cellCD[22pt]{b7d7a8}{61.6} &  \cellCD[22pt]{69a84f}{84.1} \\
exhibition\_hall &   68 &  \cellCD[22pt]{b7d7a8}{7.0e-2} &  \cellCD[22pt]{b7d7a8}{6.9e-2} &  \cellCD[22pt]{f4cccc}{2.8} &  \cellCD[22pt]{69a84f}{45.1} &  \cellCD[22pt]{f4cccc}{0.9} &  \cellCD[22pt]{69a84f}{40.9} &  \cellCD[22pt]{f4cccc}{0.1} &  \cellCD[22pt]{f4cccc}{34.3} \\
         facade &   76 &  \cellCD[22pt]{b7d7a8}{6.5e-2} &  \cellCD[22pt]{69a84f}{9.7e-5} &  \cellCD[22pt]{b7d7a8}{71.0} &  \cellCD[22pt]{69a84f}{100.0} &  \cellCD[22pt]{b7d7a8}{66.8} &  \cellCD[22pt]{69a84f}{97.4} &  \cellCD[22pt]{b7d7a8}{60.8} &  \cellCD[22pt]{69a84f}{92.4} \\
         kicker &   31 &  \cellCD[22pt]{69a84f}{5.9e-4} &  \cellCD[22pt]{b7d7a8}{1.6e-2} &  \cellCD[22pt]{69a84f}{98.5} &  \cellCD[22pt]{b7d7a8}{93.8} &  \cellCD[22pt]{b7d7a8}{86.6} &  \cellCD[22pt]{69a84f}{91.7} &  \cellCD[22pt]{b7d7a8}{65.0} &  \cellCD[22pt]{69a84f}{88.1} \\
  lecture\_room &   23 &  \cellCD[22pt]{b7d7a8}{3.0e-2} &  \cellCD[22pt]{69a84f}{2.5e-4} &  \cellCD[22pt]{b7d7a8}{84.2} &  \cellCD[22pt]{69a84f}{100.0} &  \cellCD[22pt]{b7d7a8}{71.7} &  \cellCD[22pt]{69a84f}{95.0} &  \cellCD[22pt]{b7d7a8}{55.3} &  \cellCD[22pt]{69a84f}{85.7} \\
   living\_room &   65 &  \cellCD[22pt]{b7d7a8}{1.3e-4} &  \cellCD[22pt]{69a84f}{8.4e-5} &  \cellCD[22pt]{b7d7a8}{99.7} &  \cellCD[22pt]{b7d7a8}{99.8} &  \cellCD[22pt]{b7d7a8}{95.3} &  \cellCD[22pt]{b7d7a8}{96.2} &  \cellCD[22pt]{b7d7a8}{86.8} &  \cellCD[22pt]{69a84f}{89.2} \\
         lounge &   10 &  \cellCD[22pt]{b7d7a8}{9.6e-2} &  \cellCD[22pt]{b7d7a8}{9.5e-2} &  \cellCD[22pt]{f4cccc}{33.3} &  \cellCD[22pt]{f4cccc}{33.3} &  \cellCD[22pt]{f4cccc}{32.3} &  \cellCD[22pt]{f4cccc}{32.7} &  \cellCD[22pt]{f4cccc}{30.2} &  \cellCD[22pt]{f4cccc}{31.4} \\
         meadow &   15 &  \cellCD[22pt]{b7d7a8}{1.4e-1} &  \cellCD[22pt]{b7d7a8}{1.4e-1} &  \cellCD[22pt]{f4cccc}{13.3} &  \cellCD[22pt]{69a84f}{86.7} &  \cellCD[22pt]{f4cccc}{7.7} &  \cellCD[22pt]{69a84f}{80.2} &  \cellCD[22pt]{f4cccc}{4.9} &  \cellCD[22pt]{69a84f}{68.2} \\
    observatory &   27 &  \cellCD[22pt]{b7d7a8}{6.5e-3} &  \cellCD[22pt]{69a84f}{5.8e-4} &  \cellCD[22pt]{b7d7a8}{94.9} &  \cellCD[22pt]{69a84f}{99.1} &  \cellCD[22pt]{b7d7a8}{76.5} &  \cellCD[22pt]{69a84f}{86.5} &  \cellCD[22pt]{b7d7a8}{48.5} &  \cellCD[22pt]{69a84f}{63.9} \\
         office &   26 &  \cellCD[22pt]{b7d7a8}{9.7e-3} &  \cellCD[22pt]{69a84f}{7.6e-4} &  \cellCD[22pt]{b7d7a8}{54.8} &  \cellCD[22pt]{69a84f}{95.7} &  \cellCD[22pt]{b7d7a8}{43.9} &  \cellCD[22pt]{69a84f}{82.7} &  \cellCD[22pt]{f4cccc}{34.5} &  \cellCD[22pt]{69a84f}{61.2} \\
  old\_computer &   54 &  \cellCD[22pt]{b7d7a8}{6.8e-2} &  \cellCD[22pt]{b7d7a8}{5.6e-2} &  \cellCD[22pt]{f4cccc}{21.7} &  \cellCD[22pt]{69a84f}{65.3} &  \cellCD[22pt]{f4cccc}{16.0} &  \cellCD[22pt]{69a84f}{60.9} &  \cellCD[22pt]{f4cccc}{9.8} &  \cellCD[22pt]{69a84f}{53.5} \\
          pipes &   14 &  \cellCD[22pt]{b7d7a8}{5.8e-4} &  \cellCD[22pt]{69a84f}{2.6e-4} &  \cellCD[22pt]{b7d7a8}{98.9} &  \cellCD[22pt]{b7d7a8}{100.0} &  \cellCD[22pt]{b7d7a8}{92.5} &  \cellCD[22pt]{69a84f}{97.4} &  \cellCD[22pt]{b7d7a8}{79.8} &  \cellCD[22pt]{69a84f}{92.3} \\
     playground &   38 &  \cellCD[22pt]{b7d7a8}{8.3e-4} &  \cellCD[22pt]{69a84f}{1.1e-4} &  \cellCD[22pt]{b7d7a8}{99.4} &  \cellCD[22pt]{b7d7a8}{99.9} &  \cellCD[22pt]{b7d7a8}{89.4} &  \cellCD[22pt]{69a84f}{97.1} &  \cellCD[22pt]{b7d7a8}{70.8} &  \cellCD[22pt]{69a84f}{91.7} \\
         relief &   31 &  \cellCD[22pt]{b7d7a8}{6.1e-3} &  \cellCD[22pt]{69a84f}{7.2e-5} &  \cellCD[22pt]{b7d7a8}{77.8} &  \cellCD[22pt]{69a84f}{100.0} &  \cellCD[22pt]{b7d7a8}{62.1} &  \cellCD[22pt]{69a84f}{98.4} &  \cellCD[22pt]{b7d7a8}{48.9} &  \cellCD[22pt]{69a84f}{95.2} \\
      relief\_2 &   31 &  \cellCD[22pt]{b7d7a8}{3.7e-4} &  \cellCD[22pt]{69a84f}{7.9e-5} &  \cellCD[22pt]{b7d7a8}{99.8} &  \cellCD[22pt]{b7d7a8}{100.0} &  \cellCD[22pt]{b7d7a8}{94.5} &  \cellCD[22pt]{69a84f}{98.4} &  \cellCD[22pt]{b7d7a8}{84.3} &  \cellCD[22pt]{69a84f}{95.1} \\
         statue &   11 &  \cellCD[22pt]{b7d7a8}{5.5e-5} &  \cellCD[22pt]{69a84f}{2.3e-5} &  \cellCD[22pt]{b7d7a8}{100.0} &  \cellCD[22pt]{b7d7a8}{100.0} &  \cellCD[22pt]{b7d7a8}{99.5} &  \cellCD[22pt]{b7d7a8}{99.7} &  \cellCD[22pt]{b7d7a8}{98.5} &  \cellCD[22pt]{b7d7a8}{99.0} \\
        terrace &   23 &  \cellCD[22pt]{b7d7a8}{2.1e-4} &  \cellCD[22pt]{69a84f}{1.2e-4} &  \cellCD[22pt]{b7d7a8}{100.0} &  \cellCD[22pt]{b7d7a8}{100.0} &  \cellCD[22pt]{b7d7a8}{97.5} &  \cellCD[22pt]{b7d7a8}{97.7} &  \cellCD[22pt]{b7d7a8}{92.5} &  \cellCD[22pt]{b7d7a8}{93.1} \\
     terrace\_2 &   13 &  \cellCD[22pt]{b7d7a8}{2.6e-4} &  \cellCD[22pt]{b7d7a8}{2.2e-4} &  \cellCD[22pt]{b7d7a8}{100.0} &  \cellCD[22pt]{b7d7a8}{100.0} &  \cellCD[22pt]{b7d7a8}{96.6} &  \cellCD[22pt]{b7d7a8}{96.9} &  \cellCD[22pt]{b7d7a8}{89.9} &  \cellCD[22pt]{b7d7a8}{90.8} \\
       terrains &   42 &  \cellCD[22pt]{b7d7a8}{1.3e-3} &  \cellCD[22pt]{69a84f}{2.1e-4} &  \cellCD[22pt]{b7d7a8}{94.4} &  \cellCD[22pt]{69a84f}{99.8} &  \cellCD[22pt]{b7d7a8}{70.9} &  \cellCD[22pt]{69a84f}{94.6} &  \cellCD[22pt]{b7d7a8}{39.3} &  \cellCD[22pt]{69a84f}{84.6} \\
\bottomrule
\end{tabular}
}

    }
    \caption{Per-scene camera pose metrics on ETH3D.}
    \label{tab:fastmap-eth3d}
\end{table}

\noindent\textbf{Intrinsics estimation.} The interval searches used in both distortion and focal length estimation require at least one image pair with shared intrinsics. They fail if all images have different intrinsics, and are less robust when each camera has only a small number of images. In addition, the focal length estimator relies entirely on fundamental matrices and is unreliable when the scene is dominated by homographies.

\noindent\textbf{Homography ambiguity.} Too many homography image pairs can jeopardize relative pose decomposition. Both essential and homography decomposition produce four solutions, usually disambiguated by a cheirality check. For some homography and keypoint pairs, cheirality is insufficient, and our current strategy picks the lowest-index solution when ties occur.

\noindent\textbf{Repetitive patterns and symmetric structures.} Non-learning keypoint matching is not robust to repetitive patterns and symmetric structures. These incorrect matches are hard to filter because they can have many inlier point pairs with consistent two-view geometry. This issue is most prominent in the Advanced split of Tanks and Temples (Table~\ref{tab:fastmap-supp_pose_tnt}).

\noindent\textbf{Degenerate motions.} Bundle adjustment can utilize 3D points to resolve ambiguities in global translation estimation. When cameras lie on a line, objectives that rely on relative motions or epipolar errors can fail because there is no unique way (up to scale) to determine camera distances. We evaluated on LaMAR~\cite{sarlin2022lamar}, which contains many straight and forward motions with sparse overlap between trajectories, and report results in Table~\ref{tab:fastmap-lamar}. Our method underperforms GLOMAP in this setting.

\begin{table}[t]
    \centering
    \resizebox{1.0\linewidth}{!}{
        \tablestyle{4pt}{1.1}
\begin{tabular}{l cc  cc  cc}
\toprule
 & \multicolumn{2}{c}{Recall@1m$\uparrow$} & \multicolumn{2}{c}{AUC@1m$\uparrow$} & \multicolumn{2}{c}{AUC@5m$\uparrow$} \\
\cmidrule(lr){2-3} \cmidrule(lr){4-5} \cmidrule(lr){6-7}
 & \multicolumn{1}{c}{\scriptsize \alg} & \multicolumn{1}{c}{\scriptsize GLOMAP} & \multicolumn{1}{c}{\scriptsize \alg} & \multicolumn{1}{c}{\scriptsize GLOMAP} & \multicolumn{1}{c}{\scriptsize \alg} & \multicolumn{1}{c}{\scriptsize GLOMAP} \\
\midrule 
CAB & 4.77 & 11.6 & 4.32 & 4.7 & 4.74 & 16.9 \\
HGE & 5.94 & 48.4 & 5.26 & 22.2 & 5.70 & 50.3 \\
LIN & 7.16 & 87.3 & 3.95 & 46.7 & 7.96 & 85.6 \\
\bottomrule
\end{tabular}

    }
    \caption{Per-scene camera pose metrics on LaMAR.}
    \label{tab:fastmap-lamar}
\end{table}

\subsubsection{Conclusions}
We introduce \alg, a new structure from motion method focused on simplicity and speed. Contrary to the common practice in other SfM systems, \alg uses first-order optimization extensively and is much faster than state-of-the-art methods (COLMAP and GLOMAP), while achieving comparable performance on pose accuracy and novel view synthesis quality. These improvements do come with a few drawbacks. Nevertheless, we believe it is an important step towards highly efficient camera pose estimation for real-world 3D data acquisition at scale. 

\endgroup

\chapter{Conclusion}
\label{chap:conclusion}

This thesis addressed efficient 3D content creation from
images and text. On the generative side, I introduced Instant3D, which pairs
diffusion-based multi-view synthesis with fast feed-forward reconstruction to
produce high-quality assets in seconds, and Carve3D, which improves multi-view
consistency via reinforcement-learning fine-tuning. I also explored DMV3D, a
single-stage formulation that diffuses directly over multi-view observations.
Together these efforts move text- and image-to-3D systems toward practical
latency, more faithful instruction following, better geometry and texture quality,
and higher diversity.

On the reconstruction side, I proposed FastMap, a structure-from-motion
pipeline that achieves substantial speedup by using first-order optimization
and kernel fusion while preserving accuracy. These results show that careful
algorithm and systems co-design can lead to significantly faster 3D reconstruction
without sacrificing too much quality, enabling applications that require
large-scale processing.

Looking forward, several directions are especially promising: integrating
better geometric representations to improve 3D resolution,
scaling multi-view generation and reconstruction to larger datasets, and
designing large unified multimodal systems that jointly learn to generate, reconstruct, and
edit 3D content with various control modalities. Progress along these axes
will further reduce the cost of 3D content creation and democratize the power of AI
for content creation.

\appendix

\bibliographystyle{plainnat}
\bibliography{references}

\begin{thebibliography}{325}
\providecommand{\natexlab}[1]{#1}
\providecommand{\url}[1]{\texttt{#1}}
\expandafter\ifx\csname urlstyle\endcsname\relax
  \providecommand{\doi}[1]{doi: #1}\else
  \providecommand{\doi}{doi: \begingroup \urlstyle{rm}\Url}\fi

\bibitem[Achlioptas et~al.(2018)Achlioptas, Diamanti, Mitliagkas, and
  Guibas]{pmlr-v80-achlioptas18a}
Panos Achlioptas, Olga Diamanti, Ioannis Mitliagkas, and Leonidas Guibas.
\newblock Learning representations and generative models for 3d point clouds.
\newblock In Jennifer Dy and Andreas Krause, editors, \emph{Proceedings of the
  35th International Conference on Machine Learning}, volume~80 of
  \emph{Proceedings of Machine Learning Research}, pages 40--49. PMLR, 10--15
  Jul 2018.
\newblock URL \url{https://proceedings.mlr.press/v80/achlioptas18a.html}.

\bibitem[Adobe(2023)]{adobe_firefly}
Adobe.
\newblock Adobe firefly.
\newblock \url{https://www.adobe.com/sensei/generative-ai/firefly.html}, 2023.
\newblock Accessed: 2023-11-15.

\bibitem[Agarwal et~al.(2009)Agarwal, Snavely, Simon, Seitz, and
  Szeliski]{Agarwal_2009}
Sameer Agarwal, Noah Snavely, Ian Simon, Steven~M Seitz, and Richard Szeliski.
\newblock Building rome in a day.
\newblock In \emph{2009 IEEE 12th International Conference on Computer Vision},
  pages 72--79, September 2009.
\newblock \doi{10.1109/ICCV.2009.5459148}.
\newblock URL \url{http://dx.doi.org/10.1109/ICCV.2009.5459148}.

\bibitem[Agarwal et~al.(2010)Agarwal, Snavely, Seitz, and
  Szeliski]{Agarwal_2010}
Sameer Agarwal, Noah Snavely, Steven~M. Seitz, and Richard Szeliski.
\newblock Bundle adjustment in the large.
\newblock In \emph{Computer Vision -- ECCV 2010}, pages 29--42, 2010.
\newblock \doi{10.1007/978-3-642-15552-9_3}.
\newblock URL \url{http://dx.doi.org/10.1007/978-3-642-15552-9_3}.

\bibitem[Agarwal et~al.(2011)Agarwal, Furukawa, Snavely, Simon, Curless, Seitz,
  and Szeliski]{Agarwal_2011}
Sameer Agarwal, Yasutaka Furukawa, Noah Snavely, Ian Simon, Brian Curless,
  Steven~M. Seitz, and Richard Szeliski.
\newblock Building rome in a day.
\newblock \emph{Communications of the ACM}, 54\penalty0 (10):\penalty0
  105--112, October 2011.
\newblock \doi{10.1145/2001269.2001293}.
\newblock URL \url{http://dx.doi.org/10.1145/2001269.2001293}.

\bibitem[Agarwal et~al.(2023)Agarwal, Mierle, and
  Team]{Agarwal_Ceres_Solver_2022}
Sameer Agarwal, Keir Mierle, and The Ceres~Solver Team.
\newblock {Ceres Solver}.
\newblock Software, 10 2023.
\newblock URL \url{https://github.com/ceres-solver/ceres-solver}.

\bibitem[Ali-bey et~al.(2023)Ali-bey, Chaib-draa, and
  Gigu\`ere]{Ali-bey_2023_WACV}
Amar Ali-bey, Brahim Chaib-draa, and Philippe Gigu\`ere.
\newblock Mixvpr: Feature mixing for visual place recognition.
\newblock In \emph{Proceedings of the IEEE/CVF Winter Conference on
  Applications of Computer Vision (WACV)}, pages 2998--3007, January 2023.

\bibitem[Anciukevi{\v{c}}ius et~al.(2023)Anciukevi{\v{c}}ius, Xu, Fisher,
  Henderson, Bilen, Mitra, and Guerrero]{renderdiff}
Titas Anciukevi{\v{c}}ius, Zexiang Xu, Matthew Fisher, Paul Henderson, Hakan
  Bilen, Niloy~J Mitra, and Paul Guerrero.
\newblock Renderdiffusion: Image diffusion for 3d reconstruction, inpainting
  and generation.
\newblock In \emph{CVPR}, 2023.

\bibitem[Arandjelovic et~al.(2016)Arandjelovic, Gronat, Torii, Pajdla, and
  Sivic]{Arandjelovic_2016_CVPR}
Relja Arandjelovic, Petr Gronat, Akihiko Torii, Tomas Pajdla, and Josef Sivic.
\newblock Netvlad: Cnn architecture for weakly supervised place recognition.
\newblock In \emph{Proceedings of the IEEE Conference on Computer Vision and
  Pattern Recognition (CVPR)}, June 2016.

\bibitem[Bai et~al.(2022{\natexlab{a}})Bai, Jones, Ndousse, Askell, Chen,
  DasSarma, Drain, Fort, Ganguli, Henighan, Joseph, Kadavath, Kernion, Conerly,
  El-Showk, Elhage, Hatfield-Dodds, Hernandez, Hume, Johnston, Kravec, Lovitt,
  Nanda, Olsson, Amodei, Brown, Clark, McCandlish, Olah, Mann, and
  Kaplan]{bai2022helpfulharmlessRLHF}
Yuntao Bai, Andy Jones, Kamal Ndousse, Amanda Askell, Anna Chen, Nova DasSarma,
  Dawn Drain, Stanislav Fort, Deep Ganguli, Tom Henighan, Nicholas Joseph,
  Saurav Kadavath, Jackson Kernion, Tom Conerly, Sheer El-Showk, Nelson Elhage,
  Zac Hatfield-Dodds, Danny Hernandez, Tristan Hume, Scott Johnston, Shauna
  Kravec, Liane Lovitt, Neel Nanda, Catherine Olsson, Dario Amodei, Tom Brown,
  Jack Clark, Sam McCandlish, Chris Olah, Ben Mann, and Jared Kaplan.
\newblock Training a helpful and harmless assistant with reinforcement learning
  from human feedback, 2022{\natexlab{a}}.

\bibitem[Bai et~al.(2022{\natexlab{b}})Bai, Kadavath, Kundu, Askell, Kernion,
  Jones, Chen, Goldie, Mirhoseini, McKinnon, Chen, Olsson, Olah, Hernandez,
  Drain, Ganguli, Li, Tran-Johnson, Perez, Kerr, Mueller, Ladish, Landau,
  Ndousse, Lukosuite, Lovitt, Sellitto, Elhage, Schiefer, Mercado, DasSarma,
  Lasenby, Larson, Ringer, Johnston, Kravec, Showk, Fort, Lanham,
  Telleen-Lawton, Conerly, Henighan, Hume, Bowman, Hatfield-Dodds, Mann,
  Amodei, Joseph, McCandlish, Brown, and Kaplan]{bai2022RLAIF}
Yuntao Bai, Saurav Kadavath, Sandipan Kundu, Amanda Askell, Jackson Kernion,
  Andy Jones, Anna Chen, Anna Goldie, Azalia Mirhoseini, Cameron McKinnon,
  Carol Chen, Catherine Olsson, Christopher Olah, Danny Hernandez, Dawn Drain,
  Deep Ganguli, Dustin Li, Eli Tran-Johnson, Ethan Perez, Jamie Kerr, Jared
  Mueller, Jeffrey Ladish, Joshua Landau, Kamal Ndousse, Kamile Lukosuite,
  Liane Lovitt, Michael Sellitto, Nelson Elhage, Nicholas Schiefer, Noemi
  Mercado, Nova DasSarma, Robert Lasenby, Robin Larson, Sam Ringer, Scott
  Johnston, Shauna Kravec, Sheer~El Showk, Stanislav Fort, Tamera Lanham,
  Timothy Telleen-Lawton, Tom Conerly, Tom Henighan, Tristan Hume, Samuel~R.
  Bowman, Zac Hatfield-Dodds, Ben Mann, Dario Amodei, Nicholas Joseph, Sam
  McCandlish, Tom Brown, and Jared Kaplan.
\newblock Constitutional ai: Harmlessness from ai feedback, 2022{\natexlab{b}}.

\bibitem[Barath et~al.(2017)Barath, Toth, and Hajder]{barath2017minimal}
Daniel Barath, Tekla Toth, and Levente Hajder.
\newblock A minimal solution for two-view focal-length estimation using two
  affine correspondences.
\newblock In \emph{CVPR}, 2017.

\bibitem[Barreto and Daniilidis(2005)]{barreto2005fundamental}
Jo{\~a}o~Pedro Barreto and Kostas Daniilidis.
\newblock Fundamental matrix for cameras with radial distortion.
\newblock In \emph{ICCV}, 2005.

\bibitem[Barron et~al.(2022)Barron, Mildenhall, Verbin, Srinivasan, and
  Hedman]{barron2022mipnerf360}
Jonathan~T. Barron, Ben Mildenhall, Dor Verbin, Pratul~P. Srinivasan, and Peter
  Hedman.
\newblock Mip-nerf 360: Unbounded anti-aliased neural radiance fields.
\newblock In \emph{IEEE/CVF Conference on Computer Vision and Pattern
  Recognition (CVPR)}, pages 5460--5469, 2022.
\newblock \doi{10.1109/CVPR52688.2022.00539}.
\newblock URL \url{https://doi.org/10.1109/CVPR52688.2022.00539}.

\bibitem[Barron et~al.(2023)Barron, Mildenhall, Verbin, Srinivasan, and
  Hedman]{barron2023zipnerf}
Jonathan~T. Barron, Ben Mildenhall, Dor Verbin, Pratul~P. Srinivasan, and Peter
  Hedman.
\newblock Zip-nerf: Anti-aliased grid-based neural radiance fields.
\newblock In \emph{IEEE/CVF International Conference on Computer Vision
  (ICCV)}, pages 19640--19648, 2023.
\newblock \doi{10.1109/ICCV51070.2023.01804}.
\newblock URL \url{https://doi.org/10.1109/ICCV51070.2023.01804}.

\bibitem[Barroso-Laguna et~al.(2019)Barroso-Laguna, Riba, Ponsa, and
  Mikolajczyk]{barrosolaguna2019keynetkeypointdetectionhandcrafted}
Axel Barroso-Laguna, Edgar Riba, Daniel Ponsa, and Krystian Mikolajczyk.
\newblock Key.net: Keypoint detection by handcrafted and learned cnn filters,
  2019.
\newblock URL \url{https://arxiv.org/abs/1904.00889}.

\bibitem[Berton et~al.(2022)Berton, Masone, and Caputo]{Berton_2022_CVPR}
Gabriele Berton, Carlo Masone, and Barbara Caputo.
\newblock Rethinking visual geo-localization for large-scale applications.
\newblock In \emph{Proceedings of the IEEE/CVF Conference on Computer Vision
  and Pattern Recognition (CVPR)}, pages 4878--4888, June 2022.

\bibitem[Bescos et~al.(2018)Bescos, Fácil, Civera, and
  Neira]{bescos2018dynaslamtrackingmappinginpainting}
Berta Bescos, José~M. Fácil, Javier Civera, and José Neira.
\newblock Dynaslam: Tracking, mapping and inpainting in dynamic scenes, 2018.
\newblock URL \url{https://arxiv.org/abs/1806.05620}.

\bibitem[Bhat et~al.(2023)Bhat, Birkl, Wofk, Wonka, and
  M{\"u}ller]{bhat2023zoedepth}
Shariq~Farooq Bhat, Reiner Birkl, Diana Wofk, Peter Wonka, and Matthias
  M{\"u}ller.
\newblock Zoedepth: Zero-shot transfer by combining relative and metric depth.
\newblock \emph{arXiv preprint arXiv:2302.12288}, 2023.

\bibitem[Black et~al.(2023)Black, Janner, Du, Kostrikov, and
  Levine]{black2023DDPO}
Kevin Black, Michael Janner, Yilun Du, Ilya Kostrikov, and Sergey Levine.
\newblock Training diffusion models with reinforcement learning, 2023.

\bibitem[Bogdan et~al.(2018)Bogdan, Eckstein, Rameau, and
  Bazin]{bogdan2018deepcalib}
Oleksandr Bogdan, Viktor Eckstein, Francois Rameau, and Jean-Charles Bazin.
\newblock Deepcalib: a deep learning approach for automatic intrinsic
  calibration of wide field-of-view cameras.
\newblock In \emph{Proceedings of the 15th ACM SIGGRAPH European Conference on
  Visual Media Production}, 2018.

\bibitem[Brachmann and Rother(2019)]{Brachmann_2019}
Eric Brachmann and Carsten Rother.
\newblock Neural-guided ransac: Learning where to sample model hypotheses.
\newblock In \emph{2019 IEEE/CVF International Conference on Computer Vision
  (ICCV)}, page 4321–4330. IEEE, October 2019.
\newblock \doi{10.1109/iccv.2019.00442}.
\newblock URL \url{http://dx.doi.org/10.1109/iccv.2019.00442}.

\bibitem[Brachmann et~al.(2018)Brachmann, Krull, Nowozin, Shotton, Michel,
  Gumhold, and Rother]{brachmann2018dsacdifferentiableransac}
Eric Brachmann, Alexander Krull, Sebastian Nowozin, Jamie Shotton, Frank
  Michel, Stefan Gumhold, and Carsten Rother.
\newblock Dsac - differentiable ransac for camera localization, 2018.
\newblock URL \url{https://arxiv.org/abs/1611.05705}.

\bibitem[Brachmann et~al.(2023)Brachmann, Cavallari, and
  Prisacariu]{brachmann2023ace}
Eric Brachmann, Tommaso Cavallari, and Victor~Adrian Prisacariu.
\newblock Accelerated coordinate encoding: Learning to relocalize in minutes
  using rgb and poses.
\newblock In \emph{CVPR}, 2023.

\bibitem[Brachmann et~al.(2024)Brachmann, Wynn, Chen, Cavallari, Monszpart,
  Turmukhambetov, and Prisacariu]{brachmann2024scene}
Eric Brachmann, Jamie Wynn, Shuai Chen, Tommaso Cavallari, {\'A}ron Monszpart,
  Daniyar Turmukhambetov, and Victor~Adrian Prisacariu.
\newblock Scene coordinate reconstruction: {P}osing of image collections via
  incremental learning of a relocalizer.
\newblock In \emph{ECCV}, 2024.

\bibitem[Brown et~al.(2020)Brown, Mann, Ryder, Subbiah, Kaplan, Dhariwal,
  Neelakantan, Shyam, Sastry, Askell, et~al.]{gpt}
Tom Brown, Benjamin Mann, Nick Ryder, Melanie Subbiah, Jared~D Kaplan, Prafulla
  Dhariwal, Arvind Neelakantan, Pranav Shyam, Girish Sastry, Amanda Askell,
  et~al.
\newblock Language models are few-shot learners.
\newblock \emph{Advances in neural information processing systems}, 33, 2020.

\bibitem[Cao et~al.(2020)Cao, Araujo, and Sim]{Cao_2020}
Bingyi Cao, André Araujo, and Jack Sim.
\newblock \emph{Unifying Deep Local and Global Features for Image Search}, page
  726–743.
\newblock Springer International Publishing, 2020.
\newblock ISBN 9783030585655.
\newblock \doi{10.1007/978-3-030-58565-5_43}.
\newblock URL \url{http://dx.doi.org/10.1007/978-3-030-58565-5_43}.

\bibitem[Caron et~al.(2021)Caron, Touvron, Misra, J\'egou, Mairal, Bojanowski,
  and Joulin]{caron2021emerging}
Mathilde Caron, Hugo Touvron, Ishan Misra, Herv\'e J\'egou, Julien Mairal,
  Piotr Bojanowski, and Armand Joulin.
\newblock Emerging properties in self-supervised vision transformers.
\newblock In \emph{Proceedings of the International Conference on Computer
  Vision (ICCV)}, 2021.

\bibitem[Casper et~al.(2023)Casper, Davies, Shi, Gilbert, Scheurer, Rando,
  Freedman, Korbak, Lindner, Freire, Wang, Marks, Segerie, Carroll, Peng,
  Christoffersen, Damani, Slocum, Anwar, Siththaranjan, Nadeau, Michaud, Pfau,
  Krasheninnikov, Chen, Langosco, Hase, Bıyık, Dragan, Krueger, Sadigh, and
  Hadfield-Menell]{casper2023RLHFproblems}
Stephen Casper, Xander Davies, Claudia Shi, Thomas~Krendl Gilbert, Jérémy
  Scheurer, Javier Rando, Rachel Freedman, Tomasz Korbak, David Lindner, Pedro
  Freire, Tony Wang, Samuel Marks, Charbel-Raphaël Segerie, Micah Carroll,
  Andi Peng, Phillip Christoffersen, Mehul Damani, Stewart Slocum, Usman Anwar,
  Anand Siththaranjan, Max Nadeau, Eric~J. Michaud, Jacob Pfau, Dmitrii
  Krasheninnikov, Xin Chen, Lauro Langosco, Peter Hase, Erdem Bıyık, Anca
  Dragan, David Krueger, Dorsa Sadigh, and Dylan Hadfield-Menell.
\newblock Open problems and fundamental limitations of reinforcement learning
  from human feedback, 2023.

\bibitem[Chan et~al.(2021)Chan, Monteiro, Kellnhofer, Wu, and
  Wetzstein]{Chan_2021_CVPR}
Eric~R. Chan, Mariano Monteiro, Petr Kellnhofer, Jiajun Wu, and Gordon
  Wetzstein.
\newblock pi-gan: Periodic implicit generative adversarial networks for
  3d-aware image synthesis.
\newblock In \emph{The IEEE/CVF Conference on Computer Vision and Pattern
  Recognition (CVPR)}, June 2021.

\bibitem[Chan et~al.(2022{\natexlab{a}})Chan, Lin, Chan, Nagano, Pan, De~Mello,
  Gallo, Guibas, Tremblay, Khamis, Karras, and Wetzstein]{chan2022eg3d}
Eric~R. Chan, Connor~Z. Lin, Matthew~A. Chan, Koki Nagano, Boxiao Pan, Shalini
  De~Mello, Orazio Gallo, Leonidas Guibas, Jonathan Tremblay, Sameh Khamis,
  Tero Karras, and Gordon Wetzstein.
\newblock Efficient geometry-aware 3d generative adversarial networks.
\newblock In \emph{The IEEE/CVF Conference on Computer Vision and Pattern
  Recognition (CVPR)}, June 2022{\natexlab{a}}.

\bibitem[Chan et~al.(2022{\natexlab{b}})Chan, Lin, Chan, Nagano, Pan, De~Mello,
  Gallo, Guibas, Tremblay, Khamis, et~al.]{eg3d}
Eric~R Chan, Connor~Z Lin, Matthew~A Chan, Koki Nagano, Boxiao Pan, Shalini
  De~Mello, Orazio Gallo, Leonidas~J Guibas, Jonathan Tremblay, Sameh Khamis,
  et~al.
\newblock Efficient geometry-aware 3d generative adversarial networks.
\newblock In \emph{CVPR}, 2022{\natexlab{b}}.

\bibitem[Chatterjee and Govindu(2013)]{Chatterjee_2013_ICCV}
Avishek Chatterjee and Venu~Madhav Govindu.
\newblock Efficient and robust large-scale rotation averaging.
\newblock In \emph{2013 IEEE International Conference on Computer Vision},
  pages 521--528, December 2013.
\newblock \doi{10.1109/ICCV.2013.70}.
\newblock URL \url{http://dx.doi.org/10.1109/ICCV.2013.70}.

\bibitem[Chatterjee and Govindu(2018)]{Chatterjee_2018}
Avishek Chatterjee and Venu~Madhav Govindu.
\newblock Robust relative rotation averaging.
\newblock \emph{IEEE Transactions on Pattern Analysis and Machine
  Intelligence}, 40\penalty0 (4):\penalty0 958--972, April 2018.
\newblock \doi{10.1109/TPAMI.2017.2693984}.
\newblock URL \url{http://dx.doi.org/10.1109/TPAMI.2017.2693984}.

\bibitem[Chen et~al.(2022{\natexlab{a}})Chen, Xu, Geiger, Yu, and
  Su]{Chen2022tensorf}
Anpei Chen, Zexiang Xu, Andreas Geiger, Jingyi Yu, and Hao Su.
\newblock Tensorf: Tensorial radiance fields.
\newblock In \emph{Computer Vision -- ECCV 2022 -- 17th European Conference,
  Tel Aviv, Israel, October 23-27, 2022, Proceedings, Part XXXII}, volume 13692
  of \emph{Lecture Notes in Computer Science}, pages 333--350. Springer,
  2022{\natexlab{a}}.
\newblock \doi{10.1007/978-3-031-19824-3_20}.
\newblock URL \url{https://doi.org/10.1007/978-3-031-19824-3_20}.

\bibitem[Chen et~al.(2023{\natexlab{a}})Chen, Gu, Chen, Tian, Tu, Liu, and
  Su]{chen2023single}
Hansheng Chen, Jiatao Gu, Anpei Chen, Wei Tian, Zhuowen Tu, Lingjie Liu, and
  Hao Su.
\newblock Single-stage diffusion nerf: A unified approach to 3d generation and
  reconstruction.
\newblock In \emph{ICCV}, 2023{\natexlab{a}}.

\bibitem[Chen et~al.(2022{\natexlab{b}})Chen, Luo, Zhou, Tian, Zhen, Fang,
  Mckinnon, Tsin, and Quan]{chen2022aspanformerdetectorfreeimagematching}
Hongkai Chen, Zixin Luo, Lei Zhou, Yurun Tian, Mingmin Zhen, Tian Fang, David
  Mckinnon, Yanghai Tsin, and Long Quan.
\newblock Aspanformer: Detector-free image matching with adaptive span
  transformer, 2022{\natexlab{b}}.
\newblock URL \url{https://arxiv.org/abs/2208.14201}.

\bibitem[Chen et~al.(2023{\natexlab{b}})Chen, Chen, Jiao, and
  Jia]{Chen_2023_ICCV}
Rui Chen, Yongwei Chen, Ningxin Jiao, and Kui Jia.
\newblock Fantasia3d: Disentangling geometry and appearance for high-quality
  text-to-3d content creation.
\newblock In \emph{Proceedings of the IEEE/CVF International Conference on
  Computer Vision (ICCV)}, pages 20622--20633, October 2023{\natexlab{b}}.

\bibitem[Chen et~al.(2023{\natexlab{c}})Chen, Chen, Jiao, and
  Jia]{chen2023fantasia3d}
Rui Chen, Yongwei Chen, Ningxin Jiao, and Kui Jia.
\newblock Fantasia3d: Disentangling geometry and appearance for high-quality
  text-to-3d content creation.
\newblock \emph{arXiv preprint arXiv:2303.13873}, 2023{\natexlab{c}}.

\bibitem[Chen and Zhang(2019)]{Chen_2019_CVPR}
Zhiqin Chen and Hao Zhang.
\newblock Learning implicit fields for generative shape modeling.
\newblock In \emph{The IEEE Conference on Computer Vision and Pattern
  Recognition (CVPR)}, June 2019.

\bibitem[Chen et~al.(2024)Chen, Wang, Wang, Wang, and
  Liu]{chen2024v3dvideodiffusionmodels}
Zilong Chen, Yikai Wang, Feng Wang, Zhengyi Wang, and Huaping Liu.
\newblock V3d: Video diffusion models are effective 3d generators, 2024.
\newblock URL \url{https://arxiv.org/abs/2403.06738}.

\bibitem[Chung et~al.(2023)Chung, Lee, Nam, Lee, and
  Lee]{chung2023luciddreamer}
Jaeyoung Chung, Suyoung Lee, Hyeongjin Nam, Jaerin Lee, and Kyoung~Mu Lee.
\newblock Luciddreamer: Domain-free generation of 3d gaussian splatting scenes.
\newblock \emph{arXiv preprint arXiv:2311.13384}, 2023.

\bibitem[Collins et~al.(2022)Collins, Goel, Deng, Luthra, Xu, Gundogdu, Zhang,
  Vicente, Dideriksen, Arora, et~al.]{abo}
Jasmine Collins, Shubham Goel, Kenan Deng, Achleshwar Luthra, Leon Xu, Erhan
  Gundogdu, Xi~Zhang, Tomas F~Yago Vicente, Thomas Dideriksen, Himanshu Arora,
  et~al.
\newblock Abo: Dataset and benchmarks for real-world 3d object understanding.
\newblock In \emph{CVPR}, pages 21126--21136, 2022.

\bibitem[Crandall et~al.(2011)Crandall, Owens, Snavely, and
  Huttenlocher]{Crandall_2011}
David Crandall, Andrew Owens, Noah Snavely, and Dan Huttenlocher.
\newblock Discrete-continuous optimization for large-scale structure from
  motion.
\newblock In \emph{CVPR 2011}, June 2011.
\newblock \doi{10.1109/CVPR.2011.5995626}.
\newblock URL \url{http://dx.doi.org/10.1109/CVPR.2011.5995626}.

\bibitem[Cui and Tan(2015)]{Cui_2015_ICCV}
Zhaopeng Cui and Ping Tan.
\newblock Global structure-from-motion by similarity averaging.
\newblock In \emph{2015 IEEE International Conference on Computer Vision
  (ICCV)}, pages 864--872, December 2015.
\newblock \doi{10.1109/ICCV.2015.105}.
\newblock URL \url{http://dx.doi.org/10.1109/ICCV.2015.105}.

\bibitem[Deitke et~al.(2022)Deitke, Schwenk, Salvador, Weihs, Michel,
  VanderBilt, Schmidt, Ehsani, Kembhavi, and Farhadi]{deitke2022objaverse}
Matt Deitke, Dustin Schwenk, Jordi Salvador, Luca Weihs, Oscar Michel, Eli
  VanderBilt, Ludwig Schmidt, Kiana Ehsani, Aniruddha Kembhavi, and Ali
  Farhadi.
\newblock Objaverse: A universe of annotated 3d objects, 2022.

\bibitem[Deitke et~al.(2023{\natexlab{a}})Deitke, Liu, Wallingford, Ngo,
  Michel, Kusupati, Fan, Laforte, Voleti, Gadre, et~al.]{deitke2023objaversexl}
Matt Deitke, Ruoshi Liu, Matthew Wallingford, Huong Ngo, Oscar Michel, Aditya
  Kusupati, Alan Fan, Christian Laforte, Vikram Voleti, Samir~Yitzhak Gadre,
  et~al.
\newblock Objaverse-xl: A universe of 10m+ 3d objects.
\newblock \emph{arXiv preprint arXiv:2307.05663}, 2023{\natexlab{a}}.

\bibitem[Deitke et~al.(2023{\natexlab{b}})Deitke, Schwenk, Salvador, Weihs,
  Michel, VanderBilt, Schmidt, Ehsani, Kembhavi, and
  Farhadi]{deitke2023objaverse}
Matt Deitke, Dustin Schwenk, Jordi Salvador, Luca Weihs, Oscar Michel, Eli
  VanderBilt, Ludwig Schmidt, Kiana Ehsani, Aniruddha Kembhavi, and Ali
  Farhadi.
\newblock Objaverse: A universe of annotated 3d objects.
\newblock In \emph{Proceedings of the IEEE/CVF Conference on Computer Vision
  and Pattern Recognition}, pages 13142--13153, 2023{\natexlab{b}}.

\bibitem[Deitke et~al.(2023{\natexlab{c}})Deitke, Schwenk, Salvador, Weihs,
  Michel, VanderBilt, Schmidt, Ehsani, Kembhavi, and Farhadi]{objaverse}
Matt Deitke, Dustin Schwenk, Jordi Salvador, Luca Weihs, Oscar Michel, Eli
  VanderBilt, Ludwig Schmidt, Kiana Ehsani, Aniruddha Kembhavi, and Ali
  Farhadi.
\newblock Objaverse: A universe of annotated 3d objects.
\newblock In \emph{CVPR}, pages 13142--13153, 2023{\natexlab{c}}.

\bibitem[DeTone et~al.(2018)DeTone, Malisiewicz, and Rabinovich]{DeTone_2018}
Daniel DeTone, Tomasz Malisiewicz, and Andrew Rabinovich.
\newblock Superpoint: Self-supervised interest point detection and description.
\newblock In \emph{2018 IEEE/CVF Conference on Computer Vision and Pattern
  Recognition Workshops (CVPRW)}, page 337–33712. IEEE, June 2018.
\newblock \doi{10.1109/cvprw.2018.00060}.
\newblock URL \url{http://dx.doi.org/10.1109/cvprw.2018.00060}.

\bibitem[Developers(2026)]{theia_website}
Theia Developers.
\newblock Theia: A library for structure-from-motion.
\newblock \\url{http://www.theia-sfm.org/}, 2026.
\newblock Accessed 2026-02-03.

\bibitem[Dong et~al.(2025)Dong, Wang, Liu, Cai, Fan, Kannala, and
  Yang]{Dong_2025_CVPR}
Siyan Dong, Shuzhe Wang, Shaohui Liu, Lulu Cai, Qingnan Fan, Juho Kannala, and
  Yanchao Yang.
\newblock Reloc3r: Large-scale training of relative camera pose regression for
  generalizable, fast, and accurate visual localization.
\newblock In \emph{Proceedings of the IEEE/CVF Conference on Computer Vision
  and Pattern Recognition (CVPR)}, pages 16739--16752, June 2025.

\bibitem[Downs et~al.(2022)Downs, Francis, Koenig, Kinman, Hickman, Reymann,
  McHugh, and Vanhoucke]{gso}
Laura Downs, Anthony Francis, Nate Koenig, Brandon Kinman, Ryan Hickman, Krista
  Reymann, Thomas~B McHugh, and Vincent Vanhoucke.
\newblock Google scanned objects: A high-quality dataset of 3d scanned
  household items.
\newblock In \emph{2022 International Conference on Robotics and Automation
  (ICRA)}, pages 2553--2560. IEEE, 2022.

\bibitem[Duane(1971)]{duane1971close}
C~Brown Duane.
\newblock Close-range camera calibration.
\newblock \emph{Photogramm. Eng}, 37\penalty0 (8), 1971.

\bibitem[Duisterhof et~al.(2024)Duisterhof, Zust, Weinzaepfel, Leroy, Cabon,
  and Revaud]{duisterhof2024mast3r}
Bardienus Duisterhof, Lojze Zust, Philippe Weinzaepfel, Vincent Leroy, Yohann
  Cabon, and Jerome Revaud.
\newblock {MASt3R-SfM}: {A} fully-integrated solution for unconstrained
  structure-from-motion.
\newblock \emph{arXiv preprint arXiv:2409.19152}, 2024.

\bibitem[Duisterhof et~al.(2025)Duisterhof, Zust, Weinzaepfel, Leroy, Cabon,
  and Revaud]{duisterhof2025mastrsfm}
Bardienus~Pieter Duisterhof, Lojze Zust, Philippe Weinzaepfel, Vincent Leroy,
  Yohann Cabon, and Jerome Revaud.
\newblock {MAS}t3r-sfm: a fully-integrated solution for unconstrained
  structure-from-motion.
\newblock In \emph{International Conference on 3D Vision (3DV)}, 2025.
\newblock URL \url{https://openreview.net/forum?id=5uw1GRBFoT}.

\bibitem[Dusmanu et~al.(2019)Dusmanu, Rocco, Pajdla, Pollefeys, Sivic, Torii,
  and Sattler]{dusmanu2019d2nettrainablecnnjoint}
Mihai Dusmanu, Ignacio Rocco, Tomas Pajdla, Marc Pollefeys, Josef Sivic,
  Akihiko Torii, and Torsten Sattler.
\newblock D2-net: A trainable cnn for joint detection and description of local
  features, 2019.
\newblock URL \url{https://arxiv.org/abs/1905.03561}.

\bibitem[Dusmanu et~al.(2020)Dusmanu, Schönberger, and
  Pollefeys]{Dusmanu_2020}
Mihai Dusmanu, Johannes~L. Schönberger, and Marc Pollefeys.
\newblock \emph{Multi-view Optimization of Local Feature Geometry}, page
  670–686.
\newblock Springer International Publishing, 2020.
\newblock ISBN 9783030584528.
\newblock \doi{10.1007/978-3-030-58452-8_39}.
\newblock URL \url{http://dx.doi.org/10.1007/978-3-030-58452-8_39}.

\bibitem[Eimer et~al.(2023)Eimer, Lindauer, and
  Raileanu]{eimer2023rlhyperparameters}
Theresa Eimer, Marius Lindauer, and Roberta Raileanu.
\newblock Hyperparameters in reinforcement learning and how to tune them, 2023.

\bibitem[Engstler et~al.(2025)Engstler, Xia, Trabucco, Yan, Ding, Chen, and
  Wetzstein]{Engstler_2025_ICCV}
Jakob Engstler, Gaisi Xia, Brandon Trabucco, Huicheng Yan, Rui Ding, Yun-Chun
  Chen, and Gordon Wetzstein.
\newblock Syncity: Training-free generation of 3d worlds.
\newblock In \emph{Proceedings of the IEEE/CVF International Conference on
  Computer Vision (ICCV)}, pages 18577--18587, 2025.

\bibitem[Epstein et~al.(2024)Epstein, Kim, Wang, Chen, Tseng, Chen, Yang, and
  Wang]{pmlr-v235-epstein24a}
Dave Epstein, Alisa Kim, Xueyue Wang, Hsin-Ying Chen, Hung-Yu Tseng,
  Hsiao-Yu~Fish Chen, Ming-Hsuan Yang, and Wei Wang.
\newblock Disentangled 3d scene generation with layout learning.
\newblock In Dylan Fortune, Paul Vicol, Serena Yeung, Hannah Zhang, Frank
  Hutter, and Hugo Larochelle, editors, \emph{Proceedings of the 41st
  International Conference on Machine Learning}, volume 235 of
  \emph{Proceedings of Machine Learning Research}, pages 11278--11298. PMLR,
  21--27 Jul 2024.
\newblock URL \url{https://proceedings.mlr.press/v235/epstein24a.html}.

\bibitem[Erdn{\"u}{\ss}(2021)]{erdnuss2021review}
Bastian Erdn{\"u}{\ss}.
\newblock A review of the one-parameter division undistortion model.
\newblock \emph{ISPRS Annals of the Photogrammetry, Remote Sensing and Spatial
  Information Sciences}, 1, 2021.

\bibitem[Fan et~al.(2017)Fan, Su, and Guibas]{Fan_2017_CVPR}
Haoqiang Fan, Hao Su, and Leonidas~J. Guibas.
\newblock A point set generation network for 3d object reconstruction from a
  single image.
\newblock In \emph{The IEEE Conference on Computer Vision and Pattern
  Recognition (CVPR)}, July 2017.

\bibitem[Fan et~al.(2023)Fan, Watkins, Du, Liu, Ryu, Boutilier, Abbeel,
  Ghavamzadeh, Lee, and Lee]{fan2023dpok}
Ying Fan, Olivia Watkins, Yuqing Du, Hao Liu, Moonkyung Ryu, Craig Boutilier,
  Pieter Abbeel, Mohammad Ghavamzadeh, Kangwook Lee, and Kimin Lee.
\newblock Dpok: Reinforcement learning for fine-tuning text-to-image diffusion
  models, 2023.

\bibitem[Fang et~al.(2025)Fang, Dong, Luo, Hu, Shrestha, and
  Tan]{Fang_2025_3DV}
Chuan Fang, Yuan Dong, Kunming Luo, Xiaotao Hu, Rakesh Shrestha, and Ping Tan.
\newblock Ctrl-room: Controllable text-to-3d room meshes generation with layout
  constraints.
\newblock In \emph{International Conference on 3D Vision (3DV)}, pages
  692--701, 2025.
\newblock \doi{10.1109/3DV66043.2025.00069}.

\bibitem[Faugeras(1993)]{faugeras1993three}
Olivier Faugeras.
\newblock \emph{Three-dimensional computer vision: {A} geometric viewpoint}.
\newblock MIT Press, 1993.

\bibitem[Fitzgibbon(2001)]{fitzgibbon2001simultaneous}
Andrew~W Fitzgibbon.
\newblock Simultaneous linear estimation of multiple view geometry and lens
  distortion.
\newblock In \emph{CVPR}, 2001.

\bibitem[Gao et~al.(2022)Gao, Shen, Wang, Chen, Yin, Li, Litany, Gojcic, and
  Fidler]{gao2022get3d}
Jun Gao, Tianchang Shen, Zian Wang, Wenzheng Chen, Kangxue Yin, Daiqing Li,
  Or~Litany, Zan Gojcic, and Sanja Fidler.
\newblock Get3d: A generative model of high quality 3d textured shapes learned
  from images.
\newblock \emph{Advances in Neural Information Processing Systems},
  35:\penalty0 31841--31854, 2022.

\bibitem[Ge et~al.(2024)Ge, Zhao, Zhu, Ge, Yi, Song, Li, Ding, and
  Shan]{ge2024seedx}
Yuying Ge, Sijie Zhao, Jinguo Zhu, Yixiao Ge, Kun Yi, Lin Song, Chen Li,
  Xiaohan Ding, and Ying Shan.
\newblock {SEED-X:} multimodal models with unified multi-granularity
  comprehension and generation.
\newblock \emph{CoRR}, abs/2404.14396, 2024.
\newblock \doi{10.48550/ARXIV.2404.14396}.
\newblock URL \url{https://doi.org/10.48550/arXiv.2404.14396}.

\bibitem[Genova et~al.(2020)Genova, Highlander, Sirin, Liao, Daniilidis, and
  Funkhouser]{Genova_2020_CVPR}
Kyle Genova, Ethan Highlander, Ozgur Sirin, Wonmin Liao, Kostas Daniilidis, and
  Thomas Funkhouser.
\newblock Learning shape templates with structured implicit functions.
\newblock In \emph{The IEEE/CVF Conference on Computer Vision and Pattern
  Recognition (CVPR)}, June 2020.

\bibitem[Germain et~al.(2020)Germain, Bourmaud, and
  Lepetit]{germain2020s2dnetlearningaccuratecorrespondences}
Hugo Germain, Guillaume Bourmaud, and Vincent Lepetit.
\newblock S2dnet: Learning accurate correspondences for sparse-to-dense feature
  matching, 2020.
\newblock URL \url{https://arxiv.org/abs/2004.01673}.

\bibitem[gi~Kwak et~al.(2023)gi~Kwak, Dong, Jin, Ko, Mahajan, and
  Yi]{kwak2023vivid1to3novelviewsynthesis}
Jeong gi~Kwak, Erqun Dong, Yuhe Jin, Hanseok Ko, Shweta Mahajan, and Kwang~Moo
  Yi.
\newblock Vivid-1-to-3: Novel view synthesis with video diffusion models, 2023.
\newblock URL \url{https://arxiv.org/abs/2312.01305}.

\bibitem[Goldstein et~al.(2016)Goldstein, Hand, Lee, Voroninski, and
  Soatto]{Goldstein_2016}
Thomas Goldstein, Paul Hand, Choongbum Lee, Vladislav Voroninski, and Stefano
  Soatto.
\newblock Shapefit and shapekick for robust, scalable structure from motion.
\newblock In \emph{Computer Vision -- ECCV 2016}, pages 289--304, 2016.
\newblock \doi{10.1007/978-3-319-46478-7_18}.
\newblock URL \url{http://dx.doi.org/10.1007/978-3-319-46478-7_18}.

\bibitem[G{\'o}mez et~al.(2025)G{\'o}mez, Silva, Seoane, Borr{\'a}s, Noriega,
  Ros, Iglesias-Guitian, and L{\'o}pez]{urbansyn}
Jose~L G{\'o}mez, Manuel Silva, Antonio Seoane, Agn{\`e}s Borr{\'a}s, Mario
  Noriega, Germ{\'a}n Ros, Jose~A Iglesias-Guitian, and Antonio~M L{\'o}pez.
\newblock All for one, and one for all: Urbansyn dataset, the third musketeer
  of synthetic driving scenes.
\newblock \emph{Neurocomputing}, 637:\penalty0 130038, 2025.

\bibitem[Govindu(2001)]{Govindu_2001}
V.~M. Govindu.
\newblock Combining two-view constraints for motion estimation.
\newblock In \emph{Proceedings of the 2001 IEEE Computer Society Conference on
  Computer Vision and Pattern Recognition. CVPR 2001}, volume~2, pages
  II--218--II--225, 2001.
\newblock \doi{10.1109/CVPR.2001.990963}.
\newblock URL \url{http://dx.doi.org/10.1109/CVPR.2001.990963}.

\bibitem[Govindu(2004)]{Govindu_2004}
V.~M. Govindu.
\newblock Lie-algebraic averaging for globally consistent motion estimation.
\newblock In \emph{Proceedings of the 2004 IEEE Computer Society Conference on
  Computer Vision and Pattern Recognition, 2004. CVPR 2004.}, volume~1, pages
  684--691, 2004.
\newblock \doi{10.1109/CVPR.2004.1315098}.
\newblock URL \url{http://dx.doi.org/10.1109/CVPR.2004.1315098}.

\bibitem[Greff et~al.(2022)Greff, Belletti, Beyer, Doersch, Du, Duckworth,
  Fleet, Gnanapragasam, Golemo, Herrmann, et~al.]{greff2022kubric}
Klaus Greff, Francois Belletti, Lucas Beyer, Carl Doersch, Yilun Du, Daniel
  Duckworth, David~J Fleet, Dan Gnanapragasam, Florian Golemo, Charles
  Herrmann, et~al.
\newblock Kubric: A scalable dataset generator.
\newblock In \emph{Proceedings of the IEEE/CVF conference on computer vision
  and pattern recognition}, pages 3749--3761, 2022.

\bibitem[Groueix et~al.(2018)Groueix, Fisher, Kim, Russell, and
  Aubry]{Groueix_2018_CVPR}
Thibault Groueix, Matthew Fisher, Vladimir~G. Kim, Bryan~C. Russell, and
  Mathieu Aubry.
\newblock A papier-mache approach to learning 3d surface generation.
\newblock In \emph{The IEEE Conference on Computer Vision and Pattern
  Recognition (CVPR)}, June 2018.

\bibitem[Gu et~al.(2020)Gu, Fan, Dai, Zhu, Tan, and
  Tan]{gu2020cascadecostvolumehighresolution}
Xiaodong Gu, Zhiwen Fan, Zuozhuo Dai, Siyu Zhu, Feitong Tan, and Ping Tan.
\newblock Cascade cost volume for high-resolution multi-view stereo and stereo
  matching, 2020.
\newblock URL \url{https://arxiv.org/abs/1912.06378}.

\bibitem[Guo et~al.(2023)Guo, Liu, Shao, Laforte, Voleti, Luo, Chen, Zou, Wang,
  Cao, and Zhang]{threestudio2023}
Yuan-Chen Guo, Ying-Tian Liu, Ruizhi Shao, Christian Laforte, Vikram Voleti,
  Guan Luo, Chia-Hao Chen, Zi-Xin Zou, Chen Wang, Yan-Pei Cao, and Song-Hai
  Zhang.
\newblock threestudio: A unified framework for 3d content generation.
\newblock \url{https://github.com/threestudio-project/threestudio}, 2023.

\bibitem[Gupta et~al.(2023)Gupta, Xiong, Nie, Jones, and
  O{\u{g}}uz]{gupta20233dgen}
Anchit Gupta, Wenhan Xiong, Yixin Nie, Ian Jones, and Barlas O{\u{g}}uz.
\newblock 3dgen: Triplane latent diffusion for textured mesh generation.
\newblock \emph{arXiv preprint arXiv:2303.05371}, 2023.

\bibitem[Haque et~al.(2023)Haque, Tancik, Efros, Holynski, and
  Kanazawa]{Haque_2023_ICCV}
Ayaan Haque, Matthew Tancik, Alexei~A. Efros, Aleksander Holynski, and Angjoo
  Kanazawa.
\newblock Instruct-nerf2nerf: Editing 3d scenes with instructions.
\newblock In \emph{Proceedings of the IEEE/CVF International Conference on
  Computer Vision (ICCV)}, pages 20369--20379, October 2023.

\bibitem[Harley et~al.(2025)Harley, You, Sun, Zheng, Raghuraman, Gu, Liang,
  Chu, Dave, Tokmakov, et~al.]{harley2025alltracker}
Adam~W Harley, Yang You, Xinglong Sun, Yang Zheng, Nikhil Raghuraman, Yunqi Gu,
  Sheldon Liang, Wen-Hsuan Chu, Achal Dave, Pavel Tokmakov, et~al.
\newblock Alltracker: Efficient dense point tracking at high resolution.
\newblock \emph{arXiv preprint arXiv:2506.07310}, 2025.

\bibitem[Hartley(1993)]{hartley1993extraction}
Richard Hartley.
\newblock Extraction of focal lengths from the fundamental matrix.
\newblock \emph{Unpublished manuscript}, 2, 1993.

\bibitem[Hartley and Zisserman(2003)]{hartley2003multiple}
Richard Hartley and Andrew Zisserman.
\newblock \emph{Multiple view geometry in computer vision}.
\newblock Cambridge University Press, 2003.

\bibitem[Hartley et~al.(2011)Hartley, Aftab, and Trumpf]{Hartley_2011}
Richard Hartley, Khurrum Aftab, and Jochen Trumpf.
\newblock L1 rotation averaging using the weiszfeld algorithm.
\newblock In \emph{CVPR 2011}, pages 3041--3048, June 2011.
\newblock \doi{10.1109/CVPR.2011.5995745}.
\newblock URL \url{http://dx.doi.org/10.1109/CVPR.2011.5995745}.

\bibitem[Hartley et~al.(2013)Hartley, Trumpf, Dai, and Li]{Hartley_2013}
Richard Hartley, Jochen Trumpf, Yuchao Dai, and Hongdong Li.
\newblock Rotation averaging.
\newblock \emph{International Journal of Computer Vision}, 103\penalty0
  (3):\penalty0 267--305, January 2013.
\newblock \doi{10.1007/s11263-012-0601-0}.
\newblock URL \url{http://dx.doi.org/10.1007/s11263-012-0601-0}.

\bibitem[Hausler et~al.(2021)Hausler, Garg, Xu, Milford, and
  Fischer]{hausler2021patchnetvladmultiscalefusionlocallyglobal}
Stephen Hausler, Sourav Garg, Ming Xu, Michael Milford, and Tobias Fischer.
\newblock Patch-netvlad: Multi-scale fusion of locally-global descriptors for
  place recognition, 2021.
\newblock URL \url{https://arxiv.org/abs/2103.01486}.

\bibitem[He et~al.(2018)He, Gkioxari, Dollár, and Girshick]{he2018maskrcnn}
Kaiming He, Georgia Gkioxari, Piotr Dollár, and Ross Girshick.
\newblock Mask r-cnn, 2018.
\newblock URL \url{https://arxiv.org/abs/1703.06870}.

\bibitem[Heinly et~al.(2015)Heinly, Schonberger, Dunn, and
  Frahm]{Heinly_2015_CVPR}
Jared Heinly, Johannes~L. Schonberger, Enrique Dunn, and Jan-Michael Frahm.
\newblock Reconstructing the world in six days.
\newblock In \emph{2015 IEEE Conference on Computer Vision and Pattern
  Recognition (CVPR)}, June 2015.
\newblock \doi{10.1109/CVPR.2015.7298949}.
\newblock URL \url{http://dx.doi.org/10.1109/CVPR.2015.7298949}.

\bibitem[Henrique~Brito et~al.(2013)Henrique~Brito, Angst, Koser, and
  Pollefeys]{henrique2013radial}
Jose Henrique~Brito, Roland Angst, Kevin Koser, and Marc Pollefeys.
\newblock Radial distortion self-calibration.
\newblock In \emph{CVPR}, 2013.

\bibitem[Ho et~al.(2020)Ho, Jain, and Abbeel]{ho2020denoising}
Jonathan Ho, Ajay Jain, and Pieter Abbeel.
\newblock Denoising diffusion probabilistic models.
\newblock \emph{Advances in neural information processing systems},
  33:\penalty0 6840--6851, 2020.

\bibitem[Hollein et~al.(2023)Hollein, Ceylan, Johnson, and
  Nie{\ss}ner]{Hollein_2023_ICCV}
Lukas Hollein, Duygu Ceylan, Justin Johnson, and Matthias Nie{\ss}ner.
\newblock Text2room: Extracting textured 3d meshes from 2d text-to-image
  models.
\newblock In \emph{Proceedings of the IEEE/CVF International Conference on
  Computer Vision (ICCV)}, pages 7903--7913, October 2023.

\bibitem[Hong et~al.(2023)Hong, Zhang, Gu, Bi, Zhou, Liu, Liu, Sunkavalli, Bui,
  and Tan]{hong2023lrm}
Yicong Hong, Kai Zhang, Jiuxiang Gu, Sai Bi, Yang Zhou, Difan Liu, Feng Liu,
  Kalyan Sunkavalli, Trung Bui, and Hao Tan.
\newblock Lrm: Large reconstruction model for single image to 3d.
\newblock \emph{arXiv preprint arXiv:2311.04400}, 2023.

\bibitem[Hou et~al.(2025)Hou, Li, Yang, Chen, Qian, Zhao, Jiang, Wei, Xu, and
  Zhang]{hou2025bloomscene}
Xiaolu Hou, Mingcheng Li, Dingkang Yang, Jiawei Chen, Ziyun Qian, Xiao Zhao,
  Yue Jiang, Jinjie Wei, Qingyao Xu, and Lihua Zhang.
\newblock Bloomscene: Lightweight structured 3d gaussian splatting for
  crossmodal scene generation.
\newblock In \emph{Proceedings of the AAAI Conference on Artificial
  Intelligence}, volume~39, pages 3536--3544, 2025.
\newblock \doi{10.1609/aaai.v39i4.32367}.

\bibitem[Huang and Belongie(2017)]{huang2017arbitrary}
Xun Huang and Serge Belongie.
\newblock Arbitrary style transfer in real-time with adaptive instance
  normalization.
\newblock In \emph{Proceedings of the IEEE international conference on computer
  vision}, pages 1501--1510, 2017.

\bibitem[Huang et~al.(2023)Huang, Wang, Tao, Jing, Deitke, Koltun, and
  Niu]{huang2023dreamtime}
Yukun Huang, Yiqi Wang, Zhiwei Tao, Shuang Jing, Matt Deitke, Vladlen Koltun,
  and Yiliang Niu.
\newblock Dreamtime: An improved optimization strategy for text-to-3d content
  creation, 2023.

\bibitem[Jeong et~al.(2023)Jeong, Lee, Kim, Li, Joulin, Peng, and
  Yu]{kim2023csd}
Jongheon Jeong, Dae~Young Lee, Joo~Wan Kim, Francis~F. Li, Anton Joulin, Nanyun
  Peng, and Tianhe Yu.
\newblock Collaborative score distillation for consistent visual editing, 2023.

\bibitem[Jiang et~al.(2013)Jiang, Cui, and Tan]{jiang2013global}
Nianjuan Jiang, Zhaopeng Cui, and Ping Tan.
\newblock A global linear method for camera pose registration.
\newblock In \emph{ICCV}, 2013.

\bibitem[Jiang et~al.(2021)Jiang, Trulls, Hosang, Tagliasacchi, and
  Yi]{jiang2021cotrcorrespondencetransformermatching}
Wei Jiang, Eduard Trulls, Jan Hosang, Andrea Tagliasacchi, and Kwang~Moo Yi.
\newblock Cotr: Correspondence transformer for matching across images, 2021.
\newblock URL \url{https://arxiv.org/abs/2103.14167}.

\bibitem[Jiang et~al.(2025)Jiang, Wang, Galliani, Vogel, and
  Pollefeys]{Jiang_2025_CVPR}
Xudong Jiang, Fangjinhua Wang, Silvano Galliani, Christoph Vogel, and Marc
  Pollefeys.
\newblock R-score: Revisiting scene coordinate regression for robust
  large-scale visual localization.
\newblock In \emph{Proceedings of the IEEE/CVF Conference on Computer Vision
  and Pattern Recognition (CVPR)}, pages 11536--11546, June 2025.

\bibitem[Jin et~al.(2023)Jin, Zhang, Hold-Geoffroy, Wang, Blackburn-Matzen,
  Sticha, and Fouhey]{Jin_2023_CVPR}
Linyi Jin, Jianming Zhang, Yannick Hold-Geoffroy, Oliver Wang, Kevin
  Blackburn-Matzen, Matthew Sticha, and David~F. Fouhey.
\newblock Perspective fields for single image camera calibration.
\newblock In \emph{Proceedings of the IEEE/CVF Conference on Computer Vision
  and Pattern Recognition (CVPR)}, pages 17307--17316, June 2023.

\bibitem[Jun and Nichol(2023{\natexlab{a}})]{jun2023shap}
Heewoo Jun and Alex Nichol.
\newblock Shap-e: Generating conditional 3d implicit functions.
\newblock \emph{arXiv preprint arXiv:2305.02463}, 2023{\natexlab{a}}.

\bibitem[Jun and Nichol(2023{\natexlab{b}})]{shape}
Heewoo Jun and Alex Nichol.
\newblock Shap-e: Generating conditional 3d implicit functions.
\newblock \emph{arXiv preprint arXiv:2305.02463}, 2023{\natexlab{b}}.

\bibitem[Kanatani and Matsunaga(2000)]{kanatani2000closed}
Kenichi Kanatani and Chikara Matsunaga.
\newblock Closed-form expression for focal lengths from the fundamental matrix.
\newblock In \emph{Proceedings of the Asian Conference on Computer Vision},
  2000.

\bibitem[Kaplan et~al.(2020)Kaplan, McCandlish, Henighan, Brown, Chess, Child,
  Gray, Radford, Wu, and Amodei]{kaplan2020scalinglaws}
Jared Kaplan, Sam McCandlish, Tom Henighan, Tom~B. Brown, Benjamin Chess, Rewon
  Child, Scott Gray, Alec Radford, Jeffrey Wu, and Dario Amodei.
\newblock Scaling laws for neural language models, 2020.

\bibitem[Karaev et~al.(2024)Karaev, Rocco, Graham, Neverova, Vedaldi, and
  Rupprecht]{karaev2024cotracker}
Nikita Karaev, Ignacio Rocco, Benjamin Graham, Natalia Neverova, Andrea
  Vedaldi, and Christian Rupprecht.
\newblock Cotracker: It is better to track together.
\newblock In \emph{European conference on computer vision}, pages 18--35.
  Springer, 2024.

\bibitem[Ke et~al.(2024{\natexlab{a}})Ke, Obukhov, Huang, Metzger, Daudt, and
  Schindler]{Ke_2024_CVPR}
Bingxin Ke, Anton Obukhov, Shengyu Huang, Nando Metzger, Rodrigo~Caye Daudt,
  and Konrad Schindler.
\newblock Repurposing diffusion-based image generators for monocular depth
  estimation.
\newblock In \emph{Proceedings of the IEEE/CVF Conference on Computer Vision
  and Pattern Recognition (CVPR)}, pages 9492--9502, June 2024{\natexlab{a}}.

\bibitem[Ke et~al.(2024{\natexlab{b}})Ke, Obukhov, Huang, Metzger, Daudt, and
  Schindler]{ke2024repurposing}
Bingxin Ke, Anton Obukhov, Shengyu Huang, Nando Metzger, Rodrigo~Caye Daudt,
  and Konrad Schindler.
\newblock Repurposing diffusion-based image generators for monocular depth
  estimation.
\newblock In \emph{Proceedings of the IEEE/CVF conference on computer vision
  and pattern recognition}, pages 9492--9502, 2024{\natexlab{b}}.

\bibitem[Kerbl et~al.(2023)Kerbl, Kopanas, Leimk{\"u}hler, and
  Drettakis]{kerbl20233d}
Bernhard Kerbl, Georgios Kopanas, Thomas Leimk{\"u}hler, and George Drettakis.
\newblock {3D} {G}aussian splatting for real-time radiance field rendering.
\newblock \emph{ACM Transactions on Graphics}, 42\penalty0 (4), 2023.

\bibitem[Khafizov et~al.(2025)Khafizov, Komarichev, Rakhimov, Wonka, and
  Burnaev]{khafizov2025gcut3r}
Ramil Khafizov, Artem Komarichev, Ruslan Rakhimov, Peter Wonka, and Evgeny
  Burnaev.
\newblock G-cut3r: Guided 3d reconstruction with camera and depth prior
  integration.
\newblock \emph{CoRR}, abs/2508.11379, 2025.

\bibitem[Kingma and Ba(2014)]{kingma2014adam}
Diederik~P Kingma and Jimmy Ba.
\newblock Adam: A method for stochastic optimization.
\newblock \emph{arXiv preprint arXiv:1412.6980}, 2014.

\bibitem[Kirillov et~al.(2023)Kirillov, Mintun, Ravi, Mao, Rolland, Gustafson,
  Xiao, Whitehead, Berg, Lo, et~al.]{kirillov2023segment}
Alexander Kirillov, Eric Mintun, Nikhila Ravi, Hanzi Mao, Chloe Rolland, Laura
  Gustafson, Tete Xiao, Spencer Whitehead, Alexander~C Berg, Wan-Yen Lo, et~al.
\newblock Segment anything.
\newblock \emph{arXiv preprint arXiv:2304.02643}, 2023.

\bibitem[Knapitsch et~al.(2017)Knapitsch, Park, Zhou, and
  Koltun]{knapitsch2017tanks}
Arno Knapitsch, Jaesik Park, Qian-Yi Zhou, and Vladlen Koltun.
\newblock Tanks and temples: {B}enchmarking large-scale scene reconstruction.
\newblock \emph{ACM Transactions on Graphics (ToG)}, 36\penalty0 (4), 2017.

\bibitem[Kocur et~al.(2024)Kocur, Kyselica, and Kukelova]{kocur2024robust}
Viktor Kocur, Daniel Kyselica, and Zuzana Kukelova.
\newblock Robust self-calibration of focal lengths from the fundamental matrix.
\newblock In \emph{CVPR}, 2024.

\bibitem[Koh et~al.(2023)Koh, Agrawal, Batra, Tucker, Waters, Lee, Yang,
  Baldridge, and Anderson]{Koh_2023_AAAI}
Jing~Yu Koh, Harsh Agrawal, Dhruv Batra, Richard Tucker, Austin Waters, Honglak
  Lee, Yinfei Yang, Jason Baldridge, and Peter Anderson.
\newblock Simple and effective synthesis of indoor 3d scenes.
\newblock In \emph{Proceedings of the AAAI Conference on Artificial
  Intelligence}, 2023.

\bibitem[Kondratyuk et~al.(2023)Kondratyuk, Yu, Gu, Lezama, Huang, Hornung,
  Adam, Akbari, Alon, Birodkar, Cheng, Chiu, Dillon, Essa, Gupta, Hahn, Hauth,
  Hendon, Martinez, Minnen, Ross, Schindler, Sirotenko, Sohn, Somandepalli,
  Wang, Yan, Yang, Yang, Seybold, and Jiang]{kondratyuk2023videopoet}
Dan Kondratyuk, Lijun Yu, Xiuye Gu, Jos{\'{e}} Lezama, Jonathan Huang, Rachel
  Hornung, Hartwig Adam, Hassan Akbari, Yair Alon, Vighnesh Birodkar, Yong
  Cheng, Ming{-}Chang Chiu, Joshua~V. Dillon, Irfan Essa, Agrim Gupta, Meera
  Hahn, Anja Hauth, David Hendon, Alonso Martinez, David Minnen, David~A. Ross,
  Grant Schindler, Mikhail Sirotenko, Kihyuk Sohn, Krishna Somandepalli,
  Huisheng Wang, Jimmy Yan, Ming{-}Hsuan Yang, Xuan Yang, Bryan Seybold, and
  Lu~Jiang.
\newblock Videopoet: {A} large language model for zero-shot video generation.
\newblock \emph{CoRR}, abs/2312.14125, 2023.
\newblock \doi{10.48550/ARXIV.2312.14125}.
\newblock URL \url{https://doi.org/10.48550/arXiv.2312.14125}.

\bibitem[Kulhanek and Sattler(2024)]{kulhanek2024nerfbaselines}
Jonas Kulhanek and Torsten Sattler.
\newblock {NeRFBaselines}: {C}onsistent and reproducible evaluation of novel
  view synthesis methods.
\newblock \emph{arXiv preprint arXiv:2406.17345}, 2024.

\bibitem[Leroy et~al.(2024)Leroy, Cabon, and Revaud]{mast3r_eccv24}
Vincent Leroy, Yohann Cabon, and Jerome Revaud.
\newblock Grounding image matching in 3d with mast3r, 2024.

\bibitem[Li et~al.(2023{\natexlab{a}})Li, Tan, Zhang, Xu, Luan, Xu, Hong,
  Sunkavalli, Shakhnarovich, and Bi]{li2023instant3d}
Jiahao Li, Hao Tan, Kai Zhang, Zexiang Xu, Fujun Luan, Yinghao Xu, Yicong Hong,
  Kalyan Sunkavalli, Greg Shakhnarovich, and Sai Bi.
\newblock Instant3d: Fast text-to-3d with sparse-view generation and large
  reconstruction model, 2023{\natexlab{a}}.

\bibitem[Li et~al.(2024{\natexlab{a}})Li, Tan, Zhang, Xu, Luan, Xu, Hong,
  Sunkavalli, Shakhnarovich, and Bi]{li2024instant3d}
Jiahao Li, Hao Tan, Kai Zhang, Zexiang Xu, Fujun Luan, Yinghao Xu, Yicong Hong,
  Kalyan Sunkavalli, Greg Shakhnarovich, and Sai Bi.
\newblock Instant3d: Fast text-to-3d with sparse-view generation and large
  reconstruction model.
\newblock In \emph{International Conference on Learning Representations
  (ICLR)}, 2024{\natexlab{a}}.
\newblock URL \url{https://iclr.cc/virtual/2024/poster/19538}.

\bibitem[Li et~al.(2026)Li, Wang, Irshad, Vasiljevic, Walter, Guizilini, and
  Shakhnarovich]{li2026fastmap}
Jiahao Li, Haochen Wang, Muhammad~Zubair Irshad, Igor Vasiljevic, Matthew~R.
  Walter, Vitor~Campagnolo Guizilini, and Greg Shakhnarovich.
\newblock Fastmap: Revisiting structure from motion through first-order
  optimization.
\newblock In \emph{3DV}, 2026.
\newblock URL \url{https://jiahao.ai/fastmap/}.
\newblock Oral; arXiv:2505.04612.

\bibitem[Li et~al.(2025)Li, Lee, Fan, Yadav, Peng, and Chellappa]{11125757}
Jingxing Li, Yongjae Lee, Deliang Fan, Abhay~Kumar Yadav, Cheng Peng, and Rama
  Chellappa.
\newblock Speedy mast3r.
\newblock In \emph{2025 IEEE International Conference on Omni-layer Intelligent
  Systems (COINS)}, pages 1--6, 2025.
\newblock \doi{10.1109/COINS65080.2025.11125757}.

\bibitem[Li et~al.(2023{\natexlab{b}})Li, Li, Savarese, and Hoi]{li2023blip}
Junnan Li, Dongxu Li, Silvio Savarese, and Steven Hoi.
\newblock Blip-2: Bootstrapping language-image pre-training with frozen image
  encoders and large language models.
\newblock In \emph{International Conference on Machine Learning},
  2023{\natexlab{b}}.

\bibitem[Li et~al.(2023{\natexlab{c}})Li, Duan, Zhou, and
  Lu]{li2023diffusionsdf}
Muheng Li, Yueqi Duan, Jie Zhou, and Jiwen Lu.
\newblock Diffusion-sdf: Text-to-shape via voxelized diffusion.
\newblock In \emph{Proceedings of the IEEE/CVF Conference on Computer Vision
  and Pattern Recognition (CVPR)}, pages 12642--12651, June 2023{\natexlab{c}}.
\newblock URL
  \url{https://openaccess.thecvf.com/content/CVPR2023/html/Li_Diffusion-SDF_Text-To-Shape_via_Voxelized_Diffusion_CVPR_2023_paper.html}.

\bibitem[Li et~al.(2024{\natexlab{b}})Li, Wang, Chen, Peng, Wu, Chang, and
  Li]{li2024icontrol3d}
Wei Li, Shan Wang, Wenchao Chen, Xiaotong Peng, Tianyi Wu, Chin-Yen Chang, and
  Hantao Li.
\newblock icontrol3d: An interactive system for controllable 3d scene
  generation.
\newblock \emph{CoRR}, abs/2408.01678, 2024{\natexlab{b}}.
\newblock URL \url{https://arxiv.org/abs/2408.01678}.

\bibitem[Li et~al.(2023{\natexlab{d}})Li, Xu, Zhang, Yu, Sun, and
  Luo]{li2023remax}
Ziniu Li, Tian Xu, Yushun Zhang, Yang Yu, Ruoyu Sun, and Zhi-Quan Luo.
\newblock Remax: A simple, effective, and efficient reinforcement learning
  method for aligning large language models, 2023{\natexlab{d}}.

\bibitem[Lin et~al.(2024)Lin, Zhang, Ramanan, and Tulsiani]{lin2024relposepp}
Amy Lin, Jason~Y Zhang, Deva Ramanan, and Shubham Tulsiani.
\newblock {RelPose++}: Recovering 6d poses from sparse-view observations.
\newblock In \emph{International Conference on 3D Vision (3DV)}, 2024.

\bibitem[Lin et~al.(2023{\natexlab{a}})Lin, Gao, Tang, Takikawa, Zeng, Huang,
  Kreis, Fidler, Liu, and Lin]{Lin_2023_CVPR}
Chen-Hsuan Lin, Jun Gao, Luming Tang, Towaki Takikawa, Xiaohui Zeng, Xun Huang,
  Karsten Kreis, Sanja Fidler, Ming-Yu Liu, and Tsung-Yi Lin.
\newblock Magic3d: High-resolution text-to-3d content creation.
\newblock In \emph{Proceedings of the IEEE/CVF Conference on Computer Vision
  and Pattern Recognition (CVPR)}, pages 300--309, June 2023{\natexlab{a}}.

\bibitem[Lin et~al.(2023{\natexlab{b}})Lin, Gao, Tang, Takikawa, Zeng, Huang,
  Kreis, Fidler, Liu, and Lin]{lin2023magic3d}
Chen-Hsuan Lin, Jun Gao, Luming Tang, Towaki Takikawa, Xiaohui Zeng, Xun Huang,
  Karsten Kreis, Sanja Fidler, Ming-Yu Liu, and Tsung-Yi Lin.
\newblock Magic3d: High-resolution text-to-3d content creation.
\newblock In \emph{Proceedings of the IEEE/CVF Conference on Computer Vision
  and Pattern Recognition}, pages 300--309, 2023{\natexlab{b}}.

\bibitem[Lin et~al.(2022)Lin, Liu, Hu, Yan, Xie, and Huang]{lin2022capturing}
Liqiang Lin, Yilin Liu, Yue Hu, Xingguang Yan, Ke~Xie, and Hui Huang.
\newblock Capturing, reconstructing, and simulating: the {UrbanScene3D}
  dataset.
\newblock In \emph{ECCV}, 2022.

\bibitem[Lindenberger et~al.(2021)Lindenberger, Sarlin, Larsson, and
  Pollefeys]{lindenberger2021pixel}
Philipp Lindenberger, Paul-Edouard Sarlin, Viktor Larsson, and Marc Pollefeys.
\newblock Pixel-perfect structure-from-motion with featuremetric refinement.
\newblock In \emph{ICCV}, 2021.

\bibitem[Lindenberger et~al.(2023)Lindenberger, Sarlin, and
  Pollefeys]{Lindenberger_2023_ICCV}
Philipp Lindenberger, Paul-Edouard Sarlin, and Marc Pollefeys.
\newblock Lightglue: Local feature matching at light speed.
\newblock In \emph{Proceedings of the IEEE/CVF International Conference on
  Computer Vision (ICCV)}, pages 17627--17638, October 2023.

\bibitem[Liu et~al.(2025)Liu, Sun, Wang, Wang, Sun, Ye, Zhang, and
  Duan]{liu2025reconxreconstructscenesparse}
Fangfu Liu, Wenqiang Sun, Hanyang Wang, Yikai Wang, Haowen Sun, Junliang Ye,
  Jun Zhang, and Yueqi Duan.
\newblock Reconx: Reconstruct any scene from sparse views with video diffusion
  model, 2025.
\newblock URL \url{https://arxiv.org/abs/2408.16767}.

\bibitem[Liu et~al.(2024{\natexlab{a}})Liu, Shao, and
  Lu]{liu2024novelviewextrapolationvideo}
Kunhao Liu, Ling Shao, and Shijian Lu.
\newblock Novel view extrapolation with video diffusion priors,
  2024{\natexlab{a}}.
\newblock URL \url{https://arxiv.org/abs/2411.14208}.

\bibitem[Liu et~al.(2023{\natexlab{a}})Liu, Xu, Jin, Chen, T, Xu, and
  Su]{one2345}
Minghua Liu, Chao Xu, Haian Jin, Linghao Chen, Mukund~Varma T, Zexiang Xu, and
  Hao Su.
\newblock One-2-3-45: Any single image to 3d mesh in 45 seconds without
  per-shape optimization, 2023{\natexlab{a}}.

\bibitem[Liu et~al.(2024{\natexlab{b}})Liu, Xu, Jin, Chen, Varma~T, Xu, and
  Su]{liu2023one2345}
Minghua Liu, Chao Xu, Haian Jin, Linghao Chen, Mukund Varma~T, Zexiang Xu, and
  Hao Su.
\newblock One-2-3-45: Any single image to 3d mesh in 45 seconds without
  per-shape optimization.
\newblock \emph{Advances in Neural Information Processing Systems}, 36,
  2024{\natexlab{b}}.

\bibitem[Liu et~al.(2023{\natexlab{b}})Liu, Wu, Hoorick, Tokmakov, Zakharov,
  and Vondrick]{liu2023zero1to3}
Ruoshi Liu, Rundi Wu, Basile~Van Hoorick, Pavel Tokmakov, Sergey Zakharov, and
  Carl Vondrick.
\newblock Zero-1-to-3: Zero-shot one image to 3d object, 2023{\natexlab{b}}.

\bibitem[Liu et~al.(2023{\natexlab{c}})Liu, Wu, Van~Hoorick, Tokmakov,
  Zakharov, and Vondrick]{zero123}
Ruoshi Liu, Rundi Wu, Basile Van~Hoorick, Pavel Tokmakov, Sergey Zakharov, and
  Carl Vondrick.
\newblock Zero-1-to-3: Zero-shot one image to 3d object.
\newblock \emph{arXiv preprint arXiv:2303.11328}, 2023{\natexlab{c}}.

\bibitem[Liu and Yang(2023)]{Liu_2023_CVPR}
Xin Liu and Jufeng Yang.
\newblock Progressive neighbor consistency mining for correspondence pruning.
\newblock In \emph{Proceedings of the IEEE/CVF Conference on Computer Vision
  and Pattern Recognition (CVPR)}, pages 9527--9537, June 2023.

\bibitem[Liu et~al.(2023{\natexlab{d}})Liu, Lin, Zeng, Long, Liu, Komura, and
  Wang]{liu2023syncdreamer}
Yuan Liu, Cheng Lin, Zijiao Zeng, Xiaoxiao Long, Lingjie Liu, Taku Komura, and
  Wenping Wang.
\newblock Syncdreamer: Learning to generate multiview-consistent images from a
  single-view image.
\newblock \emph{arXiv preprint arXiv:2309.03453}, 2023{\natexlab{d}}.

\bibitem[Long et~al.(2022)Long, Lin, Wang, Komura, and
  Wang]{long2022sparseneus}
Xiaoxiao Long, Cheng Lin, Peng Wang, Taku Komura, and Wenping Wang.
\newblock Sparseneus: Fast generalizable neural surface reconstruction from
  sparse views.
\newblock In \emph{European Conference on Computer Vision}, pages 210--227.
  Springer, 2022.

\bibitem[Lorraine et~al.(2023)Lorraine, Xie, Zeng, Lin, Takikawa, Sharp, Lin,
  Liu, Fidler, and Lucas]{Lorraine_2023_ICCV}
Jonathan Lorraine, Kevin Xie, Xiaohui Zeng, Chen-Hsuan Lin, Towaki Takikawa,
  Nicholas Sharp, Tsung-Yi Lin, Ming-Yu Liu, Sanja Fidler, and James Lucas.
\newblock Att3d: Amortized text-to-3d object synthesis.
\newblock In \emph{Proceedings of the IEEE/CVF International Conference on
  Computer Vision (ICCV)}, pages 17946--17956, October 2023.

\bibitem[Lourakis and Argyros(2009)]{Lourakis_2009}
Manolis I.~A. Lourakis and Antonis~A. Argyros.
\newblock Sba: A software package for generic sparse bundle adjustment.
\newblock \emph{ACM Transactions on Mathematical Software}, 36\penalty0
  (1):\penalty0 1--30, March 2009.
\newblock \doi{10.1145/1486525.1486527}.
\newblock URL \url{http://dx.doi.org/10.1145/1486525.1486527}.

\bibitem[Luo and Hu(2021)]{luo2021diffusion}
Shitong Luo and Wei Hu.
\newblock Diffusion probabilistic models for 3d point cloud generation.
\newblock In \emph{Proceedings of the IEEE/CVF Conference on Computer Vision
  and Pattern Recognition}, pages 2837--2845, 2021.

\bibitem[Luo et~al.(2023{\natexlab{a}})Luo, Rockwell, Lee, and Johnson]{cap3d}
Tiange Luo, Chris Rockwell, Honglak Lee, and Justin Johnson.
\newblock Scalable 3d captioning with pretrained models.
\newblock \emph{arXiv preprint arXiv:2306.07279}, 2023{\natexlab{a}}.

\bibitem[Luo et~al.(2023{\natexlab{b}})Luo, Rockwell, Lee, and
  Johnson]{luo2023scalable}
Tiange Luo, Chris Rockwell, Honglak Lee, and Justin Johnson.
\newblock Scalable 3d captioning with pretrained models.
\newblock \emph{arXiv preprint arXiv:2306.07279}, 2023{\natexlab{b}}.

\bibitem[Luo et~al.(2020)Luo, Zhou, Bai, Chen, Zhang, Yao, Li, Fang, and
  Quan]{luo2020aslfeatlearninglocalfeatures}
Zixin Luo, Lei Zhou, Xuyang Bai, Hongkai Chen, Jiahui Zhang, Yao Yao, Shiwei
  Li, Tian Fang, and Long Quan.
\newblock Aslfeat: Learning local features of accurate shape and localization,
  2020.
\newblock URL \url{https://arxiv.org/abs/2003.10071}.

\bibitem[Malis and Vargas(2007)]{malis2007deeper}
Ezio Malis and Manuel Vargas.
\newblock \emph{Deeper understanding of the homography decomposition for
  vision-based control}.
\newblock PhD thesis, Inria, 2007.

\bibitem[Martinec and Pajdla(2007)]{martinec2007robust}
Daniel Martinec and Tomas Pajdla.
\newblock Robust rotation and translation estimation in multiview
  reconstruction.
\newblock In \emph{CVPR}, 2007.

\bibitem[Mehl et~al.(2023)Mehl, Schmalfuss, Jahedi, Nalivayko, and
  Bruhn]{dsetspring}
Lukas Mehl, Jenny Schmalfuss, Azin Jahedi, Yaroslava Nalivayko, and Andr{\'e}s
  Bruhn.
\newblock Spring: A high-resolution high-detail dataset and benchmark for scene
  flow, optical flow and stereo.
\newblock In \emph{Proceedings of the IEEE/CVF Conference on Computer Vision
  and Pattern Recognition}, pages 4981--4991, 2023.

\bibitem[Meng et~al.(2022)Meng, He, Song, Song, Wu, Zhu, and
  Ermon]{meng2022sdedit}
Chenlin Meng, Yutong He, Yang Song, Jiaming Song, Jiajun Wu, Jun-Yan Zhu, and
  Stefano Ermon.
\newblock {SDE}dit: Guided image synthesis and editing with stochastic
  differential equations.
\newblock In \emph{International Conference on Learning Representations}, 2022.

\bibitem[Mescheder et~al.(2019)Mescheder, Oechsle, Niemeyer, Nowozin, and
  Geiger]{Mescheder_2019_CVPR}
Lars Mescheder, Michael Oechsle, Michael Niemeyer, Sebastian Nowozin, and
  Andreas Geiger.
\newblock Occupancy networks: Learning 3d reconstruction in function space.
\newblock In \emph{The IEEE Conference on Computer Vision and Pattern
  Recognition (CVPR)}, June 2019.

\bibitem[Mildenhall et~al.(2020)Mildenhall, Srinivasan, Tancik, Barron,
  Ramamoorthi, and Ng]{mildenhall2020nerf}
Ben Mildenhall, Pratul~P. Srinivasan, Matthew Tancik, Jonathan~T. Barron, Ravi
  Ramamoorthi, and Ren Ng.
\newblock Nerf: Representing scenes as neural radiance fields for view
  synthesis.
\newblock In \emph{Computer Vision -- ECCV 2020 -- 16th European Conference,
  Glasgow, UK, August 23-28, 2020, Proceedings, Part I}, volume 12346 of
  \emph{Lecture Notes in Computer Science}, pages 405--421. Springer, 2020.
\newblock \doi{10.1007/978-3-030-58452-8_24}.
\newblock URL \url{https://doi.org/10.1007/978-3-030-58452-8_24}.

\bibitem[Mildenhall et~al.(2021)Mildenhall, Srinivasan, Tancik, Barron,
  Ramamoorthi, and Ng]{mildenhall2021nerf}
Ben Mildenhall, Pratul~P Srinivasan, Matthew Tancik, Jonathan~T Barron, Ravi
  Ramamoorthi, and Ren Ng.
\newblock {NeRF}: {R}epresenting scenes as neural radiance fields for view
  synthesis.
\newblock \emph{Communications of the ACM}, 65\penalty0 (1), 2021.

\bibitem[Mishchuk et~al.(2017)Mishchuk, Mishkin, Radenovic, and Matas]{HardNet}
Anastasiya Mishchuk, Dmytro Mishkin, Filip Radenovic, and Jiri Matas.
\newblock Working hard to know your neighbor's margins: Local descriptor
  learning loss.
\newblock In \emph{Proceedings of NeurIPS}, December 2017.

\bibitem[Mishra et~al.(2022)Mishra, Khashabi, Baral, and
  Hajishirzi]{mishra2022cross}
Swaroop Mishra, Daniel Khashabi, Chitta Baral, and Hannaneh Hajishirzi.
\newblock Cross-task generalization via natural language crowdsourcing
  instructions.
\newblock In \emph{Proceedings of the 60th Annual Meeting of the Association
  for Computational Linguistics (Volume 1: Long Papers)}, pages 3470--3487,
  2022.

\bibitem[Mnih et~al.(2016)Mnih, Badia, Mirza, Graves, Lillicrap, Harley,
  Silver, and Kavukcuoglu]{mnih2016a3c}
Volodymyr Mnih, Adrià~Puigdomènech Badia, Mehdi Mirza, Alex Graves,
  Timothy~P. Lillicrap, Tim Harley, David Silver, and Koray Kavukcuoglu.
\newblock Asynchronous methods for deep reinforcement learning, 2016.

\bibitem[Moulon et~al.(2013)Moulon, Monasse, and Marlet]{Moulon_2013}
Pierre Moulon, Pascal Monasse, and Renaud Marlet.
\newblock Global fusion of relative motions for robust, accurate and scalable
  structure from motion.
\newblock In \emph{2013 IEEE International Conference on Computer Vision},
  pages 3248--3255, December 2013.
\newblock \doi{10.1109/ICCV.2013.403}.
\newblock URL \url{http://dx.doi.org/10.1109/ICCV.2013.403}.

\bibitem[Moulon et~al.(2017)Moulon, Monasse, Perrot, and Marlet]{Moulon_2017}
Pierre Moulon, Pascal Monasse, Romuald Perrot, and Renaud Marlet.
\newblock Openmvg: Open multiple view geometry.
\newblock In \emph{Reproducible Research in Pattern Recognition}, pages 60--74,
  2017.
\newblock \doi{10.1007/978-3-319-56414-2_5}.
\newblock URL \url{http://dx.doi.org/10.1007/978-3-319-56414-2_5}.

\bibitem[M\"uller et~al.(2024)M\"uller, Schwarz, R\"ossle, Porzi, Bul\`o,
  Nie{\ss}ner, and Kontschieder]{Muller_2024_CVPR}
Norman M\"uller, Katja Schwarz, Barbara R\"ossle, Lorenzo Porzi, Samuel~Rota
  Bul\`o, Matthias Nie{\ss}ner, and Peter Kontschieder.
\newblock Multidiff: Consistent novel view synthesis from a single image.
\newblock In \emph{Proceedings of the IEEE/CVF Conference on Computer Vision
  and Pattern Recognition (CVPR)}, pages 10258--10268, June 2024.

\bibitem[M\"uller et~al.(2022)M\"uller, Evans, Schied, and
  Keller]{mueller2022instantngp}
Thomas M\"uller, Alex Evans, Christoph Schied, and Alexander Keller.
\newblock Instant neural graphics primitives with a multiresolution hash
  encoding.
\newblock \emph{ACM Trans. Graph.}, 41\penalty0 (4):\penalty0 102:1--102:15,
  July 2022.
\newblock \doi{10.1145/3528223.3530127}.
\newblock URL \url{https://doi.org/10.1145/3528223.3530127}.

\bibitem[M{\"u}ller et~al.(2022)M{\"u}ller, Evans, Schied, and
  Keller]{muller2022instant}
Thomas M{\"u}ller, Alex Evans, Christoph Schied, and Alexander Keller.
\newblock Instant neural graphics primitives with a multiresolution hash
  encoding.
\newblock \emph{ACM Transactions on Graphics (TOG)}, 41\penalty0 (4), 2022.

\bibitem[Ni et~al.(2025)Ni, Huang, Zhu, Chen, Wu, and Zhang]{Ni_2025_ICCV}
Jinkun Ni, Yuanjiang Huang, Zhongxiang Zhu, Zhaoyi Chen, Zeliang Wu, and
  Jionghao Zhang.
\newblock Wonderturbo: Generating interactive 3d world in 0.72 seconds.
\newblock In \emph{Proceedings of the IEEE/CVF International Conference on
  Computer Vision (ICCV)}, pages 13584--13594, 2025.

\bibitem[Ni et~al.(2007)Ni, Steedly, and Dellaert]{Ni_2007}
Kai Ni, Drew Steedly, and Frank Dellaert.
\newblock Out-of-core bundle adjustment for large-scale 3d reconstruction.
\newblock In \emph{2007 IEEE 11th International Conference on Computer Vision},
  pages 1--8, 2007.
\newblock \doi{10.1109/ICCV.2007.4409085}.
\newblock URL \url{http://dx.doi.org/10.1109/ICCV.2007.4409085}.

\bibitem[Nichol et~al.(2022{\natexlab{a}})Nichol, Jun, Dhariwal, Mishkin, and
  Chen]{nichol2022point}
Alex Nichol, Heewoo Jun, Prafulla Dhariwal, Pamela Mishkin, and Mark Chen.
\newblock Point-e: A system for generating 3d point clouds from complex
  prompts.
\newblock \emph{arXiv preprint arXiv:2212.08751}, 2022{\natexlab{a}}.

\bibitem[Nichol et~al.(2022{\natexlab{b}})Nichol, Jun, Dhariwal, Mishkin, and
  Chen]{pointe}
Alex Nichol, Heewoo Jun, Prafulla Dhariwal, Pamela Mishkin, and Mark Chen.
\newblock Point-e: A system for generating 3d point clouds from complex
  prompts.
\newblock \emph{arXiv preprint arXiv:2212.08751}, 2022{\natexlab{b}}.

\bibitem[Noh et~al.(2017)Noh, Araujo, Sim, Weyand, and Han]{Noh_2017_ICCV}
Hyeonwoo Noh, Andre Araujo, Jack Sim, Tobias Weyand, and Bohyung Han.
\newblock Large-scale image retrieval with attentive deep local features.
\newblock In \emph{Proceedings of the IEEE International Conference on Computer
  Vision (ICCV)}, Oct 2017.

\bibitem[Olsson and Enqvist(2011)]{Olsson_2011}
Carl Olsson and Olof Enqvist.
\newblock Stable structure from motion for unordered image collections.
\newblock In \emph{Image Analysis}, pages 524--535, 2011.
\newblock \doi{10.1007/978-3-642-21227-7_49}.
\newblock URL \url{http://dx.doi.org/10.1007/978-3-642-21227-7_49}.

\bibitem[Ouyang et~al.(2023)Ouyang, Sun, Lombardi, and Heal]{ouyang2023text}
Hao Ouyang, Tiancheng Sun, Stephen Lombardi, and Kathryn Heal.
\newblock Text2immersion: Generative immersive scene with 3d gaussians.
\newblock \emph{Arxiv}, 2023.

\bibitem[Ouyang et~al.(2022{\natexlab{a}})Ouyang, Wu, Jiang, Almeida,
  Wainwright, Mishkin, Zhang, Agarwal, Slama, Ray, Schulman, Hilton, Kelton,
  Miller, Simens, Askell, Welinder, Christiano, Leike, and
  Lowe]{ouyang2022InstructGPT}
Long Ouyang, Jeff Wu, Xu~Jiang, Diogo Almeida, Carroll~L. Wainwright, Pamela
  Mishkin, Chong Zhang, Sandhini Agarwal, Katarina Slama, Alex Ray, John
  Schulman, Jacob Hilton, Fraser Kelton, Luke Miller, Maddie Simens, Amanda
  Askell, Peter Welinder, Paul Christiano, Jan Leike, and Ryan Lowe.
\newblock Training language models to follow instructions with human feedback,
  2022{\natexlab{a}}.

\bibitem[Ouyang et~al.(2022{\natexlab{b}})Ouyang, Wu, Jiang, Almeida,
  Wainwright, Mishkin, Zhang, Agarwal, Slama, Ray, et~al.]{ouyang2022training}
Long Ouyang, Jeffrey Wu, Xu~Jiang, Diogo Almeida, Carroll Wainwright, Pamela
  Mishkin, Chong Zhang, Sandhini Agarwal, Katarina Slama, Alex Ray, et~al.
\newblock Training language models to follow instructions with human feedback.
\newblock \emph{Advances in Neural Information Processing Systems},
  35:\penalty0 27730--27744, 2022{\natexlab{b}}.

\bibitem[Ozyesil and Singer(2015)]{ozyesil2015robust}
Onur Ozyesil and Amit Singer.
\newblock Robust camera location estimation by convex programming.
\newblock In \emph{CVPR}, 2015.

\bibitem[Pan et~al.(2024)Pan, Barath, Pollefeys, and
  Sch\"{o}nberger]{pan2024glomap}
Linfei Pan, Daniel Barath, Marc Pollefeys, and Johannes~Lutz Sch\"{o}nberger.
\newblock Global structure-from-motion revisited.
\newblock In \emph{European Conference on Computer Vision (ECCV)}, 2024.

\bibitem[Park et~al.(2019)Park, Florence, Straub, Newcombe, and
  Lovegrove]{Park_2019_CVPR}
Jeong~Joon Park, Peter Florence, Julian Straub, Richard Newcombe, and Steven
  Lovegrove.
\newblock Deepsdf: Learning continuous signed distance functions for shape
  representation.
\newblock In \emph{The IEEE Conference on Computer Vision and Pattern
  Recognition (CVPR)}, June 2019.

\bibitem[Park et~al.(2023)Park, Henzler, Mildenhall, Barron, and
  Martin-Brualla]{park2023camp}
Keunhong Park, Philipp Henzler, Ben Mildenhall, Jonathan~T. Barron, and Ricardo
  Martin-Brualla.
\newblock {CamP}: {C}amera preconditioning for neural radiance fields.
\newblock \emph{ACM Transactions on Graphics}, 2023.

\bibitem[Paszke et~al.(2019)Paszke, Gross, Massa, Lerer, Bradbury, Chanan,
  Killeen, Lin, Gimelshein, Antiga, et~al.]{paszke2019pytorch}
Adam Paszke, Sam Gross, Francisco Massa, Adam Lerer, James Bradbury, Gregory
  Chanan, Trevor Killeen, Zeming Lin, Natalia Gimelshein, Luca Antiga, et~al.
\newblock Pytorch: An imperative style, high-performance deep learning library.
\newblock \emph{Advances in neural information processing systems}, 32, 2019.

\bibitem[Pautrat et~al.(2021)Pautrat, Lin, Larsson, Oswald, and
  Pollefeys]{pautrat2021sold2selfsupervisedocclusionawareline}
Rémi Pautrat, Juan-Ting Lin, Viktor Larsson, Martin~R. Oswald, and Marc
  Pollefeys.
\newblock Sold2: Self-supervised occlusion-aware line description and
  detection, 2021.
\newblock URL \url{https://arxiv.org/abs/2104.03362}.

\bibitem[Pautrat et~al.(2023{\natexlab{a}})Pautrat, Barath, Larsson, Oswald,
  and Pollefeys]{pautrat2023deeplsdlinesegmentdetection}
Rémi Pautrat, Daniel Barath, Viktor Larsson, Martin~R. Oswald, and Marc
  Pollefeys.
\newblock Deeplsd: Line segment detection and refinement with deep image
  gradients, 2023{\natexlab{a}}.
\newblock URL \url{https://arxiv.org/abs/2212.07766}.

\bibitem[Pautrat et~al.(2023{\natexlab{b}})Pautrat, Suárez, Yu, Pollefeys, and
  Larsson]{pautrat2023gluestickrobustimagematching}
Rémi Pautrat, Iago Suárez, Yifan Yu, Marc Pollefeys, and Viktor Larsson.
\newblock Gluestick: Robust image matching by sticking points and lines
  together, 2023{\natexlab{b}}.
\newblock URL \url{https://arxiv.org/abs/2304.02008}.

\bibitem[Peebles and Xie(2022)]{peebles2022dit}
William Peebles and Saining Xie.
\newblock Scalable diffusion models with transformers.
\newblock \emph{arXiv preprint arXiv:2212.09748}, 2022.

\bibitem[Pilkington(2022)]{Pilkington2022}
Nicholas Pilkington.
\newblock {DroneDeploy} {NeRF} dataset.
\newblock \url{https://github.com/nickponline/dd-nerf-dataset}, May 2022.

\bibitem[Podell et~al.(2023)Podell, English, Lacey, Blattmann, Dockhorn,
  M{\"u}ller, Penna, and Rombach]{podell2023sdxl}
Dustin Podell, Zion English, Kyle Lacey, Andreas Blattmann, Tim Dockhorn, Jonas
  M{\"u}ller, Joe Penna, and Robin Rombach.
\newblock Sdxl: improving latent diffusion models for high-resolution image
  synthesis.
\newblock \emph{arXiv preprint arXiv:2307.01952}, 2023.

\bibitem[Poole et~al.(2022)Poole, Jain, Barron, and
  Mildenhall]{poole2022dreamfusion}
Ben Poole, Ajay Jain, Jonathan~T. Barron, and Ben Mildenhall.
\newblock Dreamfusion: Text-to-3d using 2d diffusion.
\newblock \emph{arXiv}, 2022.

\bibitem[Poursaeed et~al.(2018)Poursaeed, Yang, Prakash, Fang, Jiang,
  Hariharan, and Belongie]{poursaeed2018deepfundamentalmatrixestimation}
Omid Poursaeed, Guandao Yang, Aditya Prakash, Qiuren Fang, Hanqing Jiang,
  Bharath Hariharan, and Serge Belongie.
\newblock Deep fundamental matrix estimation without correspondences, 2018.
\newblock URL \url{https://arxiv.org/abs/1810.01575}.

\bibitem[Qian et~al.(2024)Qian, Mai, Hamdi, Ren, Siarohin, Li, Lee,
  Skorokhodov, Wonka, Tulyakov, and Ghanem]{Magic123}
Guocheng Qian, Jinjie Mai, Abdullah Hamdi, Jian Ren, Aliaksandr Siarohin, Bing
  Li, Hsin-Ying Lee, Ivan Skorokhodov, Peter Wonka, Sergey Tulyakov, and
  Bernard Ghanem.
\newblock Magic123: One image to high-quality 3d object generation using both
  2d and 3d diffusion priors.
\newblock In \emph{The Twelfth International Conference on Learning
  Representations (ICLR)}, 2024.
\newblock URL \url{https://openreview.net/forum?id=0jHkUDyEO9}.

\bibitem[Radenović et~al.(2018)Radenović, Tolias, and
  Chum]{radenović2018finetuningcnnimageretrieval}
Filip Radenović, Giorgos Tolias, and Ondřej Chum.
\newblock Fine-tuning cnn image retrieval with no human annotation, 2018.
\newblock URL \url{https://arxiv.org/abs/1711.02512}.

\bibitem[Radford et~al.(2021)Radford, Kim, Hallacy, Ramesh, Goh, Agarwal,
  Sastry, Askell, Mishkin, Clark, et~al.]{radford2021learning}
Alec Radford, Jong~Wook Kim, Chris Hallacy, Aditya Ramesh, Gabriel Goh,
  Sandhini Agarwal, Girish Sastry, Amanda Askell, Pamela Mishkin, Jack Clark,
  et~al.
\newblock Learning transferable visual models from natural language
  supervision.
\newblock In \emph{International conference on machine learning}, pages
  8748--8763. PMLR, 2021.

\bibitem[Raffel et~al.(2020)Raffel, Shazeer, Roberts, Lee, Narang, Matena,
  Zhou, Li, and Liu]{raffel2020exploring}
Colin Raffel, Noam Shazeer, Adam Roberts, Katherine Lee, Sharan Narang, Michael
  Matena, Yanqi Zhou, Wei Li, and Peter~J Liu.
\newblock Exploring the limits of transfer learning with a unified text-to-text
  transformer.
\newblock \emph{The Journal of Machine Learning Research}, 21\penalty0
  (1):\penalty0 5485--5551, 2020.

\bibitem[Raj et~al.(2023)Raj, Chu, Pan, Ma, Bansal, Chen, Pirk, Ramanan, and
  Ranjan]{Raj_2023_ICCV}
Amit Raj, Hongyi Chu, Yinrui Pan, Chih-Yao Ma, Aayush Bansal, Judy Chen, Soren
  Pirk, Deva Ramanan, and Anurag Ranjan.
\newblock Dreambooth3d: Subject-driven text-to-3d generation.
\newblock In \emph{Proceedings of the IEEE/CVF International Conference on
  Computer Vision (ICCV)}, pages 3004--3014, October 2023.

\bibitem[Ramesh et~al.(2022)Ramesh, Dhariwal, Nichol, Chu, and
  Chen]{ramesh2022hierarchical}
Aditya Ramesh, Prafulla Dhariwal, Alex Nichol, Casey Chu, and Mark Chen.
\newblock Hierarchical text-conditional image generation with clip latents.
\newblock \emph{arXiv preprint arXiv:2204.06125}, 1\penalty0 (2):\penalty0 3,
  2022.

\bibitem[Ranftl et~al.(2020)Ranftl, Lasinger, Hafner, Schindler, and
  Koltun]{ranftl2020towards}
Ren{\'e} Ranftl, Katrin Lasinger, David Hafner, Konrad Schindler, and Vladlen
  Koltun.
\newblock Towards robust monocular depth estimation: Mixing datasets for
  zero-shot cross-dataset transfer.
\newblock \emph{IEEE transactions on pattern analysis and machine
  intelligence}, 44\penalty0 (3):\penalty0 1623--1637, 2020.

\bibitem[Ranftl and Koltun(2018)]{Ranftl_2018}
René Ranftl and Vladlen Koltun.
\newblock \emph{Deep Fundamental Matrix Estimation}, page 292–309.
\newblock Springer International Publishing, 2018.
\newblock ISBN 9783030012465.
\newblock \doi{10.1007/978-3-030-01246-5_18}.
\newblock URL \url{http://dx.doi.org/10.1007/978-3-030-01246-5_18}.

\bibitem[Reizenstein et~al.(2021)Reizenstein, Shapovalov, Henzler, Sbordone,
  Labatut, and Novotny]{reizenstein2021common}
Jeremy Reizenstein, Roman Shapovalov, Philipp Henzler, Luca Sbordone, Patrick
  Labatut, and David Novotny.
\newblock Common objects in 3d: Large-scale learning and evaluation of
  real-life 3d category reconstruction.
\newblock In \emph{Proceedings of the IEEE/CVF International Conference on
  Computer Vision}, pages 10901--10911, 2021.

\bibitem[Ren et~al.(2025)Ren, Wang, Tan, and Han]{ren2025fin3r}
Weining Ren, Hongjun Wang, Xiao Tan, and Kai Han.
\newblock Fin3r: Fine-tuning feed-forward 3d reconstruction models via
  monocular knowledge distillation.
\newblock In \emph{The Thirty-ninth Annual Conference on Neural Information
  Processing Systems}, 2025.

\bibitem[Ren and Wang(2022)]{Ren_2022_CVPR}
Xuanchi Ren and Xiaolong Wang.
\newblock Look outside the room: Synthesizing a consistent long-term 3d scene
  video from a single image.
\newblock In \emph{Proceedings of the IEEE/CVF Conference on Computer Vision
  and Pattern Recognition (CVPR)}, pages 3563--3573, June 2022.

\bibitem[Revaud et~al.(2019)Revaud, Weinzaepfel, Souza, Pion, Csurka, Cabon,
  and Humenberger]{revaud2019r2d2repeatablereliabledetector}
Jerome Revaud, Philippe Weinzaepfel, César~De Souza, Noe Pion, Gabriela
  Csurka, Yohann Cabon, and Martin Humenberger.
\newblock R2d2: Repeatable and reliable detector and descriptor, 2019.
\newblock URL \url{https://arxiv.org/abs/1906.06195}.

\bibitem[Rocco et~al.(2018)Rocco, Cimpoi, Arandjelović, Torii, Pajdla, and
  Sivic]{rocco2018neighbourhoodconsensusnetworks}
Ignacio Rocco, Mircea Cimpoi, Relja Arandjelović, Akihiko Torii, Tomas Pajdla,
  and Josef Sivic.
\newblock Neighbourhood consensus networks, 2018.
\newblock URL \url{https://arxiv.org/abs/1810.10510}.

\bibitem[Rocco et~al.(2020)Rocco, Arandjelović, and
  Sivic]{rocco2020efficientneighbourhoodconsensusnetworks}
Ignacio Rocco, Relja Arandjelović, and Josef Sivic.
\newblock Efficient neighbourhood consensus networks via submanifold sparse
  convolutions, 2020.
\newblock URL \url{https://arxiv.org/abs/2004.10566}.

\bibitem[Rodriguez et~al.(2011)Rodriguez, L{\'o}pez-de Teruel, and
  Ruiz]{rodriguez2011gea}
Antonio~L Rodriguez, Pedro~E L{\'o}pez-de Teruel, and Alberto Ruiz.
\newblock {GEA} optimization for live structureless motion estimation.
\newblock In \emph{Proceedings of the IEEE International Conference on Computer
  Vision Workshops (ICCV Workshops)}, 2011.

\bibitem[Rodr{\'\i}guez et~al.(2011)Rodr{\'\i}guez, L{\'o}pez-de Teruel, and
  Ruiz]{rodriguez2011reduced}
Antonio~L Rodr{\'\i}guez, Pedro~E L{\'o}pez-de Teruel, and Alberto Ruiz.
\newblock Reduced epipolar cost for accelerated incremental sfm.
\newblock In \emph{CVPR}, 2011.

\bibitem[Rombach et~al.(2021)Rombach, Blattmann, Lorenz, Esser, and
  Ommer]{rombach2021highresolution}
Robin Rombach, Andreas Blattmann, Dominik Lorenz, Patrick Esser, and Björn
  Ommer.
\newblock High-resolution image synthesis with latent diffusion models, 2021.

\bibitem[Rombach et~al.(2022)Rombach, Blattmann, Lorenz, Esser, and Ommer]{ldm}
Robin Rombach, Andreas Blattmann, Dominik Lorenz, Patrick Esser, and Bj\"orn
  Ommer.
\newblock High-resolution image synthesis with latent diffusion models.
\newblock In \emph{CVPR}, 2022.

\bibitem[Rudnev et~al.(2022)Rudnev, Elgharib, Smith, Liu, Golyanik, and
  Theobalt]{rudnev2022nerf}
Viktor Rudnev, Mohamed Elgharib, William Smith, Lingjie Liu, Vladislav
  Golyanik, and Christian Theobalt.
\newblock Nerf for outdoor scene relighting.
\newblock In \emph{ECCV}, 2022.

\bibitem[Saharia et~al.(2022)Saharia, Chan, Saxena, Li, Whang, Denton,
  Ghasemipour, Gontijo~Lopes, Karagol~Ayan, Salimans,
  et~al.]{saharia2022photorealistic}
Chitwan Saharia, William Chan, Saurabh Saxena, Lala Li, Jay Whang, Emily~L
  Denton, Kamyar Ghasemipour, Raphael Gontijo~Lopes, Burcu Karagol~Ayan, Tim
  Salimans, et~al.
\newblock Photorealistic text-to-image diffusion models with deep language
  understanding.
\newblock \emph{Advances in Neural Information Processing Systems},
  35:\penalty0 36479--36494, 2022.

\bibitem[Sarlin et~al.(2019)Sarlin, Cadena, Siegwart, and
  Dymczyk]{sarlin2019coarsefinerobusthierarchical}
Paul-Edouard Sarlin, Cesar Cadena, Roland Siegwart, and Marcin Dymczyk.
\newblock From coarse to fine: Robust hierarchical localization at large scale,
  2019.
\newblock URL \url{https://arxiv.org/abs/1812.03506}.

\bibitem[Sarlin et~al.(2020)Sarlin, DeTone, Malisiewicz, and
  Rabinovich]{sarlin2020supergluelearningfeaturematching}
Paul-Edouard Sarlin, Daniel DeTone, Tomasz Malisiewicz, and Andrew Rabinovich.
\newblock Superglue: Learning feature matching with graph neural networks,
  2020.
\newblock URL \url{https://arxiv.org/abs/1911.11763}.

\bibitem[Sarlin et~al.(2022)Sarlin, Dusmanu, Sch{\"o}nberger, Speciale, Gruber,
  Larsson, Miksik, and Pollefeys]{sarlin2022lamar}
Paul-Edouard Sarlin, Mihai Dusmanu, Johannes~L Sch{\"o}nberger, Pablo Speciale,
  Lukas Gruber, Viktor Larsson, Ondrej Miksik, and Marc Pollefeys.
\newblock {LaMAR}: {B}enchmarking localization and mapping for augmented
  reality.
\newblock In \emph{ECCV}, 2022.

\bibitem[Schonberger and Frahm(2016)]{schoenberger2016sfm}
Johannes~L. Schonberger and Jan-Michael Frahm.
\newblock Structure-from-motion revisited.
\newblock In \emph{2016 IEEE Conference on Computer Vision and Pattern
  Recognition (CVPR)}, pages 4104--4113, June 2016.
\newblock \doi{10.1109/CVPR.2016.445}.
\newblock URL \url{http://dx.doi.org/10.1109/CVPR.2016.445}.

\bibitem[Schops et~al.(2017)Schops, Schonberger, Galliani, Sattler, Schindler,
  Pollefeys, and Geiger]{schops2017multi}
Thomas Schops, Johannes~L Schonberger, Silvano Galliani, Torsten Sattler,
  Konrad Schindler, Marc Pollefeys, and Andreas Geiger.
\newblock A multi-view stereo benchmark with high-resolution images and
  multi-camera videos.
\newblock In \emph{CVPR}, 2017.

\bibitem[Schuhmann et~al.(2022{\natexlab{a}})Schuhmann, Beaumont, Vencu,
  Gordon, Wightman, Cherti, Coombes, Katta, Mullis, Wortsman, Schramowski,
  Kundurthy, Crowson, Schmidt, Kaczmarczyk, and Jitsev]{schuhmann2022laion5b}
Christoph Schuhmann, Romain Beaumont, Richard Vencu, Cade Gordon, Ross
  Wightman, Mehdi Cherti, Theo Coombes, Aarush Katta, Clayton Mullis, Mitchell
  Wortsman, Patrick Schramowski, Srivatsa Kundurthy, Katherine Crowson, Ludwig
  Schmidt, Robert Kaczmarczyk, and Jenia Jitsev.
\newblock Laion-5b: An open large-scale dataset for training next generation
  image-text models, 2022{\natexlab{a}}.

\bibitem[Schuhmann et~al.(2022{\natexlab{b}})Schuhmann, Beaumont, Vencu,
  Gordon, Wightman, Cherti, Coombes, Katta, Mullis, Wortsman,
  et~al.]{schuhmann2022laion}
Christoph Schuhmann, Romain Beaumont, Richard Vencu, Cade Gordon, Ross
  Wightman, Mehdi Cherti, Theo Coombes, Aarush Katta, Clayton Mullis, Mitchell
  Wortsman, et~al.
\newblock Laion-5b: An open large-scale dataset for training next generation
  image-text models.
\newblock \emph{Advances in Neural Information Processing Systems},
  35:\penalty0 25278--25294, 2022{\natexlab{b}}.

\bibitem[Schulman(2023)]{schulman2023rlhf}
John Schulman.
\newblock Rl and truthfulness: Towards truthgpt.
\newblock YouTube, 2023.
\newblock URL \url{https://www.youtube.com/watch?v=hhiLw5Q_UFg}.

\bibitem[Schulman et~al.(2017)Schulman, Wolski, Dhariwal, Radford, and
  Klimov]{schulman2017ppo}
John Schulman, Filip Wolski, Prafulla Dhariwal, Alec Radford, and Oleg Klimov.
\newblock Proximal policy optimization algorithms, 2017.

\bibitem[Schult et~al.(2024)Schult, Torralba, Herzig, Wei, Wang, Muller, Liao,
  Zhou, and Wu]{Schult_2024_CVPR}
Jonas Schult, Antonio Torralba, Roei Herzig, Sanja Wei, Yilei Wang, Thomas
  Muller, Yao Liao, Zhenda Zhou, and Yanfeng Wu.
\newblock Controlroom3d: Room generation using semantic proxy rooms.
\newblock In \emph{Proceedings of the IEEE/CVF Conference on Computer Vision
  and Pattern Recognition (CVPR)}, pages 22864--22874, June 2024.

\bibitem[Schwarz et~al.(2020)Schwarz, Liao, Niemeyer, and
  Geiger]{schwarz2020graf}
Katja Schwarz, Yiyi Liao, Michael Niemeyer, and Andreas Geiger.
\newblock Graf: Generative radiance fields for 3d-aware image synthesis.
\newblock In \emph{Advances in Neural Information Processing Systems}, 2020.

\bibitem[Shen et~al.(2023)Shen, Yan, Qi, Najibi, Deng, Guibas, Zhou, and
  Anguelov]{shen2023gina}
Bokui Shen, Xinchen Yan, Charles~R Qi, Mahyar Najibi, Boyang Deng, Leonidas
  Guibas, Yin Zhou, and Dragomir Anguelov.
\newblock Gina-3d: Learning to generate implicit neural assets in the wild.
\newblock In \emph{Proceedings of the IEEE/CVF conference on computer vision
  and pattern recognition}, pages 4913--4926, 2023.

\bibitem[Shi et~al.(2023)Shi, Wang, Ye, Mai, Li, and Yang]{shi2023mvdream}
Yichun Shi, Peng Wang, Jianglong Ye, Long Mai, Kejie Li, and Xiao Yang.
\newblock Mvdream: Multi-view diffusion for 3d generation.
\newblock \emph{arXiv:2308.16512}, 2023.

\bibitem[{Shi et al.}(2023)]{shi2023zero123plus}
Ruoxi {Shi et al.}
\newblock Zero123++: a single image to consistent multi-view diffusion base
  model, 2023.

\bibitem[Shriram et~al.(2025)Shriram, Trevithick, Liu, and
  Ramamoorthi]{shriram2024realmdreamer}
Jaidev Shriram, Alex Trevithick, Lingjie Liu, and Ravi Ramamoorthi.
\newblock Realmdreamer: Text-driven 3d scene generation with inpainting and
  depth diffusion.
\newblock In \emph{International Conference on 3D Vision (3DV)}, 2025.

\bibitem[Shu et~al.(2019)Shu, Park, and Kwon]{Shu_2019_ICCV}
Zhijing Shu, Su-jin Park, and Junseok Kwon.
\newblock 3d point cloud generative adversarial network based on tree
  structured graph convolutions.
\newblock In \emph{The IEEE International Conference on Computer Vision
  (ICCV)}, October 2019.

\bibitem[Shue et~al.(2023)Shue, Chan, Zheng, Gu, and
  Su]{Shue2023triplanediffusion}
Zaiwei Shue, Ching-Yao Chan, Wenzheng Zheng, Jiatao Gu, and Hao Su.
\newblock 3d neural field generation using triplane diffusion.
\newblock In \emph{Proceedings of the IEEE/CVF Conference on Computer Vision
  and Pattern Recognition (CVPR)}, 2023.

\bibitem[Sinha et~al.(2023)Sinha, Zhang, Tagliasacchi, Gilitschenski, and
  Lindell]{sinha2023sparsepose}
Samarth Sinha, Jason~Y Zhang, Andrea Tagliasacchi, Igor Gilitschenski, and
  David~B Lindell.
\newblock {SparsePose}: Sparse-view camera pose regression and refinement.
\newblock In \emph{Computer Vision and Pattern Recognition (CVPR)}, 2023.

\bibitem[Sitzmann et~al.(2021)Sitzmann, Rezchikov, Freeman, Tenenbaum, and
  Durand]{sitzmann2021light}
Vincent Sitzmann, Semon Rezchikov, Bill Freeman, Josh Tenenbaum, and Fredo
  Durand.
\newblock Light field networks: Neural scene representations with
  single-evaluation rendering.
\newblock \emph{Advances in Neural Information Processing Systems},
  34:\penalty0 19313--19325, 2021.

\bibitem[Snavely et~al.(2006)Snavely, Seitz, and Szeliski]{Snavely_2006}
Noah Snavely, Steven~M. Seitz, and Richard Szeliski.
\newblock Photo tourism: exploring photo collections in 3d.
\newblock \emph{ACM Transactions on Graphics}, 25\penalty0 (3):\penalty0
  835--846, July 2006.
\newblock \doi{10.1145/1141911.1141964}.
\newblock URL \url{http://dx.doi.org/10.1145/1141911.1141964}.

\bibitem[Snavely et~al.(2007)Snavely, Seitz, and Szeliski]{Snavely_2007}
Noah Snavely, Steven~M. Seitz, and Richard Szeliski.
\newblock Modeling the world from internet photo collections.
\newblock \emph{International Journal of Computer Vision}, 80\penalty0
  (2):\penalty0 189--210, December 2007.
\newblock \doi{10.1007/s11263-007-0107-3}.
\newblock URL \url{http://dx.doi.org/10.1007/s11263-007-0107-3}.

\bibitem[Snavely et~al.(2008)Snavely, Seitz, and Szeliski]{Snavely_2008}
Noah Snavely, Steven~M. Seitz, and Richard Szeliski.
\newblock Skeletal graphs for efficient structure from motion.
\newblock In \emph{2008 IEEE Conference on Computer Vision and Pattern
  Recognition}, pages 1--8, June 2008.
\newblock \doi{10.1109/CVPR.2008.4587678}.
\newblock URL \url{http://dx.doi.org/10.1109/CVPR.2008.4587678}.

\bibitem[Song et~al.(2020)Song, Meng, and Ermon]{ddim}
Jiaming Song, Chenlin Meng, and Stefano Ermon.
\newblock Denoising diffusion implicit models.
\newblock \emph{arXiv preprint arXiv:2010.02502}, 2020.

\bibitem[Song et~al.(2021)Song, Meng, and Ermon]{song2021denoising}
Jiaming Song, Chenlin Meng, and Stefano Ermon.
\newblock Denoising diffusion implicit models.
\newblock In \emph{International Conference on Learning Representations}, 2021.
\newblock URL \url{https://openreview.net/forum?id=St1giarCHLP}.

\bibitem[Song et~al.(2022)Song, Meng, and Ermon]{song2022ddim}
Jiaming Song, Chenlin Meng, and Stefano Ermon.
\newblock Denoising diffusion implicit models, 2022.

\bibitem[Sun et~al.(2021)Sun, Shen, Wang, Bao, and
  Zhou]{sun2021loftrdetectorfreelocalfeature}
Jiaming Sun, Zehong Shen, Yuang Wang, Hujun Bao, and Xiaowei Zhou.
\newblock Loftr: Detector-free local feature matching with transformers, 2021.
\newblock URL \url{https://arxiv.org/abs/2104.00680}.

\bibitem[Sun et~al.(2020)Sun, Trinh, and Lee]{Sun_2020_WACV}
Weiyue Sun, Khanh Trinh, and Daeheung Lee.
\newblock Pointgrow: Autoregressively learned point cloud generation with
  self-attention.
\newblock In \emph{The IEEE Winter Conference on Applications of Computer
  Vision (WACV)}, March 2020.

\bibitem[Sweeney et~al.(2015)Sweeney, Sattler, Hollerer, Turk, and
  Pollefeys]{sweeney2015optimizing}
Chris Sweeney, Torsten Sattler, Tobias Hollerer, Matthew Turk, and Marc
  Pollefeys.
\newblock Optimizing the viewing graph for structure-from-motion.
\newblock In \emph{ICCV}, 2015.

\bibitem[Szymanowicz et~al.(2024)Szymanowicz, Rupprecht, and
  Vedaldi]{Szymanowicz_2024_CVPR}
Stanislaw Szymanowicz, Christian Rupprecht, and Andrea Vedaldi.
\newblock Splatter image: Ultra-fast single-view 3d reconstruction.
\newblock In \emph{Proceedings of the IEEE/CVF Conference on Computer Vision
  and Pattern Recognition (CVPR)}, pages 10208--10217, June 2024.

\bibitem[Szymanowicz et~al.(2025)Szymanowicz, Zhang, Srinivasan, Gao, Brussee,
  Holynski, Martin-Brualla, Barron, and Henzler]{szymanowicz2025bolt3d}
Stanislaw Szymanowicz, Jason~Y. Zhang, Pratul Srinivasan, Ruiqi Gao, Arthur
  Brussee, Aleksander Holynski, Ricardo Martin-Brualla, Jonathan~T. Barron, and
  Philipp Henzler.
\newblock {Bolt3D: Generating 3D Scenes in Seconds}.
\newblock \emph{International Conference on Computer Vision}, 2025.

\bibitem[Tang and Tan(2019)]{tang2019banetdensebundleadjustment}
Chengzhou Tang and Ping Tan.
\newblock Ba-net: Dense bundle adjustment network, 2019.
\newblock URL \url{https://arxiv.org/abs/1806.04807}.

\bibitem[Tang et~al.(2024)Tang, Ren, and Zhou]{DBLP:conf/iclr/TangRZ0Z24}
Jiaxiang Tang, Jiawei Ren, and Hang Zhou.
\newblock Dreamgaussian: Generative gaussian splatting for efficient 3d content
  creation.
\newblock In \emph{The Twelfth International Conference on Learning
  Representations, {ICLR} 2024, Vienna, Austria, May 7-11, 2024}.
  OpenReview.net, 2024.
\newblock URL \url{https://openreview.net/forum?id=DBmRFGYXzW}.

\bibitem[Tang et~al.(2023)Tang, Wang, Zhang, Zhang, Yi, Ma, and
  Chen]{Tang_2023_ICCV}
Junshu Tang, Tengfei Wang, Bo~Zhang, Ting Zhang, Ran Yi, Lizhuang Ma, and Dong
  Chen.
\newblock Make-it-3d: High-fidelity 3d creation from a single image with
  diffusion prior.
\newblock In \emph{Proceedings of the IEEE/CVF International Conference on
  Computer Vision (ICCV)}, pages 22819--22829, October 2023.

\bibitem[Teed and Deng(2021)]{teed2021droid}
Zachary Teed and Jia Deng.
\newblock Droid-slam: Deep visual slam for monocular, stereo, and rgb-d
  cameras.
\newblock \emph{Advances in neural information processing systems},
  34:\penalty0 16558--16569, 2021.

\bibitem[Tirado-Garín and
  Civera(2025)]{tiradogarín2025anycalibonmanifoldlearningmodelagnostic}
Javier Tirado-Garín and Javier Civera.
\newblock Anycalib: On-manifold learning for model-agnostic single-view camera
  calibration, 2025.
\newblock URL \url{https://arxiv.org/abs/2503.12701}.

\bibitem[Trevithick and Yang(2021)]{trevithick2021grf}
Alex Trevithick and Bo~Yang.
\newblock Grf: Learning a general radiance field for 3d representation and
  rendering.
\newblock In \emph{Proceedings of the IEEE/CVF International Conference on
  Computer Vision}, pages 15182--15192, 2021.

\bibitem[Triggs et~al.(2000)Triggs, McLauchlan, Hartley, and
  Fitzgibbon]{triggs2000bundle}
Bill Triggs, Philip~F. McLauchlan, Richard~I. Hartley, and Andrew~W.
  Fitzgibbon.
\newblock Bundle adjustment --- a modern synthesis.
\newblock In \emph{Vision Algorithms: Theory and Practice}, pages 298--372,
  2000.
\newblock \doi{10.1007/3-540-44480-7_21}.
\newblock URL \url{http://dx.doi.org/10.1007/3-540-44480-7_21}.

\bibitem[Turki et~al.(2022)Turki, Ramanan, and Satyanarayanan]{turki2022mega}
Haithem Turki, Deva Ramanan, and Mahadev Satyanarayanan.
\newblock Mega-{NeRF}: {S}calable construction of large-scale {NeRFs} for
  virtual fly-throughs.
\newblock In \emph{CVPR}, 2022.

\bibitem[Tyszkiewicz et~al.(2020)Tyszkiewicz, Fua, and
  Trulls]{tyszkiewicz2020disklearninglocalfeatures}
Michał~J. Tyszkiewicz, Pascal Fua, and Eduard Trulls.
\newblock Disk: Learning local features with policy gradient, 2020.
\newblock URL \url{https://arxiv.org/abs/2006.13566}.

\bibitem[Van~Hoorick et~al.(2024)Van~Hoorick, Wu, Ozguroglu, Sargent, Liu,
  Tokmakov, Dave, Zheng, and Vondrick]{vanhoorick2024gcd}
Basile Van~Hoorick, Rundi Wu, Ege Ozguroglu, Kyle Sargent, Ruoshi Liu, Pavel
  Tokmakov, Achal Dave, Changxi Zheng, and Carl Vondrick.
\newblock Generative camera dolly: Extreme monocular dynamic novel view
  synthesis.
\newblock In \emph{European Conference on Computer Vision (ECCV)}, 2024.

\bibitem[Voleti et~al.(2024)Voleti, Yao, Boss, Letts, Pankratz, Tochilkin,
  Laforte, Rombach, and Jampani]{voleti2024sv3d}
Vikram Voleti, Chun-Han Yao, Mark Boss, Adam Letts, David Pankratz, Dmitry
  Tochilkin, Christian Laforte, Robin Rombach, and Varun Jampani.
\newblock Sv3d: Novel multi-view synthesis and 3d generation from a single
  image using latent video diffusion.
\newblock In \emph{European Conference on Computer Vision}, pages 439--457.
  Springer, 2024.

\bibitem[von Werra et~al.(2020)von Werra, Belkada, Tunstall, Beeching, Thrush,
  Lambert, and Huang]{vonwerra2022trl}
Leandro von Werra, Younes Belkada, Lewis Tunstall, Edward Beeching, Tristan
  Thrush, Nathan Lambert, and Shengyi Huang.
\newblock Trl: Transformer reinforcement learning.
\newblock \url{https://github.com/huggingface/trl}, 2020.

\bibitem[Wang et~al.(2020{\natexlab{a}})Wang, Galliani, Vogel, Speciale, and
  Pollefeys]{wang2020patchmatchnetlearnedmultiviewpatchmatch}
Fangjinhua Wang, Silvano Galliani, Christoph Vogel, Pablo Speciale, and Marc
  Pollefeys.
\newblock Patchmatchnet: Learned multi-view patchmatch stereo,
  2020{\natexlab{a}}.
\newblock URL \url{https://arxiv.org/abs/2012.01411}.

\bibitem[Wang et~al.(2023{\natexlab{a}})Wang, Du, Li, Yeh, and
  Shakhnarovich]{Wang_2023_CVPR}
Haochen Wang, Xiaodan Du, Jiahao Li, Raymond Yeh, and Gregory Shakhnarovich.
\newblock Score jacobian chaining: Lifting pretrained 2d diffusion models for
  3d generation.
\newblock In \emph{Proceedings of the IEEE/CVF Conference on Computer Vision
  and Pattern Recognition (CVPR)}, pages 12619--12629, June 2023{\natexlab{a}}.

\bibitem[Wang et~al.(2023{\natexlab{b}})Wang, Du, Li, Yeh, and
  Shakhnarovich]{swang2023score}
Haochen Wang, Xiaodan Du, Jiahao Li, Raymond~A Yeh, and Greg Shakhnarovich.
\newblock Score jacobian chaining: Lifting pretrained 2d diffusion models for
  3d generation.
\newblock In \emph{Proceedings of the IEEE/CVF Conference on Computer Vision
  and Pattern Recognition}, pages 12619--12629, 2023{\natexlab{b}}.

\bibitem[Wang et~al.(2023{\natexlab{c}})Wang, Du, Li, Yeh, and
  Shakhnarovich]{wang2023score}
Haochen Wang, Xiaodan Du, Jiahao Li, Raymond~A Yeh, and Greg Shakhnarovich.
\newblock Score jacobian chaining: Lifting pretrained 2d diffusion models for
  3d generation.
\newblock In \emph{Proceedings of the IEEE/CVF Conference on Computer Vision
  and Pattern Recognition}, pages 12619--12629, 2023{\natexlab{c}}.

\bibitem[Wang et~al.(2023{\natexlab{d}})Wang, Rupprecht, and
  Novotny]{Wang_2023_ICCV}
Jianyuan Wang, Christian Rupprecht, and David Novotny.
\newblock Posediffusion: Solving pose estimation via diffusion-aided bundle
  adjustment.
\newblock In \emph{Proceedings of the IEEE/CVF International Conference on
  Computer Vision (ICCV)}, pages 9773--9783, October 2023{\natexlab{d}}.

\bibitem[Wang et~al.(2024{\natexlab{a}})Wang, Karaev, Rupprecht, and
  Novotny]{wang2024vggsfm}
Jianyuan Wang, Nikita Karaev, Christian Rupprecht, and David Novotny.
\newblock Vggsfm: Visual geometry grounded deep structure from motion.
\newblock In \emph{Proceedings of the IEEE/CVF Conference on Computer Vision
  and Pattern Recognition (CVPR)}, pages 21686--21697, June 2024{\natexlab{a}}.

\bibitem[Wang et~al.(2025{\natexlab{a}})Wang, Chen, Karaev, Vedaldi, Rupprecht,
  and Novotny]{Wang_2025_CVPR}
Jianyuan Wang, Minghao Chen, Nikita Karaev, Andrea Vedaldi, Christian
  Rupprecht, and David Novotny.
\newblock Vggt: Visual geometry grounded transformer.
\newblock In \emph{Proceedings of the IEEE/CVF Conference on Computer Vision
  and Pattern Recognition (CVPR)}, pages 5294--5306, June 2025{\natexlab{a}}.

\bibitem[Wang et~al.(2018)Wang, Zhang, Li, Fu, Liu, and Jiang]{Wang_2018_ECCV}
Nanyang Wang, Yinda Zhang, Zhuwen Li, Yanwei Fu, Wei Liu, and Yu-Gang Jiang.
\newblock Pixel2mesh: Generating 3d mesh models from single rgb images.
\newblock In \emph{Proceedings of the European Conference on Computer Vision
  (ECCV)}, 2018.

\bibitem[Wang et~al.(2025{\natexlab{b}})Wang, Zhang, Holynski, Efros, and
  Kanazawa]{cut3r}
Qianqian Wang, Yifei Zhang, Aleksander Holynski, Alexei~A. Efros, and Angjoo
  Kanazawa.
\newblock Continuous 3d perception model with persistent state.
\newblock In \emph{CVPR}, 2025{\natexlab{b}}.

\bibitem[Wang et~al.(2024{\natexlab{b}})Wang, Leroy, Cabon, Chidlovskii, and
  Revaud]{wang2024dust3r}
Shuzhe Wang, Vincent Leroy, Yohann Cabon, Boris Chidlovskii, and Jerome Revaud.
\newblock Dust3r: Geometric 3d vision made easy.
\newblock In \emph{Proceedings of the IEEE/CVF Conference on Computer Vision
  and Pattern Recognition (CVPR)}, pages 20697--20709, June 2024{\natexlab{b}}.

\bibitem[Wang et~al.(2020{\natexlab{b}})Wang, Zhu, Wang, Hu, Qiu, Wang, Hu,
  Kapoor, and Scherer]{tartanair}
Wenshan Wang, Delong Zhu, Xiangwei Wang, Yaoyu Hu, Yuheng Qiu, Chen Wang, Yafei
  Hu, Ashish Kapoor, and Sebastian Scherer.
\newblock Tartanair: A dataset to push the limits of visual slam.
\newblock In \emph{2020 IEEE/RSJ International Conference on Intelligent Robots
  and Systems (IROS)}, pages 4909--4916. IEEE, 2020{\natexlab{b}}.

\bibitem[Wang et~al.(2025{\natexlab{c}})Wang, Zhai, and Zuo]{pmlr-v258-wang25j}
Yikai Wang, Xinyu Zhai, and Wangmeng Zuo.
\newblock Steindreamer: Variance reduction for text-to-3d score distillation
  via stein identity.
\newblock In Aarti Singh and Genevieve Fried, editors, \emph{Proceedings of the
  42nd International Conference on Machine Learning}, volume 258 of
  \emph{Proceedings of Machine Learning Research}, pages 52547--52567. PMLR,
  13--19 Jul 2025{\natexlab{c}}.
\newblock URL \url{https://proceedings.mlr.press/v258/wang25j.html}.

\bibitem[Wang et~al.(2023{\natexlab{e}})Wang, Lu, Wang, Bao, Li, Su, and
  Zhu]{wang2023prolificdreamer}
Zhengyi Wang, Cheng Lu, Yikai Wang, Fan Bao, Chongxuan Li, Hang Su, and Jun
  Zhu.
\newblock Prolificdreamer: High-fidelity and diverse text-to-3d generation with
  variational score distillation.
\newblock \emph{arXiv preprint arXiv:2305.16213}, 2023{\natexlab{e}}.

\bibitem[Wang and Xu(2025)]{wang2025flashvggt}
Zipeng Wang and Dan Xu.
\newblock Flashvggt: Efficient and scalable visual geometry transformers with
  compressed descriptor attention.
\newblock \emph{arXiv preprint arXiv:2512.01540}, 2025.

\bibitem[Watson et~al.(2022)Watson, Chan, Martin-Brualla, Ho, Tagliasacchi, and
  Norouzi]{watson2022novel}
Daniel Watson, William Chan, Ricardo Martin-Brualla, Jonathan Ho, Andrea
  Tagliasacchi, and Mohammad Norouzi.
\newblock Novel view synthesis with diffusion models, 2022.

\bibitem[Weber et~al.(2024)Weber, Glaser, He, Sun, Ho, Johnson, Tulyakov, and
  Li]{Weber_2024_CVPR}
Sam Weber, Jonathan Glaser, Yingqing He, Weicheng Sun, Joshua Ho, Justin
  Johnson, Sergey Tulyakov, and Xiuming Li.
\newblock Nerfiller: Completing scenes via generative 3d inpainting.
\newblock In \emph{Proceedings of the IEEE/CVF Conference on Computer Vision
  and Pattern Recognition (CVPR)}, pages 20441--20452, June 2024.

\bibitem[Wei et~al.(2021)Wei, Bosma, Zhao, Guu, Yu, Lester, Du, Dai, and
  Le]{wei2021finetuned}
Jason Wei, Maarten Bosma, Vincent Zhao, Kelvin Guu, Adams~Wei Yu, Brian Lester,
  Nan Du, Andrew~M Dai, and Quoc~V Le.
\newblock Finetuned language models are zero-shot learners.
\newblock In \emph{International Conference on Learning Representations}, 2021.

\bibitem[Williams(1992)]{williams1992REINFORCE}
Ronald~J Williams.
\newblock Simple statistical gradient-following algorithms for connectionist
  reinforcement learning.
\newblock \emph{Machine learning}, 8:\penalty0 229--256, 1992.

\bibitem[Wilson and Snavely(2014{\natexlab{a}})]{wilson2014robust}
Kyle Wilson and Noah Snavely.
\newblock Robust global translations with {1DSfM}.
\newblock In \emph{ECCV}, 2014{\natexlab{a}}.

\bibitem[Wilson and Snavely(2014{\natexlab{b}})]{wilson_eccv2014_1dsfm}
Kyle Wilson and Noah Snavely.
\newblock Robust global translations with 1dsfm.
\newblock In \emph{Computer Vision -- ECCV 2014}, pages 61--75,
  2014{\natexlab{b}}.
\newblock \doi{10.1007/978-3-319-10578-9_5}.
\newblock URL \url{http://dx.doi.org/10.1007/978-3-319-10578-9_5}.

\bibitem[Wilson et~al.(2016)Wilson, Bindel, and Snavely]{Wilson_2016}
Kyle Wilson, David Bindel, and Noah Snavely.
\newblock When is rotations averaging hard?
\newblock In \emph{Computer Vision -- ECCV 2016}, pages 255--270, 2016.
\newblock \doi{10.1007/978-3-319-46478-7_16}.
\newblock URL \url{http://dx.doi.org/10.1007/978-3-319-46478-7_16}.

\bibitem[Workman et~al.(2015)Workman, Greenwell, Zhai, Baltenberger, and
  Jacobs]{09525ae48a2041789462b93743c9e0d3}
Scott Workman, Connor Greenwell, Menghua Zhai, Ryan Baltenberger, and Nathan
  Jacobs.
\newblock Deepfocal: A method for direct focal length estimation.
\newblock In \emph{2015 IEEE International Conference on Image Processing, ICIP
  2015 - Proceedings}, pages 1369--1373, 2015.
\newblock \doi{10.1109/ICIP.2015.7351024}.

\bibitem[Wu(2013)]{Wu_2013}
Changchang Wu.
\newblock Towards linear-time incremental structure from motion.
\newblock In \emph{2013 International Conference on 3D Vision}, pages 127--134,
  June 2013.
\newblock \doi{10.1109/3DV.2013.25}.
\newblock URL \url{http://dx.doi.org/10.1109/3DV.2013.25}.

\bibitem[Wu et~al.(2011)Wu, Agarwal, Curless, and Seitz]{Wu_2011}
Changchang Wu, Sameer Agarwal, Brian Curless, and Steven~M. Seitz.
\newblock Multicore bundle adjustment.
\newblock In \emph{CVPR 2011}, pages 3057--3064, June 2011.
\newblock \doi{10.1109/CVPR.2011.5995552}.
\newblock URL \url{http://dx.doi.org/10.1109/CVPR.2011.5995552}.

\bibitem[Wu et~al.(2016)Wu, Zhang, Xue, Freeman, and Tenenbaum]{wu2016learning}
Jiajun Wu, Chengkai Zhang, Tianfan Xue, Bill Freeman, and Joshua~B. Tenenbaum.
\newblock Learning a probabilistic latent space of object shapes via 3d
  generative-adversarial modeling.
\newblock In \emph{Advances in Neural Information Processing Systems}, pages
  82--90, 2016.

\bibitem[Wu et~al.(2024)Wu, Mildenhall, Henzler, Park, Gao, Watson, Srinivasan,
  Verbin, Barron, Poole, and Ho?y?ski]{Wu_2024_CVPR}
Rundi Wu, Ben Mildenhall, Philipp Henzler, Keunhong Park, Ruiqi Gao, Daniel
  Watson, Pratul~P. Srinivasan, Dor Verbin, Jonathan~T. Barron, Ben Poole, and
  Aleksander Ho?y?ski.
\newblock Reconfusion: 3d reconstruction with diffusion priors.
\newblock In \emph{Proceedings of the IEEE/CVF Conference on Computer Vision
  and Pattern Recognition (CVPR)}, pages 21551--21561, June 2024.

\bibitem[Wu et~al.(2023)Wu, Fei, Qu, Ji, and Chua]{wu2023nextgpt}
Shengqiong Wu, Hao Fei, Leigang Qu, Wei Ji, and Tat{-}Seng Chua.
\newblock Next-gpt: Any-to-any multimodal {LLM}.
\newblock \emph{CoRR}, abs/2309.05519, 2023.
\newblock \doi{10.48550/ARXIV.2309.05519}.
\newblock URL \url{https://doi.org/10.48550/arXiv.2309.05519}.

\bibitem[Wu et~al.(2015)Wu, Song, Khosla, Yu, Zhang, Tang, and
  Xiao]{Wu:2015:3SA}
Zhirong Wu, Shuran Song, Aditya Khosla, Fisher Yu, Linguang Zhang, Xiaoou Tang,
  and Jianxiong Xiao.
\newblock 3d shapenets: A deep representation for volumetric shapes.
\newblock In \emph{IEEE Conference on Computer Vision and Pattern Recognition
  (CVPR) oral presentation}, June 2015.

\bibitem[Xiang et~al.(2025)Xiang, Lv, Xu, Deng, Wang, Zhang, Chen, Tong, and
  Yang]{trellis}
Jianfeng Xiang, Zelong Lv, Sicheng Xu, Yu~Deng, Ruicheng Wang, Bowen Zhang,
  Dong Chen, Xin Tong, and Jiaolong Yang.
\newblock Structured 3d latents for scalable and versatile 3d generation.
\newblock In \emph{Proceedings of the Computer Vision and Pattern Recognition
  Conference}, pages 21469--21480, 2025.

\bibitem[Xie et~al.(2024)Xie, Li, Tan, Sun, Shu, Zhou, Bi, Pirk, and
  Kaufman]{xie2024carve3d}
Desai Xie, Jiahao Li, Hao Tan, Xin Sun, Zhixin Shu, Yi~Zhou, Sai Bi, S\"{o}ren
  Pirk, and Arie~E. Kaufman.
\newblock Carve3d: Improving multi-view reconstruction consistency for
  diffusion models with rl finetuning.
\newblock In \emph{Proceedings of the IEEE/CVF Conference on Computer Vision
  and Pattern Recognition (CVPR)}, pages 6369--6379, June 2024.
\newblock URL
  \url{https://openaccess.thecvf.com/content/CVPR2024/html/Xie_Carve3D_Improving_Multi-view_Reconstruction_Consistency_for_Diffusion_Models_with_RL_CVPR_2024_paper.html}.

\bibitem[Xie et~al.(2021)Xie, Wang, Yu, Anandkumar, Alvarez, and
  Luo]{xie2021segformersimpleefficientdesign}
Enze Xie, Wenhai Wang, Zhiding Yu, Anima Anandkumar, Jose~M. Alvarez, and Ping
  Luo.
\newblock Segformer: Simple and efficient design for semantic segmentation with
  transformers, 2021.
\newblock URL \url{https://arxiv.org/abs/2105.15203}.

\bibitem[Xie et~al.(2023)Xie, Zhang, Lin, Hinz, and Zhang]{xie2023smartbrush}
Shaoan Xie, Zhifei Zhang, Zhe Lin, Tobias Hinz, and Kun Zhang.
\newblock Smartbrush: Text and shape guided object inpainting with diffusion
  model.
\newblock In \emph{Proceedings of the IEEE/CVF Conference on Computer Vision
  and Pattern Recognition}, pages 22428--22437, 2023.

\bibitem[Xie et~al.(2025)Xie, Yao, Voleti, Jiang, and
  Jampani]{xie2025sv4ddynamic3dcontent}
Yiming Xie, Chun-Han Yao, Vikram Voleti, Huaizu Jiang, and Varun Jampani.
\newblock Sv4d: Dynamic 3d content generation with multi-frame and multi-view
  consistency, 2025.
\newblock URL \url{https://arxiv.org/abs/2407.17470}.

\bibitem[Xu et~al.(2025)Xu, Guo, He, Hu, He, Bai, Chen, Wang, Fan, Dang, Zhang,
  Wang, Chu, and Lin]{xu2025qwen25omni}
Jin Xu, Zhifang Guo, Jinzheng He, Hangrui Hu, Ting He, Shuai Bai, Keqin Chen,
  Jialin Wang, Yang Fan, Kai Dang, Bin Zhang, Xiong Wang, Yunfei Chu, and
  Junyang Lin.
\newblock Qwen2.5-omni technical report.
\newblock \emph{CoRR}, abs/2503.20215, 2025.
\newblock \doi{10.48550/ARXIV.2503.20215}.
\newblock URL \url{https://doi.org/10.48550/arXiv.2503.20215}.

\bibitem[Xu et~al.(2023{\natexlab{a}})Xu, Agrawal, Laney, Garcia, Bansal, Kim,
  Rota~Bulò, Porzi, Kontschieder, Božič, Lin, Zollhöfer, and
  Richardt]{VRNeRF}
Linning Xu, Vasu Agrawal, William Laney, Tony Garcia, Aayush Bansal, Changil
  Kim, Samuel Rota~Bulò, Lorenzo Porzi, Peter Kontschieder, Aljaž Božič,
  Dahua Lin, Michael Zollhöfer, and Christian Richardt.
\newblock {VR-NeRF}: {H}igh-fidelity virtualized walkable spaces.
\newblock In \emph{SIGGRAPH Asia Conference Proceedings}, 2023{\natexlab{a}}.
\newblock \doi{10.1145/3610548.3618139}.

\bibitem[Xu et~al.(2023{\natexlab{b}})Xu, Tan, Luan, Bi, Wang, Li, Shi,
  Sunkavalli, Wetzstein, Xu, et~al.]{xu2023dmv3d}
Yinghao Xu, Hao Tan, Fujun Luan, Sai Bi, Peng Wang, Jiahao Li, Zifan Shi,
  Kalyan Sunkavalli, Gordon Wetzstein, Zexiang Xu, et~al.
\newblock Dmv3d: Denoising multi-view diffusion using 3d large reconstruction
  model.
\newblock \emph{arXiv preprint arXiv:2311.09217}, 2023{\natexlab{b}}.

\bibitem[Xue et~al.(2020)Xue, Wu, Bai, Wang, Xia, Zhang, and
  Torr]{xue2020holisticallyattractedwireframeparsing}
Nan Xue, Tianfu Wu, Song Bai, Fu-Dong Wang, Gui-Song Xia, Liangpei Zhang, and
  Philip H.~S. Torr.
\newblock Holistically-attracted wireframe parsing, 2020.
\newblock URL \url{https://arxiv.org/abs/2003.01663}.

\bibitem[Yan et~al.(2025)Yan, Xu, Lin, Jin, Guo, Wang, Zhan, Lang, Bao, Zhou,
  and Peng]{yan2024streetcrafter}
Yunzhi Yan, Zhen Xu, Haotong Lin, Haian Jin, Haoyu Guo, Yida Wang, Kun Zhan,
  Xianpeng Lang, Hujun Bao, Xiaowei Zhou, and Sida Peng.
\newblock Streetcrafter: Street view synthesis with controllable video
  diffusion models.
\newblock In \emph{Proceedings of the IEEE/CVF Conference on Computer Vision
  and Pattern Recognition (CVPR)}, 2025.

\bibitem[Yang et~al.(2019)Yang, Huang, Hao, Liu, Belongie, and
  Hariharan]{Yang_2019_ICCV}
Guandao Yang, Xun Huang, Zekun Hao, Ming-Yu Liu, Serge Belongie, and Bharath
  Hariharan.
\newblock Pointflow: 3d point cloud generation with continuous normalizing
  flows.
\newblock In \emph{The IEEE International Conference on Computer Vision
  (ICCV)}, October 2019.

\bibitem[Yang et~al.(2025)Yang, Sax, Liang, Henaff, Tang, Cao, Chai, Meier, and
  Feiszli]{yang2025fast3r}
Jianing Yang, Alexander Sax, Kevin~J Liang, Mikael Henaff, Hao Tang, Ang Cao,
  Joyce Chai, Franziska Meier, and Matt Feiszli.
\newblock Fast3r: Towards 3d reconstruction of 1000+ images in one forward
  pass.
\newblock \emph{arXiv preprint arXiv:2501.13928}, 2025.

\bibitem[Yang et~al.(2018)Yang, Feng, Shen, and Tian]{Yang_2018_CVPR}
Yaoqing Yang, Chen Feng, Yiru Shen, and Dong Tian.
\newblock Foldingnet: Point cloud auto-encoder via deep grid deformation.
\newblock In \emph{The IEEE Conference on Computer Vision and Pattern
  Recognition (CVPR)}, June 2018.

\bibitem[Yang et~al.(2024)Yang, Yin, Fan, Chen, Li, and
  Yu]{yang2024scene123prompt3dscene}
Yiying Yang, Fukun Yin, Jiayuan Fan, Xin Chen, Wanzhang Li, and Gang Yu.
\newblock Scene123: One prompt to 3d scene generation via video-assisted and
  consistency-enhanced mae, 2024.
\newblock URL \url{https://arxiv.org/abs/2408.05477}.

\bibitem[Yao et~al.(2018)Yao, Luo, Li, Fang, and
  Quan]{yao2018mvsnetdepthinferenceunstructured}
Yao Yao, Zixin Luo, Shiwei Li, Tian Fang, and Long Quan.
\newblock Mvsnet: Depth inference for unstructured multi-view stereo, 2018.
\newblock URL \url{https://arxiv.org/abs/1804.02505}.

\bibitem[Yao et~al.(2019)Yao, Luo, Li, Shen, Fang, and
  Quan]{yao2019recurrentmvsnethighresolutionmultiview}
Yao Yao, Zixin Luo, Shiwei Li, Tianwei Shen, Tian Fang, and Long Quan.
\newblock Recurrent mvsnet for high-resolution multi-view stereo depth
  inference, 2019.
\newblock URL \url{https://arxiv.org/abs/1902.10556}.

\bibitem[Yao et~al.(2023)Yao, Aminabadi, Ruwase, Rajbhandari, Wu, Awan, Rasley,
  Zhang, Li, Holmes, Zhou, Wyatt, Smith, Kurilenko, Qin, Tanaka, Che, Song, and
  He]{yao2023deepspeedchat}
Zhewei Yao, Reza~Yazdani Aminabadi, Olatunji Ruwase, Samyam Rajbhandari,
  Xiaoxia Wu, Ammar~Ahmad Awan, Jeff Rasley, Minjia Zhang, Conglong Li, Connor
  Holmes, Zhongzhu Zhou, Michael Wyatt, Molly Smith, Lev Kurilenko, Heyang Qin,
  Masahiro Tanaka, Shuai Che, Shuaiwen~Leon Song, and Yuxiong He.
\newblock Deepspeed-chat: Easy, fast and affordable rlhf training of
  chatgpt-like models at all scales, 2023.

\bibitem[Yi et~al.(2016)Yi, Trulls, Lepetit, and
  Fua]{yi2016liftlearnedinvariantfeature}
Kwang~Moo Yi, Eduard Trulls, Vincent Lepetit, and Pascal Fua.
\newblock Lift: Learned invariant feature transform, 2016.
\newblock URL \url{https://arxiv.org/abs/1603.09114}.

\bibitem[Yi et~al.(2024)Yi, Xu, Wang, and Bo]{Yi_2024_CVPR}
Ziyi Yi, Huizhuo Xu, Yuxi Wang, and Liefeng Bo.
\newblock Gaussiandreamer: Fast generation from text to 3d gaussian splatting
  with point cloud priors.
\newblock In \emph{Proceedings of the IEEE/CVF Conference on Computer Vision
  and Pattern Recognition (CVPR)}, pages 11207--11216, June 2024.

\bibitem[You et~al.(2025)You, Zhu, Liu, and Hou]{you2025nvs}
Meng You, Zhiyu Zhu, Hui Liu, and Junhui Hou.
\newblock Nvs-solver: Video diffusion model as zero-shot novel view
  synthesizer.
\newblock In \emph{International Conference on Learning Representations}, 2025.

\bibitem[Yu et~al.(2021)Yu, Ye, Tancik, and Kanazawa]{yu2020pixelnerf}
Alex Yu, Vickie Ye, Matthew Tancik, and Angjoo Kanazawa.
\newblock pixelnerf: Neural radiance fields from one or few images.
\newblock In \emph{IEEE Conference on Computer Vision and Pattern Recognition
  (CVPR)}, pages 4578--4587, 2021.
\newblock \doi{10.1109/CVPR46437.2021.00455}.
\newblock URL
  \url{https://openaccess.thecvf.com/content/CVPR2021/html/Yu_pixelNeRF_Neural_Radiance_Fields_From_One_or_Few_Images_CVPR_2021_paper.html}.

\bibitem[Yu et~al.(2023{\natexlab{a}})Yu, Shi, Pasunuru, Muller, Golovneva,
  Wang, Babu, Tang, Karrer, Sheynin, Ross, Polyak, Howes, Sharma, Xu, Tamoyan,
  Ashual, Singer, Li, Zhang, James, Ghosh, Taigman, Fazel{-}Zarandi,
  Celikyilmaz, Zettlemoyer, and Aghajanyan]{yu2023cm3leon}
Lili Yu, Bowen Shi, Ramakanth Pasunuru, Benjamin Muller, Olga Golovneva, Tianlu
  Wang, Arun Babu, Binh Tang, Brian Karrer, Shelly Sheynin, Candace Ross, Adam
  Polyak, Russell Howes, Vasu Sharma, Puxin Xu, Hovhannes Tamoyan, Oron Ashual,
  Uriel Singer, Shang{-}Wen Li, Susan Zhang, Richard James, Gargi Ghosh, Yaniv
  Taigman, Maryam Fazel{-}Zarandi, Asli Celikyilmaz, Luke Zettlemoyer, and
  Armen Aghajanyan.
\newblock Scaling autoregressive multi-modal models: Pretraining and
  instruction tuning.
\newblock \emph{CoRR}, abs/2309.02591, 2023{\natexlab{a}}.
\newblock \doi{10.48550/ARXIV.2309.02591}.
\newblock URL \url{https://doi.org/10.48550/arXiv.2309.02591}.

\bibitem[Yu et~al.(2024)Yu, Xing, Yuan, Hu, Li, Huang, Gao, Wong, Shan, and
  Tian]{yu2024viewcraftertamingvideodiffusion}
Wangbo Yu, Jinbo Xing, Li~Yuan, Wenbo Hu, Xiaoyu Li, Zhipeng Huang, Xiangjun
  Gao, Tien-Tsin Wong, Ying Shan, and Yonghong Tian.
\newblock Viewcrafter: Taming video diffusion models for high-fidelity novel
  view synthesis, 2024.
\newblock URL \url{https://arxiv.org/abs/2409.02048}.

\bibitem[Yu et~al.(2023{\natexlab{b}})Yu, Xu, Zhang, Liu, Ye, Wu, Yan, Zhu,
  Xiong, Liang, et~al.]{mvimgnet}
Xianggang Yu, Mutian Xu, Yidan Zhang, Haolin Liu, Chongjie Ye, Yushuang Wu,
  Zizheng Yan, Chenming Zhu, Zhangyang Xiong, Tianyou Liang, et~al.
\newblock Mvimgnet: A large-scale dataset of multi-view images.
\newblock In \emph{CVPR}, pages 9150--9161, 2023{\natexlab{b}}.

\bibitem[Yu et~al.(2025)Yu, Hu, Zhan, Wang, Xie, Zhang, and Qin]{Yu_2025_CVPR}
Yunbo Yu, Xiaoyang Hu, Xin Zhan, Zhe Wang, Lixin Xie, Tong Zhang, and Pengda
  Qin.
\newblock Wonderworld: Interactive 3d scene generation from a single image.
\newblock In \emph{Proceedings of the IEEE/CVF Conference on Computer Vision
  and Pattern Recognition (CVPR)}, pages 24508--24518, 2025.

\bibitem[Yuan et~al.(2026)Yuan, Yang, Yang, Zhang, Zhao, Zhang, and
  Zhang]{yuan2026infinitevggt}
Shuai Yuan, Yantai Yang, Xiaotian Yang, Xupeng Zhang, Zhonghao Zhao, Lingming
  Zhang, and Zhipeng Zhang.
\newblock Infinitevggt: Visual geometry grounded transformer for endless
  streams, 2026.

\bibitem[Zeng et~al.(2022)Zeng, Vahdat, Williams, Gojcic, Litany, Fidler, and
  Kreis]{zeng2022lion}
Xiaohui Zeng, Arash Vahdat, Francis Williams, Zan Gojcic, Or~Litany, Sanja
  Fidler, and Karsten Kreis.
\newblock Lion: Latent point diffusion models for 3d shape generation.
\newblock In \emph{Advances in Neural Information Processing Systems
  (NeurIPS)}, 2022.

\bibitem[Zhang et~al.(2022)Zhang, Ramanan, and Tulsiani]{zhang2022relpose}
Jason~Y. Zhang, Deva Ramanan, and Shubham Tulsiani.
\newblock {RelPose}: Predicting probabilistic relative rotation for single
  objects in the wild.
\newblock In \emph{European Conference on Computer Vision (ECCV)}, 2022.

\bibitem[Zhang et~al.(2024{\natexlab{a}})Zhang, Lin, Kumar, Yang, Ramanan, and
  Tulsiani]{zhang2024raydiffusion}
Jason~Y Zhang, Amy Lin, Moneish Kumar, Tzu-Hsuan Yang, Deva Ramanan, and
  Shubham Tulsiani.
\newblock Cameras as rays: Pose estimation via ray diffusion.
\newblock In \emph{International Conference on Learning Representations
  (ICLR)}, 2024{\natexlab{a}}.

\bibitem[Zhang et~al.(2019)Zhang, Sun, Luo, Yao, Zhou, Shen, Chen, Quan, and
  Liao]{zhang2019learningtwoviewcorrespondencesgeometry}
Jiahui Zhang, Dawei Sun, Zixin Luo, Anbang Yao, Lei Zhou, Tianwei Shen, Yurong
  Chen, Long Quan, and Hongen Liao.
\newblock Learning two-view correspondences and geometry using order-aware
  network, 2019.
\newblock URL \url{https://arxiv.org/abs/1908.04964}.

\bibitem[Zhang et~al.(2023)Zhang, Li, Wan, Wang, and Liao]{zhang2023text2nerf}
Jingbo Zhang, Xiaoyu Li, Ziyu Wan, Can Wang, and Jing Liao.
\newblock Text2nerf: Text-driven 3d scene generation with neural radiance
  fields.
\newblock \emph{arXiv preprint arXiv:2305.11588}, 2023.

\bibitem[Zhang et~al.(2025{\natexlab{a}})Zhang, Xia, Dong, Shen, Yue, and
  Zheng]{Zhang_2025_CVPR}
Jintao Zhang, Zimin Xia, Mingyue Dong, Shuhan Shen, Linwei Yue, and Xianwei
  Zheng.
\newblock Comatcher: Multi-view collaborative feature matching.
\newblock In \emph{Proceedings of the IEEE/CVF Conference on Computer Vision
  and Pattern Recognition (CVPR)}, pages 21970--21980, June 2025{\natexlab{a}}.

\bibitem[Zhang et~al.(2024{\natexlab{b}})Zhang, Bi, Tan, Xiangli, Zhao,
  Sunkavalli, and Xu]{gslrm2024}
Kai Zhang, Sai Bi, Hao Tan, Yuanbo Xiangli, Nanxuan Zhao, Kalyan Sunkavalli,
  and Zexiang Xu.
\newblock Gs-lrm: Large reconstruction model for 3d gaussian splatting.
\newblock \emph{European Conference on Computer Vision}, 2024{\natexlab{b}}.

\bibitem[Zhang et~al.(2018{\natexlab{a}})Zhang, Isola, Efros, Shechtman, and
  Wang]{zhang2018lpips}
Richard Zhang, Phillip Isola, Alexei~A Efros, Eli Shechtman, and Oliver Wang.
\newblock The unreasonable effectiveness of deep features as a perceptual
  metric.
\newblock In \emph{CVPR}, 2018{\natexlab{a}}.

\bibitem[Zhang et~al.(2018{\natexlab{b}})Zhang, Isola, Efros, Shechtman, and
  Wang]{zhang2018perceptual}
Richard Zhang, Phillip Isola, Alexei~A Efros, Eli Shechtman, and Oliver Wang.
\newblock The unreasonable effectiveness of deep features as a perceptual
  metric.
\newblock In \emph{CVPR}, 2018{\natexlab{b}}.

\bibitem[Zhang et~al.(2025{\natexlab{b}})Zhang, Zhang, Mehl, Gross, and
  Schroers]{zhang2025highfidelitynovelviewsynthesis}
Xiang Zhang, Yang Zhang, Lukas Mehl, Markus Gross, and Christopher Schroers.
\newblock High-fidelity novel view synthesis via splatting-guided diffusion,
  2025{\natexlab{b}}.
\newblock URL \url{https://arxiv.org/abs/2502.12752}.

\bibitem[Zhang et~al.(2024{\natexlab{c}})Zhang, Zhang, Zhang, Mu, Huang,
  B{\"u}scher, and Xu]{Zhang_2024_CVPR}
Zhongxiang Zhang, Zhe Zhang, Yue Zhang, Jiahui Mu, Yisong Huang, Daniel
  B{\"u}scher, and Ziwei Xu.
\newblock 3d-scenedreamer: Text-driven 3d-consistent scene generation.
\newblock In \emph{Proceedings of the IEEE/CVF Conference on Computer Vision
  and Pattern Recognition (CVPR)}, pages 20270--20280, June 2024{\natexlab{c}}.

\bibitem[Zhao et~al.(2021)Zhao, Ge, Zhu, Zhao, Li, and
  Salzmann]{Zhao_2021_ICCV}
Chen Zhao, Yixiao Ge, Feng Zhu, Rui Zhao, Hongsheng Li, and Mathieu Salzmann.
\newblock Progressive correspondence pruning by consensus learning.
\newblock In \emph{Proceedings of the IEEE/CVF International Conference on
  Computer Vision (ICCV)}, pages 6464--6473, October 2021.

\bibitem[Zhao et~al.(2023)Zhao, Zhao, Liang, Li, Zhao, Hu, Fan, and
  Yu]{zhao2023efficientdreamer}
Minda Zhao, Chaoyi Zhao, Xinyue Liang, Lincheng Li, Zeng Zhao, Zhipeng Hu,
  Changjie Fan, and Xin Yu.
\newblock Efficientdreamer: High-fidelity and robust 3d creation via
  orthogonal-view diffusion prior.
\newblock \emph{arXiv preprint arXiv:2308.13223}, 2023.

\bibitem[Zhao et~al.(2025)Zhao, Lin, Tan, Zhang, Ramanan, and
  Tulsiani]{zhao2025diffusionsfm}
Qitao Zhao, Amy Lin, Jeff Tan, Jason~Y Zhang, Deva Ramanan, and Shubham
  Tulsiani.
\newblock Diffusionsfm: Predicting structure and motion via ray origin and
  endpoint diffusion.
\newblock In \emph{Proceedings of the Computer Vision and Pattern Recognition
  Conference}, pages 6317--6326, 2025.

\bibitem[Zheng et~al.(2023{\natexlab{a}})Zheng, Dou, Gao, Hua, Shen, Wang, Liu,
  Jin, Liu, Zhou, Xiong, Chen, Xi, Xu, Lai, Zhu, Chang, Yin, Weng, Cheng,
  Huang, Sun, Yan, Gui, Zhang, Qiu, and Huang]{zheng2023RLHFsecrets}
Rui Zheng, Shihan Dou, Songyang Gao, Yuan Hua, Wei Shen, Binghai Wang, Yan Liu,
  Senjie Jin, Qin Liu, Yuhao Zhou, Limao Xiong, Lu~Chen, Zhiheng Xi, Nuo Xu,
  Wenbin Lai, Minghao Zhu, Cheng Chang, Zhangyue Yin, Rongxiang Weng, Wensen
  Cheng, Haoran Huang, Tianxiang Sun, Hang Yan, Tao Gui, Qi~Zhang, Xipeng Qiu,
  and Xuanjing Huang.
\newblock Secrets of rlhf in large language models part i: Ppo,
  2023{\natexlab{a}}.

\bibitem[Zheng et~al.(2023{\natexlab{b}})Zheng, Harley, Shen, Wetzstein, and
  Guibas]{pointodyssey}
Yang Zheng, Adam~W Harley, Bokui Shen, Gordon Wetzstein, and Leonidas~J Guibas.
\newblock Pointodyssey: A large-scale synthetic dataset for long-term point
  tracking.
\newblock In \emph{Proceedings of the IEEE/CVF International Conference on
  Computer Vision}, pages 19855--19865, 2023{\natexlab{b}}.

\bibitem[Zhong et~al.(2025)Zhong, Zhan, Gao, Chen, Lou, Mao, Neumann, and
  Wang]{zhong2025instantsfmfullysparseparallel}
Jiankun Zhong, Zitong Zhan, Quankai Gao, Ziyu Chen, Haozhe Lou, Jiageng Mao,
  Ulrich Neumann, and Yue Wang.
\newblock Instantsfm: Fully sparse and parallel structure-from-motion, 2025.
\newblock URL \url{https://arxiv.org/abs/2510.13310}.

\bibitem[Zhou et~al.(2023)Zhou, Liu, Xu, Iyer, Sun, Mao, Ma, Efrat, Yu, Yu,
  et~al.]{zhou2023lima}
Chunting Zhou, Pengfei Liu, Puxin Xu, Srini Iyer, Jiao Sun, Yuning Mao, Xuezhe
  Ma, Avia Efrat, Ping Yu, Lili Yu, et~al.
\newblock Lima: Less is more for alignment.
\newblock \emph{arXiv preprint arXiv:2305.11206}, 2023.

\bibitem[Zhou et~al.(2025)Zhou, Gao, Voleti, Vasishta, Yao, Boss, Torr,
  Rupprecht, and Jampani]{zhou2025stablevirtualcameragenerative}
Jensen Zhou, Hang Gao, Vikram Voleti, Aaryaman Vasishta, Chun-Han Yao, Mark
  Boss, Philip Torr, Christian Rupprecht, and Varun Jampani.
\newblock Stable virtual camera: Generative view synthesis with diffusion
  models, 2025.
\newblock URL \url{https://arxiv.org/abs/2503.14489}.

\bibitem[Zhou et~al.(2021)Zhou, Sattler, and
  Leal-Taixe]{zhou2021patch2pixepipolarguidedpixellevelcorrespondences}
Qunjie Zhou, Torsten Sattler, and Laura Leal-Taixe.
\newblock Patch2pix: Epipolar-guided pixel-level correspondences, 2021.
\newblock URL \url{https://arxiv.org/abs/2012.01909}.

\bibitem[Zhou et~al.(2019)Zhou, Barnes, Lu, Yang, and Li]{zhou2019continuity}
Yi~Zhou, Connelly Barnes, Jingwan Lu, Jimei Yang, and Hao Li.
\newblock On the continuity of rotation representations in neural networks.
\newblock In \emph{CVPR}, 2019.

\bibitem[Zhu et~al.(2025)Zhu, Zhang, Zhang, Liu, Wen, Cheng, Li, Tang, and
  Huang]{zhu2025light3rsfm}
Meidong Zhu, Jijie Zhang, Fangyun Zhang, Jiabao Liu, Chenglu Wen, Ziyuan Cheng,
  Haojie Li, Yanjun Tang, and Can Huang.
\newblock Light3r-sfm: Towards feed-forward structure-from-motion.
\newblock \emph{arXiv preprint arXiv:2504.07855}, 2025.

\bibitem[Zhuang et~al.(2018)Zhuang, Cheong, and Lee]{zhuang2018baseline}
Bingbing Zhuang, Loong-Fah Cheong, and Gim~Hee Lee.
\newblock Baseline desensitizing in translation averaging.
\newblock In \emph{CVPR}, 2018.

\bibitem[Zhuo et~al.(2025)Zhuo, Zheng, Guo, Wu, Zhou, and Lu]{streamVGGT}
Dong Zhuo, Wenzhao Zheng, Jiahe Guo, Yuqi Wu, Jie Zhou, and Jiwen Lu.
\newblock Streaming 4d visual geometry transformer.
\newblock \emph{arXiv preprint arXiv:2507.11539}, 2025.

\end{thebibliography}

\end{document}